%% file: ms.tex
\documentclass[10pt,twocolumn,letterpaper]{article}

\usepackage{3dv}
\usepackage{times}
\usepackage{epsfig}
\usepackage{graphicx}
\usepackage{amsmath}
\usepackage{amssymb}

\usepackage{booktabs}
\usepackage{multicol}
\usepackage{subcaption}
\usepackage{textcomp}
\usepackage{amsthm}
\usepackage{multirow}
\usepackage[toc,page]{appendix}

\usepackage{balance}

\newtheorem{theorem}{Theorem}

\makeatletter
\@namedef{ver@everyshi.sty}{}
\makeatother
\usepackage{tikz}

\usepackage[pagebackref=true,breaklinks=true,colorlinks,bookmarks=false]{hyperref}

\threedvfinalcopy 

\def\threedvPaperID{312} 

\ifthreedvfinal\fi

\newcommand\copyrighttext{%
	\footnotesize \textcopyright 2021 IEEE. Personal use of this material is permitted.
	Permission from IEEE must be obtained for all other uses, in any current or future
	media, including reprinting/republishing this material for advertising or promotional
	purposes, creating new collective works, for resale or redistribution to servers or
	lists, or reuse of any copyrighted component of this work in other works.
	Presented at the 2021 International Conference on 3D Vision (3DV). DOI: 10.1109/3DV53792.2021.00020. Publisher version: https://ieeexplore.ieee.org/document/9665839.
	}
\newcommand\copyrightnotice{%
	\begin{tikzpicture}[remember picture,overlay]
	\node[anchor=south west,yshift=10pt,xshift=1.6cm] at (current page.south west) {\parbox{\textwidth}{\copyrighttext}};
	\end{tikzpicture}%
}

\begin{document}

\title{DiffSDFSim: Differentiable Rigid-Body Dynamics With Implicit Shapes}

\author{Michael Strecke and Joerg Stueckler\\
Embodied Vision Group, Max Planck Institute for Intelligent Systems, T\"ubingen, Germany\\
{\tt\small \{michael.strecke,joerg.stueckler\}@tuebingen.mpg.de}
}

\maketitle
\copyrightnotice
\thispagestyle{empty}

\begin{abstract}
\input{abstract}

\end{abstract}

\section{Introduction}

\input{introduction}

\section{Related work}
\input{related_work}

\section{Background}
\input{background}

\section{Method}

\input{method}

\section{Experiments}
\input{experiments}

\section{Conclusion}
\input{conclusion}

\paragraph{Acknowledgements}
This work was supported by Cyber Valley, the Max Planck Society and the German Federal Ministry of Education and Research (BMBF) through the Tuebingen AI Center  (FKZ: 01IS18039B). 
The authors thank the International Max Planck Research School for Intelligent Systems (IMPRS-IS) for supporting Michael Strecke.

{\small
\bibliographystyle{ieee_fullname}
\bibliography{references/references}
\balance
}

\newpage
\nobalance
\begin{appendices}
\input{supplement}

\end{appendices}

\end{document}

%% file: abstract.tex
Differentiable physics is a powerful tool in computer vision and robotics for scene understanding and reasoning about interactions.
Existing approaches have frequently been limited to objects with simple shape or shapes that are known in advance.
In this paper, we propose a novel approach to differentiable physics with frictional contacts which represents object shapes implicitly using signed distance fields (SDFs).
Our simulation supports contact point calculation even when the involved shapes are nonconvex.
Moreover, we propose ways for differentiating the dynamics for the object shape to facilitate shape optimization using gradient-based methods.
In our experiments, we demonstrate that our approach allows for model-based inference of physical parameters such as friction coefficients, mass, forces or shape parameters from trajectory and depth image observations in several challenging synthetic scenarios and a real image sequence.

%% file: introduction.tex
Physical and dynamic scene understanding is an important capability for intelligent agents to perceive and reason about the environment.
Against this backdrop, differentiable physics simulations have recently attracted interest in the computer vision and robotics communities as they allow for the identification of physical parameters~\cite{hu2019_difftaichi,krishnamurthy2021_gradsim,geilinger2020_add}.
This potentially enables scene models that allow autonomous agents to reason about interactions with the environment.

\begin{figure}
	\centering
	\includegraphics[width=0.99\linewidth]{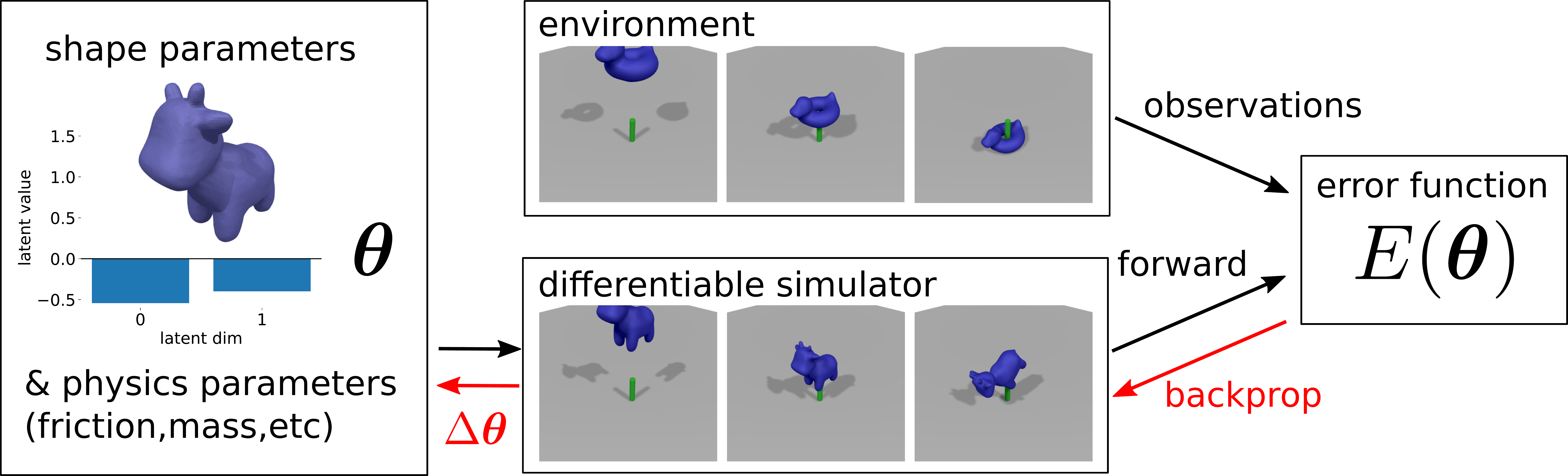}
	\caption{Shape and physics parameter optimization through differentiable physics simulation. We represent arbitary watertight shapes by signed distance functions (SDF) and optimize the parameters through the physics-based dynamics to fit trajectories and depth image observations.}
	\label{fig:teaser}
\end{figure}

However, existing approaches have frequently been limited to objects with primitive shapes or shapes that are known in advance.
They are thus not applicable if only a coarse estimate of the shape is available and the refinement of that shape as part of the physical properties is needed.
To address this issue, we develop a framework capable of simulating objects of complex shapes and optimizing the shape and other physical parameters of these objects such as mass and friction (see Fig.~\ref{fig:teaser}).
We represent shapes using signed distance functions (SDFs), implicit shape representations that allow for low-dimensional description and detection of collisions for complex shapes.
We propose a new differential through the physical dynamics with contacts that allows us to optimize shape parameters.

We base our approach on a class of constraint velocity-based dynamics simulation methods~\cite{stewart1996_implicit} which lead to linear complementarity problems (LCPs) that can be differentiated at the solution~\cite{avilabelbuteperes2018_end}.
We represent shapes using signed distance functions (SDFs), which represent the object shape as the zero-level set of the signed distance of 3D points to the surface.
The sign of the distance defines the inside or outside of the object.
In our approach, we assume that this mapping is differentiable for the coordinates of the 3D points and shape parameters.
We devise methods for differentiating time of contact and mass-inertia tensors which enables gradient-based shape optimization.
In experiments, we demonstrate that our approach can be used in optimization approaches which identify physical parameters such as shape, forces, mass and friction from sample trajectories and depth image observations.

In summary, the contributions of our work are:
\begin{itemize}\setlength\itemsep{0em}
	\item We propose a novel differentiable physics simulation approach which supports contact handling and differentiable mass-inertia tensor calculation for arbitrary watertight shapes represented as SDFs.
	\item We develop a novel formulation of differentiable time of contact which enables gradient-based shape optimization through collision constraints.
	\item We demonstrate that our approach makes shape optimization and system identification feasible for shapes modelled by SDFs from object trajectories and depth image observations in several synthetic scenarios and a real RGB-D image sequence.
\end{itemize}

%% file: related_work.tex
\paragraph{Physics simulation}
\label{sec:sota_physicssimulation}

Over the last decades, a large body of methods for physical simulation has been developed in the computer graphics and mechanical engineering communities~\cite{bender2013_interactive}.
Physics-based simulation methods can be distinguished as following different paradigms: time stepping or impulse-based.
Time stepping methods formulate the dynamics by Newton-Euler equations and add equality and inequality constraints to model joints, contacts and collisions.
The methods can be phrased as position-based~\cite{mueller2007_position}, velocity-based~\cite{stewart1996_implicit,anitescu1997_formulating}, or acceleration-based~\cite{baraff1996_linear} optimization problems.
Position-based methods often solve for collision and joint constraints using Gauss-Seidel and fast projection methods.
Friction and impulse-conservation laws have to be included in a post-processing step which augment the velocities of the objects.
While the methods have been demonstrated to yield visibly plausible results, they are typically difficult to tune for accuracy.
Position-based methods can also be extended to simulate deformable objects and fluids by solving collision and deformation constraints between particles.
Recently, a differentiable position-based approach has been proposed~\cite{macklin2020_local}.

Acceleration-based approaches estimate constraint forces through solving complementarity problems.
They can suffer from indeterminancy for Coulomb friction which can be solved by reformulating the optimization problem with contact impulses, leading to a velocity-based formulation~\cite{stewart1996_implicit}.
In this paper, we also follow a constraint-based velocity-based approach as proposed in~\cite{anitescu1997_formulating}.
Belbutes \etal \cite{avilabelbuteperes2018_end} make this formulation differentiable at the solution using the OptNet approach~\cite{amos2017_optnet}.
Recently, the method has also been extended to increase efficiency for many objects and mesh-based collision detection~\cite{qiao2020_scalable}.
\cite{geilinger2020_add} propose a mollified contact model for frictional contacts which can be applied for rigid and deformable objects.
Differently, we represent shapes using signed distance functions in a differentiable way and derive a novel time of contact differential which allows for shape optimization from collision constraints.

Another line of research are event-driven impulse-based methods~\cite{mirtich1995_impulse,weinstein2006_dynamic,bender2006_fast} which have been introduced in the seminal work of~\cite{mirtich1995_impulse}.
Impulse-based methods iteratively update the velocities of the rigid bodies at the events of contacts until all joint and collision constraints are satisfied.
Other related simulation methods are finite element~\cite{terzopoulos1988_modeling} or meshless methods such as~\cite{sulsky1995_application,hu2019_chainqueen} which are used to simulate deformables and fluids.
In contrast to our LCP-based rigid-body physics formulation, these methods cannot model strictly rigid objects or hard collision constraints.

\paragraph{Physical system identification}
\label{sec:sota_physicalsceneunderstanding}

Physical system identification from observations has recently attracted attention in the machine learning and computer vision communities.
Early approaches use non-differentiable physics engines~\cite{wu2015_galileo,wu2017_learning} such as Bullet~\cite{coumans2010_bullet} or integrate specific physical laws for each scenario~\cite{wu2016_physics}.
Lelidec \etal \cite{lelidec2021_differentiable} propose a variant of the staggered projections method to identify Coulomb friction coefficients from objects observed in video via markers.
Weiss \etal \cite{weiss2020_materialprop} estimate material properties of deformables by matching a differentiable deformable simulation to point cloud observations.
In Kandukuri \etal \cite{kandukuri2020_learning} a differentiable physics simulation based on~\cite{avilabelbuteperes2018_end} is embedded as layer into a deep neural network which infers the physical state from images and predicts the next states. 
Recently,~\cite{krishnamurthy2021_gradsim} proposed gradSim, a framework that combines differentiable simulation and differentiable rendering for system identification from video and visual control.
The method uses penalty based resolution of contacts for rigid body modeling.
Yet, these approaches lack a physical model which supports differentiation for arbitrary watertight shapes like ours.

%% file: background.tex
We base our formulation on a differentiable velocity-based constraint-based time stepping method.
We extend the approach with differentiable SDF shape representations, inertia tensors and time of contact.

\subsection{Constraint-based time stepping dynamics}
\label{sec:lcp}
Constraint-based time stepping methods formulate the dynamics as solving the Newton-Euler equations and add equality and inequality constraints to model joints, contacts and collisions~\cite{anitescu1997_formulating,avilabelbuteperes2018_end}.
The Newton-Euler equations relate wrenches (\ie torques and forces) acting on the objects in the scene with their motion, \ie
	$\mathbf{f} = \mathbf{M}\ddot{\mathbf{x}} + \textrm{Coriolis forces}$.
	We denote wrenches by a time dependent mapping $\mathbf{f}: [0,\infty[ ~ \rightarrow ~ \mathbb{R}^{6N}$.
	The $N$ objects in the scene are described by their mass-inertia collected in the block-diagonal matrix $\mathbf{M} \in \mathbb{R}^{6N \times 6N}$ and their poses (positions and orientations) $\mathbf{x} \in SE(3)^N$, where $SE(3)$ is the special Euclidean group.
	We represent the velocities $\dot{\mathbf{x}}_i = \boldsymbol{\xi}_i  = \left( \boldsymbol{\omega}_i^\top, \mathbf{v}_i^\top \right)^\top$ of object $i$ by twist coordinates which stack rotational and linear velocities $\boldsymbol{\omega}_i, \mathbf{v}_i: [0,\infty[ \rightarrow \mathbb{R}^3$, respectively~\cite{cline2002_rigid}.
	
To simulate joints, contacts and friction, corresponding constraints are included.	
In the event of collisions, the direction of velocity changes suddenly in infinitesimal time which cannot be expressed with the acceleration-based dynamics.
Hence, acceleration is discretized as $\dot{\boldsymbol{\xi}} = (\boldsymbol{\xi}_{t+h} - \boldsymbol{\xi}_t)/h$, where $\boldsymbol{\xi}_{t+h}$ and $\boldsymbol{\xi}_t$ are the velocities in successive time steps at times $t+h$ and $t$, and $h$ is the time-step size to yield
$\mathbf{M}\boldsymbol{\xi}_{t+h} = \mathbf{M}\boldsymbol{\xi}_{t} + \mathbf{f}_{\text{ext}} \cdot h$.
This velocity-based formulation is also better suited for modeling friction~\cite{cline2002_rigid}. 

Joints between objects impose equality constraints $g_e( \mathbf{x} ) = 0$ on their poses and restrict degrees of freedom in their motion. 
The constraints $\mathbf{J}_e\boldsymbol{\xi}_{t+h}=0$ are obtained in terms of the object velocities by deriving the pose constraints with corresponding Jacobian $\mathbf{J}_e$.
The constraints are added using Lagrange multipliers $\boldsymbol{\lambda}_e$ which determine the magnitude of the constraint force.
	
Contacts give rise to inequality constraints $g_c( \mathbf{x} ) \geq 0$ in the poses which prevent objects from interpenetrating each others.
In our 3D simulation, the collision pose constraint function is
$g_c( \mathbf{x} ) = \mathbf{n}^\top \left( \mathbf{p}^w_i - \mathbf{p}^w_j \right) - \epsilon$,
where $\mathbf{p}^w_{i} := \mathbf{x}_{i} + \mathbf{p}^i_{i} = \mathbf{x}_{i} + \mathbf{R}_{i} \mathbf{p}_{i}$ and $\mathbf{n}$ are contact point on object $i$ and normal in the world frame (indicated by the superscript $w$), and $\mathbf{R}_{i} \in SO(3)$ is the rotation of the object frames relative to the world frame.
As indicated by the superscript $i$, the contact point $\mathbf{p}^i_{i}$ is given in relative coordinates of body $i$.
We find a velocity constraint through time differentiation of the pose constraint as $\mathbf{J}_c\boldsymbol{\xi}_{t+h} \geq -k\mathbf{J}_c\boldsymbol{\xi}_{t} = -\mathbf{c}$~\cite{cline2002_rigid}.
Here, parameter $k$ is the coefficient of restitution and $\mathbf{J}_c$ is the Jacobian of the contact pose constraint function.
We provide further details on the contact constraints and its derivatives in the supplementary material.
We add the inequality constraint using a Lagrange multiplier $\boldsymbol{\lambda}_c$  and introduce slack variables $\mathbf{a}$, which lead to complementarity constraints \cite{boyd2004_convex}.

Friction also leads to inequality constraints.
It is modeled using a maximum energy dissipation principle.
This leads to two inequality constraints, $\mathbf{J}_f \boldsymbol{\xi}_{t+h} + \mathbf{E} \boldsymbol{\gamma} \geq 0$ and $\mu \boldsymbol{\lambda}_c \geq \mathbf{E}^\top \boldsymbol{\lambda}_f$,
where $\mathbf{J}_f$ is the Jacobian of the friction constraint in the object pose.
The constraints are governed by the coefficient of friction $\mu$. The matrix $\mathbf{E}$ is a binary matrix which makes the equation linearly independent at multiple contacts.
We refer to the supplementary material for further details.
The constraint forces are given by the Lagrange multipliers $\boldsymbol{\lambda}_f$ and $\boldsymbol{\gamma}$ with corresponding slack variables $\boldsymbol{\sigma},\boldsymbol{\zeta}$ and two complementarity constraints.
	
This constrained dynamics model leads to a linear complementarity problem (LCP), see supplementary material for details.
The LCP is solved using a primal-dual algorithm as described in~\cite{boyd2004_convex} and can be differentiated for the input states, forces and physical parameters at the solution~\cite{avilabelbuteperes2018_end}.

\subsection{Shape representation by SDFs}
Signed distance functions (SDFs)
	$\phi: \mathbb{R}^3 \rightarrow \mathbb{R}, \mathbf{x} \mapsto \phi(\mathbf{x})$
map 3D points to the signed Euclidean distance $d(\mathbf{x},\mathcal{S}) = \min_{\mathbf{y} \in \mathcal{S}} \left\|  \mathbf{x} - \mathbf{y} \right\|_2$ to the closest point on a surface $\mathcal{S}$ in 3D.
Points inside the objects have negative sign, \ie $\phi(\mathbf{x}) = -d(\mathbf{x},\mathcal{S})$ while points outside have positive signed distance $\phi(\mathbf{x}) = d(\mathbf{x},\mathcal{S})$.
The surface is given implicitly by the zero level-set $\mathcal{S} = \{ \mathbf{x} \in \mathbb{R}^3 \mid \phi(\mathbf{x}) = 0 \}$.

Recent works on neural embeddings of shape spaces learn implicit shape representations which additionally depend on a $d$-dimensional shape encoding $\mathbf{z} \in \mathbb{R}^d$. 
In DeepSDF~\cite{park2019_deepsdf}, a signed distance function $\phi_\theta: \mathbb{R}^3 \times \mathbb{R}^d \rightarrow \mathbb{R}, ( \mathbf{x}, \mathbf{z} ) \mapsto \phi_\theta( \mathbf{x}, \mathbf{z} )$ is learned from a collection of shapes.
The approach in~\cite{gropp2020_implicit} regularizes the function to be a valid SDF away from the surface by enforcing the Eikonal constraint by an additional loss function.
We prefer the latter method for modeling shape spaces over DeepSDF in this paper to obtain better SDF gradients for collision detection and modeling.
In the supplementary material, we illustrate the shape spaces used for our experiments.

\subsection{Differentiable SDF-based mesh representation}
\label{sec:meshsdf}

In~\cite{remelli2020_meshsdf}, Remelli \etal propose an approach to extract a mesh from a SDF representation using marching cubes and to differentiate the mesh vertex position $\mathbf{v}$ for the underlying SDF by
	$\frac{\partial \mathbf{v}}{\partial \phi}( \mathbf{v} ) = -\nabla \phi( \mathbf{v} )$.

\subsection{Contact detection between SDFs}
\label{sec:sdfcontacts}

We detect contacts and estimate contact points and normals from the SDFs of both objects based on the approach of \cite{macklin2020_local}.
We first transform the object SDF representations into their differentiable mesh representations (see sec. \ref{sec:meshsdf}).
The culling and starting point strategy of \cite{macklin2020_local} determines an initial set of mesh triangles.
This involves determining for each triangle with vertices $\mathcal{V} = \left\{ \mathbf{v}_1,\ldots,\mathbf{v}_3 \right\}$, if the signed distance $\phi_j(\mathbf{c})$ at the triangle's centroid position $\mathbf{c} = \frac{1}{3} \sum_l \mathbf{v}_l$ in the other object is below the radius $\max_k \left\| \mathbf{v}_k - \mathbf{c} \right\|_2$ of the triangle.
The contact point for such a triangle is found iteratively from the triangle vertex position $\mathbf{p}_0$ with the smallest SDF value in the other object using the Frank-Wolfe algorithm~\cite{frank1956_frankwolfealgo} based approach in~\cite{macklin2020_local}.

In each iteration $k$, the algorithm determines a point $\mathbf{s}_k$ that minimizes
	$\operatorname{argmin}_{\mathbf{s}_k} \mathbf{s}_k^\top \nabla \phi(\mathbf{p}_k)~s.t.~\mathbf{s}_k \in \mathcal{T}$,
where $\mathcal{T} \subset \mathbb{R}^3$ is the surface spanned by the triangle vertices.
The contact point is updated according to
	$\mathbf{p}_{k+1} = (1-\alpha_k) \mathbf{p}_k + \alpha_k \mathbf{s}_k$,
where $\alpha_k = \frac{2}{k+2}$.
We iterate the algorithm until the criterion $\left| (\mathbf{p}_{k} - \mathbf{s}_k )^\top \nabla \phi(\mathbf{p}_k) \right| < \tau$ converges below a threshold $\tau$ or a maximum iteration count is reached.

%% file: method.tex
We now detail our simulation framework and novel contributions to include a differentiable SDF shape representation and differentiable time of contact into a velocity- and constraint-based time-stepping 3D simulation method.
Our overall goal is to develop a differentiable model $\mathbf{x}_t = g( \mathbf{x}_{t-1}, \boldsymbol{\theta}, \mathbf{f}_{\text{ext}}, h )$ for physics simulation which can be used for predicting the next scene state $\mathbf{x}_t$ (object poses) based on physics parameters $\boldsymbol{\theta}$ (including shape parameters $\mathbf{z}$), initial states (object poses and velocities) $\mathbf{x}_{t-1}$, external wrench $\mathbf{f}_{\text{ext}}$, and a simulation time step $h$.
In our approach, the function $g$ is given by the solution of the LCP in section \ref{sec:lcp}.
The property of differentiability of $g$ for its inputs  allows us to formulate optimization problems
to fit simulated to target trajectories by adjusting parameters $\boldsymbol{\theta}$.
For instance, 
\begin{equation}
\label{eq:nllsobjective}
	E( \boldsymbol{\theta} ) = \frac{1}{2} \sum_{i=0}^{N-1} \left\| g( \mathbf{x}_{i}, \boldsymbol{\theta}, \mathbf{f}_{\text{ext},i}, h_{i} ) - \widehat{\mathbf{x}}_{i+1} \right\|_2^2
\end{equation}
fits a target trajectory $\widehat{\mathbf{x}}_{1}, \ldots, \widehat{\mathbf{x}}_{N}$ of duration $T$ with appropriate time discretization $t_i = t_0 + \sum_{j=0}^{i-1} h_j$ where $h_i \geq 0 \wedge \sum_{i=0}^{N-1} h_i = T$.

\subsection{Differentiable inertia tensors}

The mass-inertia matrix $\mathbf{M}$ stacks angular inertia tensor $\mathcal{I}$ and object mass $m$ for the angular and linear parts of the Newton-Euler equations.
We determine the angular inertia tensor of the object from the shape represented implicitly in an SDF.
The tensor $\mathcal{I}$ is found efficiently from a triangular surface mesh using \cite{mirtich1996_fast} which we implement using differentiable operations.
We extract the mesh from the SDF using the marching cubes algorithm~\cite{lorensen1987_marching}.
Using~\cite{remelli2020_meshsdf} (see Sec.~\ref{sec:meshsdf}), we differentiate the inertia tensor for the underlying SDF and subsequently the shape encoding $\mathbf{z}$ through the mesh,
	$\frac{d \mathcal{I}}{d \mathbf{z}} = \sum_{\mathbf{v} \in \mathcal{V}} \frac{\partial \mathcal{I}}{\partial \mathbf{v}} \frac{\partial \mathbf{v}}{\partial \phi} \frac{\partial \phi}{\partial \mathbf{z}}$,
with mesh vertices $\mathcal{V}$.

\subsection{Differentiable contact modeling}

Our differentiable physics simulation supports friction and collision contact constraints.
Both constraints are formulated in terms of contacts with associated contact points $\mathbf{p}_i, \mathbf{p}_j$ relative to the reference frames of both objects $i$ and $j$, contact normal $\mathbf{n}_{i/j}$ and penetration distance $d_{i/j}$ (see supplementary material for details).

\subsubsection{Differentiable contact points and normals}

We extend the approach in \cite{macklin2020_local} (Sec~\ref{sec:sdfcontacts}) to develop efficient differentiable contact detection for SDF shapes.
The contact detection algorithm provides the contact point $\mathbf{p}^*$ on a mesh triangle through Frank-Wolfe iterations \cite{frank1956_frankwolfealgo}.
The solution $\mathbf{p}^*$ of the algorithm is a linear combination $\mathbf{p}^* = \sum_l w_l \mathbf{v}_l $ of the triangle vertices since $\mathbf{s}_k \subset \mathcal{V}$, where
\begin{equation}
 w_{l,k+1} = (1-\alpha_k) w_{l,k} + \begin{cases}
 			     \alpha_k, & \text{if } \mathbf{s}_k = \mathbf{v}_l\\ 
 			    0,  & \text{otherwise.}\\ 
 			 \end{cases}
\end{equation}
We directly differentiate the resulting linear combination of the Frank-Wolfe algorithm,
and find the derivative of this linear combination for the underlying SDF  using the mesh to SDF differential in Sec.~\ref{sec:meshsdf}.
This also enables us to propagate gradients through the contact points onto the shape encoding $\mathbf{z}$ of SDF shape spaces to optimize the object shapes for physical plausibility in contact situations.

Finally, we find the contact normal and penetration distance directly from the object SDFs.
The normal is given by the SDF gradient at the contact point $\mathbf{n}_i = \pm \nabla \phi_{j/i}( \mathbf{p}_i ) / \left\| \nabla \phi_{j/i}( \mathbf{p}_i ) \right\|_2$ in either the other or the own object, while the penetration is the signed distance value $d_i = \phi_j(\mathbf{p}_i)$.
We choose the contact normal from the object with the smallest mean curvature of the SDF at the contact point $\mathbf{p}_i$.
The corresponding contact point on the other object $j$ is detected at $\mathbf{p}_j = \mathbf{p}_i - \phi_j( \mathbf{p}_i ) \nabla \phi_j( \mathbf{p}_i ) $.

Since the runtime complexity for solving the LCP increases with the number of contact points, we determine redundant contact points by clustering the points according to normal similarity and reducing the set to the points on its convex hull in the common plane. 
Further details and a proof that this process does not change the contact constraints are provided in the supplementary material.

\begin{figure}
	\centering
	\includegraphics[width=0.89\linewidth]{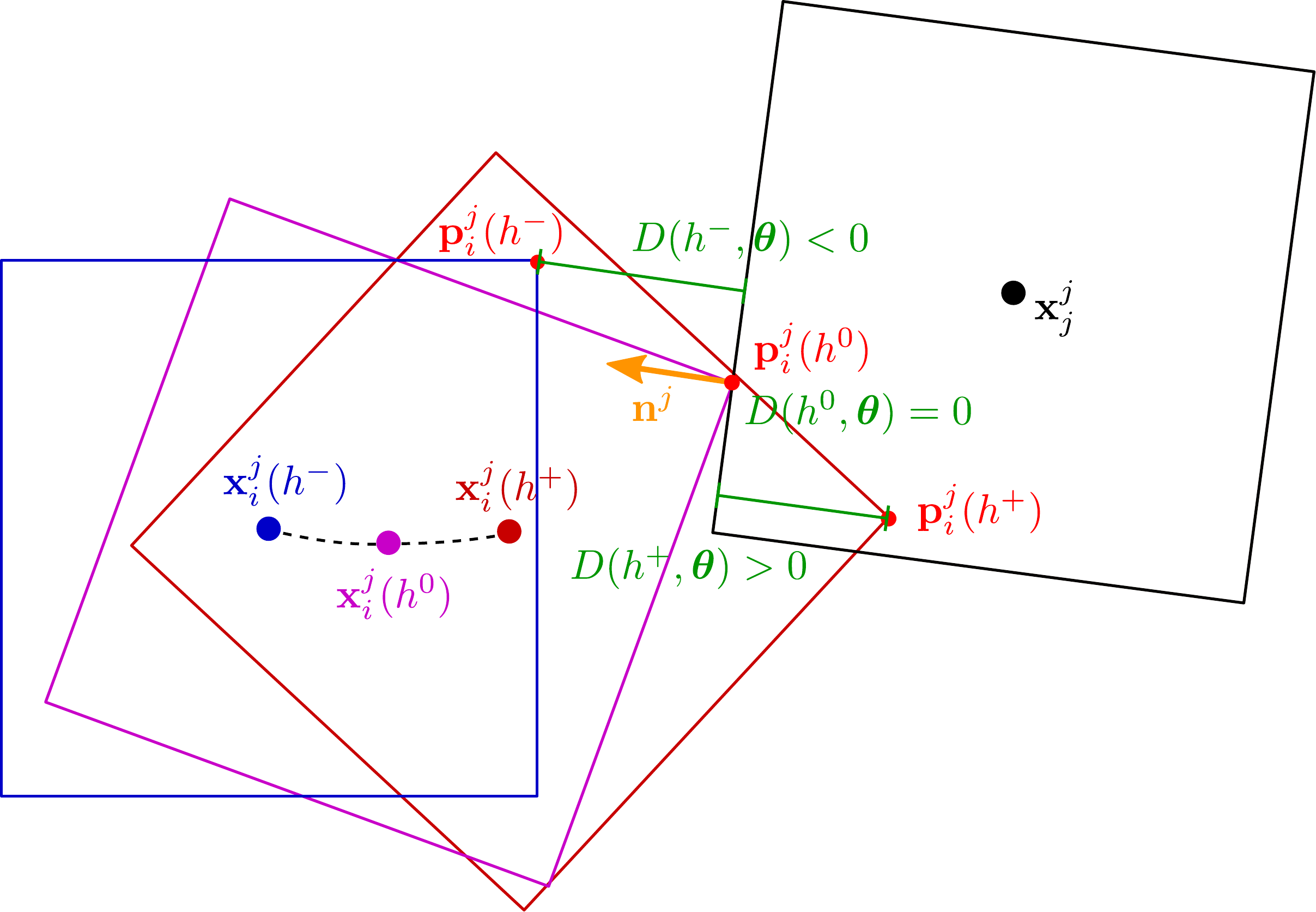}
	\caption{Differentiable time of contact. We differentiate the time of contact $h$ for the contact point to enable shape optimization through collisions.  The constraint $D(h,\boldsymbol{\theta}) = 0$ requires the distance of the contact point $\mathbf{p}_i^j$ of object $i$ to its corresponding contact point of object $j$ along the contact normal direction $\mathbf{n}^j$ to be zero. Shape, physical states and parameters which $D$ depends on are summarized in $\boldsymbol{\theta}$. The constraint defines an implicit relationship between $h$ and $\boldsymbol{\theta}$.  The distance is defined with body frame of object $j$ as reference frame (denoted by superscript $j$ on points and vectors). In this frame, object $i$ and its contact point move along trajectories $\mathbf{x}_i^j(h)$ and $\mathbf{p}_i^j(h)$, respectively. We use implicit differentiation to determine $\partial h/\partial \boldsymbol{\theta}$.}
	\label{fig:toidiff}
\end{figure}

\subsubsection{Differentiable time of contact}
The LCP contains the contact points in the contact and friction constraint Jacobians by which the LCP solution can also be differentiated for the contact points and hence for the shape parameters of the objects.
Due to the time discretization and time derivatives of the position constraint, there is no direct dependency of the contact points on the linear velocity of the bodies.
Observe, that the shape and the induced collisions influence the time of contact: the larger the shape in the direction of motion, the earlier the contact (see Fig.~\ref{fig:toidiff}).
Hence, changes in object shape induce changes in time of contact.
We propose an approach to model this dependency of the simulation on the shape parameters.

The LCP from section \ref{sec:lcp} assumes the time step size~$h$ constant and does not model its dependency on other parameters of the simulation.
The time step is found through a separate optimization process which determines the largest step size $h \leq H$ until the next collision with maximum step size $H$.
The dependency of $h$ on the shape parameters is implicitly defined by the solution of the optimization problem.
It determines the time step for which corresponding points of contact $\mathbf{p}_i, \mathbf{p}_j$ on both objects $i$, $j$ coincide, \ie
\begin{multline}
\label{eq:constraint_d}
		D\left( h, \mathbf{p}_i^j(h, \Theta(h)), \mathbf{p}_j^j, \Theta(h) \right)\\ ={\mathbf{n}^j}^{\top} \left( \mathbf{p}_j^j - \mathbf{p}_i^j(h, \Theta(h)) \right) = 0.
\end{multline}
The distance $D$ also depends on the contact normal $\mathbf{n}$ and variables $\Theta(h)$ such as the object shape parameters, poses, and velocities at the time of collision.
We express the distance in the body frame of object $j$ (indicated by the superscript $j$) so that the contact point on object $j$ and the contact normal remain constant, while the contact point on object~$i$ depends on time~$h$ via the pose $\mathbf{x}$, velocity $\mathbf{v}$ and acceleration $\mathbf{a}$ of the objects,
\begin{multline}
		\mathbf{p}_i^j(h, \Theta(h)) = (\exp(\widehat{\boldsymbol{\omega}}_jh) \mathbf{R}_j)^\top \left( \exp(\widehat{\boldsymbol{\omega}}_ih) \mathbf{p}_i^I + \mathbf{x}_i\right.\\ \left.+ \mathbf{v}_i h + \tfrac{1}{2} \mathbf{a}_i h^2 - \left( \mathbf{x}_j + \mathbf{v}_j h + \tfrac{1}{2} \mathbf{a}_j h^2 \right) \right)
\end{multline}
and $\mathbf{p}_j^j = \exp(\widehat{\boldsymbol{\omega}}_jh) \mathbf{R}_j^\top \mathbf{p}_j^J$, where $\mathbf{R}_j$ is the rotation of object in the world frame.
The operator $\,\widehat{\cdot}\,$ maps twist coordinates $\omega$ to $so(3)$ Lie algebra elements and $\exp$ is the exponential map of $SO(3)$.
By the constraint in eq.~\eqref{eq:constraint_d}, the time step $h( \mathbf{p}_i, \mathbf{p}_j, \Theta )$ implicitly becomes a function of the contact points, the contact normal, and the simulation state and parameters (see Fig.~\ref{fig:toidiff}).

When optimizing system identification objectives such as Eq.~\eqref{eq:nllsobjective} using gradient-based methods, we require the derivative of the time of contact $h$ for its parameters.
For instance, differentiating the objective in Eq.~\eqref{eq:nllsobjective} 
yields a Jacobian depending on three derivative terms
\begin{multline}
	\frac{dE( \boldsymbol{\theta} )}{d\boldsymbol{\theta}} = \sum_{i=0}^{N-1}  \left( g( \mathbf{x}_{i}, \boldsymbol{\theta}, \mathbf{f}_{\text{ext},i}, h_{i}(\boldsymbol{\theta}) ) - \widehat{\mathbf{x}}_{i+1} \right)\\ \left( \frac{\partial g}{\partial \boldsymbol{\theta}} + \frac{\partial g}{\partial \mathbf{x}_{i}} \frac{\partial \mathbf{x}_{i}}{\partial \boldsymbol{\theta}} + \frac{\partial g}{\partial h_{i}} \frac{\partial h_{i}}{\partial \boldsymbol{\theta}} \right).
\end{multline}
The first term is obtained by differentiating the LCP at its solution for the physics parameters.
The second term is the derivative of the LCP for the input state $\mathbf{x}_i$ multiplied with the derivative of $g$ for the parameters from the previous time step which follows by recursion (due to $\mathbf{x}_i = g( \mathbf{x}_{i-1}, \boldsymbol{\theta}, \mathbf{f}_{\text{ext},i-1}, h_{i-1}(\boldsymbol{\theta}) )$).
For the third term, we require the time of contact derivative $\partial h_{i}/\partial \boldsymbol{\theta}$.

We distinguish contacts with impact and resting contacts and determine the time of contact differential only for contacts with impact.
Contacts with impact are those which are newly found between two objects in a time step (\ie the previous time step did not have contacts between these objects).
Resting contacts are contacts in successive time steps after the first contact time step between two objects.

In practice, in each simulation step, the step size $h$ is either chosen as a maximum time step $H$ or the time until the first contact occurs.
To this end, the simulation step size $h$ is iteratively halved until all penetrations between objects are below a contact threshold distance $\epsilon$.
This yields the approximate time of contact $h$ and a set of contact points $\mathbf{p}_i^k$, $\mathbf{p}_j^k$ with contact normals $\mathbf{n}^k$ determined as the surface normal on one of the objects with $k \in \{ 1, \ldots, K \}$.
The constraints form an overdetermined set of equations,
\begin{equation}
	\mathbf{D}( h, \boldsymbol{\theta} ) = \left(\begin{array}{c} D_1( h, \boldsymbol{\theta} )\\ \vdots \\ D_K( h, \boldsymbol{\theta} ) \end{array}\right) = \mathbf{0},
\end{equation}
where $\boldsymbol{\theta}$ subsumes the parameters on which $h$ depends on and $D_k( h, \boldsymbol{\theta} ) = 0$ is the time of contact constraint for the $k$-th contact point.
From this set of equations, we find the time of contact derivative through implicit differentiation,
	$\frac{d\mathbf{D}( h, \boldsymbol{\theta} )}{d\boldsymbol{\theta}} = \frac{\partial \mathbf{D}( h, \boldsymbol{\theta} )}{\partial \boldsymbol{\theta}} + \frac{\partial \mathbf{D}( h, \boldsymbol{\theta} )}{\partial h} \, \frac{\partial h}{\partial \boldsymbol{\theta}} = \mathbf{0}$,
which gives
\begin{equation}
	\frac{\partial h}{\partial \boldsymbol{\theta}} = -\frac{\partial \mathbf{D}( h, \boldsymbol{\theta} )}{\partial h}^+ \frac{\partial \mathbf{D}( h, \boldsymbol{\theta} )}{\partial \boldsymbol{\theta}}
\end{equation}
By applying the Moore-Penrose pseudo-inverse $\frac{\partial \mathbf{D}( h, \boldsymbol{\theta} )}{\partial h}^+$ we find a least squares fit to the overdetermined set of equations.
The time of contact derivative effectively allows for optimizing the shape of objects through the contact points so that collisions occur earlier or later and the trajectories of the objects are adapted.
Components $k$ contribute only meaningfully if the bodies move towards each other according to the relative velocity, \ie if $\frac{\partial D_k( h, \boldsymbol{\theta} )}{\partial h} \geq 0$.
Otherwise, the component is excluded from $\mathbf{D}$.
Both time steps $h_i( \boldsymbol{\theta})$ and $h_{i+1}( \boldsymbol{\theta})$ before and after the collision are dependent on the time of contact, since $\overline{h} = h_i( \boldsymbol{\theta}) + h_{i+1}( \boldsymbol{\theta})$ is constant and $h_{i+1}( \boldsymbol{\theta}) = \overline{h} - h_i( \boldsymbol{\theta})$.

%% file: experiments.tex
We evaluate and demonstrate our method in several physical system identification scenarios which involve inference on shape, friction coefficient, mass or forces.
The observations are generated using the LCP physics engine with ground truth parameters with a time step size of $H=\frac{1}{30}$. 
For shape primitives, \ie spheres, boxes and cylinders, we compute the SDF analytically.
We evaluate the accuracy of our novel approach quantitatively and provide an ablation study.
Additional details and supplementary results are provided in the supplementary material alongside a video.

\begin{figure}
	\centering
	\includegraphics[trim=100 110 50 0,clip,width=.19\linewidth]{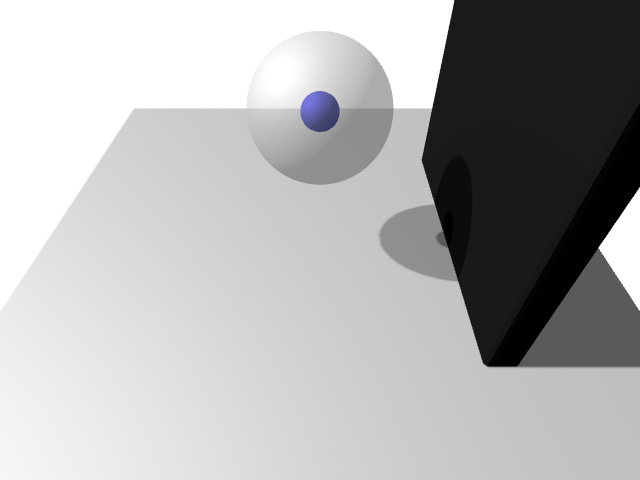}
	\includegraphics[trim=100 110 50 0,clip,width=.19\linewidth]{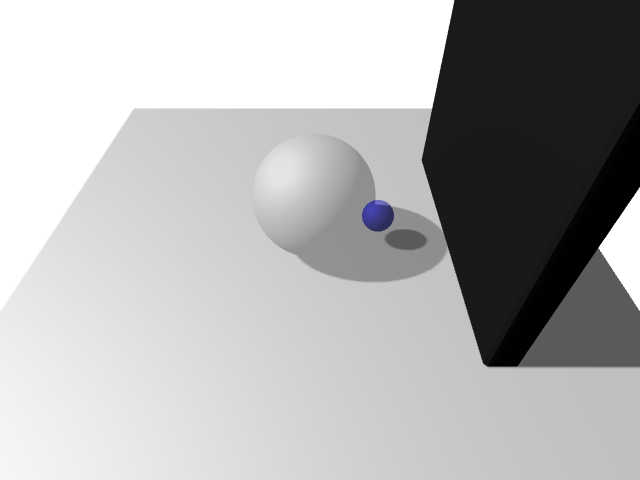}\hfill
	\includegraphics[trim=100 150 100 0,clip,width=.19\linewidth]{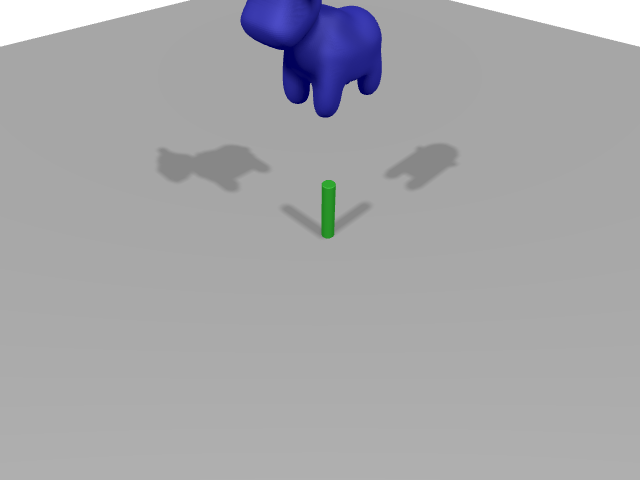}
	\includegraphics[trim=100 150 100 0,clip,width=.19\linewidth]{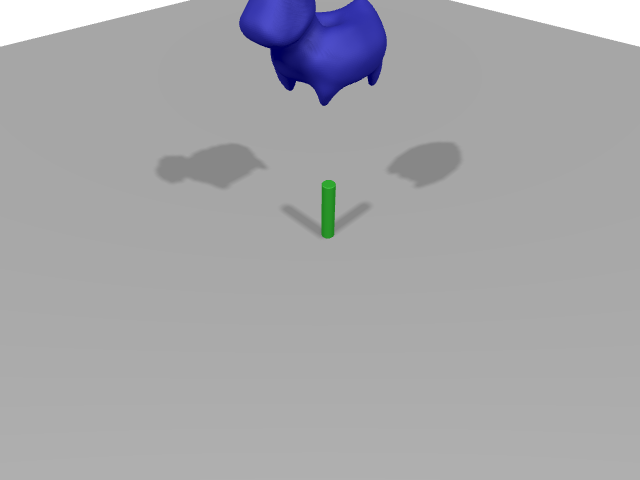}
	\includegraphics[trim=100 150 100 0,clip,width=.19\linewidth]{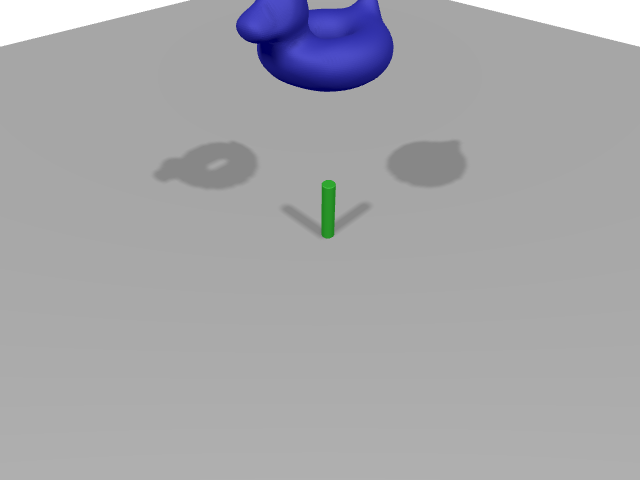}\\
	\includegraphics[trim=100 110 50 0,clip,width=.19\linewidth]{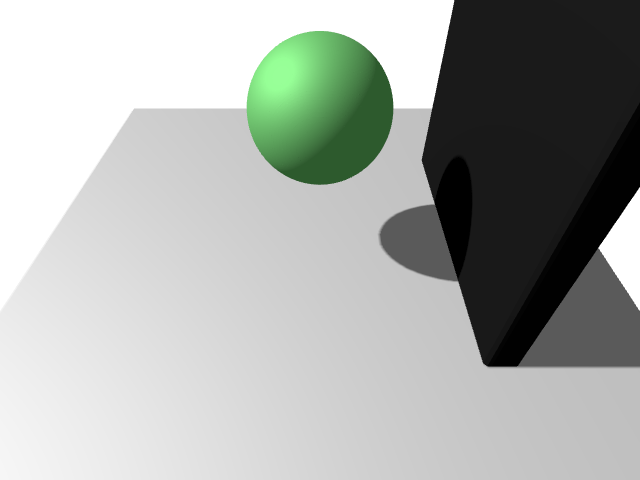}
	\includegraphics[trim=100 110 50 0,clip,width=.19\linewidth]{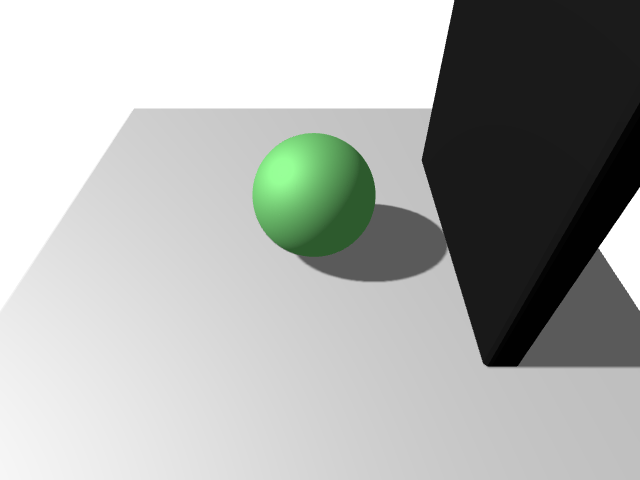}\hfill
	\includegraphics[trim=100 150 100 0,clip,width=0.19\linewidth]{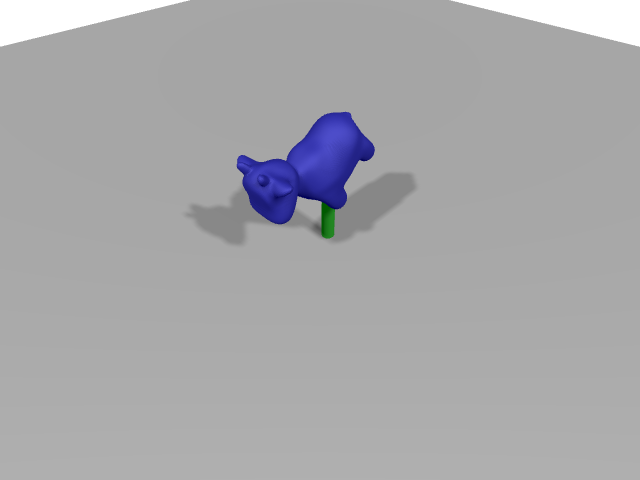}
	\includegraphics[trim=100 150 100 0,clip,width=0.19\linewidth]{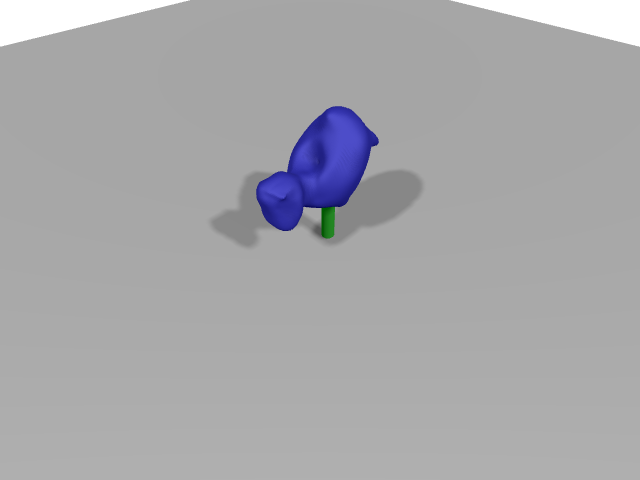}
	\includegraphics[trim=100 150 100 0,clip,width=0.19\linewidth]{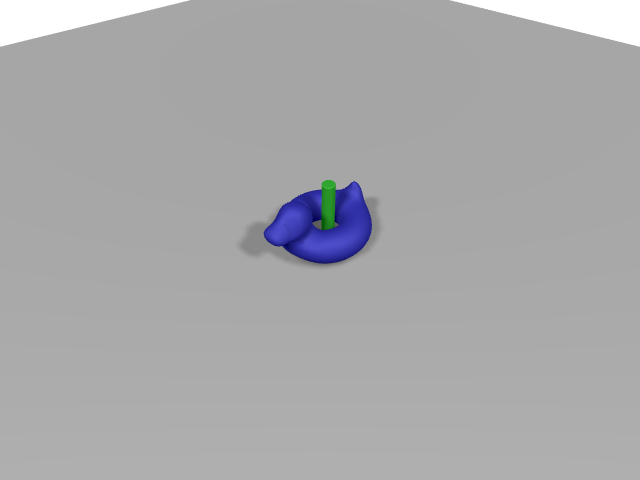}
	\caption{Collision-based shape optimization. Left: Sphere radius optimization through two bounces with a wall and the floor by our approach. Blue: initial, gray overlay: target, green: resulting shape. Right: Complex topological shape changes can be achieved by our approach. The shape is initialized with the genus-0 spot shape that falls onto a stick (left col.). The target pose for the object is on the floor through the stick (bottom right). This pose requires adapting the latent code to the genus-1 spot shape (middle col.: intermediate result (4 its.), right col.: final result (44 its.).}
	\label{fig:exp_scenarios}
\end{figure}

\begin{figure}
	\centering
	\includegraphics[trim=12 12 12 12,clip,width=0.92\linewidth]{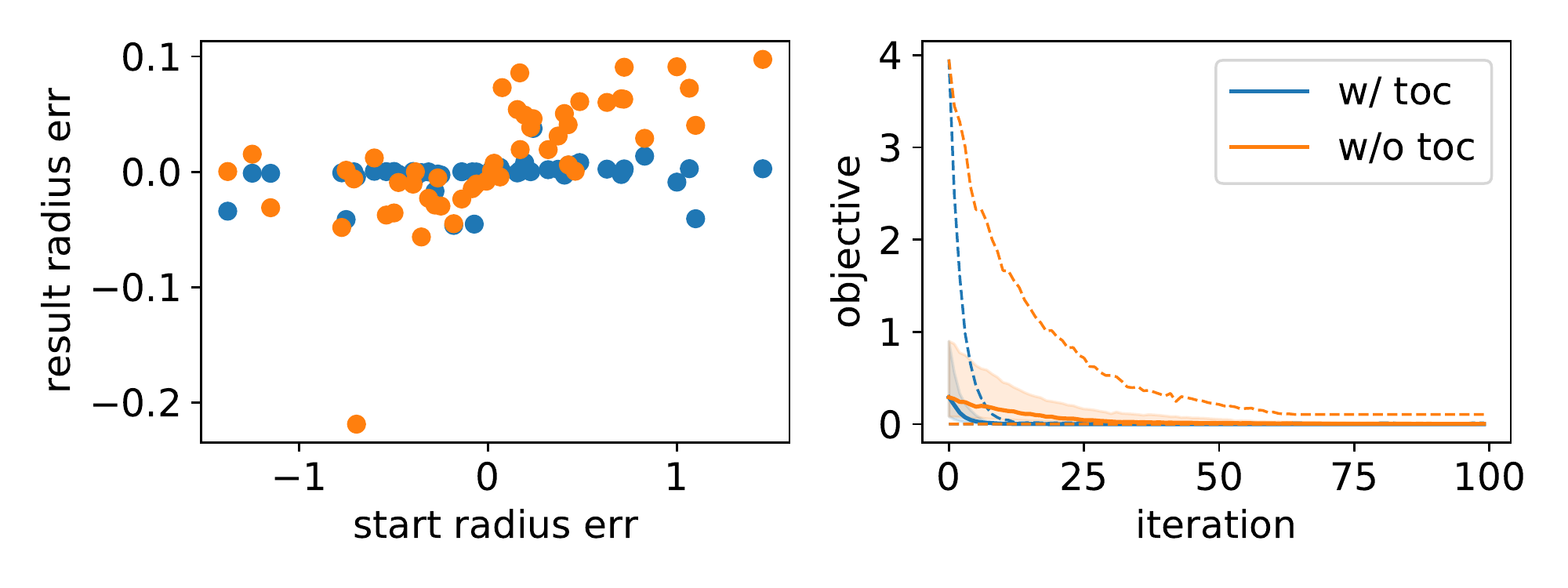}
	\caption{Bouncing sphere scenario with gravity. Left: start radius error vs. resulting radius error in sphere bounce experiment with gravity enabled with (blue) and without (orange) time of contact (toc) differential. The toc differential yields faster convergence and more accurate results. Right: median (solid lines), quartiles (shaded areas) and min/max (dashed lines) for the objective over time (blue: with toc, orange: without toc differential).}
	\label{fig:exp_sphereonwallplots}
\end{figure}

\subsection{Shape identification}
\label{sec:shape_id}

The following experiments analyze the accuracy of our method for estimating shape from collisions and inertia in several scenarios such as bouncing shapes, topological shape changes, and shape from inertia for rotating objects.

\begin{table}
\footnotesize
	\begin{center}
	\begin{tabular}{lcccc}
   	    \toprule
   	    & & \multicolumn{3}{c}{resulting radius error}\\
   	    \cmidrule(lr){3-5}
   	    scenario & variant & min & mean & max \\
		\midrule
		w/ gravity & w/o toc & 6e-5 & 0.038 & 0.219 \\
		 & w/ toc & 2e-6 & 0.007 & 0.046 \\
		\midrule
		w/o gravity & w/o toc & fails & fails & fails \\
		 & w/ toc & 2e-4 & 0.002 & 0.006 \\
		\bottomrule
	\end{tabular}
	\end{center}
	\caption{Resulting radius error for variants in the bouncing sphere scenarios. Time of contact differentials clearly improve the results and make the ``no gravity'' case work.}
	\label{tab:exp_sphereonwall}
\end{table}

\paragraph{Bouncing objects}
We evaluate our time of contact differential for estimating the radius of sphere that bounces against a wall and the floor (see Fig.~\ref{fig:exp_scenarios}, left).
We generate 50 scenes with randomly sampled sphere radii between 0.4 and 2.0 with an initial velocity of 5 towards the wall in two scenarios with gravity enabled and disabled.
We optimize the radius to match a target position trajectory also generated with a random radius between 0.4 and 2.0 with the same initial conditions otherwise.
This position depends on the shape parameters $\theta$.
Trajectories are recorded as observation sequences of length 1.5\,s.
In each scenario, a new sphere radius is sampled randomly and the predicted trajectory is fit to the observations using gradient descent on the mean squared error along the whole trajectory.
In Table~\ref{tab:exp_sphereonwall} we compare accuracy for different variants.
If gravity is disabled, the sphere hits the wall in a direction along the contact normal and the shape receives no gradient if the time of contact differential is not used.
If gravity is enabled, velocity at the contact has a downwards component, which transmits into a rotational velocity after the first bounce.
Together with the second bounce, this renders the trajectory more complex and we found it beneficial to optimize the trajectory in chunks including each contact separately by detaching poses and velocities before the second bounce. 
The rotational velocity component leads to a gradient on the shape through the inertia tensor.
In this case, using the time of contact differential improves convergence and accuracy of the radius estimate as displayed in Table~\ref{tab:exp_sphereonwall} and Fig.~\ref{fig:exp_sphereonwallplots}.

\begin{figure}
\begin{center}
	\includegraphics[width=.89\linewidth]{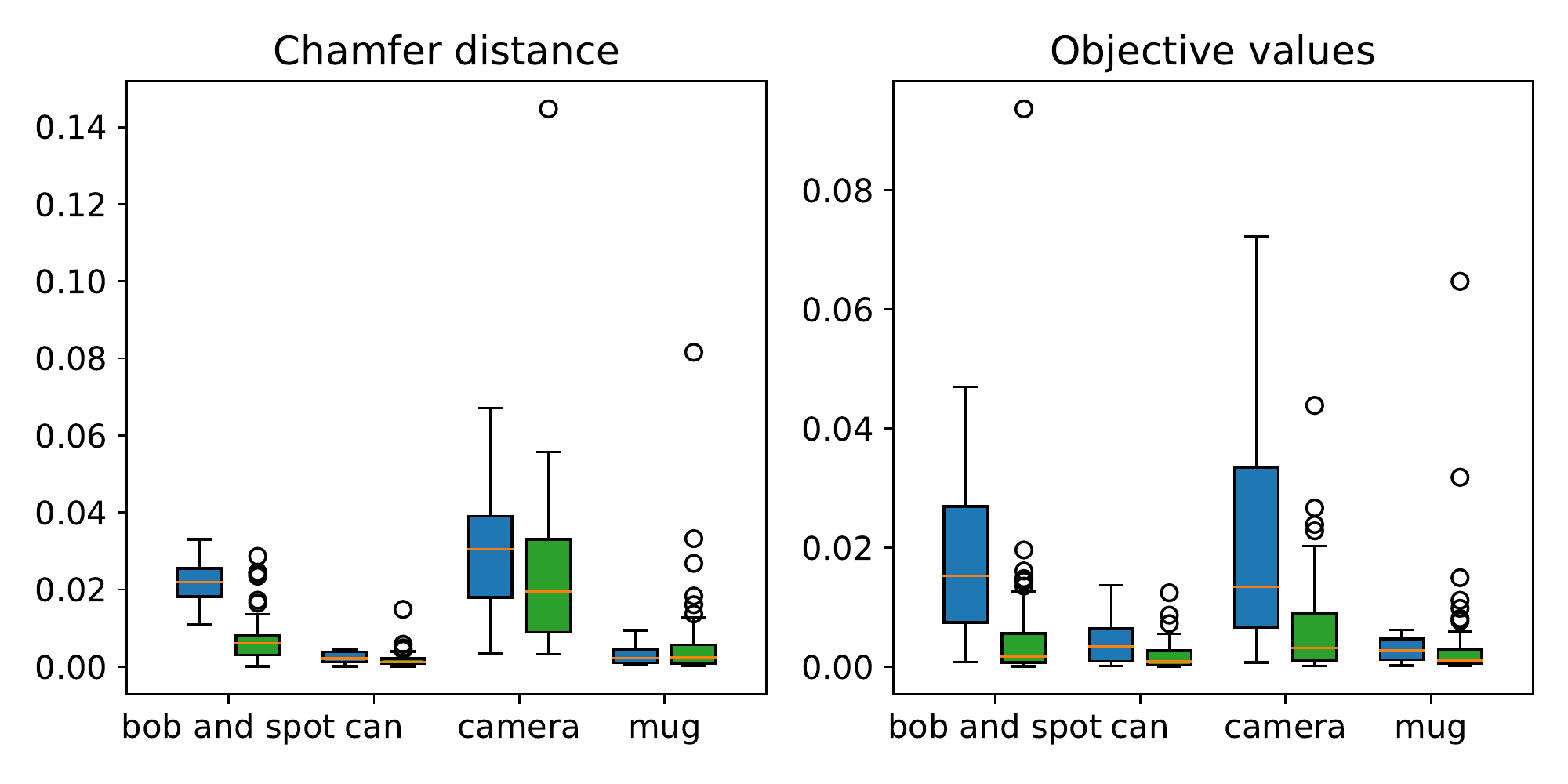}
	\caption{Trajectory fitting for learned shape spaces. Shape and trajectory accuracy improve in most cases. Left blue boxes: initialization, right green: optimization results. Initialization outliers are not shown for better graph scaling. See supp.~mat for different scaled versions.}
	\label{fig:exp_traj_shapespace}
\end{center}
\end{figure}

In Fig. \ref{fig:exp_traj_shapespace}, we show the results for optimizing the shape of several DeepSDF shape spaces (bob and spot trained with latent dimension 2; can, camera and mug with latent dimension 4) via a single bounce against the wall without gravity.
The average chamfer distance (measuring shape accuracy) over all objects is reduced from initial 0.016 to 0.010 by our approach.
Without the time of contact differential, the average is 0.017.
One can see that while the optimization objective generally decreases, in some cases this does not hold for the shape accuracy. 
For these outlier runs, the objective exhibits a local optimum in this challenging scenario which does not contain the true shape.

We also test our approach for a challenging shape optimization scenario in which the DeepSDF shape space of bob and spot is used (see Fig.~\ref{fig:exp_scenarios}, right).
The object is dropped on a stick. One of the objects is a genus-0 type cow-like shape (spot), while the other is genus-1 with a hole in the duck-like body (bob).
Only the latter will fall through the stick and reach the target position.
We optimize the shape latents in this scenario to reach the target position at the end of a 1.1\,s sequence.

\paragraph{Shape from inertia}

\begin{figure}
	\centering
	\includegraphics[trim=12 12 12 12,clip,width=0.89\linewidth]{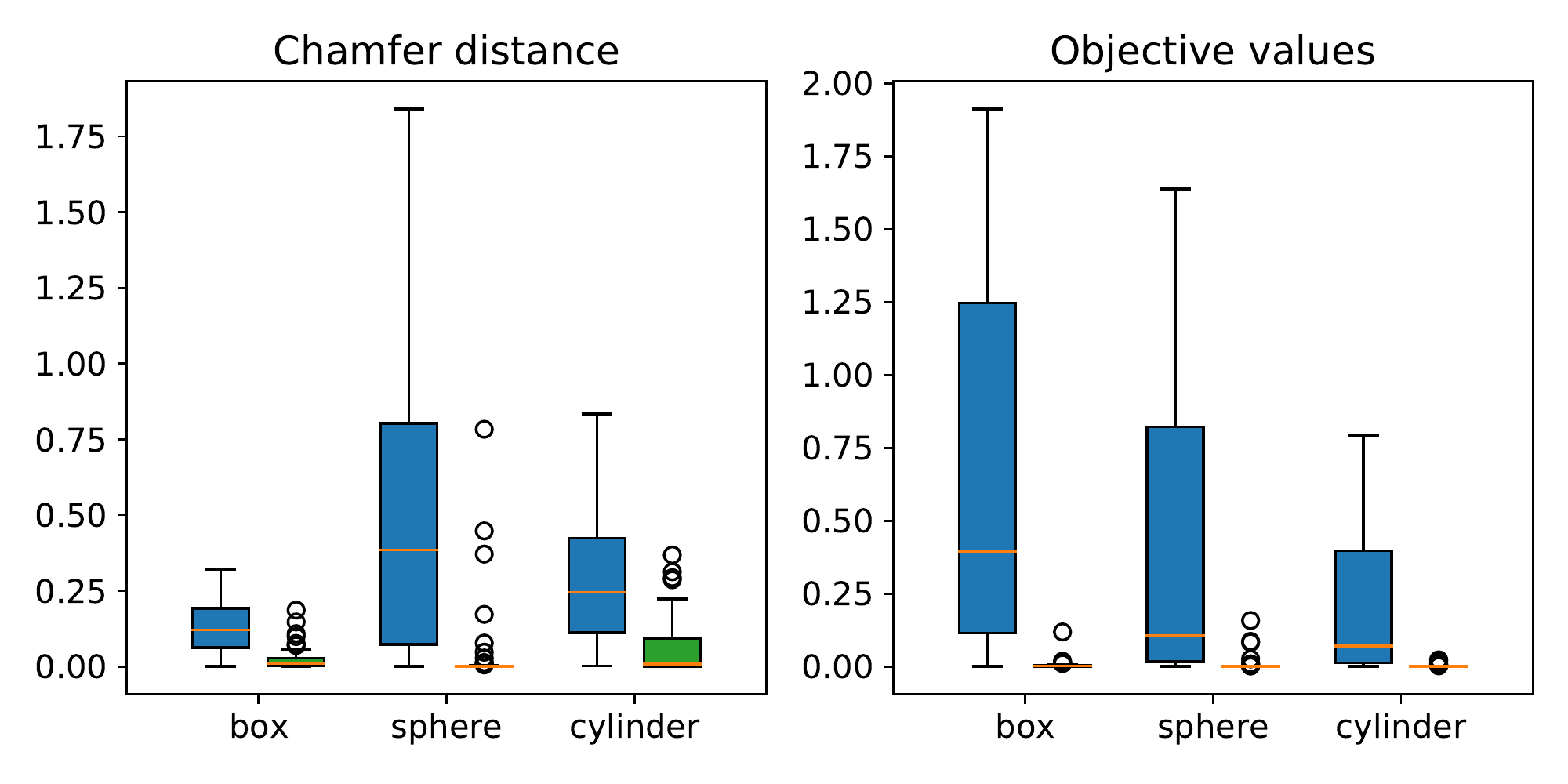}\\
	\includegraphics[trim=12 12 12 12,clip,width=0.89\linewidth]{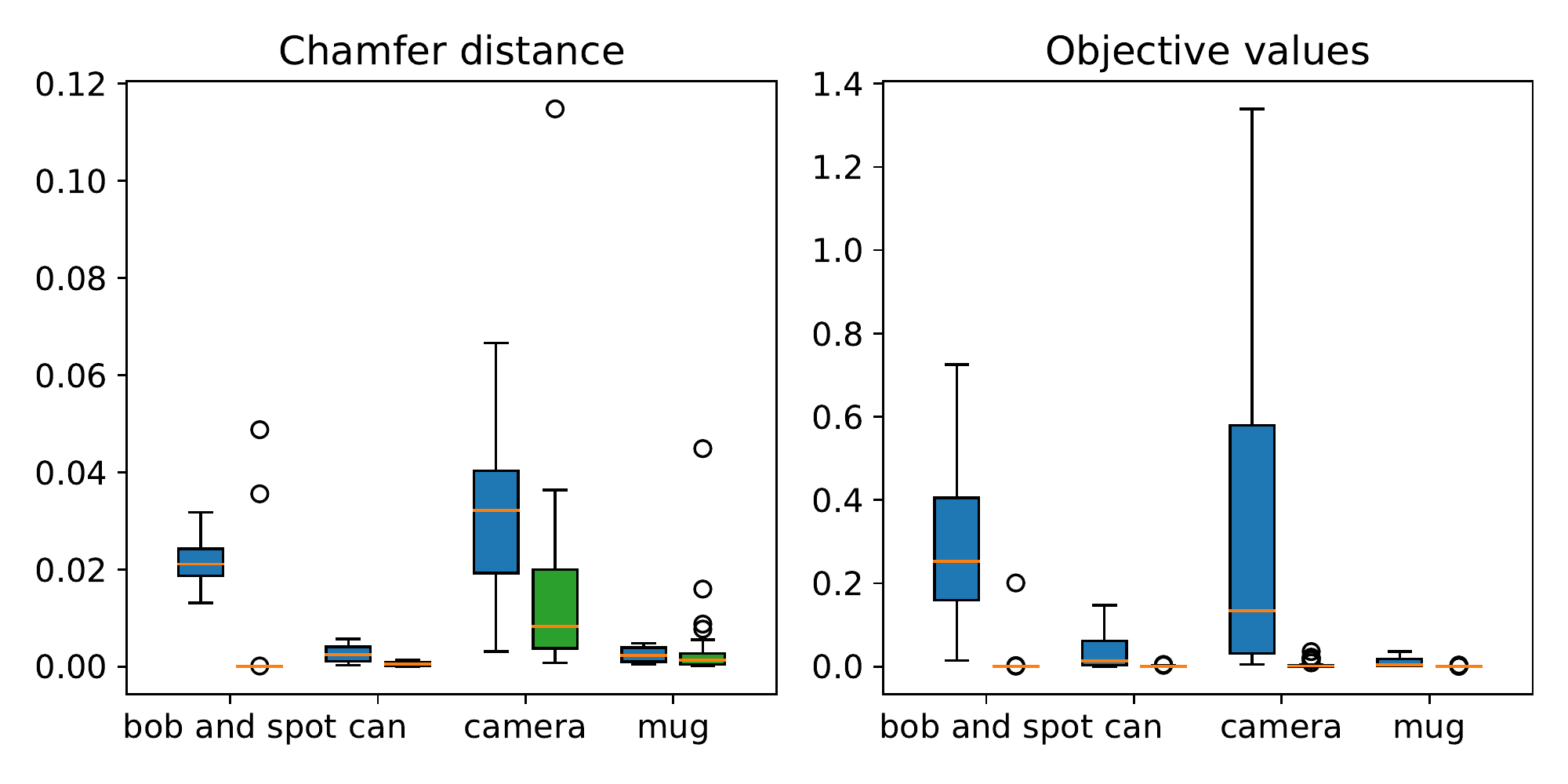}
	\caption{Shape from inertia. Left: chamfer distance. Right: optimization objective value. Left blue boxes: initialization, right green: optimization results. Initialization outliers are not shown for better graph scaling. See supp.~mat for different scaled versions.}
	\label{fig:exp_shapefrominertia}
\end{figure}

Fig.~\ref{fig:exp_shapefrominertia} gives results for shape from inertia optimization.
We apply a random torque with unit norm on the objects in the first 0.3\,s and simulate for 2\,s.
The optimization objective in this experiment is the mean squared error of the object's rotational velocity to the ground truth.
For the learned shape spaces, we additionally add the squared $\ell^2$ norm of the latent code with weight 1e-4 as regularizer.
We sample 50 scenes with random initial and target shape parameters for each shape type and evaluate our approach for sphere, box and cylinder shapes, and the DeepSDF shape spaces from the bouncing objects experiment.
We observe that the shape is well recovered in most of the runs. 
The average in chamfer distance over all objects drops from 0.164 to 0.021.
In a few outlier runs, the difference between result and target shape is large, while the optimization objective still achieves a very low value.
This hints at local minima in the objective landscape, which could possibly be alleviated by choosing other torques.

\subsection{Friction and mass identification, force optimization}

\begin{figure}
	\centering
	\includegraphics[width=.89\linewidth]{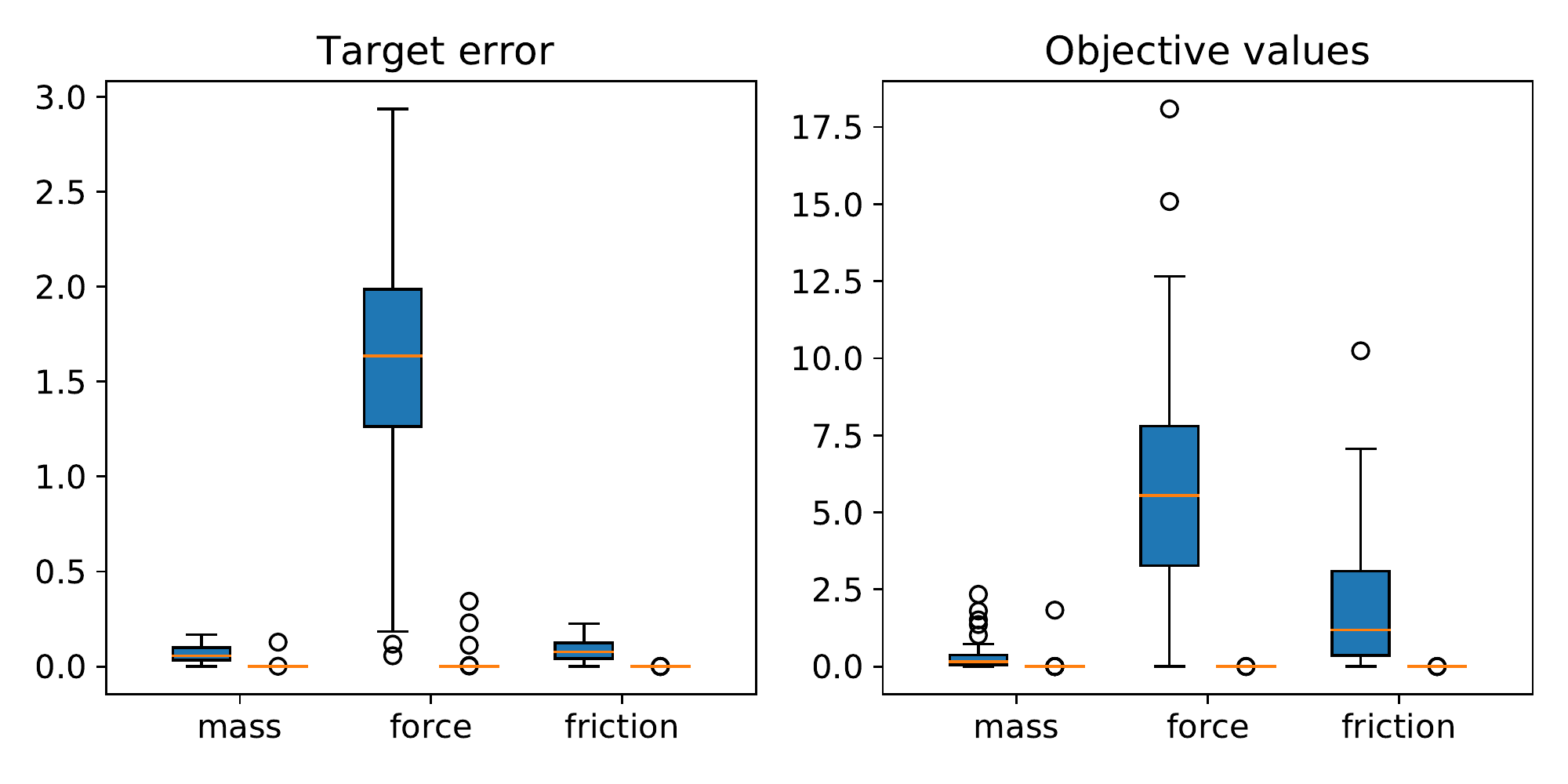}
	\caption{System identification results. Mass, force and friction are estimated with high accuracy. Left blue boxes: initialization, right green: optimization results.}
	\label{fig:exp_sysid}
\end{figure}

In Fig.~\ref{fig:exp_sysid}, we show results for friction, mass and force estimation in which either the duck-like ``bob'', or the cow ``spot'' is placed on a plane pushed with a constant force along the plane for 1\,s.
We generate 50 runs for each optimization target (mass, force and friction).
We uniformly sample the force between 2 and 5 for each of the two dimensions along the plane, mass between 0.9 and 1.1, and friction between 0.01 and 0.25.
Depending on the setting, one of these parameters is sampled anew to create a varied initialization.
We optimize the mean squared position error of the simulation towards the ground-truth.
The physical quantities are recovered with high accuracy in most settings.
The average error drops from 1.55 to 0.01 for force, 0.07 to 0.002 for mass, and 0.08 to 1e-4 for friction.

\subsection{Fitting to depth image observations}

\begin{figure}
	\centering
	\includegraphics[width=0.19\linewidth]{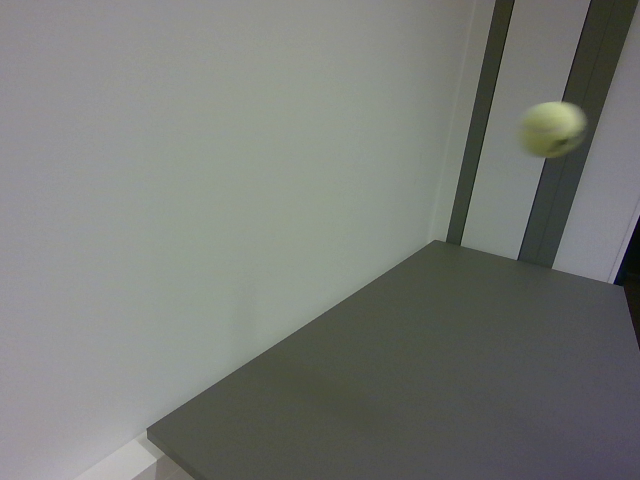} 
	\includegraphics[width=0.19\linewidth]{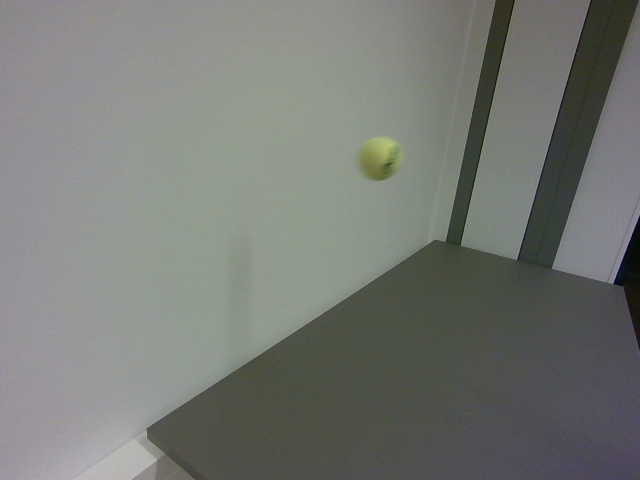}
	\includegraphics[width=0.19\linewidth]{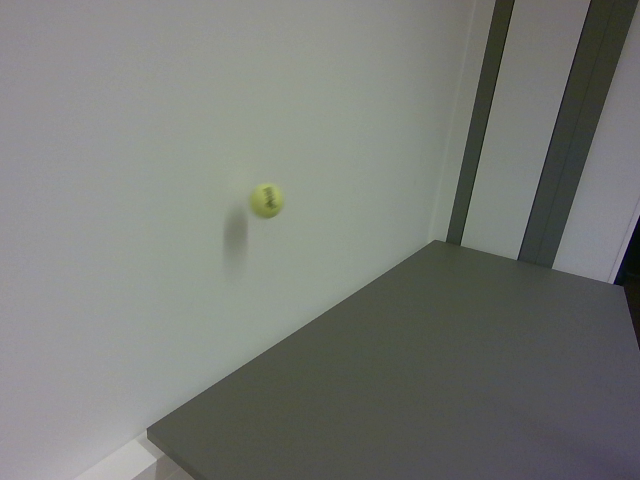}
	\includegraphics[width=0.19\linewidth]{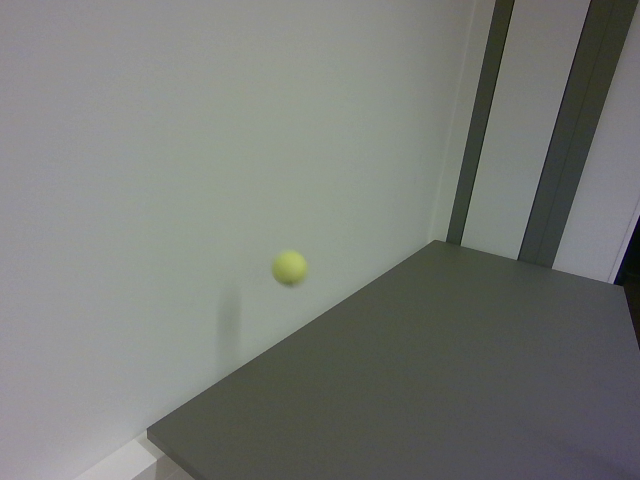}
	\includegraphics[width=0.19\linewidth]{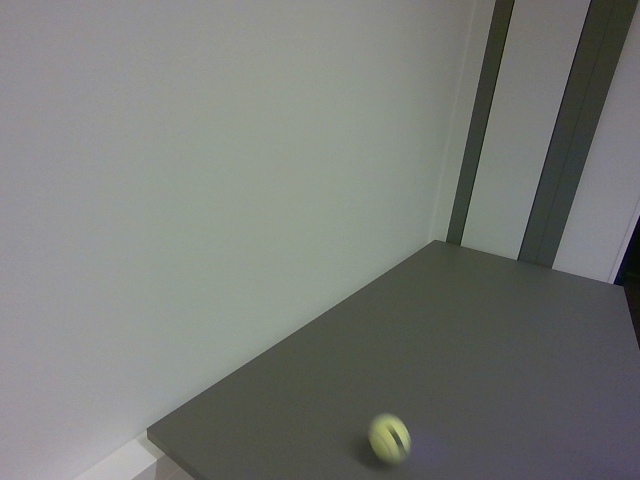}\\
	\includegraphics[width=0.19\linewidth]{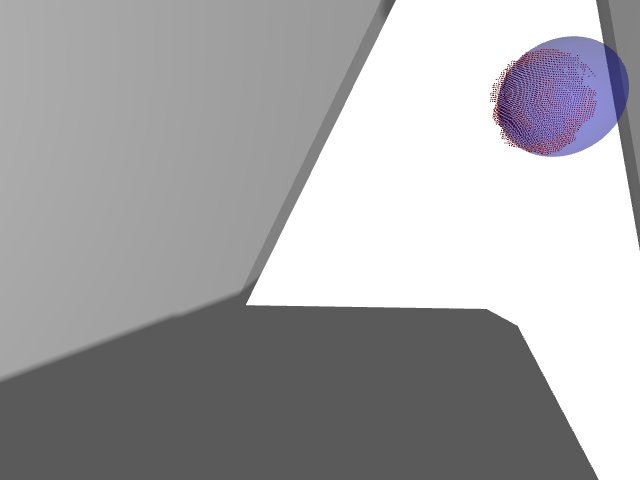} 
	\includegraphics[width=0.19\linewidth]{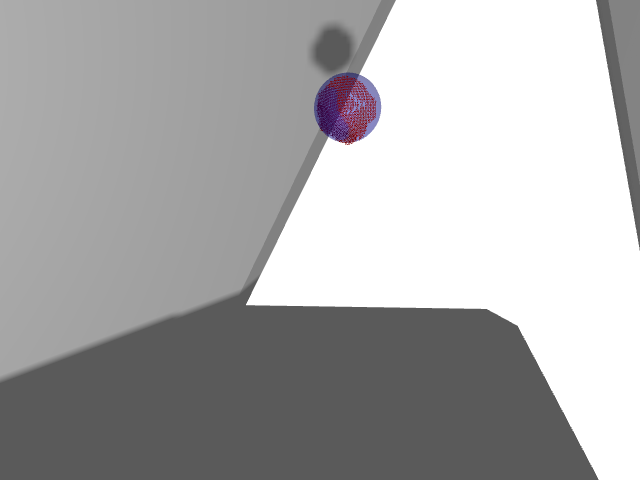}
	\includegraphics[width=0.19\linewidth]{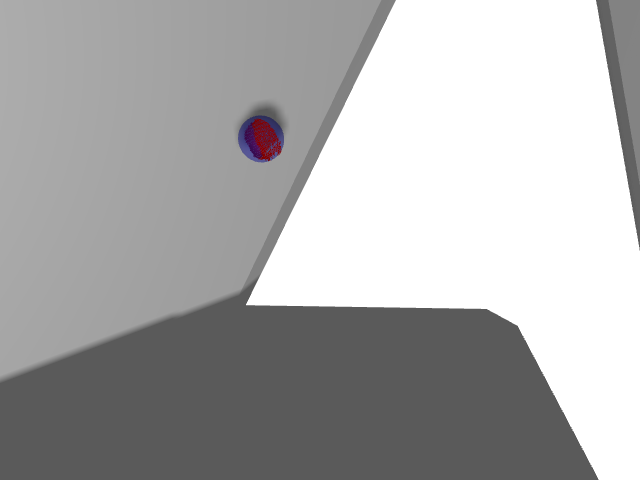}
	\includegraphics[width=0.19\linewidth]{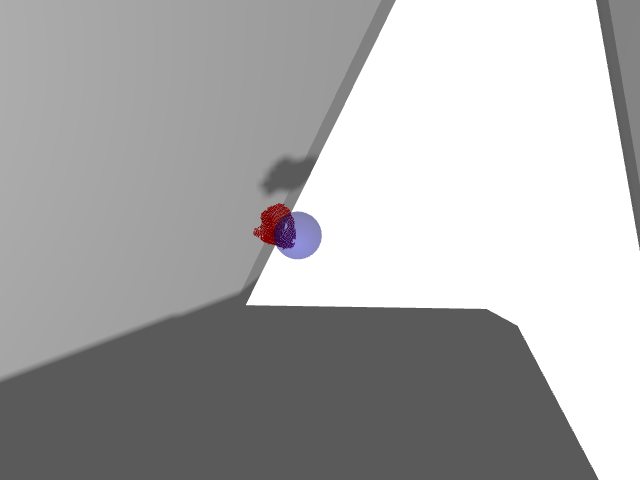}
	\includegraphics[width=0.19\linewidth]{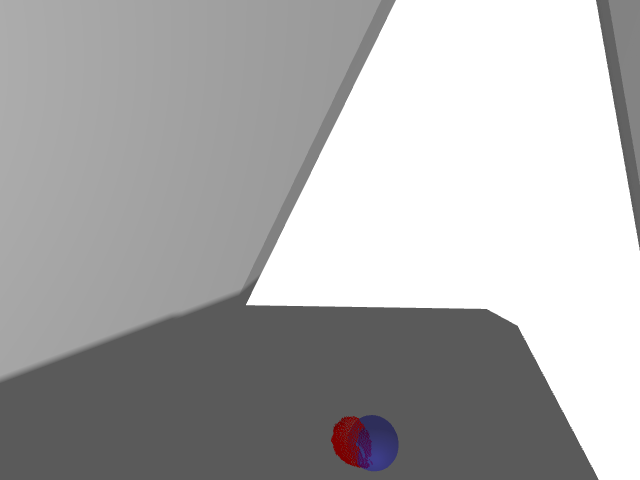}\\
	\includegraphics[width=0.19\linewidth]{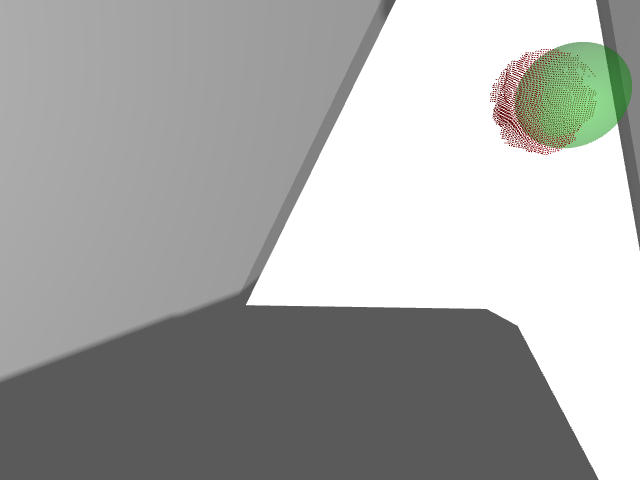}
	\includegraphics[width=0.19\linewidth]{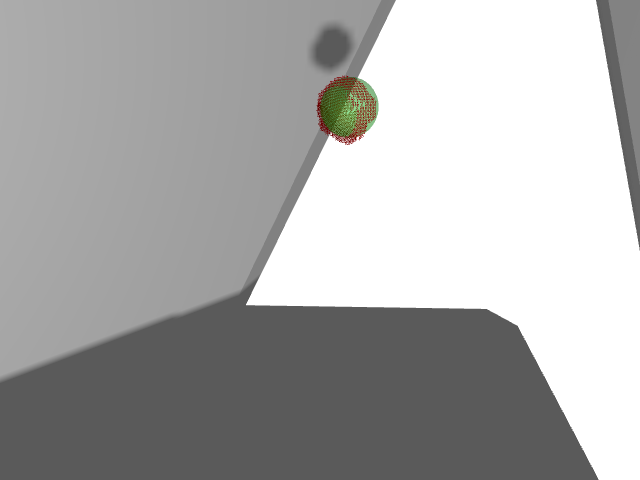}
	\includegraphics[width=0.19\linewidth]{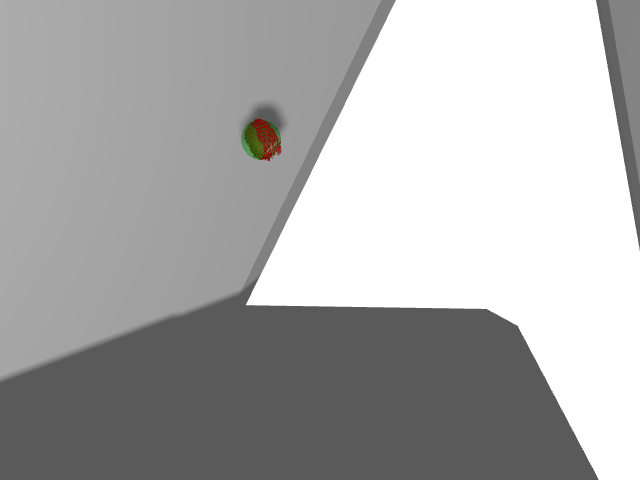}
	\includegraphics[width=0.19\linewidth]{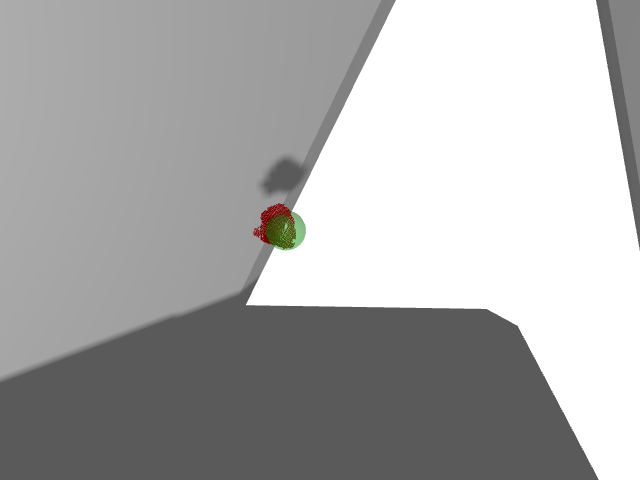}
	\includegraphics[width=0.19\linewidth]{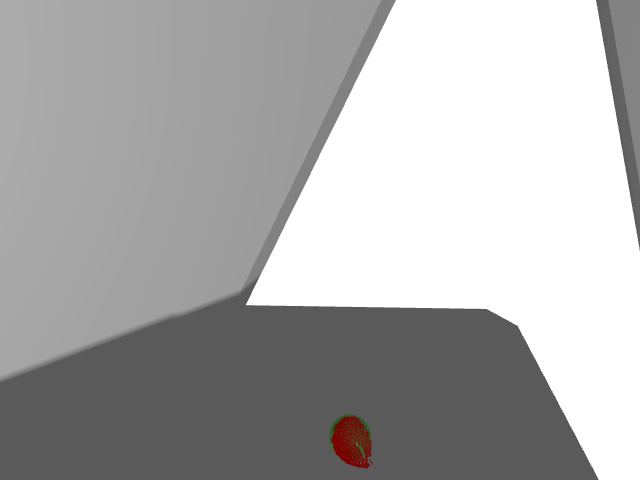}\\
	\caption{Results on a real-world scene (frame 0, 2, 4, 6 and 9). Blue: fit to first two frames. Green: fit of simulated trajectory. Red: Recorded point cloud segment. Our result (green) fits the trajectory better overall, especially after the bounce. The radius is improved from 3.68\,cm (initial 2 frames fit) to 3.13\,cm (trajectory fit, ground truth 3.24\,cm).}
	\label{fig:real_exp}
\end{figure}

We also test our method for fitting the physics simulation based on synthetic depth image observations from a static camera.
We render depth images and object segmentation masks at resolution of $640 \times 480$.
The depth images are augmented with synthetic per-pixel Gaussian noise with mean $\mu = d$ and std. dev. $\sigma = 0.0001 d^2$.
We evaluate using the bounce scenarios (Sec.~\ref{sec:shape_id}) with spheres and cubes with rounded edges in randomized poses and sizes.
We run the scenarios 20 times each for different settings (sphere with and without gravity and cube without gravity).
The ground truth pose is set by adding Gaussian noise with mean 0 and standard deviation 0.1 to a standard rotation (represented as unit quaternion) and position.
We initialize the first pose by adding the same noise distribution to the ground-truth pose.
Spheres and cubes are sampled with radii/edge lengths in range 0.5 to 1.5.
For pose and trajectory fitting, we use the mean squared SDF value of the measured 3D points after transforming them to the object frame using its pose estimate and the known camera view pose.
Fitting of pose and object size in a single frame is prone to local minima and typically overestimates the object size. 
We sample object sizes which range from 0 to 1.0 larger than the target size.
We compare the single frame pose fit with its refinement through fitting the trajectory on all depth image observations in each sequence using the differentiable simulation.
Table \ref{tab:exp_pointcloud} shows that fitting pose and size of spheres and cubes to a depth map falls into a local optimum, making the pose estimate worse while improving the shape estimates.
Refining this result via trajectory optimization improves radius and pose estimate.
In Fig.~\ref{fig:real_exp} we provide results for a bounce of a tennis ball on a wall recorded with an Intel Realsense D455 camera (640$\times$480 at 30\,Hz).
We optimize initial position and velocity, radius, restitution and friction based on the SDF alignment objective.
The result demonstrates that trajectory and radius estimates can be improved by our physics-based approach in this challenging scenario.
Further details are provided in the supplementary material.

\begin{table}
\footnotesize
	\centering
	\begin{tabular}{lccc}\toprule
		 & \multicolumn{2}{c}{sphere} & \multicolumn{1}{c}{cube} \\\cmidrule(lr){2-3}\cmidrule(lr){4-4}
		error & w/o gravity & w/  gravity &  w/o gravity \\\midrule
		init pos & 0.040 & 0.040 & 0.040 \\
		pos frame fit & 0.056 & 0.056 & 0.077  \\
		pos traj. fit & 0.031 & 0.044 & 0.023  \\
		init rot & -- & -- & 0.135  \\
		rot frame fit & -- & -- & 0.001  \\
		rot traj. fit & -- & -- & 0.002  \\
		init size & 0.512 & 0.512 & 0.512  \\
		size frame fit & 0.163 & 0.163 & 0.137  \\
		size traj. fit & 0.022 & 0.014 & 0.029  \\
		\bottomrule
	\end{tabular}
	\caption{Position and shape parameter errors for single-frame fitting and trajectory fitting to depth observations.}
	\label{tab:exp_pointcloud}
\end{table}

\subsection{Runtime}

We implemented our method in PyTorch \cite{NEURIPS2019_9015} and have not yet tuned our implementation for efficiency.
Currently, our method requires several seconds of computation time for one second in simulation.
Major room for run-time improvements is in collision detection and contact point estimation which we have not yet optimized on the GPU.
Note that~\cite{macklin2020_local} have demonstrated that these steps can be significantly sped up with timings in the microseconds.
The LCP optimization could be sped up by approaches such as~\cite{shao2021_accurately}.

\subsection{Limitations}

We observed that bouncing boxes with sharp corners and edges can have strong variations in contact points and number of contacts which makes system identification challenging due to varying contact situations (\eg a box collides with different corners/edges).
First-order gradient based based optimization can be difficult in such cases for the trajectory alignment.
In the shape from inertia experiment we observed that in some outlier cases, depending on the start and target configurations, the torque and the shape space, the velocity loss can be reduced while the shape is converging to a wrong local minimum.
Non-convex shapes can render the system identification problem itself non-convex for which gradient descent will retrieve local minima.

%% file: conclusion.tex
We propose a novel approach for differentiable rigid-body physics simulation that models arbitrary watertight shapes using SDF representations.
We devise differentiable inertia tensors and time of contact in a velocity-based constraint-based time-stepping method.
Our experimental results demonstrate that physical system identification including shape inference is possible for several challenging scenarios with non-convex shapes and collisions via gradient descent.
We fit our model on sample trajectories and depth image observations of synthetic scenes.
Further scaling our approach for system identification, 3D vision and control in more complex scenarios through more elaborate optimization methods than gradient descent is an interesting direction for future research.

%% file: supplement.tex
\renewcommand{\thefigure}{A.\arabic{figure}}
\setcounter{figure}{0}
\renewcommand{\thetable}{A.\arabic{table}}
\setcounter{table}{0}
\renewcommand{\theequation}{A.\arabic{equation}}
\setcounter{equation}{0}

\section{Introduction}

In this supplementary material, we provide further details on our method and experiments.

\section{Differentiable Physics Simulation}

In this section we provide continued details for the differentiable physics engine used in our approach.

\subsection{Collision Constraints}
As given in the main paper, the collision pose constraint function is
$g_c( \mathbf{x} ) = \mathbf{n}^\top \left( \mathbf{p}^w_i - \mathbf{p}^w_j \right) - \epsilon$,
where $\mathbf{p}^w_{i} := \mathbf{x}_{i} + \mathbf{p}^i_{i} = \mathbf{x}_{i} + \mathbf{R}_{i} \mathbf{p}_{i}$ and $\mathbf{n}$ are contact point and normal in the world frame, and $\mathbf{R}_{i} \in SO(3)$ is the rotation of the object frames relative to the world frame.
The contact Jacobian $\mathbf{J}_c$ of this constraint follows from
\begin{align}
        \dot{g}(\mathbf{x}) &= \mathbf{n}^\top (\mathbf{\dot{p}}^w_i - \mathbf{\dot{p}}^w_j)\\
        &= \mathbf{n}^\top ((\mathbf{v}_i + \boldsymbol{\omega}_i \times \mathbf{p}^i_i) - (\mathbf{v}_j + \boldsymbol{\omega_j} \times \mathbf{p}^j_j))\\
        &= \underbrace{\begin{pmatrix}(\mathbf{p}^i_{i} \times \mathbf{n})^T & \mathbf{n}^T
        \end{pmatrix}}_{\mathbf{J}_i}\underbrace{\begin{pmatrix}\boldsymbol{\omega}_i \\ \mathbf{v}_i \end{pmatrix}}_{\boldsymbol{\xi}_i}\\ 
        &\hspace{5ex}+ 
        \underbrace{\begin{pmatrix}-(\mathbf{p}^j_{j} \times \mathbf{n})^T & -\mathbf{n}^T
        \end{pmatrix}}_{\mathbf{J}_j}\underbrace{\begin{pmatrix}\boldsymbol{\omega}_j \\ \mathbf{v}_j \end{pmatrix}}_{\boldsymbol{\xi}_j}\\
        &= \underbrace{\begin{pmatrix}\mathbf{J}_i & 0\\ 0 & \mathbf{J}_j \end{pmatrix}}_{\mathbf{J}_c}
        \underbrace{\begin{pmatrix}\boldsymbol{\xi}_i\\ \boldsymbol{\xi}_j\end{pmatrix}}_{\boldsymbol{\xi}} = \mathbf{J}_c \boldsymbol{\xi}.\label{eq:collision_jac}
\end{align}

\subsection{Friction Jacobians}

The friction Jacobian $\mathbf{J}_f$ is calculated similar to $\mathbf{J}_c$ (eq.~\eqref{eq:collision_jac}), now replacing the contact normal direction $\mathbf{n}$ with directions along the tangential contact surface.
We approximate the spherical friction cone with a polyhedral cone using 8 directions $\mathbf{d}_1,\ldots,\mathbf{d}_8 \in \mathbb{R}^3$ with unit norm and equal angular spacing on the tangential plane.
We use 8 directions as a trade off between efficiency and accuracy and determine the directions as
\begin{equation}
	\begin{split}
		\mathbf{p}^\perp &= \mathbf{p} \times \mathbf{e}_{j^*}, j^* = \operatorname{arg\,min}_{j \in \{ 0, 1, 2 \}} \left| p_j \right|\\
		\mathbf{d}_1 &= \frac{\left(\mathbf{p}_i^i\right)^\perp}{\left\|\left(\mathbf{p}_i^i\right)^\perp\right\|_2}\\
		\mathbf{d}_2 &= \frac{ \mathbf{d}_1 \times \mathbf{p}_i^i }{ \left\| \mathbf{d}_1 \times \mathbf{p}_i^i \right\|_2 }\\
		\mathbf{d}_3 &= \frac{ \mathbf{d}_1 + \mathbf{d}_2 }{ \left\| \mathbf{d}_1 + \mathbf{d}_2 \right\|_2 }\\
		\mathbf{d}_4 &= \frac{ \mathbf{d}_3 \times \mathbf{p}_i^i }{ \left\| \mathbf{d}_3 \times \mathbf{p}_i^i \right\|_2 }\\
		\mathbf{d}_{4+i} &= -\mathbf{d}_{i}, i \in \{ 1,\ldots,4 \}
	\end{split},
\end{equation}
where $\mathbf{e}_{j}$ is the j-th unit vector.

\subsection{Linear Complementarity Problem}

The constrained dynamics model can be written as the following linear complementarity problem (LCP) with the variables defined as in the main paper
	\begin{multline}
		\begin{pmatrix}
			0\\
			0\\
			\mathbf{a}\\
			\boldsymbol{\sigma}\\
			\boldsymbol{\zeta}
		\end{pmatrix}
		-
		\begin{pmatrix}
			\mathbf{M} & -\mathbf{J}_{e}^T & -\mathbf{J}_{c}^T & -\mathbf{J}_{f}^T & 0\\
			\mathbf{J}_{e} & 0 & 0 & 0 & 0\\
			\mathbf{J}_{c} & 0 & 0 & 0 & 0\\
			\mathbf{J}_{f} & 0 & 0 & 0 & \mathbf{E}\\
			0 & 0 & \mu & -\mathbf{E}^T & 0
		\end{pmatrix}
		\begin{pmatrix}
			\boldsymbol{\xi}_{t+h}\\
			\boldsymbol{\lambda}_{eq}\\
			\boldsymbol{\lambda}_{c}\\
			\boldsymbol{\lambda}_{f}\\
			\boldsymbol{\gamma}
		\end{pmatrix}\\
		=
		\begin{pmatrix}
			-\mathbf{M}\boldsymbol{\xi}_{t} - h\mathbf{f}_{ext}\\
			0\\
			\mathbf{c}\\
			0\\
			0
		\end{pmatrix},\\
	\label{eq: friction matrix dynamics impulse}
	\end{multline}
	\begin{equation*}
		\textrm{subject to}\:\:\begin{pmatrix}
			\mathbf{a}\\
			\boldsymbol{\sigma}\\
			\boldsymbol{\zeta}
		\end{pmatrix}\geq 0,
		\:\:\begin{pmatrix}
			\boldsymbol{\lambda}_c\\
			\boldsymbol{\lambda}_f\\
			\boldsymbol{\gamma}
		\end{pmatrix}\geq 0,
		\:\:\begin{pmatrix}
			\mathbf{a}\\
			\boldsymbol{\sigma}\\
			\boldsymbol{\zeta}
		\end{pmatrix}^T\begin{pmatrix}
			\boldsymbol{\lambda}_c\\
			\boldsymbol{\lambda}_f\\
			\boldsymbol{\gamma}
		\end{pmatrix}=0.
	\end{equation*}

\subsection{Dependency of LCP solution derivative on contact points}

The derivative of the physics simulation is found at the solution of the LCP from the previous section~\cite{avilabelbuteperes2018_end}.
The contact points appear in the contact and friction Jacobians in a cross product with the contact normal which is then multiplied with the rotational velocity in a dot product (see Eq.~\eqref{eq:collision_jac}).
Direct dependency on linear velocity is not incorporated in the velocity constraint.
Moreover, the LCP does not solve for the time step $h$ through the collision constraint, but assumes it constant.
That means that also no implicit dependency of the time step on other states and parameters is modeled.
With our time of contact differential based on the contact position constraint, we include these dependencies.

\section{Differentiable Collision Handling}

\subsection{Collision Detection}
\begin{figure*}
	\centering
	\includegraphics[width=0.3\linewidth]{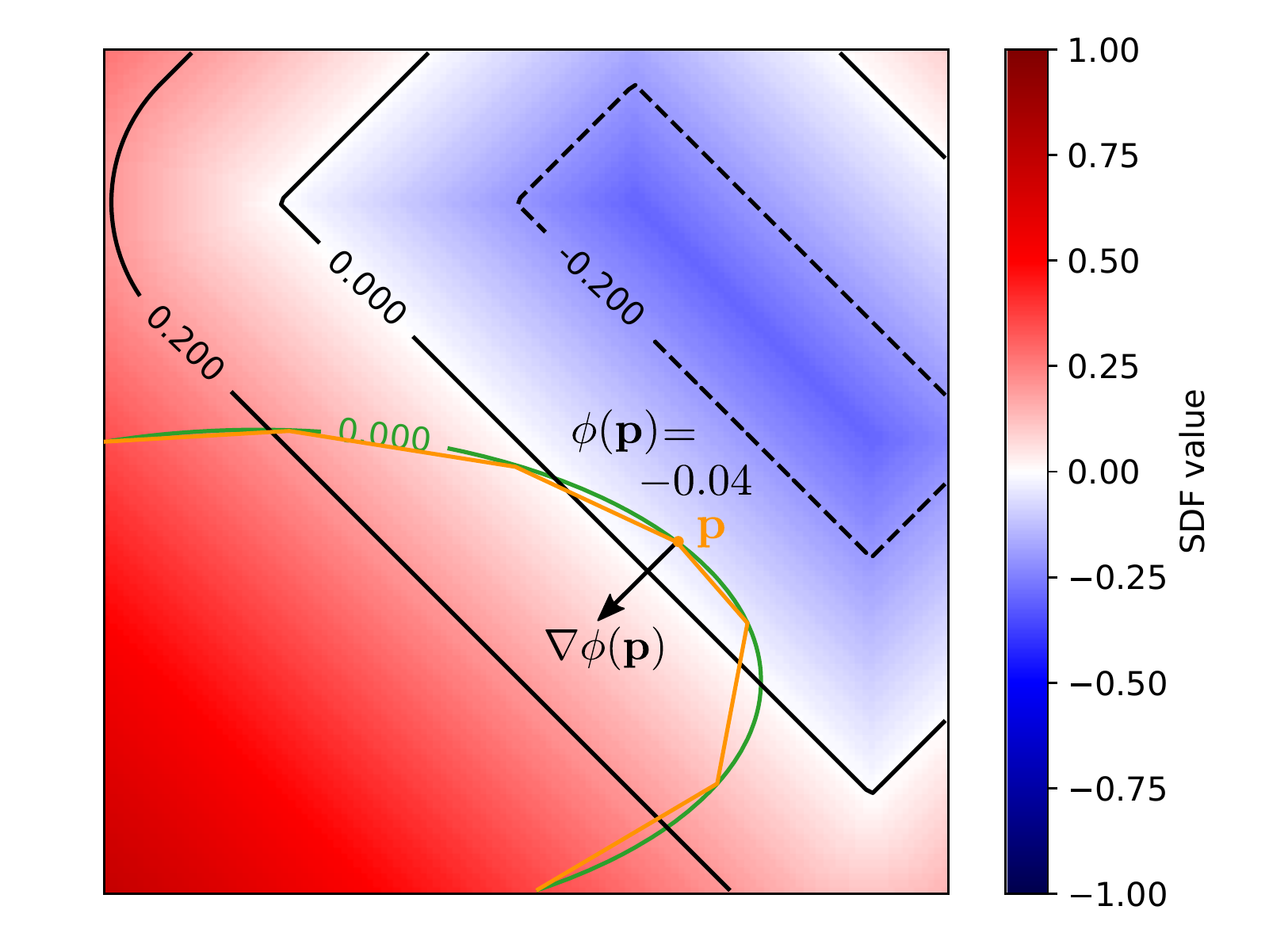}\hfill
	\includegraphics[width=0.3\linewidth]{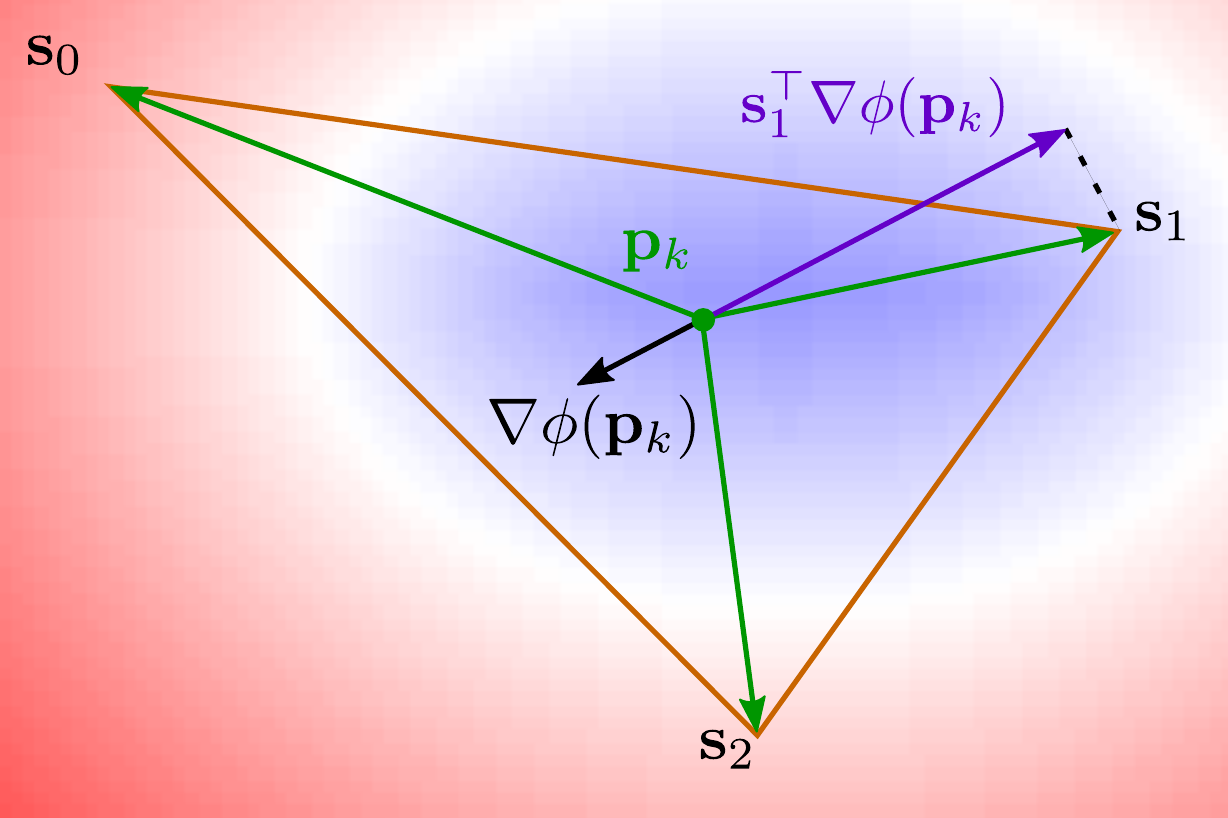}\hfill
	\includegraphics[width=0.3\linewidth]{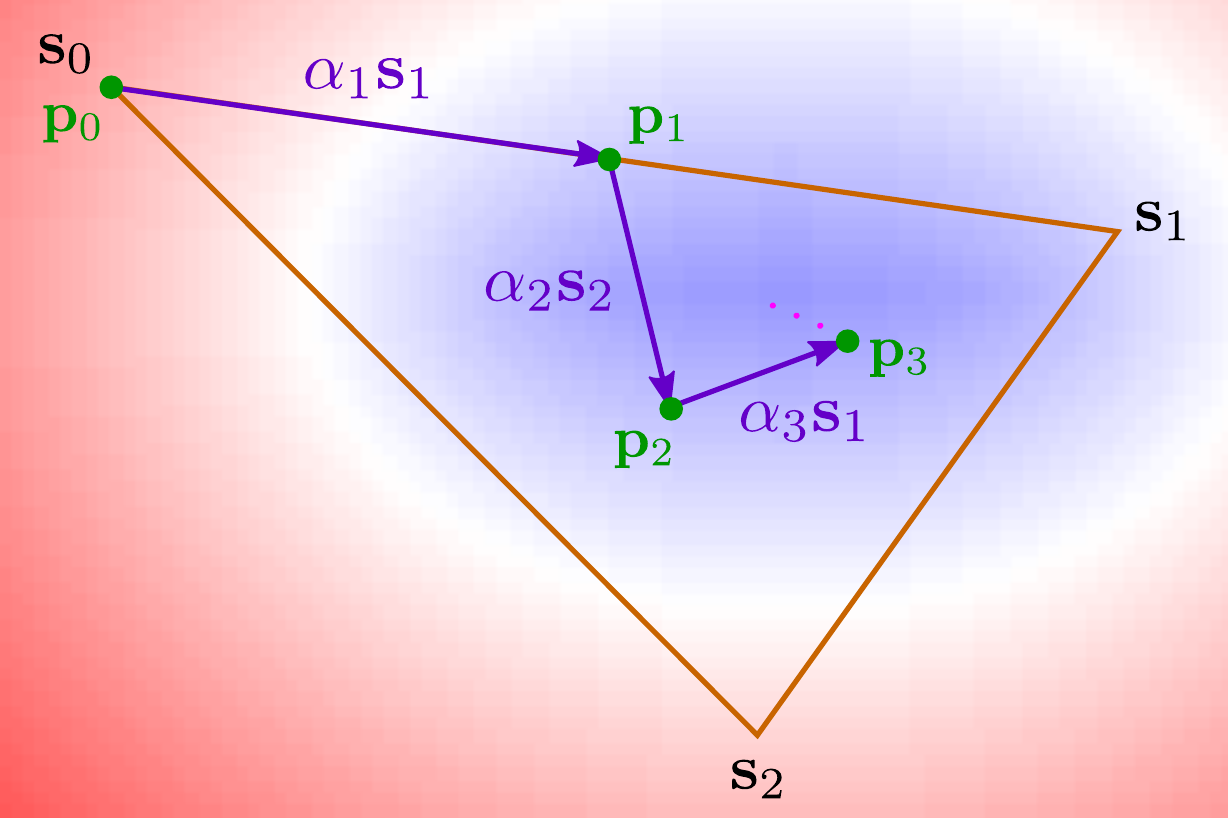}
	\caption{Collision detection with SDFs. Left: we extract a differentiable mesh (orange) from an SDF (ellipse, green) and find contact points $\mathbf{p}$ (points of maximum penetration) and normal directions $\nabla \phi(\mathbf{p})$ (direction towards closest point on penetrated surface) on the mesh faces towards a second SDF $\phi$ (box, black). The differentiable mesh allows for propagating gradients from the contact points and normals onto the SDF shape parametrization. Middle: We localize contact points using a Frank-Wolfe algorithm which iteratively select the vertex $\mathbf{s_i}$ with smallest signed distance $\mathbf{s_i}^\top \nabla \phi(\mathbf{p_k})$ from the current point $\mathbf{p_k}$. Right: The solution of the Frank-Wolfe algorithm can be written as a linear combination of the vertex positions. The found contact point is differentiable for the underlying SDF through the vertex positions.}
	\label{fig:sdfcontacts}
\end{figure*}

As explained in the main paper, we use the Frank-Wolfe-based approach proposed in \cite{macklin2020_local} for contact detection.
As illustrated in Fig.~\ref{fig:sdfcontacts}, this optimization yields a linear interpolation between the mesh vertices.
We utilize this fact and store the interpolation parameters $\alpha_i$, so we can recompute the contact points directly based on the mesh vertices.
Thus, we avoid differentiating through the optimization iterations of the Frank-Wolfe algorithm.

\subsection{Proof for reducing contact points to their convex hull}
Our contact detection might generate an overcomplete set of contact points, which in turn might make the LCP very large.
We thus propose to reduce contact points which share a common surface normal to their convex hull.
We introduce the following theroem to show that this does not change the constraints implemented through the contact and friction jacobians.

\begin{theorem}
	Let $C$ be a set of contact points between two bodies $i$ and $j$ sharing a common surface normal. 
	Let further $\mathcal{C} = \{(\mathbf{p}_i^1, \mathbf{p}_j^1), \dots, (\mathbf{p}_i^n, \mathbf{p}_j^n)\}$ be the convex hull of $C$ and let the contact constraint
	\begin{equation}
	\mathbf{J}_c\boldsymbol{\xi}_{t + h} \ge -k \mathbf{J}_c\boldsymbol{\xi}_t = - \mathbf{c}
	\label{eq:contact_constr}
	\end{equation}
	be satisfied for all contact points in $\mathcal{C}$.
	Then the contact constraint \eqref{eq:contact_constr} is also satisfied for all points of $C$.
	\begin{proof}
		We start by expressing the contact Jacobians in their full form:
		\begin{equation}
			\mathbf{J}_c = \begin{pmatrix}
			\mathbf{J}_i & 0 \\
			0 & \mathbf{J}_j
			\end{pmatrix}
		\end{equation}
		where
		\begin{equation}
		\mathbf{J}_i = \begin{bmatrix}
		(\mathbf{p}_i \times \mathbf{n})^T & \mathbf{n}^T
		\end{bmatrix}, \quad \mathbf{J}_j = -\begin{bmatrix}
		(\mathbf{p}_j \times \mathbf{n})^T & \mathbf{n}^T
		\end{bmatrix}
		\end{equation}
		By definition of the convex hull, we can express any point that lies inside the convex hull as a linear combination of the hull points:
		\begin{equation}
		\mathbf{p}_i = \sum_{k=1}^n a_k \mathbf{p}_i^k, \text{ where } \sum_{k=1}^n a_k = 1; \forall k : a_k \ge 0.
		\label{eq:conv_hull}
		\end{equation}
		
		Now, we apply these properties to show that the contact constraint is automatically satisfied for all contact points in $C$ if it is satisfied for all points in $\mathcal{C}$:
		\begin{align}
		&\mathbf{J}_c\boldsymbol{\xi}_{t+h} \\
		&\quad\begin{aligned} = \begin{bmatrix}
		(\mathbf{p}_i \times \mathbf{n})^T & \mathbf{n}^T
		\end{bmatrix}\boldsymbol{\xi}_i^{t+h}
		- \begin{bmatrix}
		(\mathbf{p}_j \times \mathbf{n})^T & \mathbf{n}^T
		\end{bmatrix}\boldsymbol{\xi}_j^{t+h}\end{aligned}\\
		&\quad \begin{aligned}
		= \begin{bmatrix}
		(\sum_{l=1}^{n}a_l\mathbf{p}_i^l \times \mathbf{n})^T & \mathbf{n}^T
		\end{bmatrix}\boldsymbol{\xi}_i^{t+h}&\\
		- \begin{bmatrix}
		(\sum_{l=1}^{n}a_l\mathbf{p}_j^l \times \mathbf{n})^T & \mathbf{n}^T
		\end{bmatrix}&\boldsymbol{\xi}_j^{t+h}
		\end{aligned}\label{eq:conv_hull_appl_1}\\
		&\quad\begin{aligned}
		=\sum_{l=1}^{n}a_l \left(\begin{bmatrix}
		(\mathbf{p}_i^l \times \mathbf{n})^T & \mathbf{n}^T
		\end{bmatrix}\boldsymbol{\xi}_i^{t+h}\right.&\\
		- \begin{bmatrix}
		(\mathbf{p}_j^l \times \mathbf{n})^T & \mathbf{n}^T
		\end{bmatrix}&\boldsymbol{\xi}_j^{t+h}\Big)
		\end{aligned}\label{eq:distr_law_1}\\	
		&\quad\begin{aligned}
		\ge \sum_{l=1}^n a_l \Big(-k\Big[\begin{bmatrix}
		(\mathbf{p}_i^l \times \mathbf{n})^T & \mathbf{n}^T
		\end{bmatrix}\boldsymbol{\xi}_i^{t}&\\
		- \begin{bmatrix}
		(\mathbf{p}_j^l \times \mathbf{n})^T & \mathbf{n}^T
		\end{bmatrix}&\boldsymbol{\xi}_j^{t}\Big]\Big)
		\end{aligned}\label{eq:constr_conv_hull}\\
		&\quad\begin{aligned}
		= -k\Big[\begin{bmatrix}
		(\sum_{l=1}^{n}a_l\mathbf{p}_i^l \times \mathbf{n})^T & \mathbf{n}^T
		\end{bmatrix}\boldsymbol{\xi}_i^{t}&\\
		- \begin{bmatrix}
		(\sum_{l=1}^{n}a_l\mathbf{p}_j^l \times \mathbf{n})^T & \mathbf{n}^T
		\end{bmatrix}&\boldsymbol{\xi}_j^{t}\Big]
		\end{aligned}\label{eq:distr_law_2}\\
		&\quad\begin{aligned}
		= -k\left[\begin{bmatrix}
		(\mathbf{p}_i \times \mathbf{n})^T & \mathbf{n}^T
		\end{bmatrix}\boldsymbol{\xi}_i^{t}
		- \begin{bmatrix}
		(\mathbf{p}_j \times \mathbf{n})^T & \mathbf{n}^T
		\end{bmatrix}\boldsymbol{\xi}_j^{t}\right]
		\end{aligned}\label{eq:conv_hull_appl_2}\\
		&\quad = -\mathbf{c}
		\end{align}
		In eqs. \eqref{eq:conv_hull_appl_1} and \eqref{eq:conv_hull_appl_2}, we applied the definition of the convex hull (eq. \eqref{eq:conv_hull}).
		Eqs. \eqref{eq:distr_law_1} and \eqref{eq:distr_law_2} follow from the distributive law and the fact that
		\begin{equation}
		\mathbf{n} = \sum_{l=1}^n a_l \mathbf{n} \text{ if } \sum_{l=1}^n a_l = 1.
		\end{equation}
		In eq. \eqref{eq:constr_conv_hull}, we applied the constraint inequality for the individual hull points.
		We have thus shown that satisfying the contact constraint \eqref{eq:contact_constr} for all points in $\mathcal{C}$ also satisfies it for all points in $C$.
	\end{proof}
	\label{thm:conv_hull_contacts}
\end{theorem}

The proof for the friction Jacobians follows analogous to the proof of Theorem \ref{thm:conv_hull_contacts} by replacing the contact normals with the friction directions.

\section{Additional Experiment Details}
In this section, we provide additional details on the experiments in the main paper.
We explain the evaluation metrics used in the experiments in section \ref{sec:metrics} and show numeric values, as well as different version of the plots in the main paper in section \ref{sec:supp_res}.

\subsection{Evaluation metrics}
\label{sec:metrics}
\paragraph{Shape identification (sec.~5.1 in main paper)} The shape accuracy in the bouncing objects and shape from inertia experiments is evaluated by the symmetric chamfer distance between the mesh estimate $\mathcal{M}_e = \{V_e, F_e\}$ and the target mesh $\mathcal{M}_t = \{V_t, F_t\}$.
The meshes consist of sets of vertices $V$ and faces $F$ and are extracted from the SDF using the marching cubes algorithm \cite{lorensen1987_marching}.
The symmetric chamfer distance is defined as
\begin{multline}
	CD(V_e, V_t) = \frac{1}{\|V_e\|}\sum_{v_e\in V_e} \min_{v_t\in V_t} \| v_e - v_t \|^2\\ + \frac{1}{\|V_t\|}\sum_{v_t\in V_t} \min_{v_e\in V_e} \| v_t - v_e \|^2,
\end{multline}
where $\|\cdot\|$ denotes the euclidean norm.

\paragraph{Friction and mass identification, force optimization (sec.~5.2 in main paper)}
Friction coefficient and mass are evaluated by the absolute difference between the estimated and target values.
The force vector is evaluated by the euclidean norm between the estimated and target force vectors.

\paragraph{Fitting to depth image observations (sec.~5.3 in main paper)}
The position error is the euclidean norm between the locations of the center of the estimated and target objects.
The rotation error is measured as the relative angle between the estimated and target rotation of the object.
The size error is the absolute difference between the estimated and target objects radius for the sphere and edge length for the cube.

\subsection{Supplementary Results for Experiments in Main Paper}
\label{sec:supp_res}

\begin{figure*}[tb]
	\centering
	\begin{subfigure}{.49\linewidth}
		\centering
	\includegraphics[trim=12 12 12 12,clip,width=.89\linewidth]{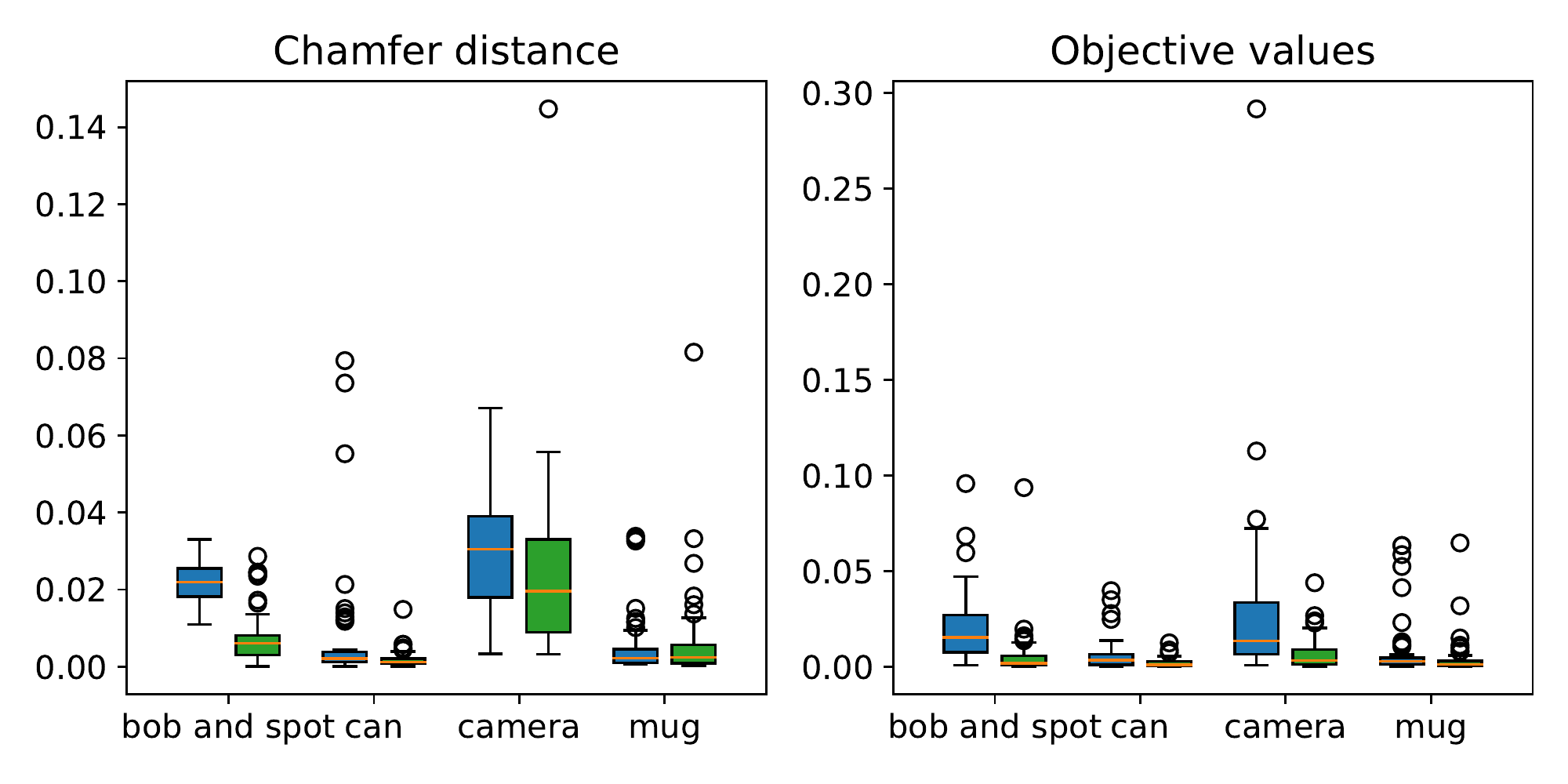}
	\end{subfigure}
	\begin{subfigure}{.49\linewidth}
		\centering
		\includegraphics[trim=12 12 12 12,clip,width=.89\linewidth]{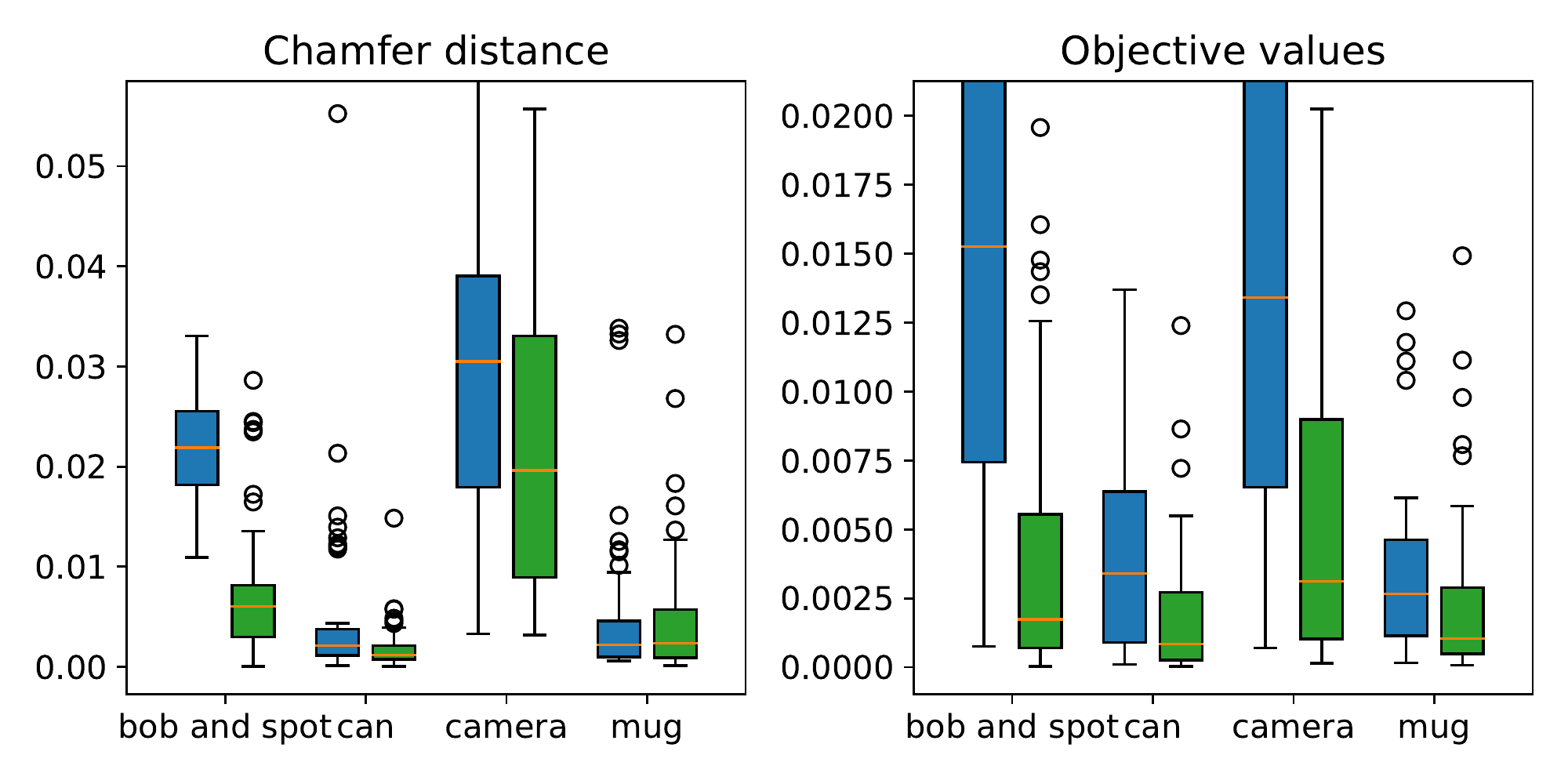}
	\end{subfigure}
	\caption{Trajectory fitting with learned shape spaces. This figure shows different scalings for Fig.~5 from the main paper. Left: including the outliers for the initializations. Right: scaling by the result error values. Left/blue boxes per object category indicate initialization while the right/green ones show the optimization result (position trajectory error).}
	\label{fig:exp_trajectory_shapespace}
\end{figure*}

\begin{figure*}
	\centering
	\begin{subfigure}{.49\linewidth}
		\centering
		\includegraphics[trim=12 12 12 12,clip,width=0.89\linewidth]{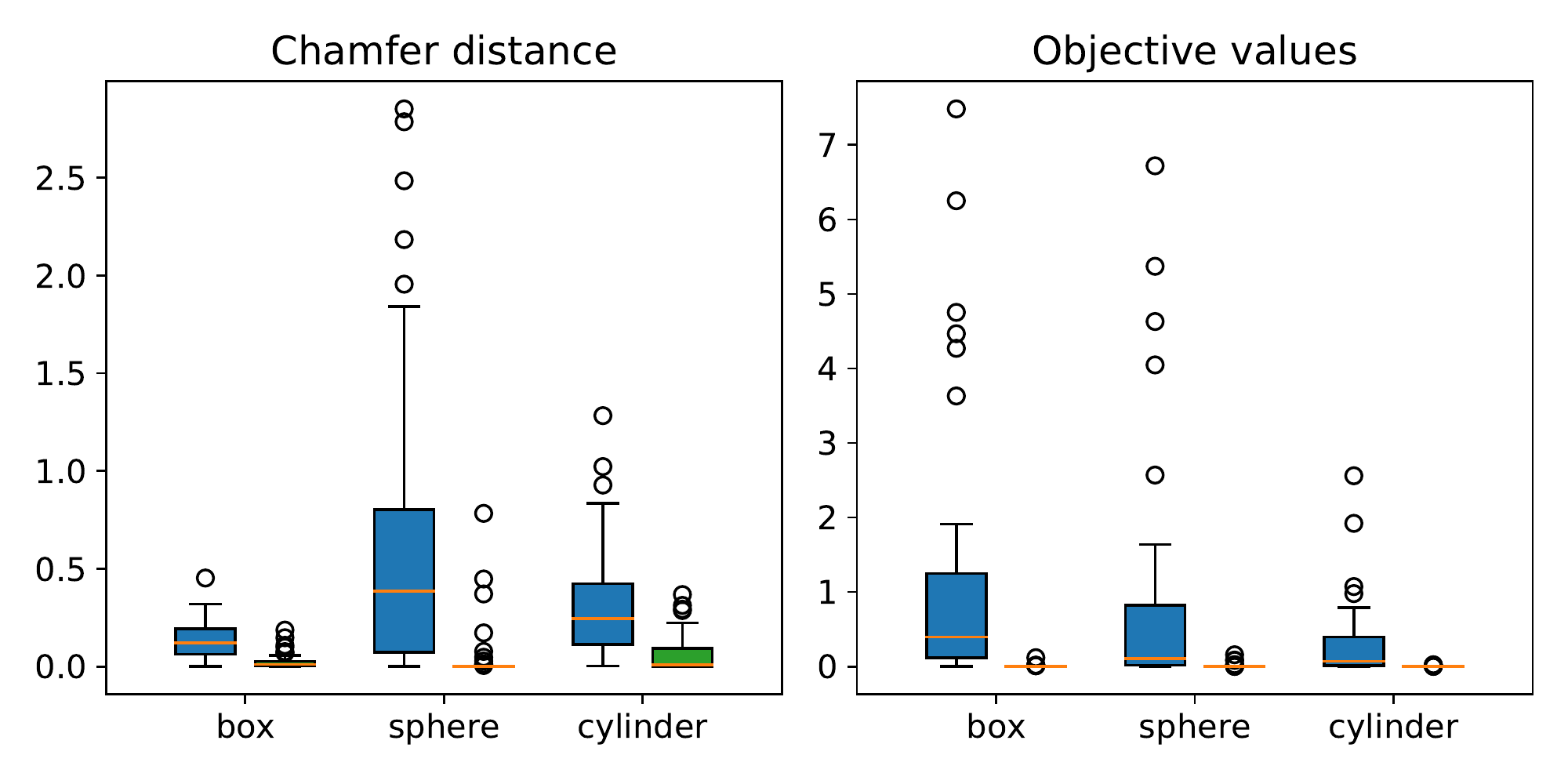}\\
		\includegraphics[trim=12 12 12 12,clip,width=0.89\linewidth]{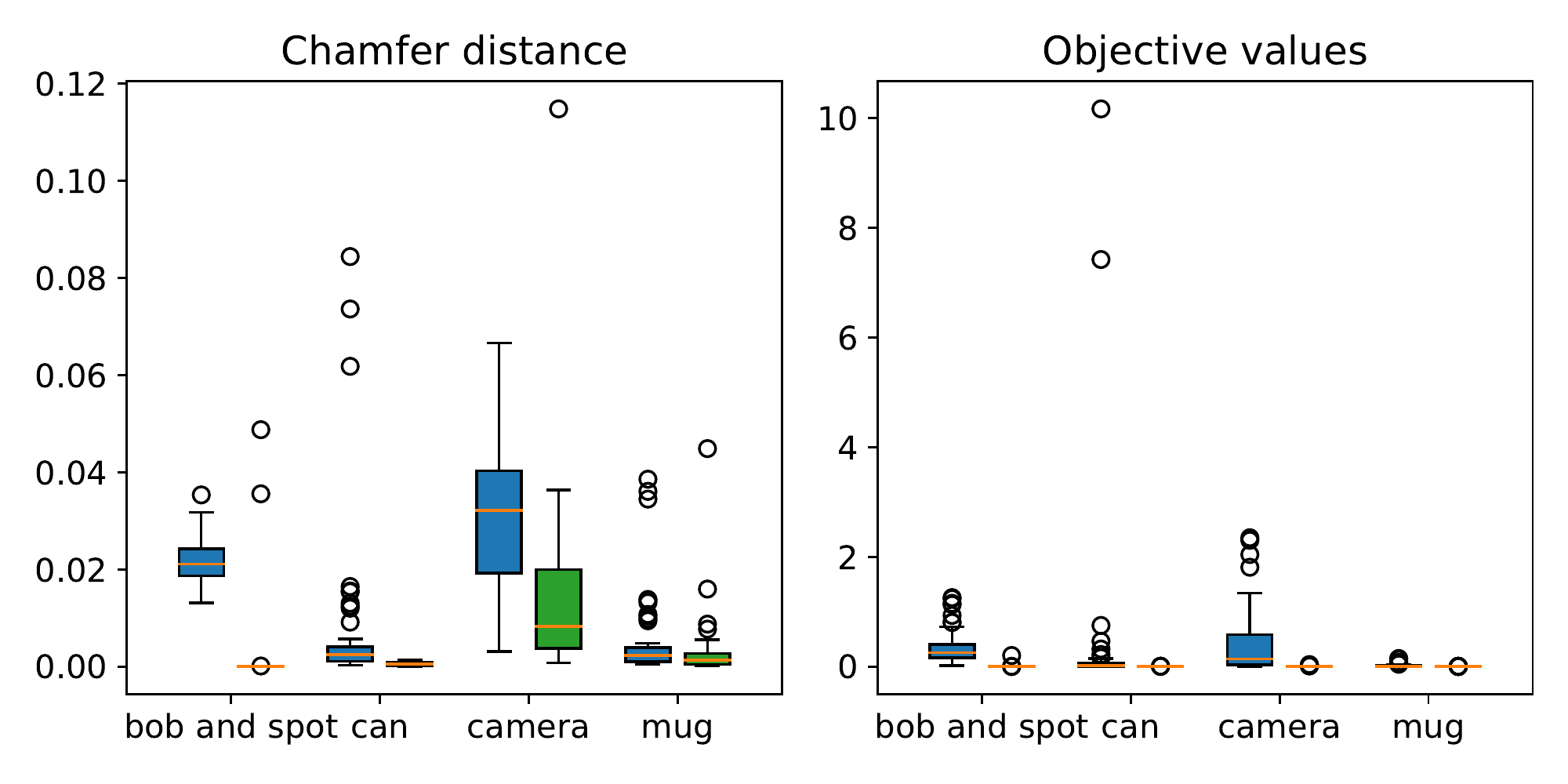}
	\end{subfigure}
	\begin{subfigure}{.49\linewidth}
		\centering
		\includegraphics[trim=12 12 12 12,clip,width=0.89\linewidth]{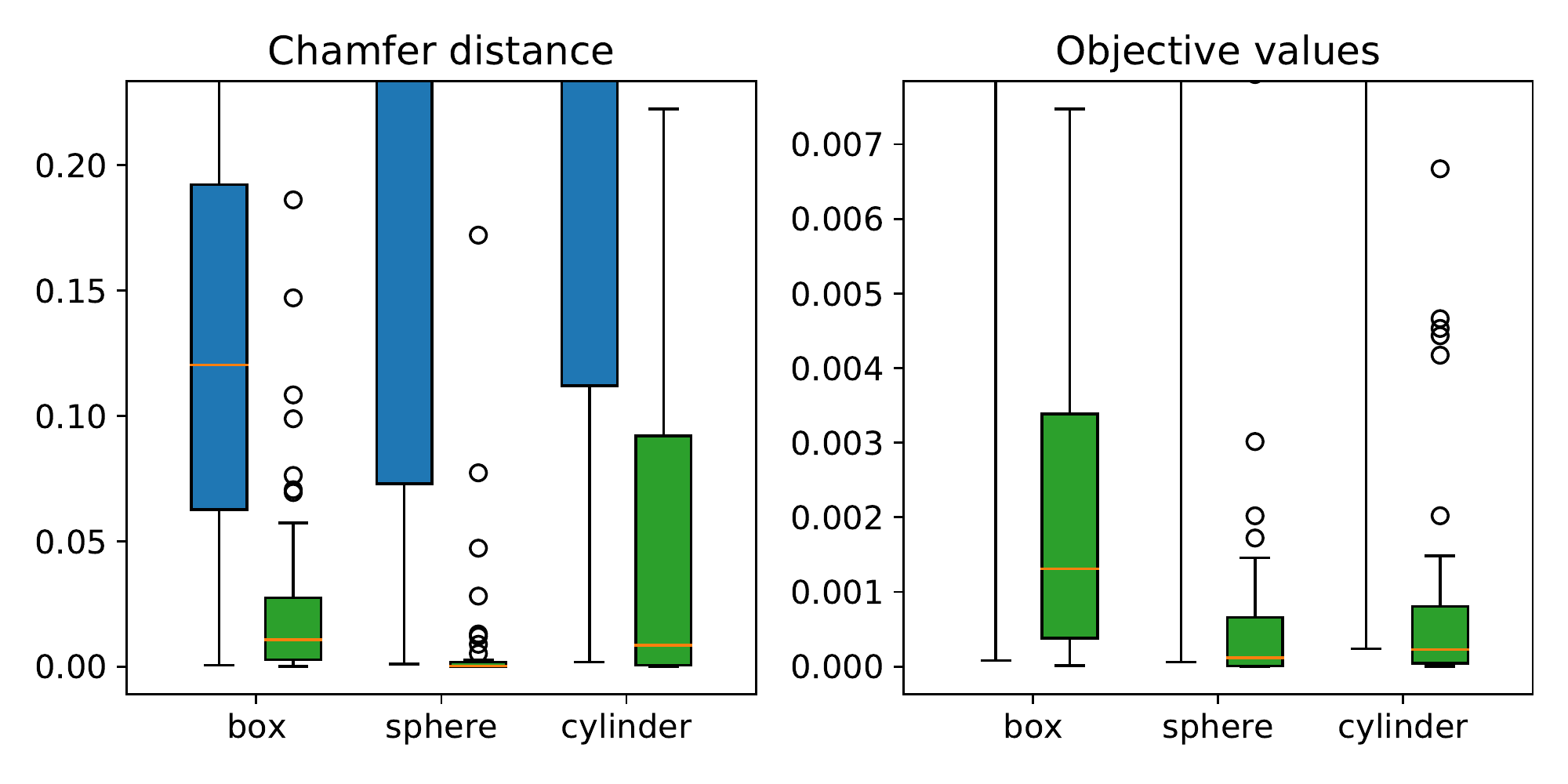}\\
		\includegraphics[trim=12 12 12 12,clip,width=0.89\linewidth]{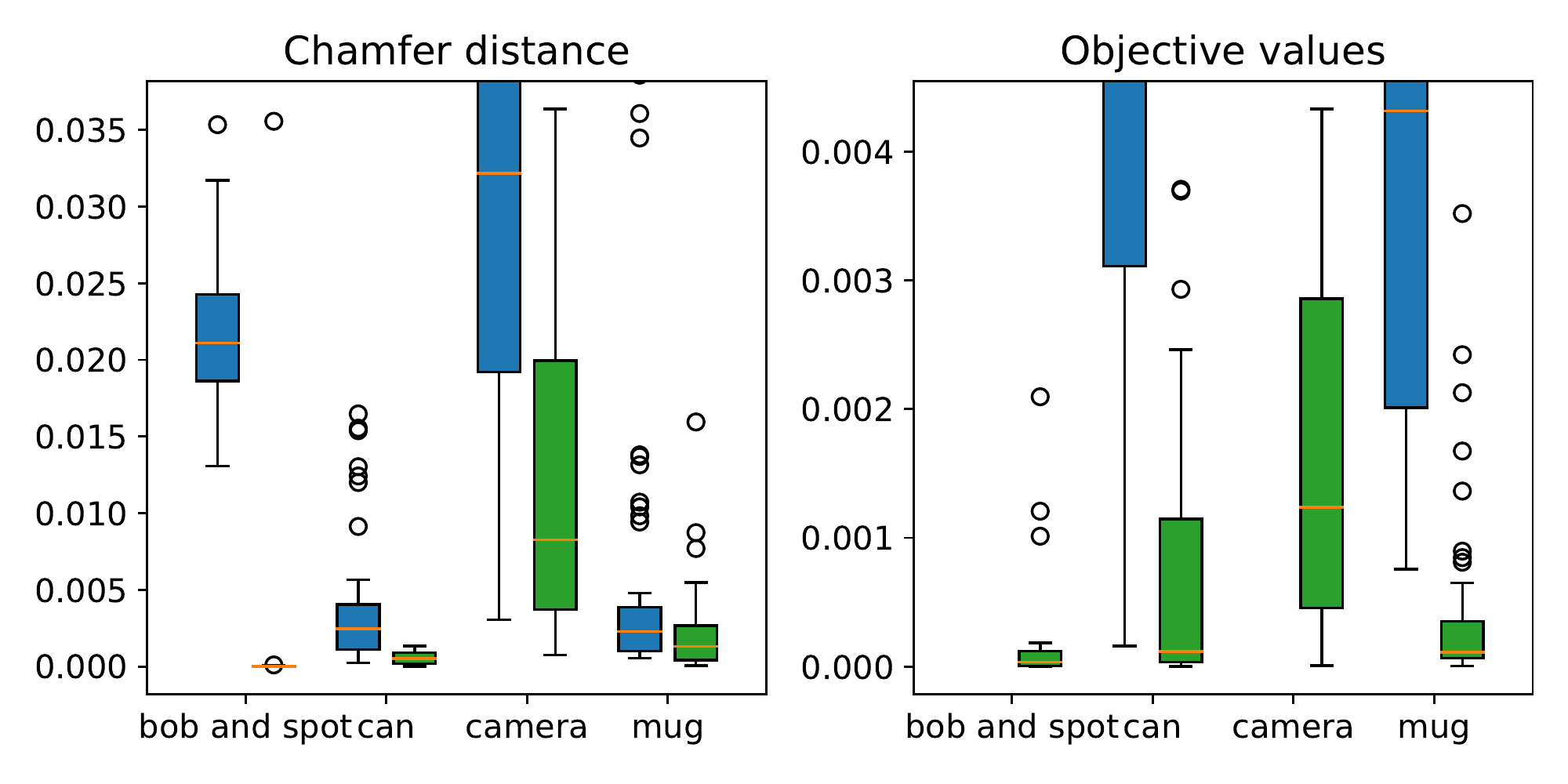}
	\end{subfigure}
	\caption{Shape from inertia. We show different scalings for Fig.~6 from the main paper. Left: including the outliers for the initializations. Right: scaling by the result errors. Box plots illustrating the distribution between initialization (left, blue for each object) and resulting values (right, green) for the chamfer distance and optimization objective (rotational velocity error).}
	\label{fig:exp_shapefrominertia}
\end{figure*}

\begin{figure}
	\centering
	\includegraphics[trim=12 12 12 12,clip,width=0.89\linewidth]{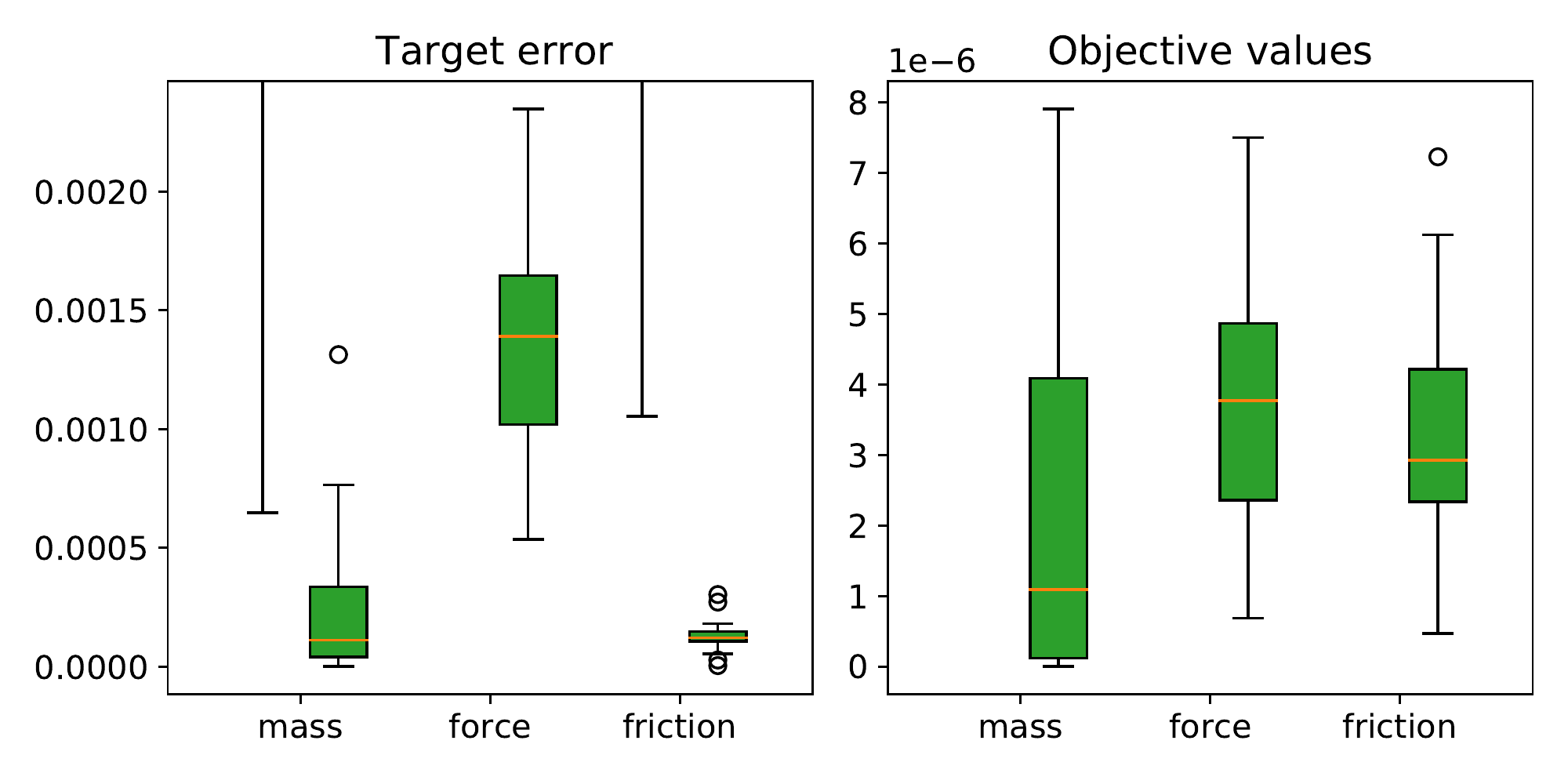}\\
	\caption{System identification results. This figure shows the same data as Fig.~7 from the main paper scaled by result error values. Mass, force and friction are estimated with high accuracy. Left/blue boxes per parameter category indicate initialization while the right/green ones show the optimization results (position trajectory error).}
	\label{fig:exp_systemid}
\end{figure}

\begin{table*}[tbh]
	\centering
	\footnotesize
	\begin{tabular}{cccccccccccccc}\toprule
		\multicolumn{2}{c}{\multirow{2}{*}{Error}} & \multicolumn{3}{c}{bob and spot} & \multicolumn{3}{c}{can} & \multicolumn{3}{c}{camera} & \multicolumn{3}{c}{mug} \\\cmidrule(lr){3-5}\cmidrule(lr){6-8}\cmidrule(lr){9-11}\cmidrule(lr){12-14}
		& & init & result & w/o toc & init & result & w/o toc & init & result & w/o toc & init & result & w/o toc \\\midrule
\multirow{6}{*}{\rotatebox{90}{CD}} & mean & 0.0216 & \textbf{0.0081} & 0.0157 & 0.0076 & \textbf{0.0019} & 0.0041 & 0.0290 & \textbf{0.0244} & 0.0285 & \textbf{0.0052} & 0.0064 & 0.0190 \\\cmidrule(lr){2-14}
& min & 0.0109 & \textbf{0.0000} & 0.0002 & 0.0001 & \textbf{0.0000} & \textbf{0.0000} & 0.0033 & \textbf{0.0032} & 0.0033 & 0.0006 & \textbf{0.0001} & 0.0002 \\
& Q25 & 0.0182 & \textbf{0.0030} & 0.0044 & 0.0012 & \textbf{0.0008} & 0.0012 & 0.0179 & \textbf{0.0089} & 0.0137 & 0.0010 & \textbf{0.0009} & 0.0010 \\
& median & 0.0219 & \textbf{0.0060} & 0.0091 & 0.0021 & \textbf{0.0012} & 0.0017 & 0.0305 & \textbf{0.0196} & 0.0255 & \textbf{0.0022} & 0.0023 & 0.0022 \\
& Q75 & 0.0255 & \textbf{0.0081} & 0.0199 & 0.0038 & \textbf{0.0021} & 0.0033 & 0.0390 & \textbf{0.0330} & 0.0337 & \textbf{0.0046} & 0.0057 & 0.0064 \\
& max & 0.0330 & \textbf{0.0286} & 0.1052 & 0.0794 & \textbf{0.0148} & 0.0382 & \textbf{0.0671} & 0.1447 & 0.1251 & \textbf{0.0338} & 0.0816 & 0.6794 \\\midrule
\multirow{6}{*}{\rotatebox{90}{obj}} & mean & 0.0199 & \textbf{0.0059} & 0.0130 & 0.0058 & \textbf{0.0020} & 0.0027 & 0.0279 & \textbf{0.0070} & 0.0141 & 0.0087 & \textbf{0.0042} & 0.0063 \\\cmidrule(lr){2-14}
& min & 0.0008 & \textbf{0.0000} & 0.0002 & 0.0001 & \textbf{0.0000} & 0.0001 & 0.0007 & \textbf{0.0001} & 0.0004 & 0.0002 & 0.0001 & \textbf{0.0000} \\
& Q25 & 0.0074 & \textbf{0.0007} & 0.0011 & 0.0009 & \textbf{0.0002} & 0.0004 & 0.0065 & \textbf{0.0010} & 0.0020 & 0.0011 & 0.0005 & \textbf{0.0004} \\
& median & 0.0153 & \textbf{0.0017} & 0.0044 & 0.0034 & \textbf{0.0008} & 0.0011 & 0.0134 & \textbf{0.0031} & 0.0061 & 0.0027 & \textbf{0.0010} & 0.0012 \\
& Q75 & 0.0269 & \textbf{0.0056} & 0.0101 & 0.0064 & \textbf{0.0027} & 0.0032 & 0.0334 & \textbf{0.0090} & 0.0161 & 0.0046 & \textbf{0.0029} & 0.0037 \\
& max & 0.0957 & \textbf{0.0936} & 0.1254 & 0.0397 & \textbf{0.0124} & 0.0333 & 0.2917 & \textbf{0.0438} & 0.1112 & \textbf{0.0633} & 0.0647 & 0.0666 \\\midrule
\multirow{6}{*}{\rotatebox{90}{pos err}} & mean & 0.3409 & \textbf{0.1648} & 0.2356 & 0.1601 & \textbf{0.0962} & 0.1065 & 0.3472 & \textbf{0.2018} & 0.2519 & 0.1799 & \textbf{0.1399} & 0.1404 \\\cmidrule(lr){2-14}
& min & 0.0794 & \textbf{0.0091} & 0.0249 & 0.0125 & \textbf{0.0047} & 0.0147 & 0.0344 & 0.0368 & \textbf{0.0163} & 0.0224 & 0.0145 & \textbf{0.0127} \\
& Q25 & 0.2188 & \textbf{0.0617} & 0.0669 & 0.0721 & \textbf{0.0435} & 0.0447 & 0.2142 & \textbf{0.0929} & 0.1109 & 0.0746 & 0.0558 & \textbf{0.0415} \\
& median & 0.3212 & \textbf{0.1215} & 0.1709 & 0.1456 & \textbf{0.0726} & 0.0798 & 0.3336 & \textbf{0.1531} & 0.2038 & 0.1351 & 0.0908 & \textbf{0.0890} \\
& Q75 & 0.4080 & \textbf{0.2327} & 0.2714 & 0.2291 & \textbf{0.1430} & 0.1556 & 0.4442 & \textbf{0.2838} & 0.3321 & 0.1945 & \textbf{0.1721} & 0.1756 \\
& max & 0.9399 & \textbf{0.9301} & 1.0076 & 0.5049 & \textbf{0.3238} & 0.3469 & 1.0767 & \textbf{0.6934} & 0.8276 & 0.8014 & 0.8032 & \textbf{0.7997} \\\midrule
	\end{tabular}
	\caption{Numerical results for trajectory fitting with learned shape spaces. ``CD'' denotes the chamfer distance of the object to the target shape and ``obj'' denotes the optimization objective (position trajectory error). The time of contact differential shows improvements in accuracy in most cases over not using the differential. For the mug objects, the median accuracy is similar for using or not using the time of contact differential, while for chamfer distance using time of contact differential makes the fitting more robust (see max measure).}
	\label{tab:results_trajectory_shapespace}
\end{table*}

\begin{table*}
	\centering
	\footnotesize
	\setlength{\tabcolsep}{3.5pt}
	\begin{tabular}{cccccccccccccccc}\toprule
		\multicolumn{2}{c}{\multirow{2}{*}{Error}} & \multicolumn{2}{c}{box} & \multicolumn{2}{c}{sphere} & \multicolumn{2}{c}{cylinder} & \multicolumn{2}{c}{bob and spot} & \multicolumn{2}{c}{can} & \multicolumn{2}{c}{camera} & \multicolumn{2}{c}{mug} \\\cmidrule(lr){3-4}\cmidrule(lr){5-6}\cmidrule(lr){7-8}\cmidrule(lr){9-10}\cmidrule(lr){11-12}\cmidrule(lr){13-14}\cmidrule(lr){15-16}
		& & init & result & init & result & init & result & init & result & init & result & init & result & init & result \\\midrule
\multirow{6}{*}{\rotatebox{90}{CD}} & mean & 1.4e-01 & 2.6e-02 & 6.3e-01 & 4.0e-02 & 3.1e-01 & 6.2e-02 & 2.2e-02 & 1.7e-03 & 7.9e-03 & 5.6e-04 & 3.0e-02 & 1.4e-02 & 5.4e-03 & 2.9e-03 \\\cmidrule(lr){2-16}
& min & 5.0e-04 & 1.0e-05 & 9.3e-04 & 2.3e-08 & 1.8e-03 & 1.2e-05 & 1.3e-02 & 1.3e-06 & 2.3e-04 & 9.1e-06 & 3.1e-03 & 7.5e-04 & 5.4e-04 & 6.5e-05 \\
& Q25 & 6.3e-02 & 2.8e-03 & 7.3e-02 & 8.2e-06 & 1.1e-01 & 5.5e-04 & 1.9e-02 & 8.2e-06 & 1.1e-03 & 1.8e-04 & 1.9e-02 & 3.7e-03 & 1.0e-03 & 4.1e-04 \\
& median & 1.2e-01 & 1.1e-02 & 3.9e-01 & 1.5e-04 & 2.4e-01 & 8.5e-03 & 2.1e-02 & 1.3e-05 & 2.5e-03 & 5.3e-04 & 3.2e-02 & 8.3e-03 & 2.3e-03 & 1.3e-03 \\
& Q75 & 1.9e-01 & 2.7e-02 & 8.0e-01 & 1.7e-03 & 4.2e-01 & 9.2e-02 & 2.4e-02 & 3.3e-05 & 4.0e-03 & 9.0e-04 & 4.0e-02 & 2.0e-02 & 3.9e-03 & 2.7e-03 \\
& max & 4.5e-01 & 1.9e-01 & 2.9e+00 & 7.8e-01 & 1.3e+00 & 3.7e-01 & 3.5e-02 & 4.9e-02 & 8.4e-02 & 1.3e-03 & 6.7e-02 & 1.1e-01 & 3.9e-02 & 4.5e-02 \\\midrule
\multirow{6}{*}{\rotatebox{90}{obj}} & mean & 1.1e+00 & 5.2e-03 & 7.6e-01 & 7.8e-03 & 2.9e-01 & 2.1e-03 & 3.6e-01 & 4.2e-03 & 4.1e-01 & 7.3e-04 & 4.3e-01 & 4.0e-03 & 1.7e-02 & 4.0e-04 \\\cmidrule(lr){2-16}
& min & 8.0e-05 & 1.2e-05 & 6.1e-05 & 2.3e-08 & 2.4e-04 & 8.4e-08 & 1.4e-02 & 9.9e-07 & 1.6e-04 & 2.7e-06 & 5.1e-03 & 8.8e-06 & 7.6e-04 & 4.0e-06 \\
& Q25 & 1.1e-01 & 3.8e-04 & 1.7e-02 & 4.6e-06 & 1.2e-02 & 4.2e-05 & 1.6e-01 & 4.6e-06 & 3.1e-03 & 3.5e-05 & 3.2e-02 & 4.6e-04 & 2.0e-03 & 6.5e-05 \\
& median & 4.0e-01 & 1.3e-03 & 1.1e-01 & 1.2e-04 & 7.0e-02 & 2.3e-04 & 2.5e-01 & 3.4e-05 & 1.4e-02 & 1.2e-04 & 1.3e-01 & 1.2e-03 & 4.3e-03 & 1.1e-04 \\
& Q75 & 1.2e+00 & 3.4e-03 & 8.2e-01 & 6.6e-04 & 4.0e-01 & 8.1e-04 & 4.0e-01 & 1.2e-04 & 6.2e-02 & 1.1e-03 & 5.8e-01 & 2.9e-03 & 1.8e-02 & 3.5e-04 \\
& max & 7.5e+00 & 1.2e-01 & 6.7e+00 & 1.6e-01 & 2.6e+00 & 2.3e-02 & 1.3e+00 & 2.0e-01 & 1.0e+01 & 4.9e-03 & 2.3e+00 & 3.6e-02 & 1.5e-01 & 3.5e-03 \\\bottomrule
	\end{tabular}
	\caption{Numerical results for shape fitting by inertia.  ``CD'' denotes the chamfer distance of the object to the target shape and ``obj'' denotes the optimization objective (rotational velocity error).}
	\label{tab:results_inertia}
\end{table*}

\begin{table}
	\centering
	\footnotesize
	\setlength{\tabcolsep}{3pt}
	\begin{tabular}{cccccccc}\toprule
		 \multicolumn{2}{c}{\multirow{2}{*}{Error}}& \multicolumn{2}{c}{mass} & \multicolumn{2}{c}{force} & \multicolumn{2}{c}{friction}\\\cmidrule(lr){3-4}\cmidrule(lr){5-6}\cmidrule(lr){7-8}
		 & & init & result & init & result & init & result \\\midrule
\multirow{6}{*}{\rotatebox{90}{\parbox{3em}{target error}}} & mean & 6.6e-02 & 2.8e-03 & 1.6e+00 & 1.5e-02 & 8.3e-02 & 1.3e-04 \\\cmidrule(lr){2-8}
& min & 6.5e-04 & 1.4e-07 & 5.6e-02 & 5.3e-04 & 1.1e-03 & 4.0e-06 \\
& Q25 & 3.3e-02 & 4.1e-05 & 1.3e+00 & 1.0e-03 & 4.1e-02 & 1.1e-04 \\
& median & 5.6e-02 & 1.1e-04 & 1.6e+00 & 1.4e-03 & 7.7e-02 & 1.2e-04 \\
& Q75 & 1.0e-01 & 3.4e-04 & 2.0e+00 & 1.6e-03 & 1.2e-01 & 1.5e-04 \\
& max & 1.7e-01 & 1.3e-01 & 2.9e+00 & 3.4e-01 & 2.3e-01 & 3.0e-04 \\\midrule
\multirow{6}{*}{\rotatebox{90}{\parbox{3em}{objective}}} & mean & 3.3e-01 & 3.7e-02 & 5.7e+00 & 1.8e-05 & 1.9e+00 & 5.0e-06 \\\cmidrule(lr){2-8}
& min & 8.5e-06 & 1.2e-12 & 5.6e-03 & 6.9e-07 & 2.1e-04 & 4.7e-07 \\
& Q25 & 5.6e-02 & 1.2e-07 & 3.3e+00 & 2.4e-06 & 3.5e-01 & 2.3e-06 \\
& median & 1.5e-01 & 1.1e-06 & 5.6e+00 & 3.8e-06 & 1.2e+00 & 2.9e-06 \\
& Q75 & 3.7e-01 & 4.1e-06 & 7.8e+00 & 4.9e-06 & 3.1e+00 & 4.2e-06 \\
& max & 2.3e+00 & 1.8e+00 & 1.8e+01 & 3.4e-04 & 1.0e+01 & 4.4e-05 \\\bottomrule
	\end{tabular}
	\caption{Numerical results for system identification. The target error is the difference to the target value and ``obj'' denotes the optimization objective (position trajectory error).}
	\label{tab:results_systemid}
\end{table}

For completeness, we provide other scaled versions of the plots in Figs.~\ref{fig:exp_trajectory_shapespace}, \ref{fig:exp_shapefrominertia} and \ref{fig:exp_systemid} which show outliers of the initialization and zoomed in versions of the optimization results.
In Tables~\ref{tab:results_trajectory_shapespace}, \ref{tab:results_inertia} and \ref{tab:results_systemid} we provide the numeric results for the plots.
Table \ref{tab:results_trajectory_shapespace} additionally shows an ablation study for not using the time of contact differential for trajectory-based optimization of DeepSDF shape space objects.

\subsubsection{Shape identification (sec.~5.1 in main paper)}
\paragraph{Bouncing objects}
In all the bouncing object experiments, the friction coefficient is set to 0.25 and the restitution to 0.5.
For numerical evaluation, we take 50 runs in each setting (with/without gravity and with/without time of contact differential).

\begin{figure*}
	\centering
	\begin{subfigure}{.49\linewidth}
		\centering
		\includegraphics[width=.3\linewidth]{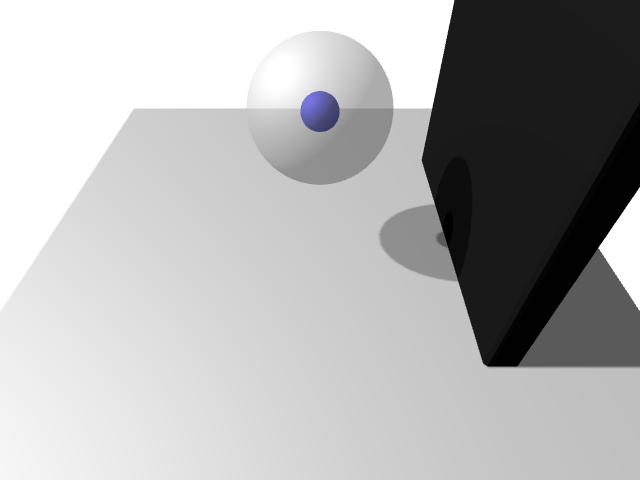}
		\includegraphics[width=.3\linewidth]{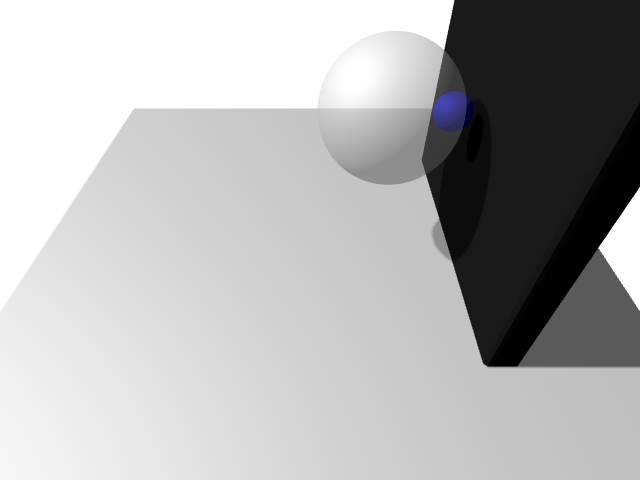}
		\includegraphics[width=.3\linewidth]{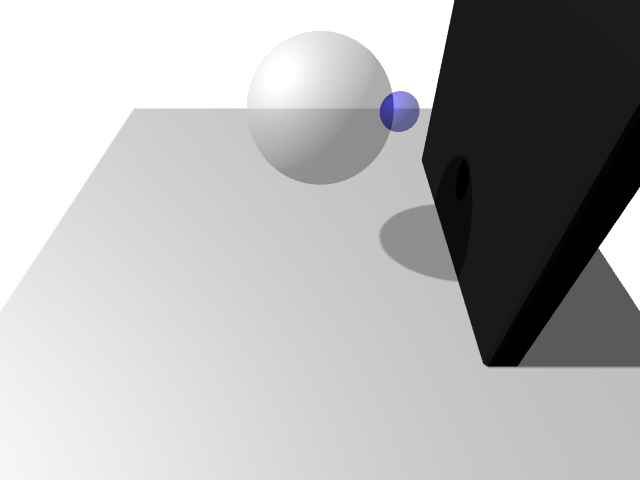} \\
		\includegraphics[width=.3\linewidth]{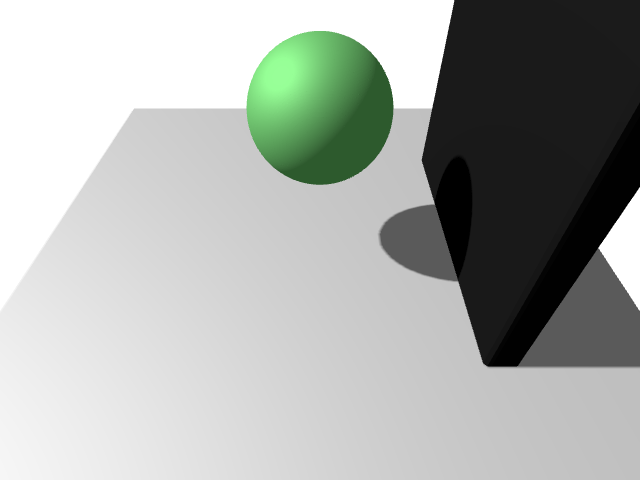}
		\includegraphics[width=.3\linewidth]{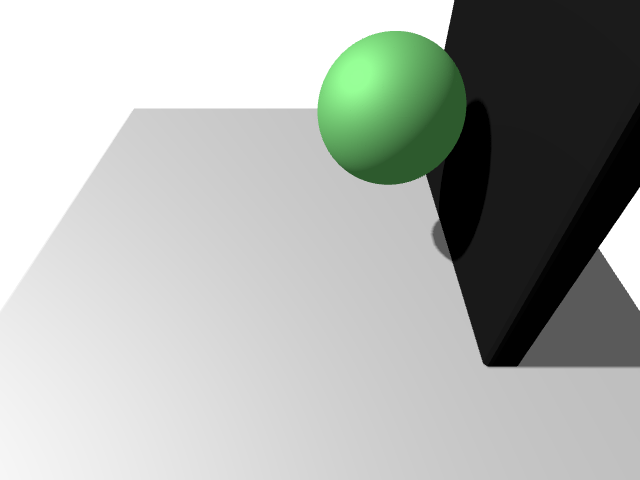}
		\includegraphics[width=.3\linewidth]{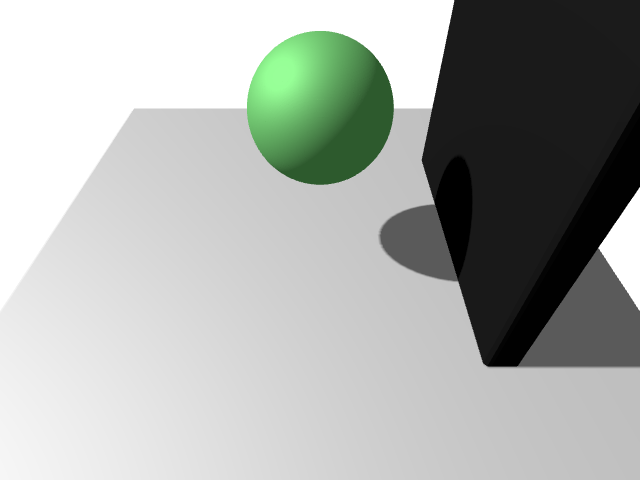}
	\end{subfigure}
	\begin{subfigure}{.49\linewidth}
		\centering
		\includegraphics[width=.3\linewidth]{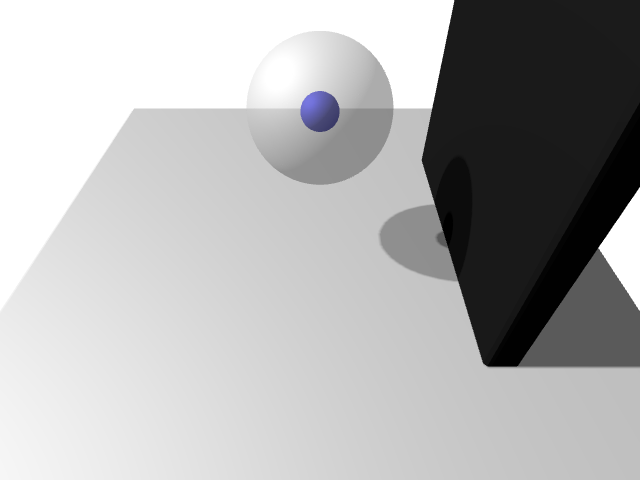}
		\includegraphics[width=.3\linewidth]{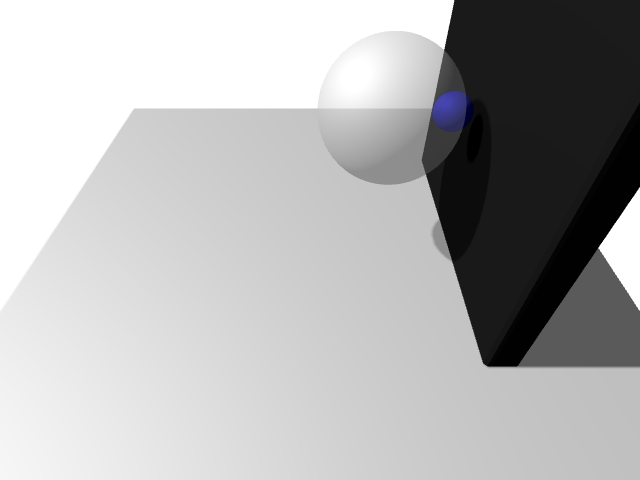}
		\includegraphics[width=.3\linewidth]{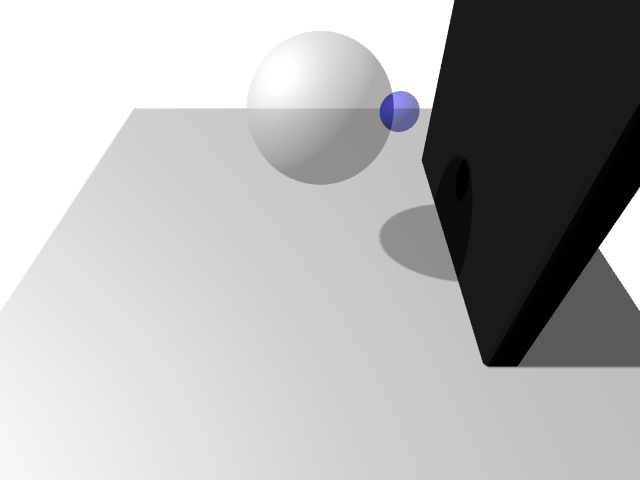} \\
		\includegraphics[width=.3\linewidth]{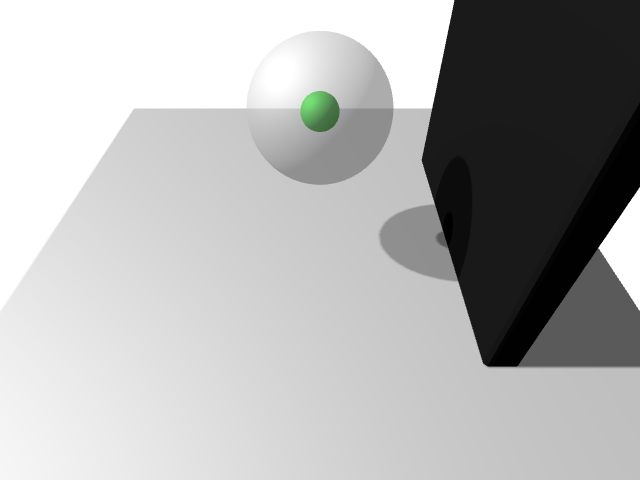}
		\includegraphics[width=.3\linewidth]{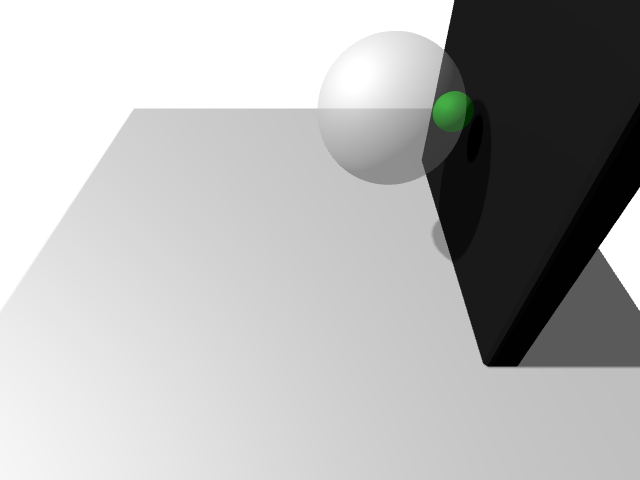}
		\includegraphics[width=.3\linewidth]{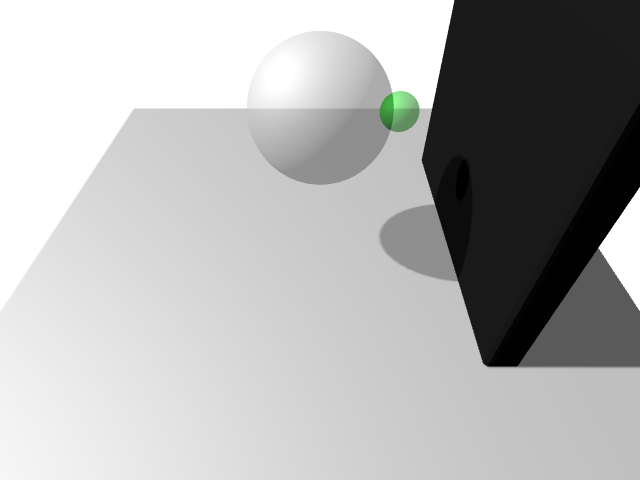}
	\end{subfigure}\\[1em]
	\begin{subfigure}{.49\linewidth}
		\centering
		\includegraphics[width=.3\linewidth]{figures/renderings/trajectory_sphere_gr_on_toc_on/overlay_start_0}
		\includegraphics[width=.3\linewidth]{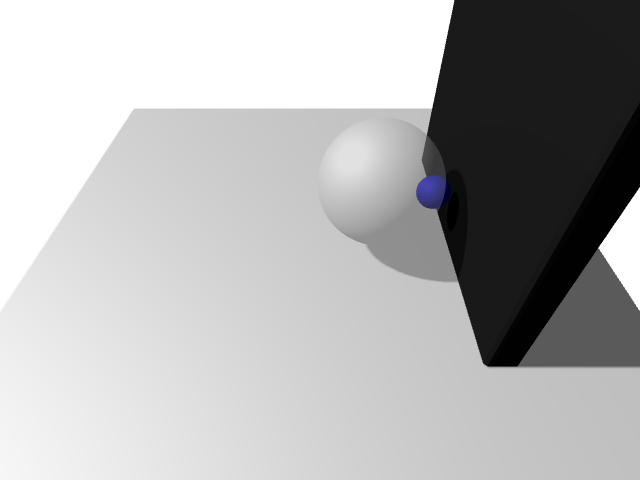}
		\includegraphics[width=.3\linewidth]{figures/renderings/trajectory_sphere_gr_on_toc_on/overlay_start_45} \\
		\includegraphics[width=.3\linewidth]{figures/renderings/trajectory_sphere_gr_on_toc_on/overlay_final_0}
		\includegraphics[width=.3\linewidth]{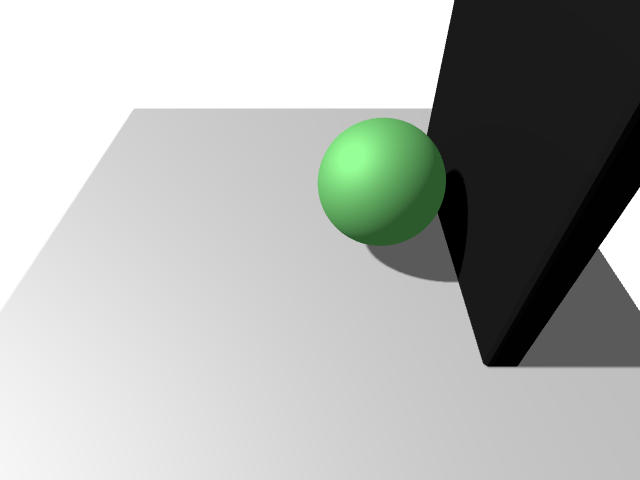}
		\includegraphics[width=.3\linewidth]{figures/renderings/trajectory_sphere_gr_on_toc_on/overlay_final_45}
	\end{subfigure}
	\begin{subfigure}{.49\linewidth}
		\centering
		\includegraphics[width=.3\linewidth]{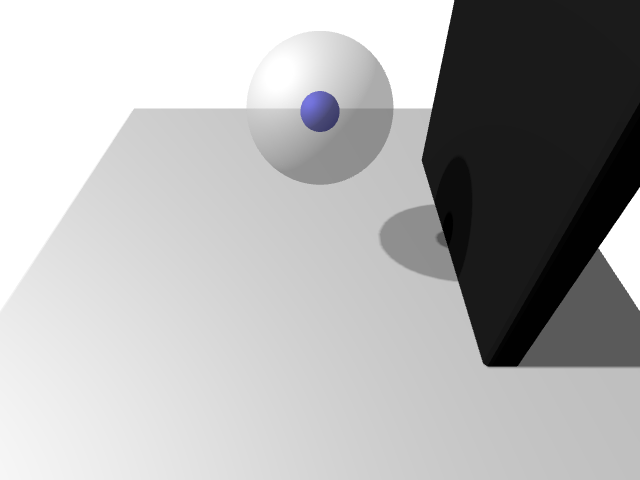}
		\includegraphics[width=.3\linewidth]{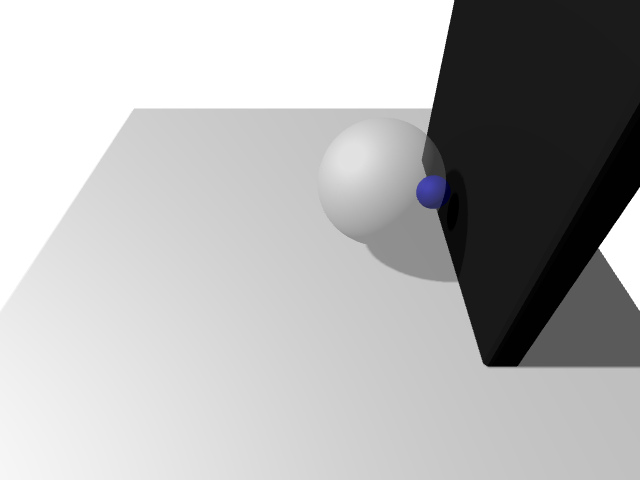}
		\includegraphics[width=.3\linewidth]{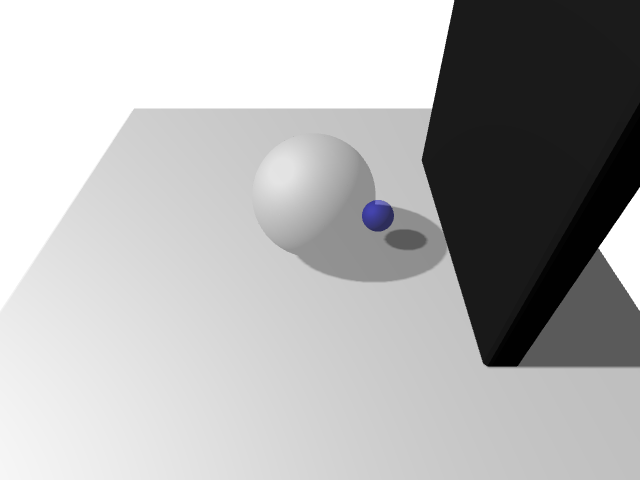} \\
		\includegraphics[width=.3\linewidth]{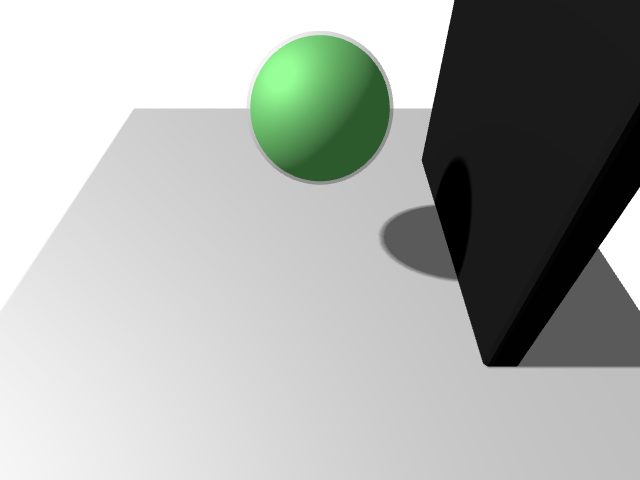}
		\includegraphics[width=.3\linewidth]{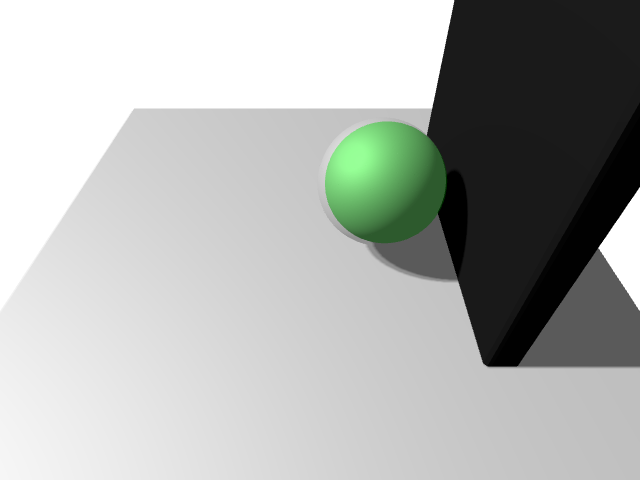}
		\includegraphics[width=.3\linewidth]{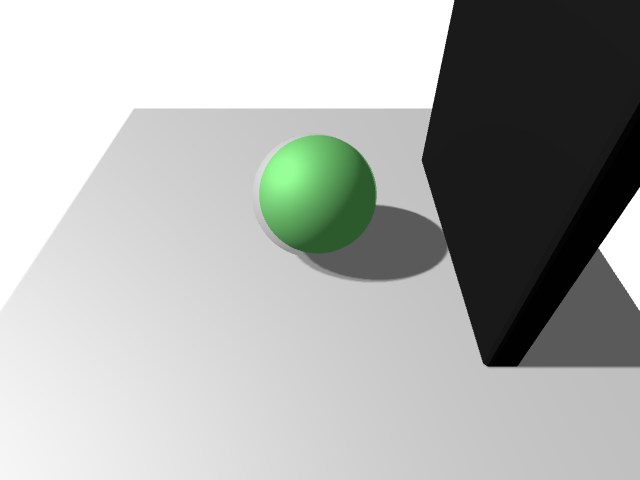}
	\end{subfigure}
	\caption{Trajectory fitting for a sphere. From left to right, each group shows 3 frames from the start, middle and end of the trajectory. We optimize the radius of a sphere from an initial value (blue), by comparing the simulated trajectory to that of a target sphere (gray, overlaid) and arrive at the result in green (also overlaid with gt in gray). Our formulation (left) works even in the case of a head on-collision without gravity (top two rows), while the engine without the time-of-contact differential fails. In the setting with gravity enabled (bottom two rows), our formulation achieves more accurate results.}
	\label{fig:trajectory_sphere}
\end{figure*}

Fig.~\ref{fig:exp_trajectory_shapespace} shows different scaled version of the box plots shown in the main paper.
Table~\ref{tab:results_trajectory_shapespace} provides numeric results for the plots and the ablation study for not using the time of contact differential for trajectory-based optimization of DeepSDF shape space objects.
One of the mug examples diverged and lead to an invalid state for the physics engine without the time of contact differential.
It is thus excluded from the statistics in the w/o toc column and the overall average reported in the main paper.
Fig.~\ref{fig:trajectory_sphere} shows qualitative results for the sphere bouncing scenario.

In the sphere bouncing scenarios, the optimization objective is the mean squared position error over the entire trajectory in all timesteps $t \in T$:
\begin{equation}
	E(\theta) = \frac{1}{\|T\|}\sum_{t\in T} \|\mathbf{t}(\theta)^t_e - \mathbf{t}(\theta)^t_g\|^2
	\label{eq:trajectory_error}
\end{equation}
where $\mathbf{t}(\theta)^t_e$ and $\mathbf{t}(\theta)^t_g$ denote the center location of the estimated and goal spheres at time $t$, respectively.

In the setting without gravity (top two rows in Fig.~\ref{fig:trajectory_sphere}), the sphere bounces straight back and touches the wall earlier or later depending on the radius.
With gravity, the sphere bounces off the floor after touching the wall and gets rotation motion due to friction from the first bounce.
This makes the trajectory more complex, and we found it beneficial to optimize the trajectory in chunks including each contact separately by detaching poses
and velocities before the second bounce to avoid exploding gradients that might build up by recursively computing gradients through this complex trajectory.

\begin{figure*}
	\centering
	\begin{subfigure}{.49\linewidth}
		\includegraphics[width=.3\linewidth]{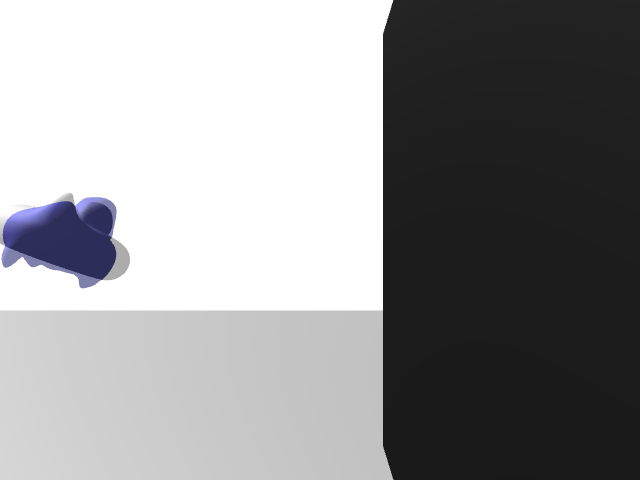}
		\includegraphics[width=.3\linewidth]{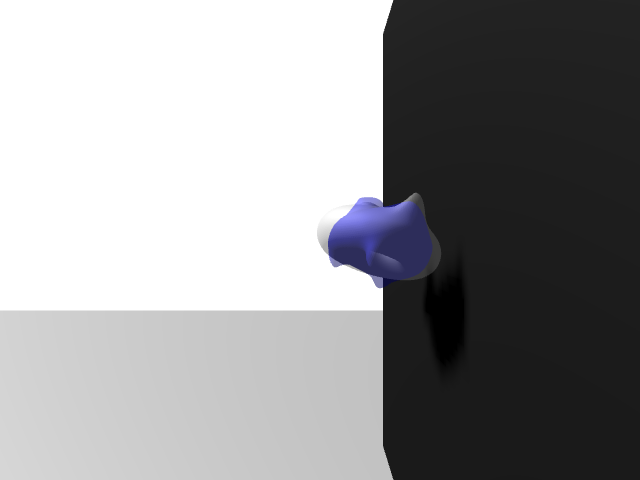}
		\includegraphics[width=.3\linewidth]{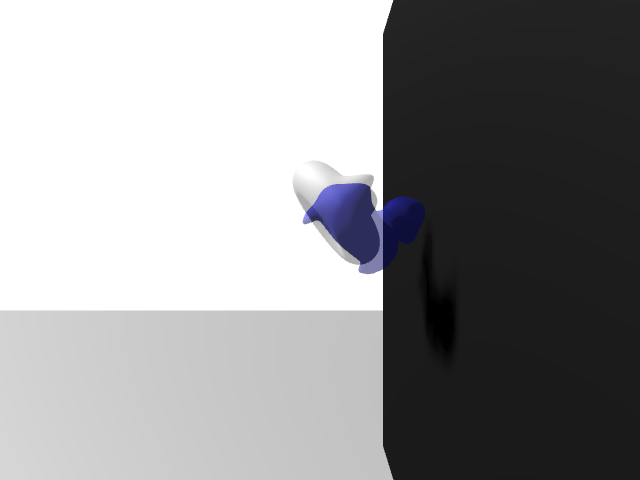}\\
		\includegraphics[width=.3\linewidth]{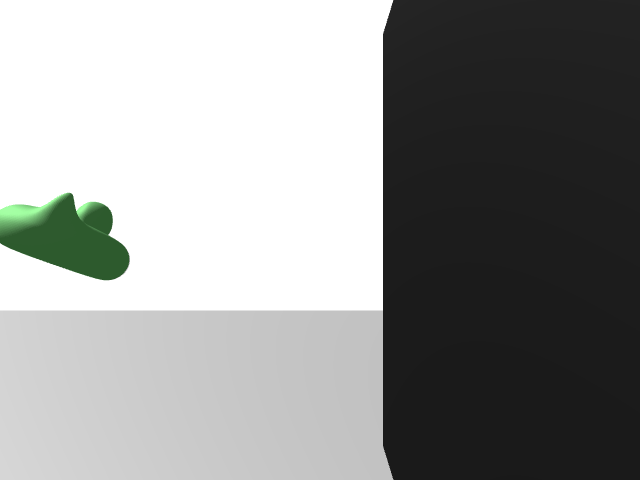}
		\includegraphics[width=.3\linewidth]{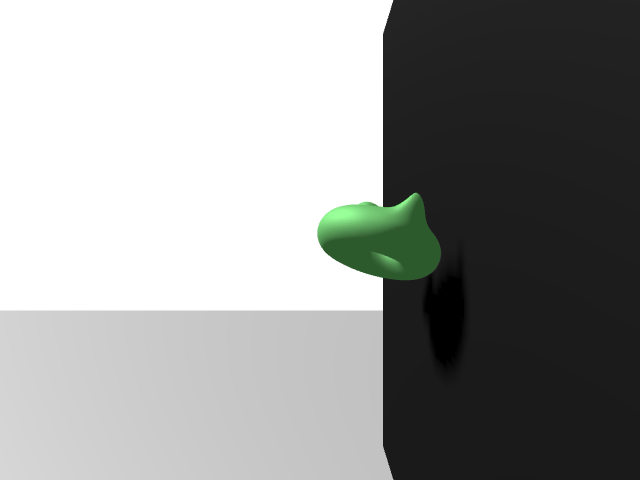}
		\includegraphics[width=.3\linewidth]{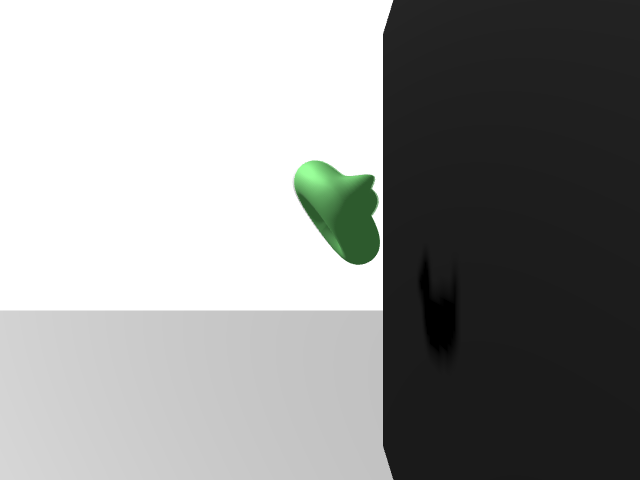}
	\end{subfigure}
	\begin{subfigure}{.49\linewidth}
		\includegraphics[width=.3\linewidth]{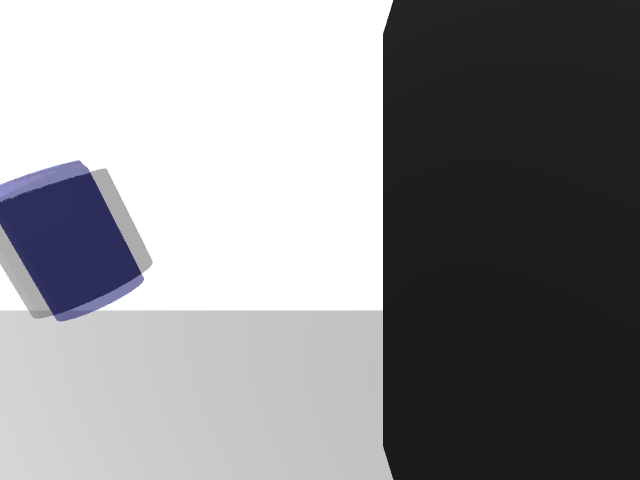}
		\includegraphics[width=.3\linewidth]{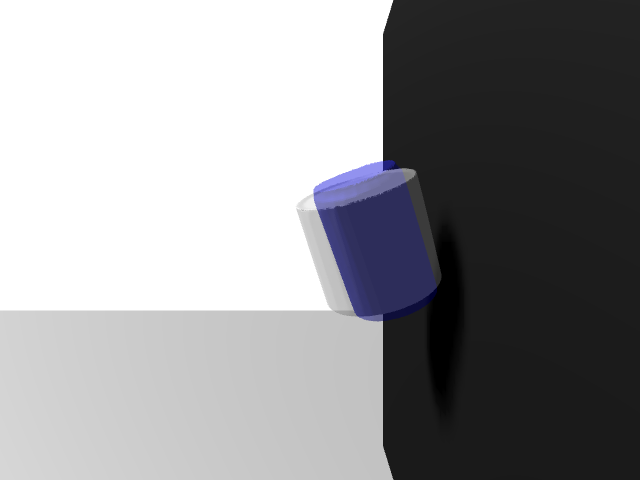}
		\includegraphics[width=.3\linewidth]{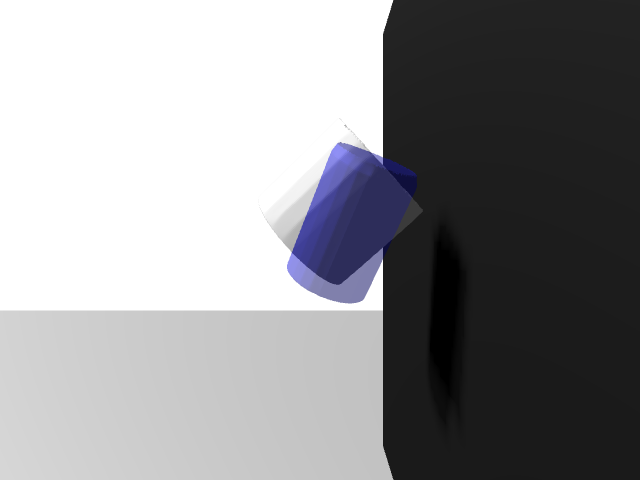}\\
		\includegraphics[width=.3\linewidth]{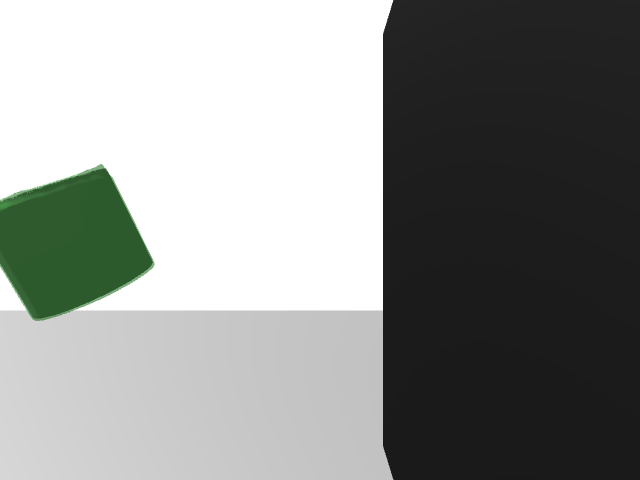}
		\includegraphics[width=.3\linewidth]{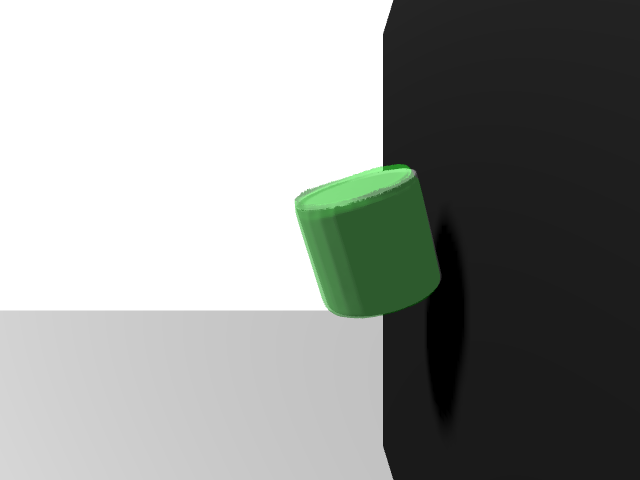}
		\includegraphics[width=.3\linewidth]{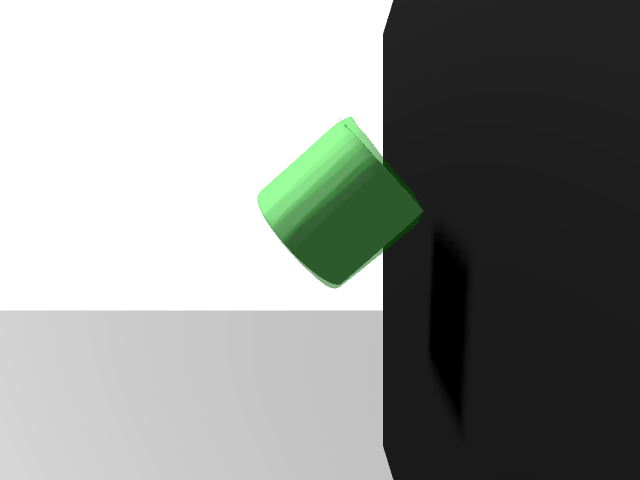}
	\end{subfigure}\\[1em]
	\begin{subfigure}{.49\linewidth}
		\includegraphics[width=.3\linewidth]{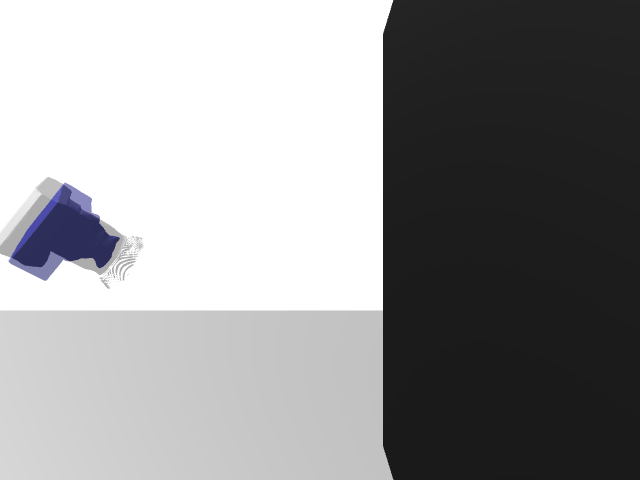}
		\includegraphics[width=.3\linewidth]{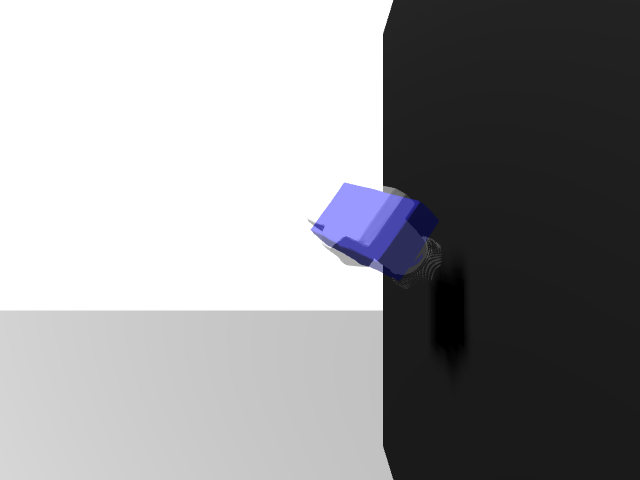}
		\includegraphics[width=.3\linewidth]{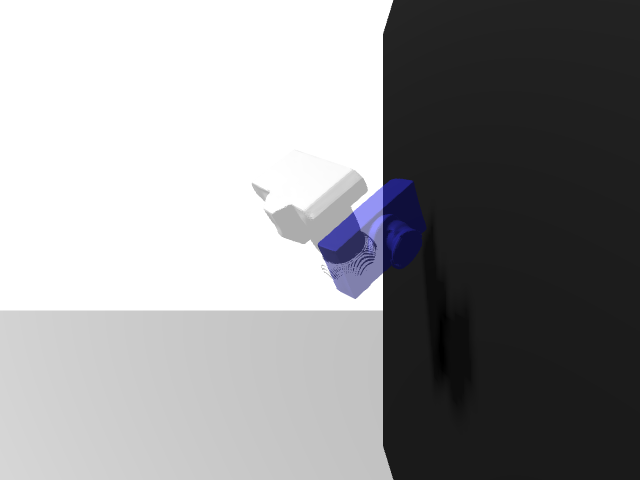}\\
		\includegraphics[width=.3\linewidth]{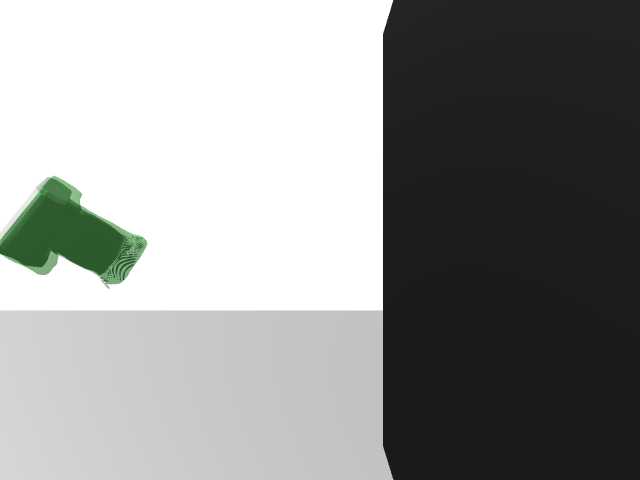}
		\includegraphics[width=.3\linewidth]{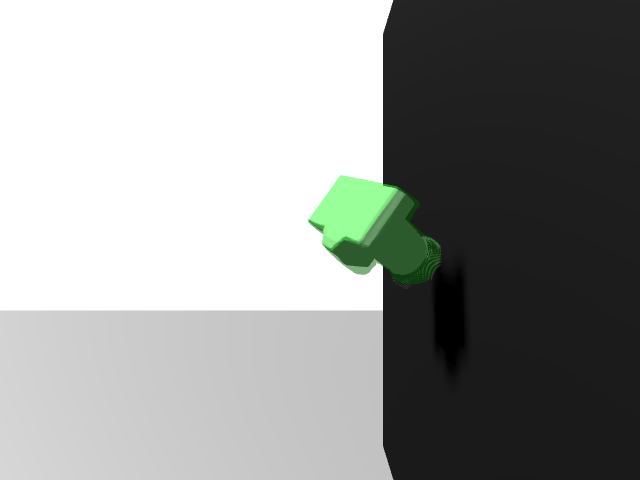}
		\includegraphics[width=.3\linewidth]{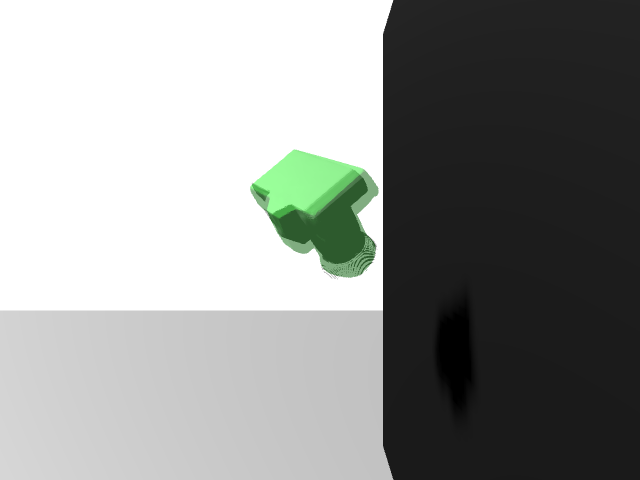}
	\end{subfigure}
	\begin{subfigure}{.49\linewidth}
		\includegraphics[width=.3\linewidth]{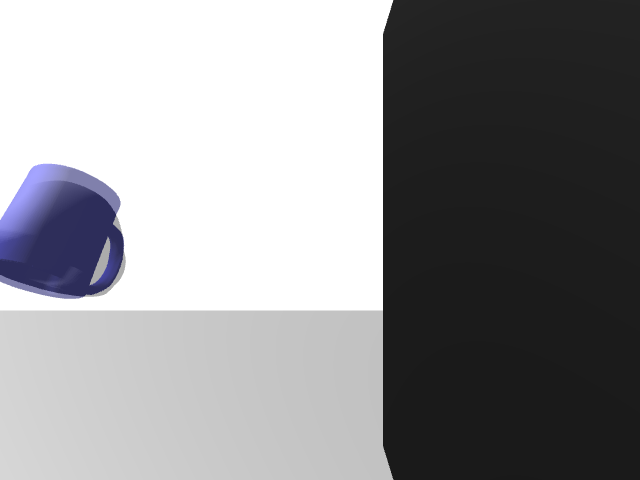}
		\includegraphics[width=.3\linewidth]{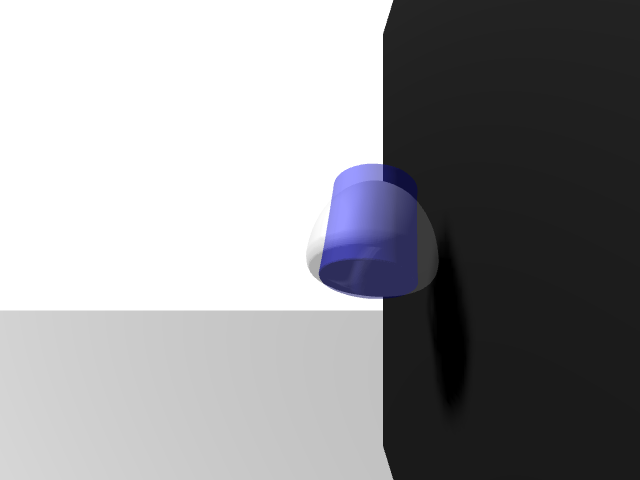}
		\includegraphics[width=.3\linewidth]{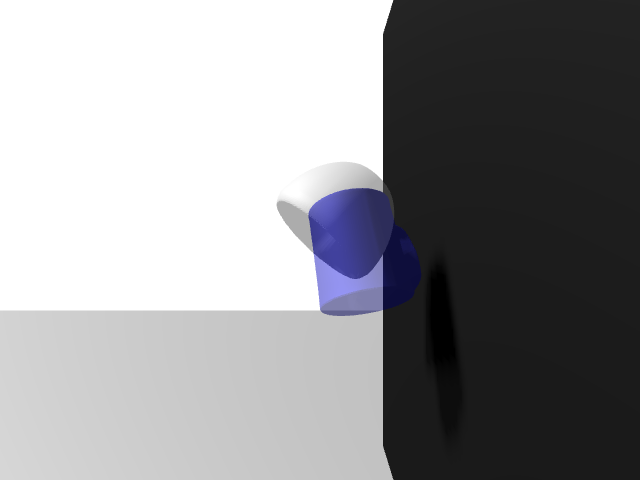}\\
		\includegraphics[width=.3\linewidth]{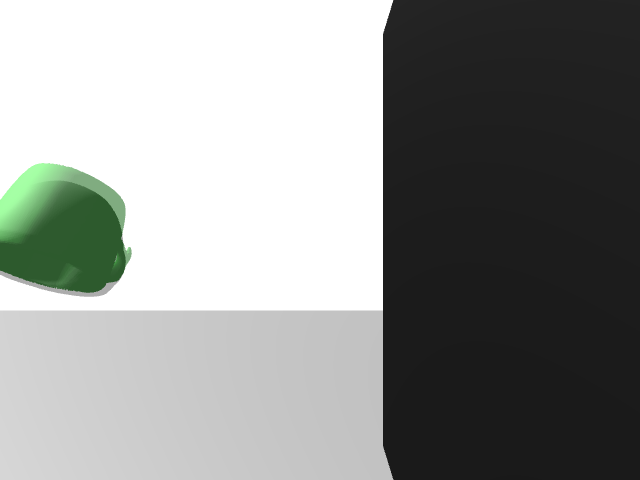}
		\includegraphics[width=.3\linewidth]{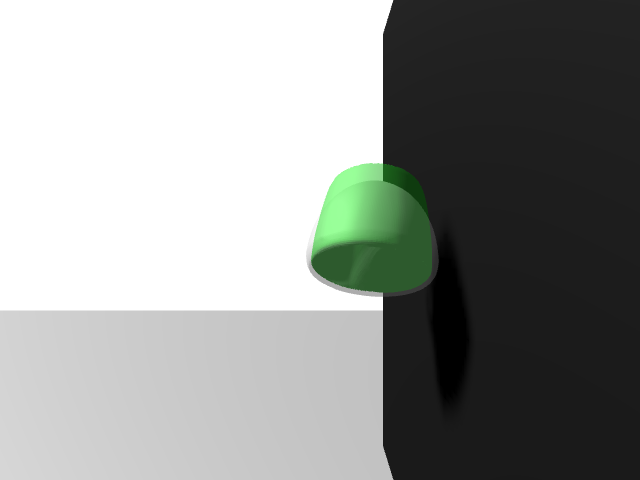}
		\includegraphics[width=.3\linewidth]{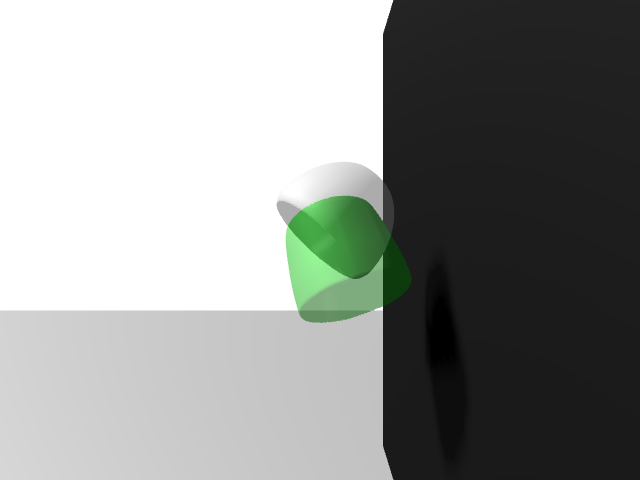}
	\end{subfigure}
	\caption{Trajectory fitting for learned shape spaces. From left to right, each group shows 3 frames from the start, middle and end of the trajectory. Initializations (blue) and results (green) overlaid with targets in gray. Each group shows one of the 4 learned shape spaces: bob and spot (top left), can (top right), camera (bottom left) and mug (bottom right). In most cases the estimated shapes are very accurate, except for the example shown for mugs.}
	\label{fig:trajectory_shapespace}
\end{figure*}

In the setting without gravity for learned shape spaces, there is always just a single bounce in this setting, so the simulated trajectory is shortened to 1s.
Due to the more complex shapes, even the single bounce can generate complex trajectories after it.
The optimization objective is the same as in eq.~\eqref{eq:trajectory_error}, plus the squared $\ell^2$ norm of the latent code with weight 1e-4 as a regularizer.
Our approach manages to recover the shape well in most cases.
Supplemental qualitative example results are shown in Fig.~\ref{fig:trajectory_shapespace}.

\paragraph{Shape from inertia}

\begin{figure*}
	\centering
	\begin{subfigure}{.33\linewidth}
		\centering
		\includegraphics[width=.49\linewidth]{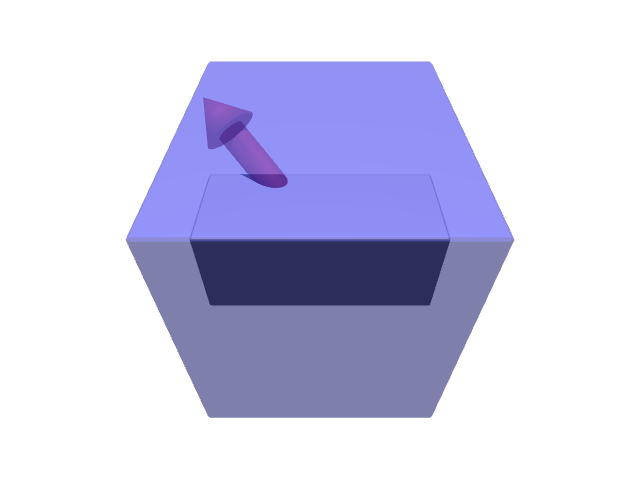}
		\includegraphics[width=.49\linewidth]{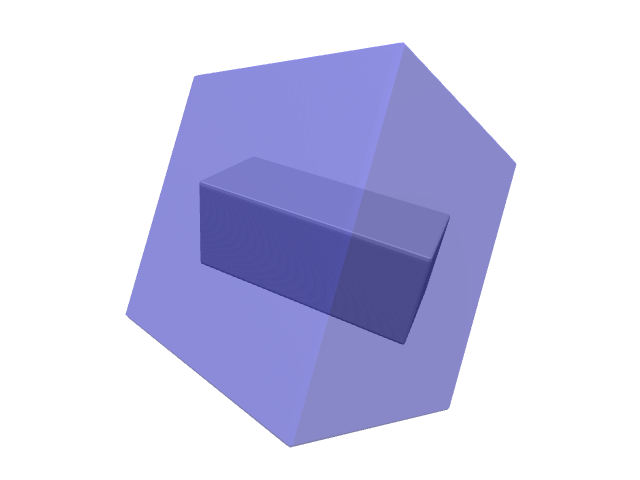}\\
		\includegraphics[width=.49\linewidth]{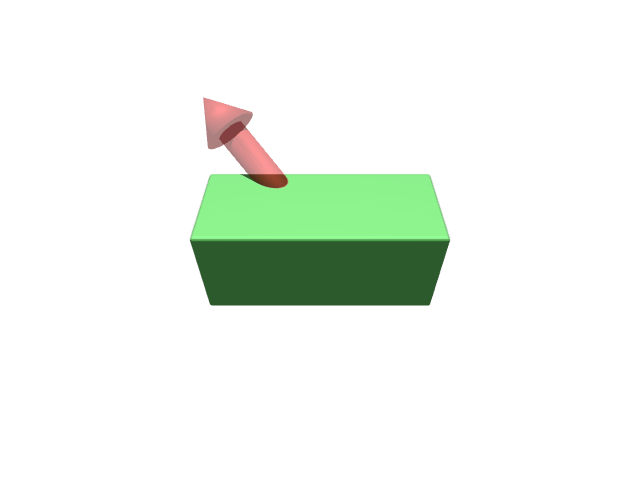}
		\includegraphics[width=.49\linewidth]{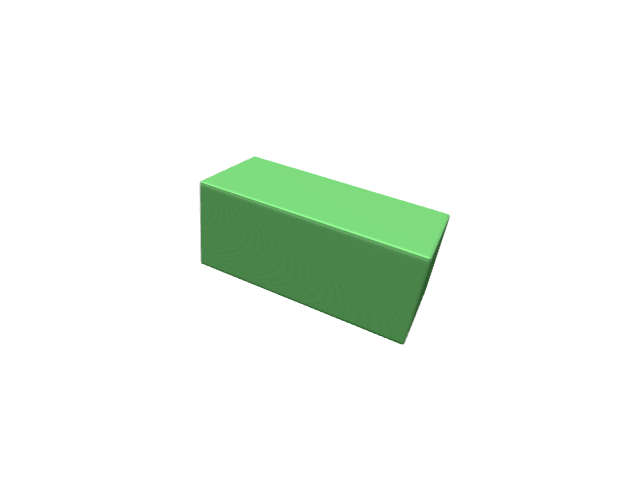}
	\end{subfigure}
	\begin{subfigure}{.33\linewidth}
		\centering
		\includegraphics[width=.49\linewidth]{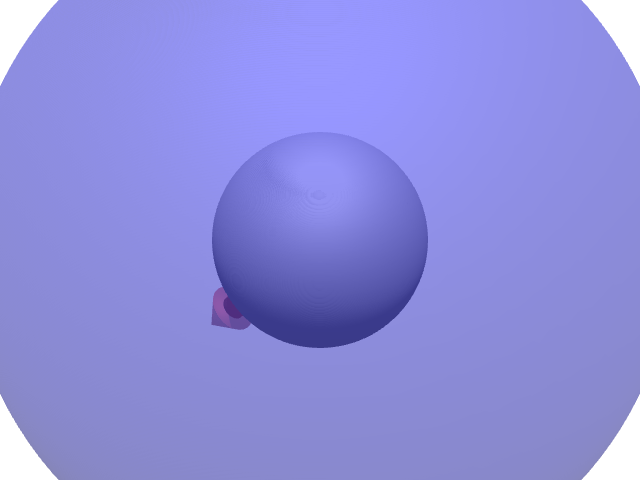}	\includegraphics[width=.49\linewidth]{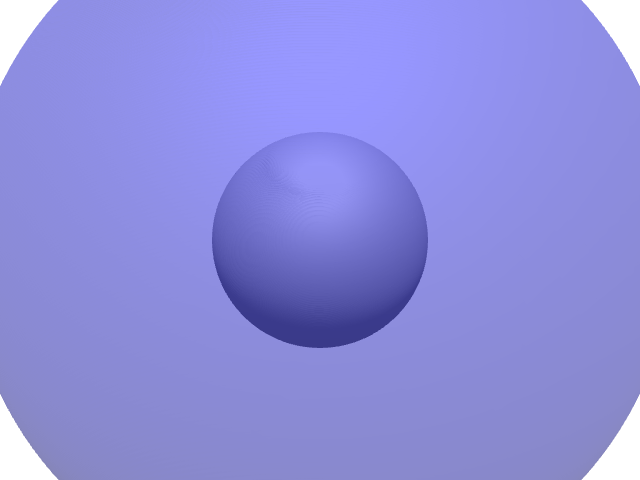}\\
		\includegraphics[width=.49\linewidth]{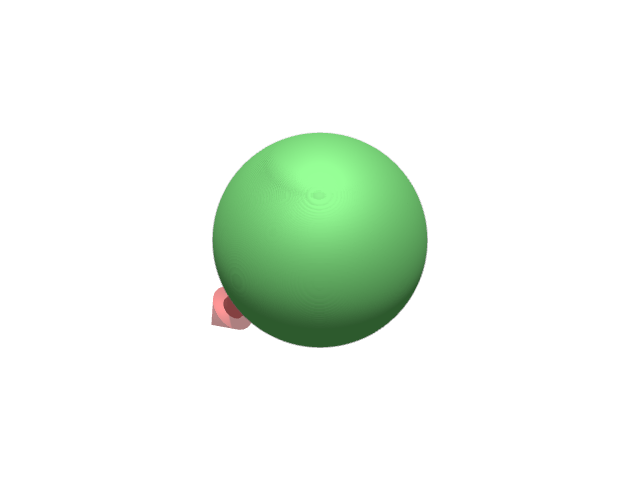}	\includegraphics[width=.49\linewidth]{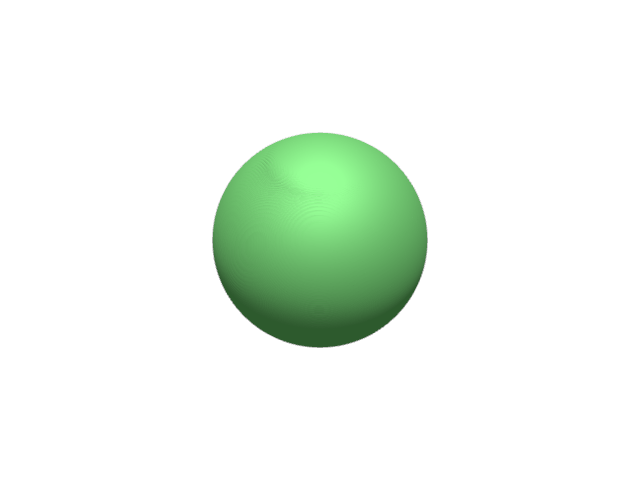}
	\end{subfigure}
	\begin{subfigure}{.33\linewidth}
		\centering
		\includegraphics[width=.49\linewidth]{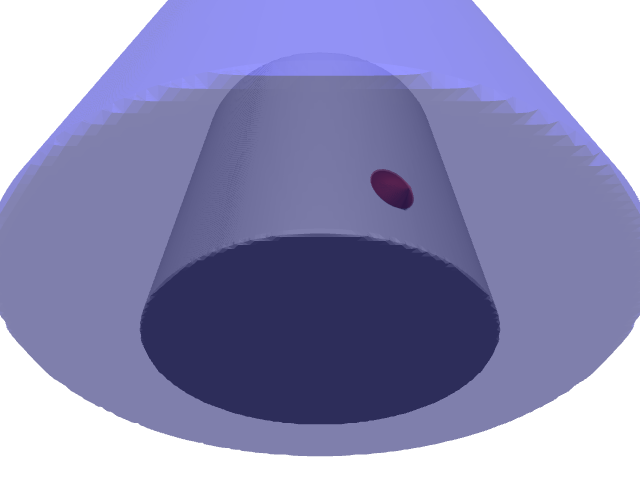}	\includegraphics[width=.49\linewidth]{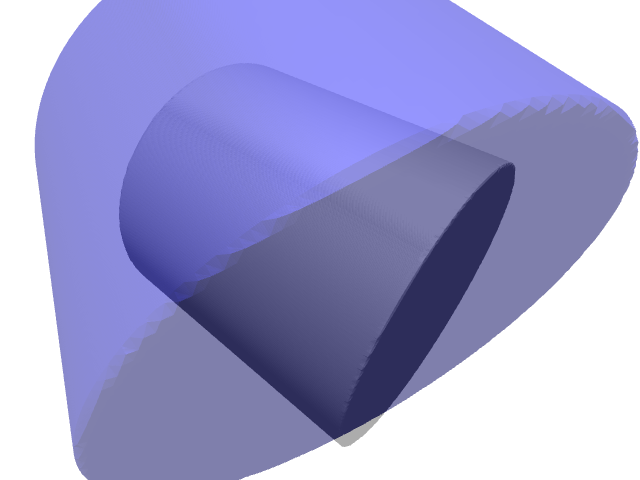}\\
		\includegraphics[width=.49\linewidth]{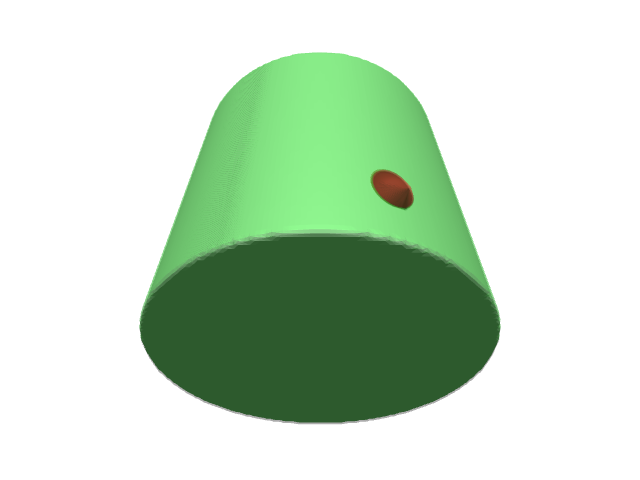}	\includegraphics[width=.49\linewidth]{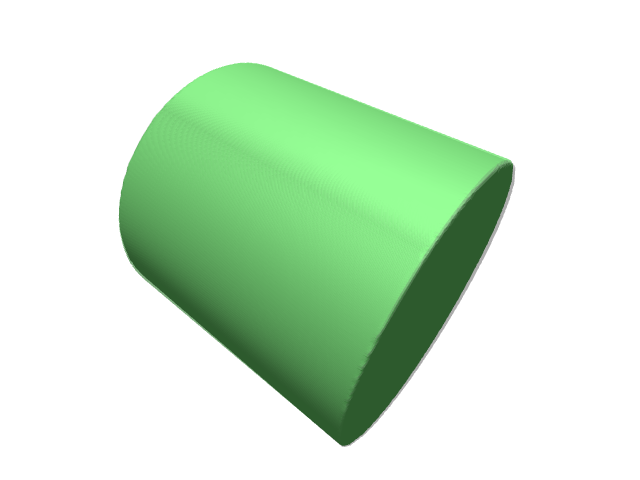}
	\end{subfigure}\\[1em]
	\begin{subfigure}{.24\linewidth}
		\centering
		\includegraphics[width=.49\linewidth]{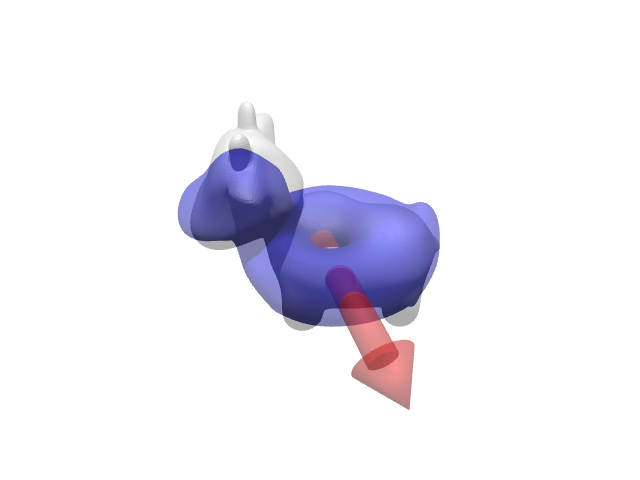}
		\includegraphics[width=.49\linewidth]{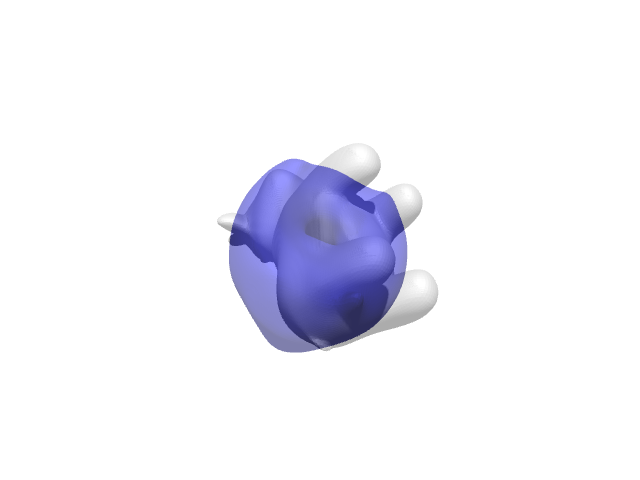}\\
		\includegraphics[width=.49\linewidth]{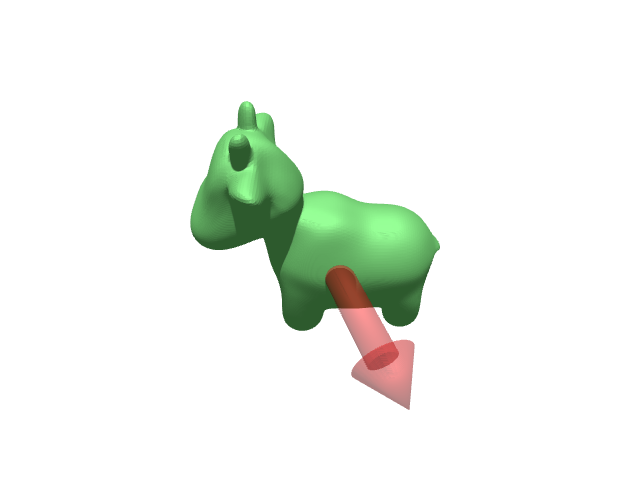}
		\includegraphics[width=.49\linewidth]{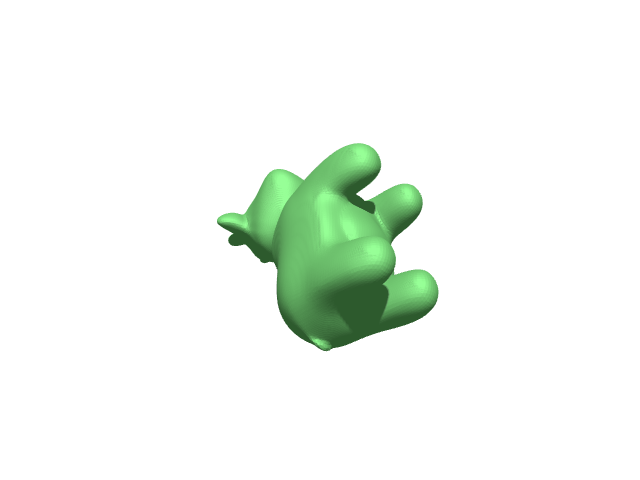}
	\end{subfigure}
	\begin{subfigure}{.24\linewidth}
		\centering
		\includegraphics[width=.49\linewidth]{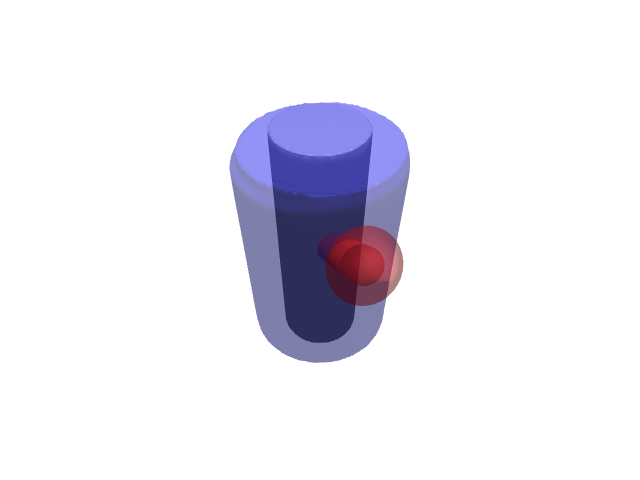}
		\includegraphics[width=.49\linewidth]{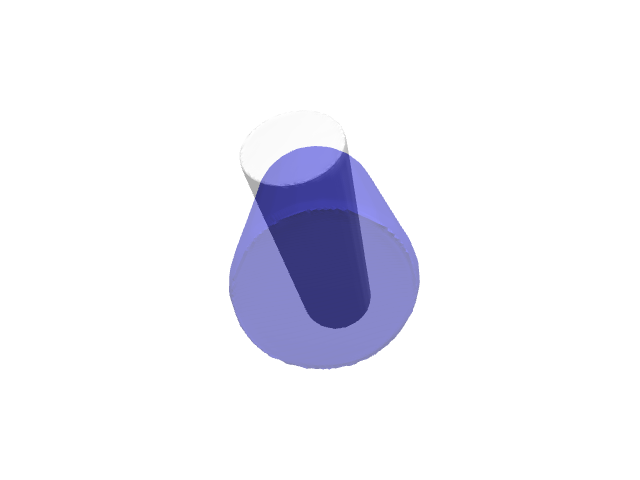}\\
		\includegraphics[width=.49\linewidth]{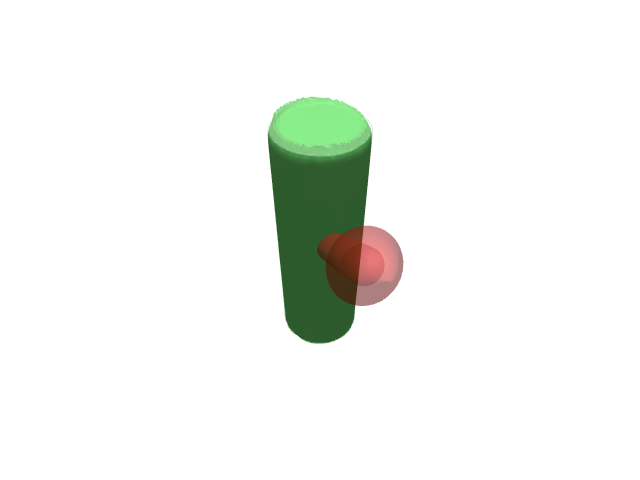}
		\includegraphics[width=.49\linewidth]{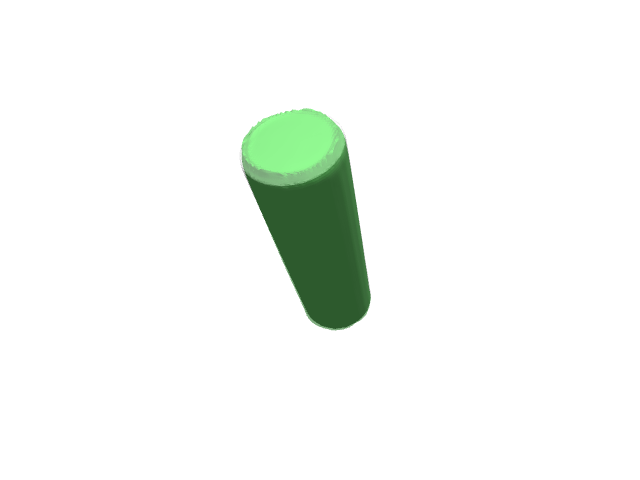}
	\end{subfigure}
	\begin{subfigure}{.24\linewidth}
		\centering
		\includegraphics[width=.49\linewidth]{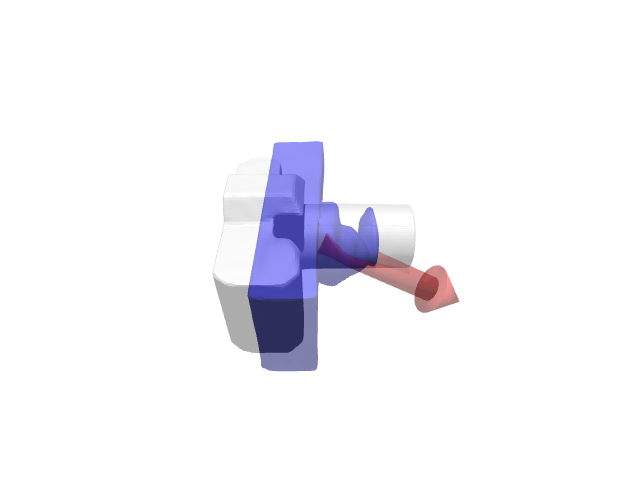}
		\includegraphics[width=.49\linewidth]{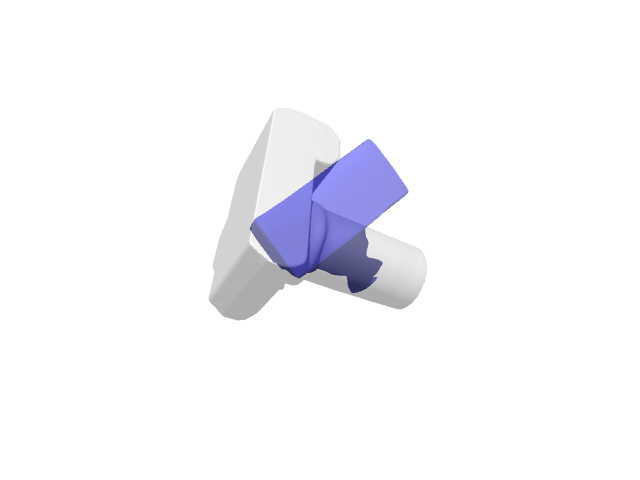}\\
		\includegraphics[width=.49\linewidth]{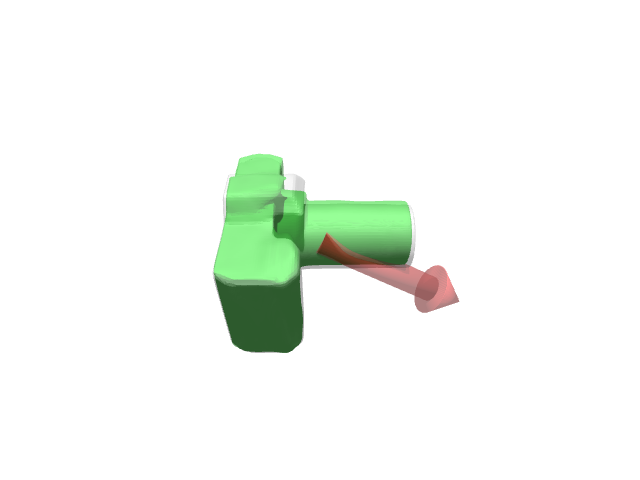}
		\includegraphics[width=.49\linewidth]{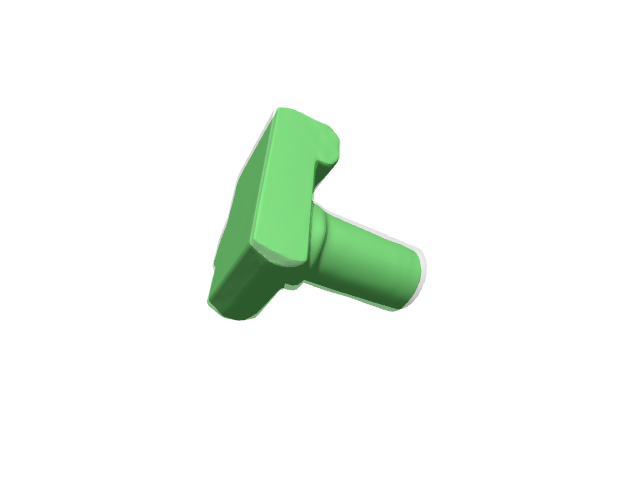}
	\end{subfigure}
	\begin{subfigure}{.24\linewidth}
		\centering
		\includegraphics[width=.49\linewidth]{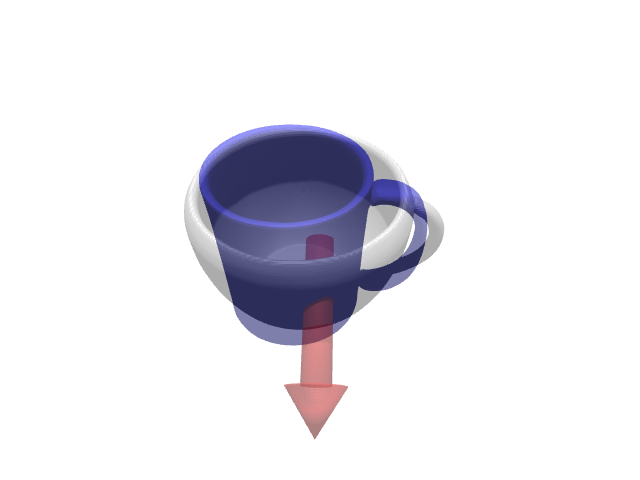}
		\includegraphics[width=.49\linewidth]{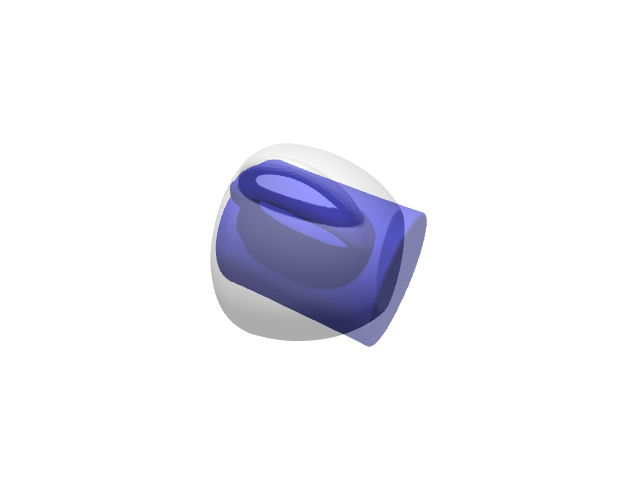}\\
		\includegraphics[width=.49\linewidth]{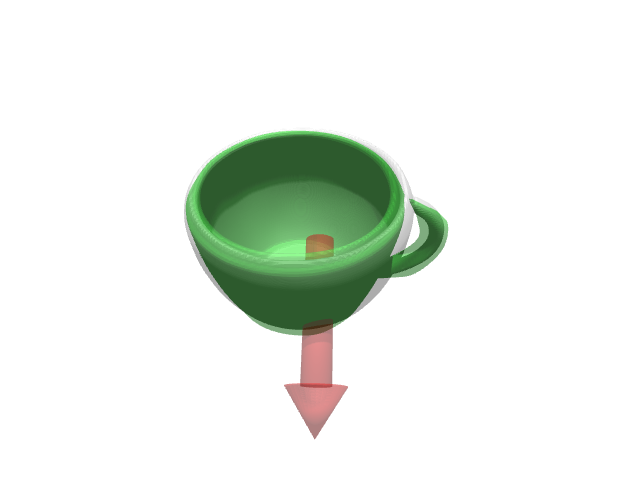}
		\includegraphics[width=.49\linewidth]{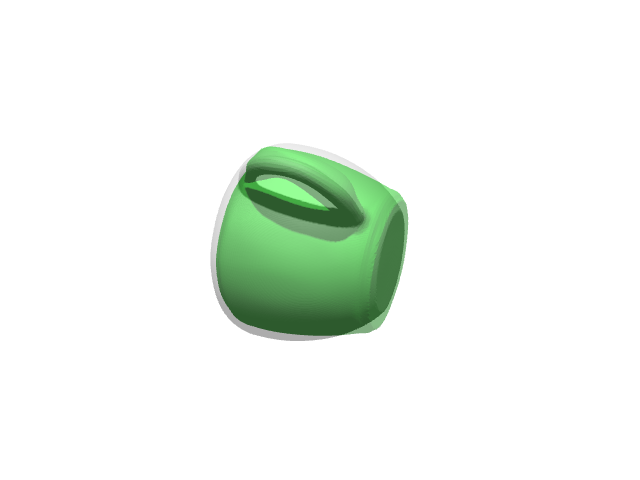}
	\end{subfigure}
	\caption{Shape from inertia. For each object, we show the initialization (blue) and result (green) overlaid with the ground truth in gray. The difference in inertia results in a different pose after 2 seconds of simulation (right rendering for each object). The red arrows indicate the torque that is applied at the beginning of the trajectory.}
	\label{fig:inertia}
\end{figure*}

Fig.~\ref{fig:exp_shapefrominertia} shows different scaled version of the box plots shown in the main paper for the shape from inertia experiment.
Table~\ref{tab:results_inertia} provides numeric results for this experiment.
Fig.~\ref{fig:inertia} shows qualitative example results for this experiment.

\subsubsection{Friction and mass identification, force optimization (sec.~5.2 in main paper)}
\begin{figure*}
	\centering
	\begin{subfigure}{.33\linewidth}
		\centering
		\includegraphics[width=.3\linewidth]{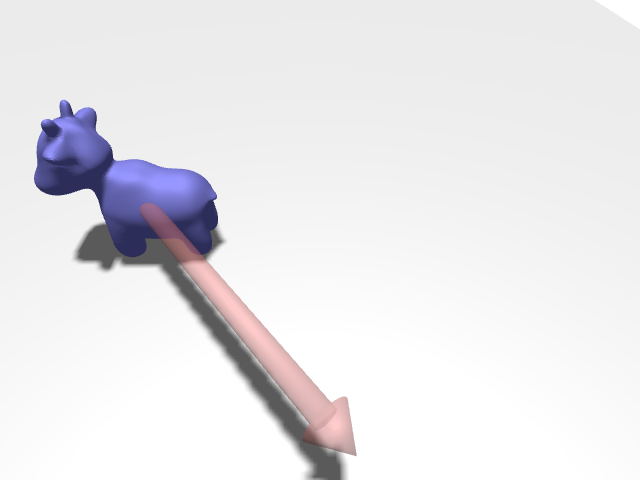}
		\includegraphics[width=.3\linewidth]{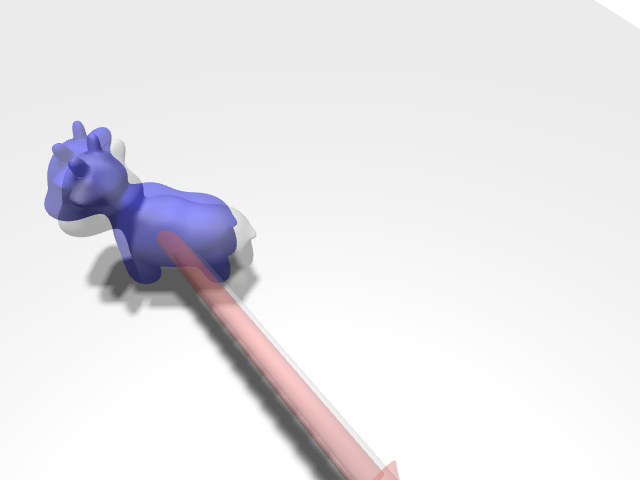}
		\includegraphics[width=.3\linewidth]{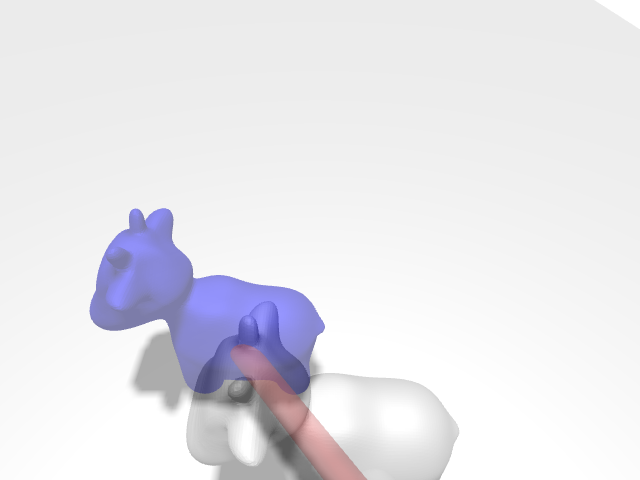}\\
		\includegraphics[width=.3\linewidth]{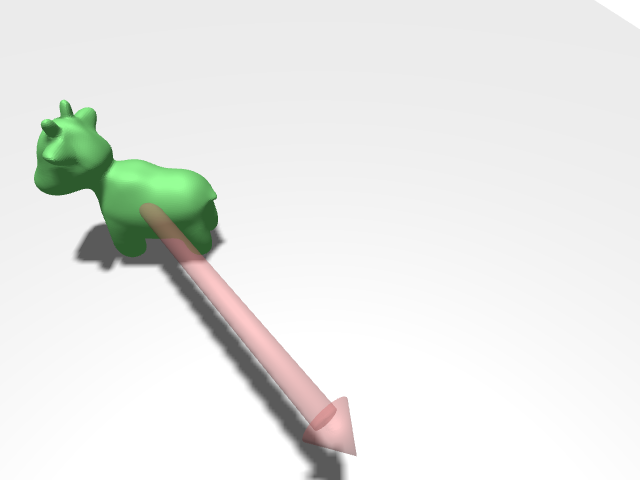}
		\includegraphics[width=.3\linewidth]{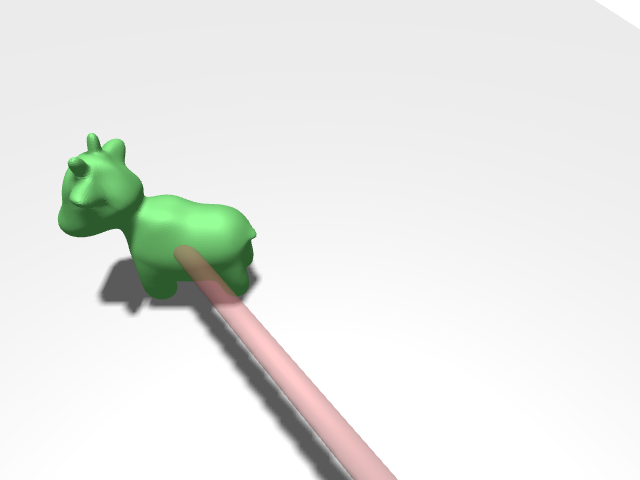}
		\includegraphics[width=.3\linewidth]{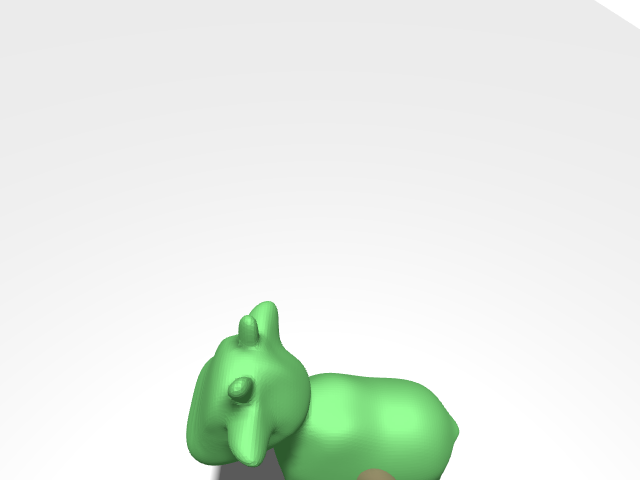}\\
	\end{subfigure}
	\begin{subfigure}{.33\linewidth}
		\centering
		\includegraphics[width=.3\linewidth]{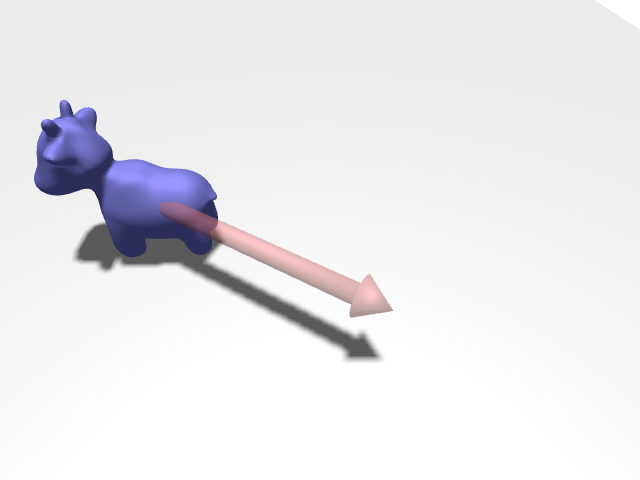}
		\includegraphics[width=.3\linewidth]{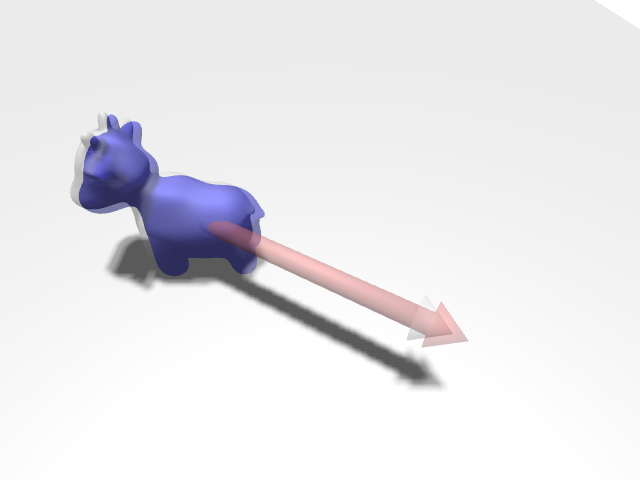}
		\includegraphics[width=.3\linewidth]{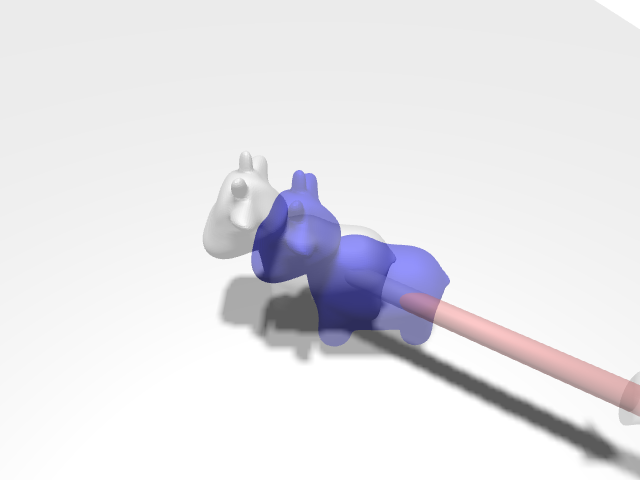}\\
		\includegraphics[width=.3\linewidth]{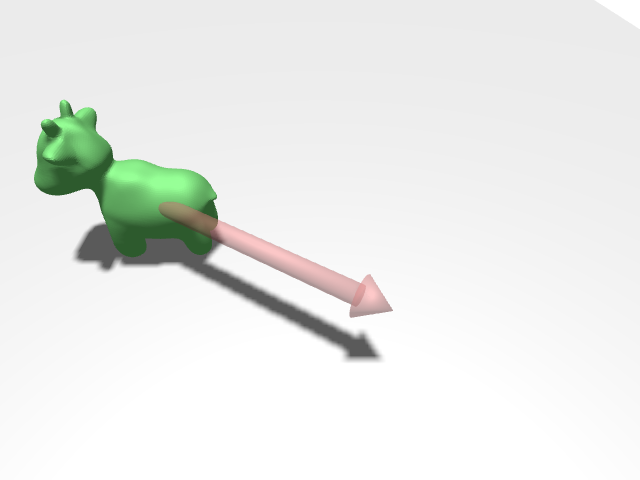}
		\includegraphics[width=.3\linewidth]{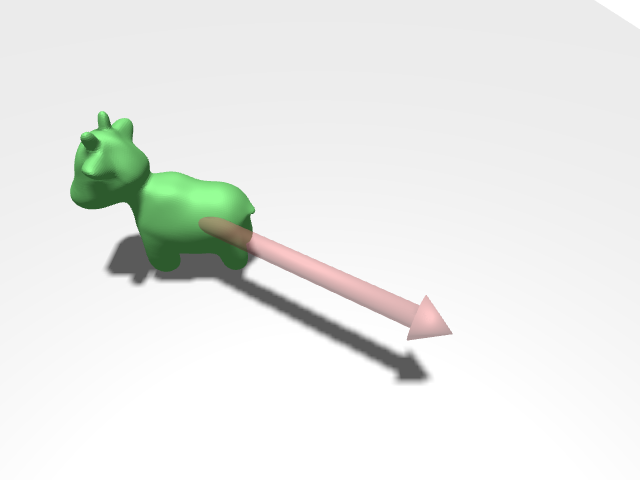}
		\includegraphics[width=.3\linewidth]{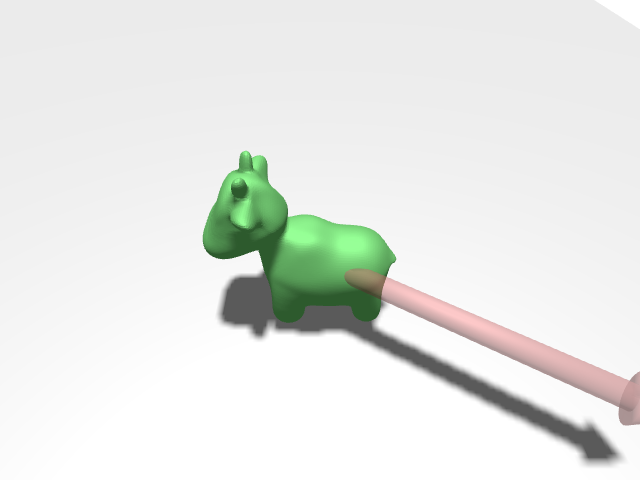}\\
	\end{subfigure}
	\begin{subfigure}{.33\linewidth}
		\centering
		\includegraphics[width=.3\linewidth]{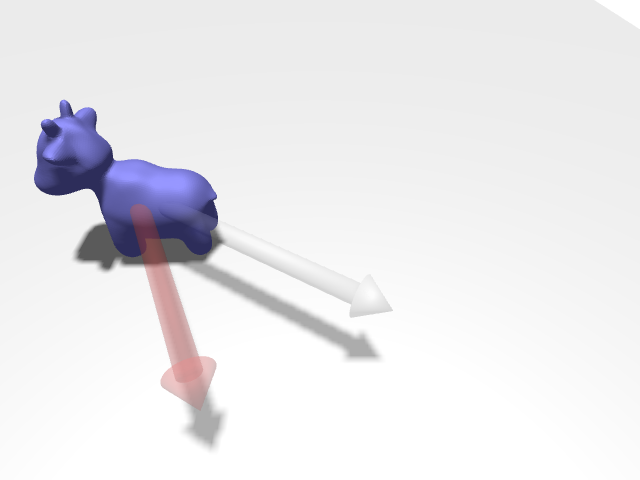}
		\includegraphics[width=.3\linewidth]{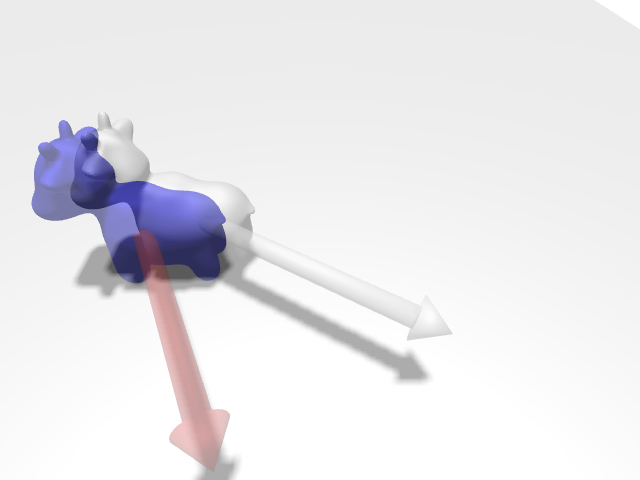}
		\includegraphics[width=.3\linewidth]{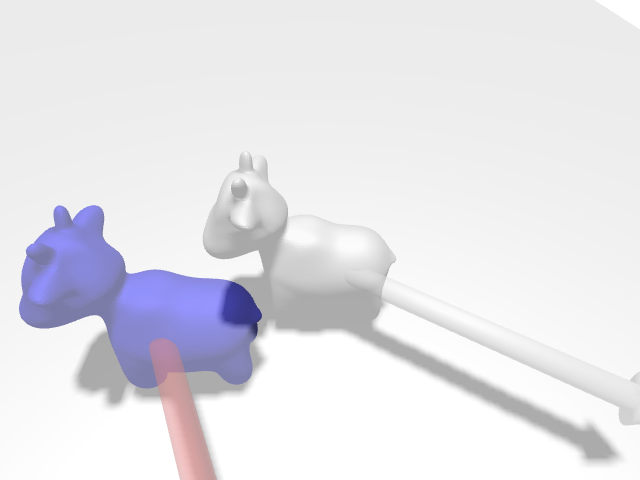}\\
		\includegraphics[width=.3\linewidth]{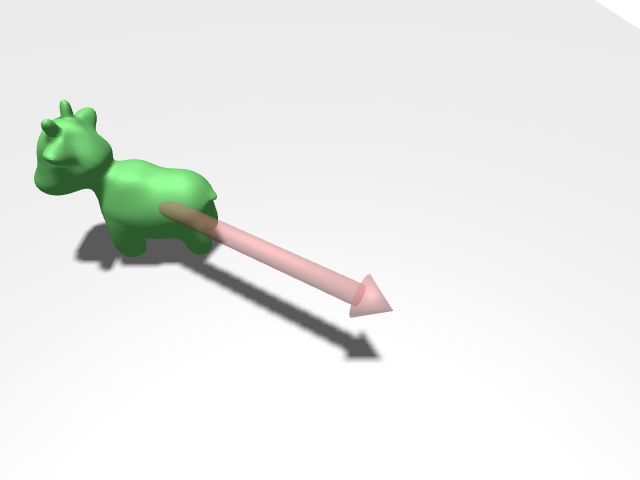}
		\includegraphics[width=.3\linewidth]{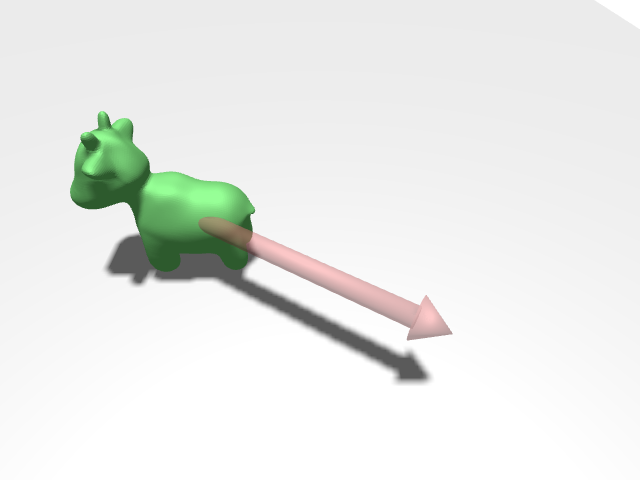}
		\includegraphics[width=.3\linewidth]{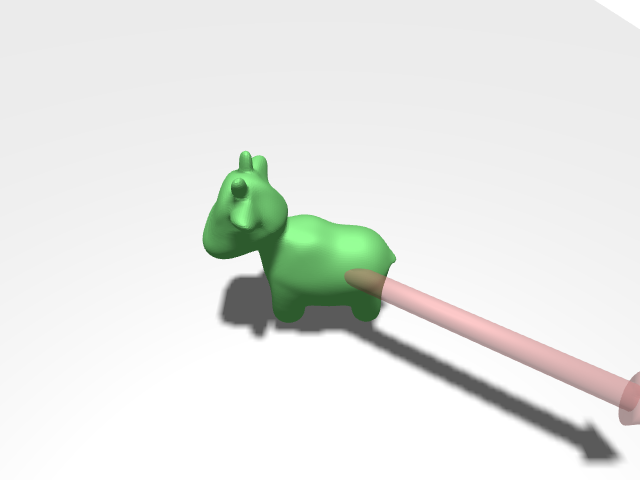}\\
	\end{subfigure}
	\caption{Friction (left) and mass (center) identification and force optimization (right). Each group show 3 frames from the beginning, middle, and end of a 1s trajectory. With the initial estimates (blue), the simulated trajectory differs from the target one (overlay in gray). By optimizing the parameters to match the trajectory, we manage to recover them well. The red arrows indicate the pushing force parallel to the plane that is applied to the object (scaled by a factor of 0.5 for better visibility).}
	\label{fig:system_id}
\end{figure*}

Fig.~\ref{fig:exp_systemid} shows different scaled version of the box plots shown in the main paper for the friction, mass and force identification experiment.
Table~\ref{tab:results_systemid} provides numeric results for this experiment.
Fig.~\ref{fig:system_id} shows qualitative results, showing that the simulation with the resulting parameters (green) matches the target trajectory (overlay in gray) much better than the initialization (blue).
The optimization objective is the position trajectory error from eq.~\eqref{eq:trajectory_error}, but this time $\theta$ denotes the physical parameter to be optimized (mass, friction or force).

\subsubsection{Fitting to depth observations (sec.~5.3 in main paper)}
In this experiment, we render depth and segmentation images of a target trajectory using pyrender\footnote{\url{https://github.com/mmatl/pyrender}} at a resolution of $640 \times 480$ pixels.
After adding synthetic Gaussian noise to the depth image with mean $\mu = d$ and std. dev. $\sigma = 0.0001d^2$, where $d$ is the depth value, we compute point clouds from the depth map using the stored camera intrinsic matrix and the known camera pose.
Segmenting this point cloud using the segmentation mask gives us a point set $\mathcal{P}$ for the object.

\begin{figure}
	\centering
	\includegraphics[width=.49\linewidth]{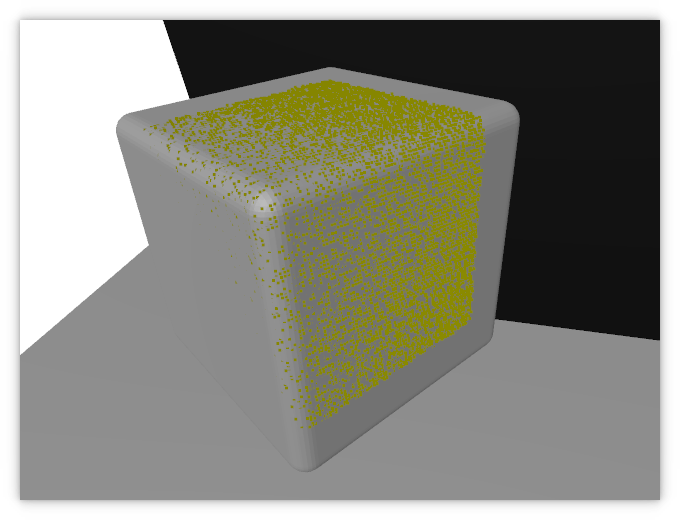}\includegraphics[width=.49\linewidth]{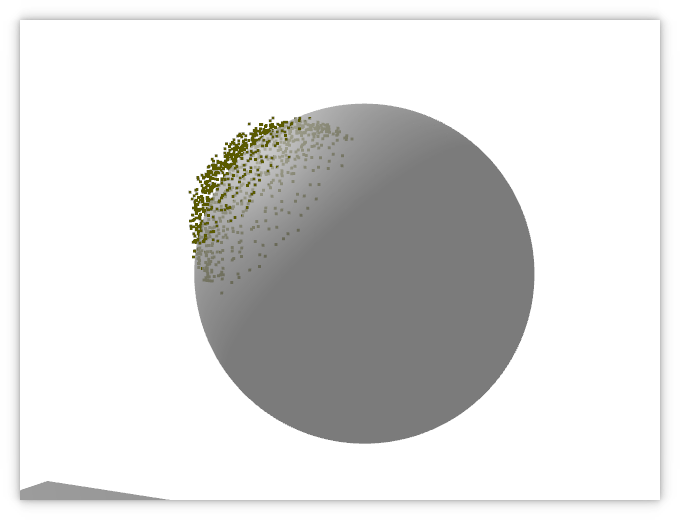}
	\caption{Fitting an object (gray) to a point cloud (yellow) can yield a wrong local optimum for both pose and shape estimates since the pointcloud does not constrain the unseen side of the object. For the box (left) this local optimum yields a perfect fit, for the sphere (right), some of the points end up outside the object (darker) and some inside it (lighter/gray points).}
	\label{fig:pointcloud_fit_local_optimum}
\end{figure}

\begin{figure*}
	\centering
	\begin{subfigure}{.49\linewidth}
		\centering
		\includegraphics[width=.3\linewidth]{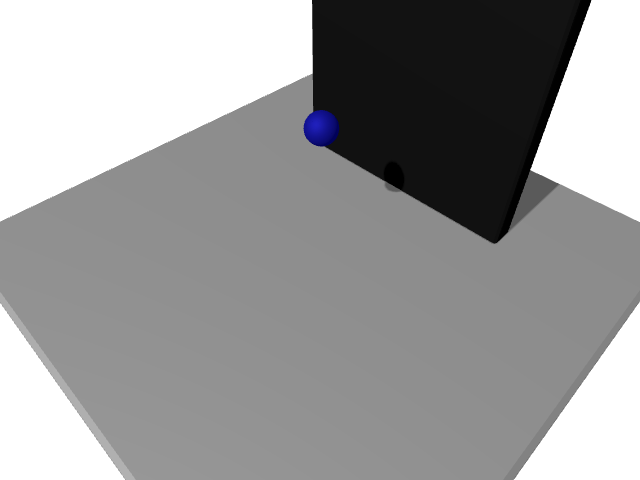}
		\includegraphics[width=.3\linewidth]{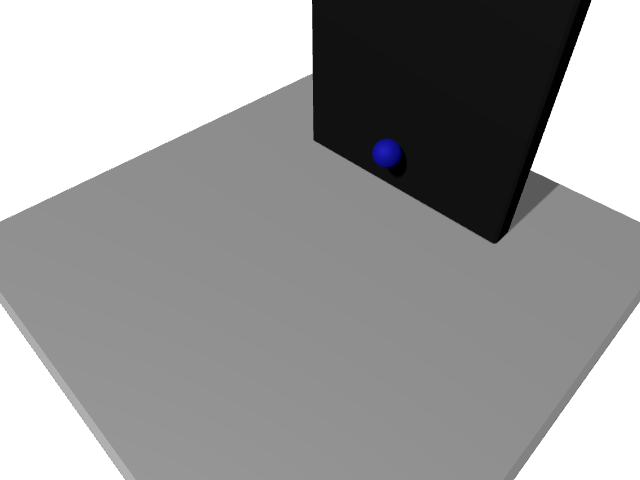}
		\includegraphics[width=.3\linewidth]{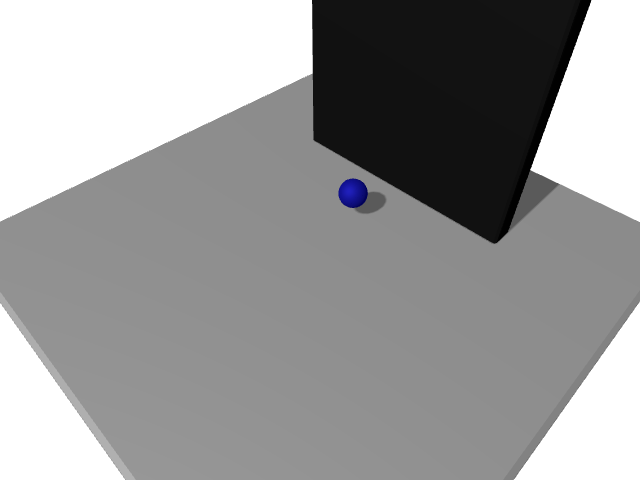}\\
		\includegraphics[width=.3\linewidth]{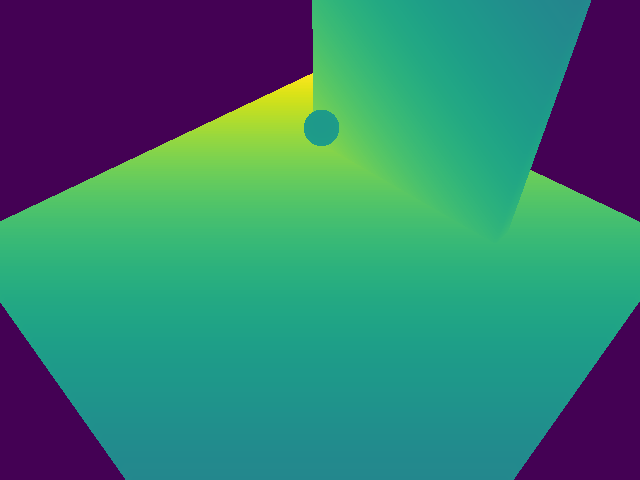}
		\includegraphics[width=.3\linewidth]{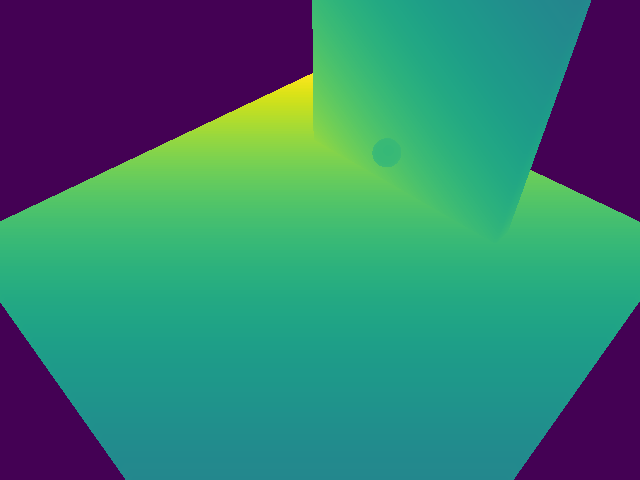}
		\includegraphics[width=.3\linewidth]{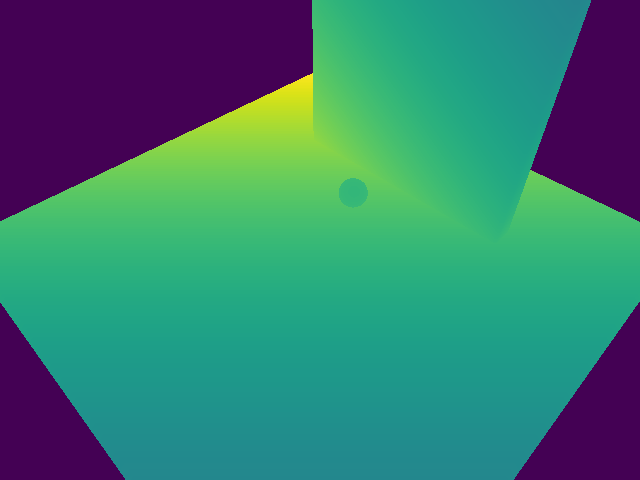}\\
		\includegraphics[width=.3\linewidth]{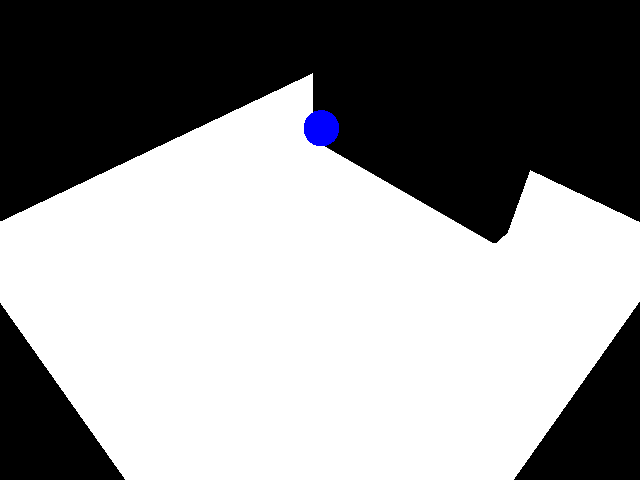}
		\includegraphics[width=.3\linewidth]{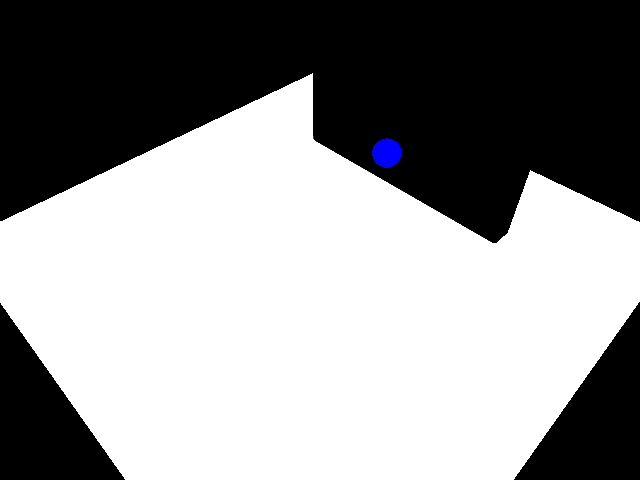}
		\includegraphics[width=.3\linewidth]{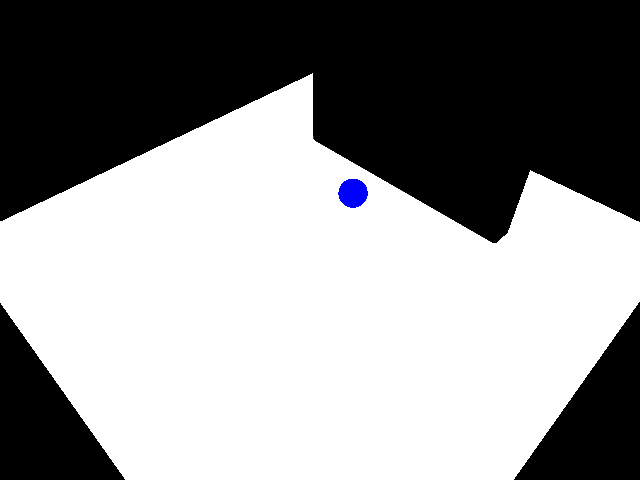}
	\end{subfigure}
	\begin{subfigure}{.49\linewidth}
		\centering
		\includegraphics[width=.3\linewidth]{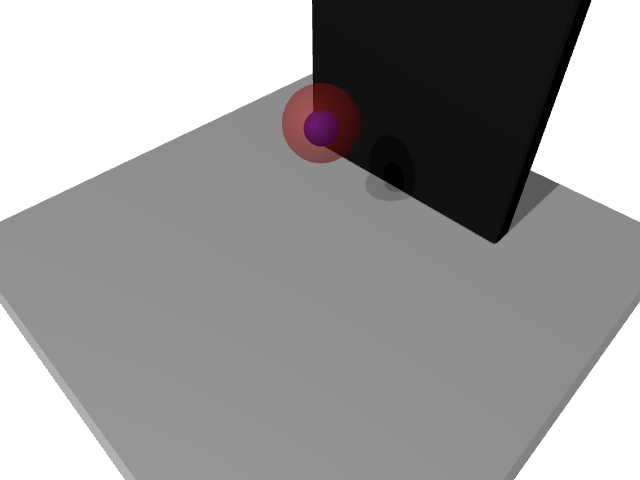}
		\includegraphics[width=.3\linewidth]{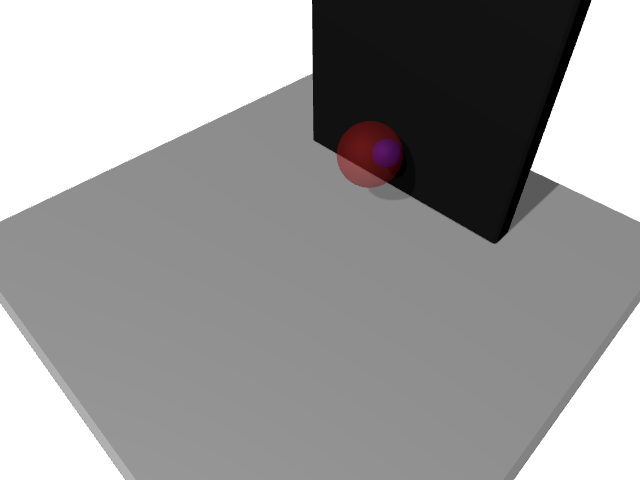}
		\includegraphics[width=.3\linewidth]{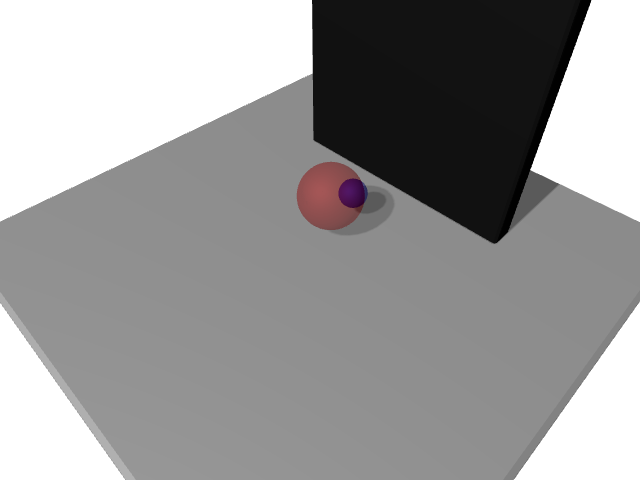}\\
		\includegraphics[width=.3\linewidth]{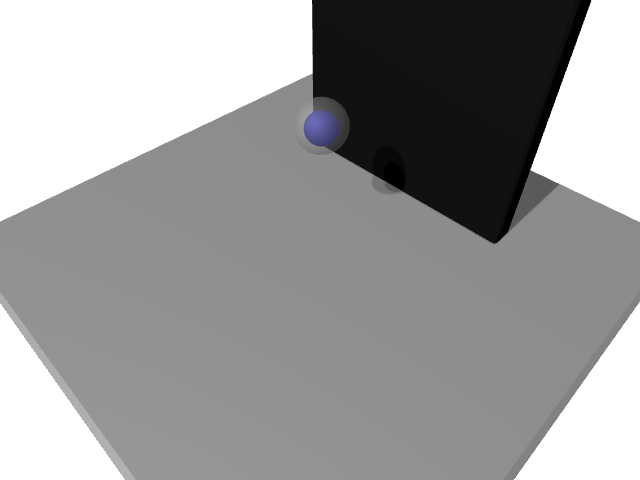}
		\includegraphics[width=.3\linewidth]{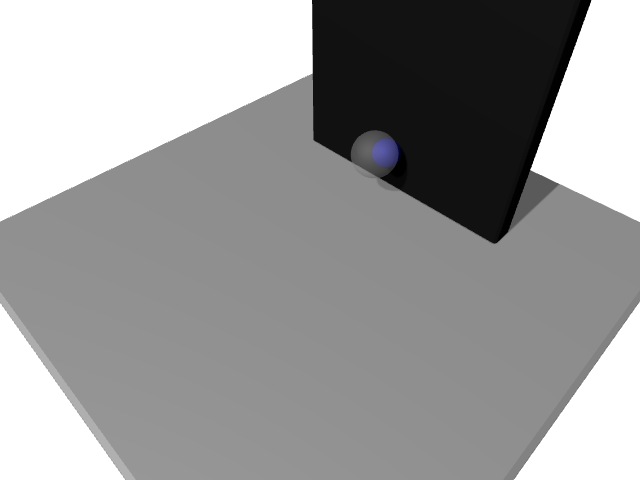}
		\includegraphics[width=.3\linewidth]{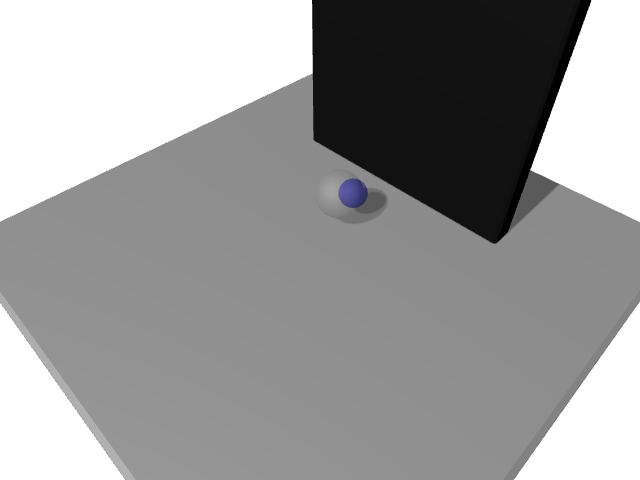}\\
		\includegraphics[width=.3\linewidth]{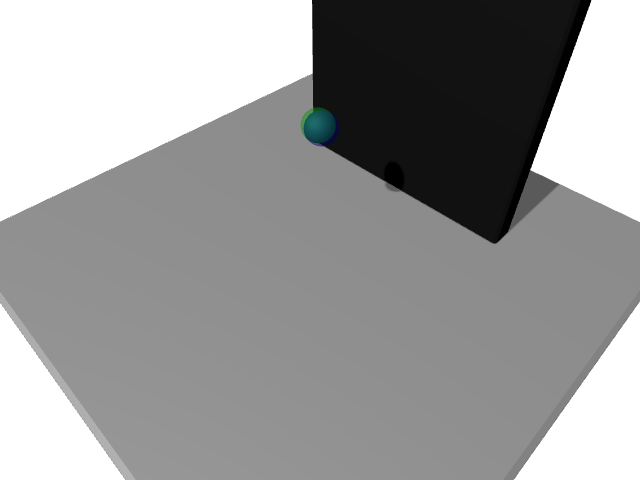}
		\includegraphics[width=.3\linewidth]{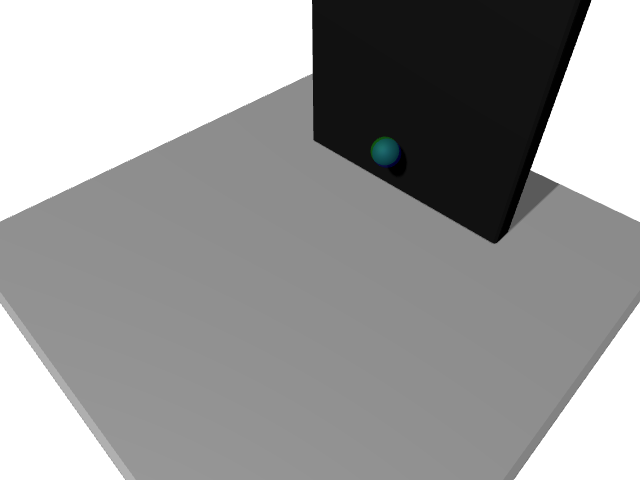}
		\includegraphics[width=.3\linewidth]{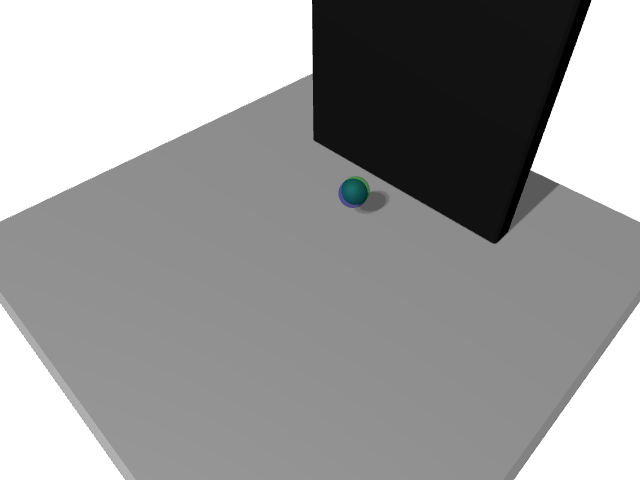}
	\end{subfigure}\\[1em]
	\begin{subfigure}{.49\linewidth}
		\centering
		\includegraphics[width=.3\linewidth]{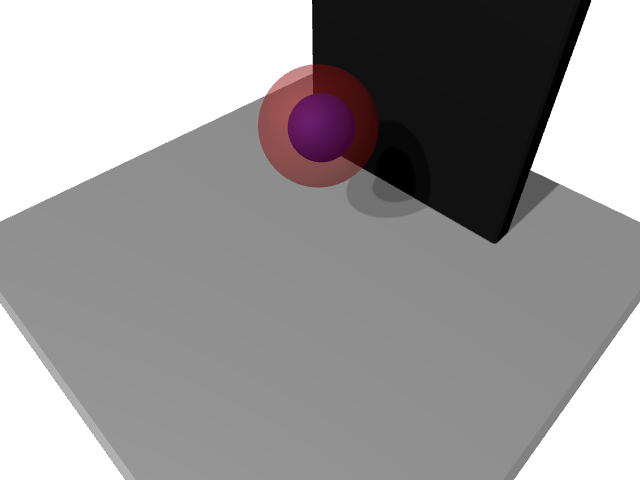}
		\includegraphics[width=.3\linewidth]{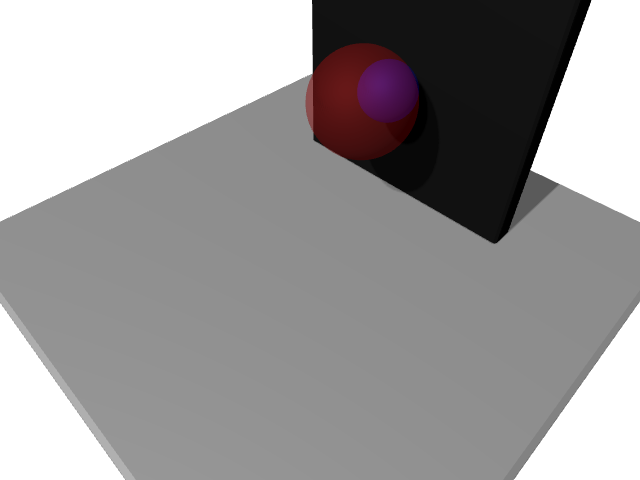}
		\includegraphics[width=.3\linewidth]{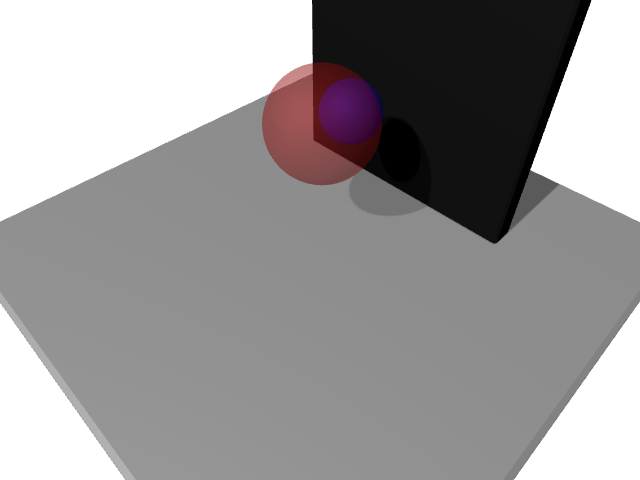}\\
		\includegraphics[width=.3\linewidth]{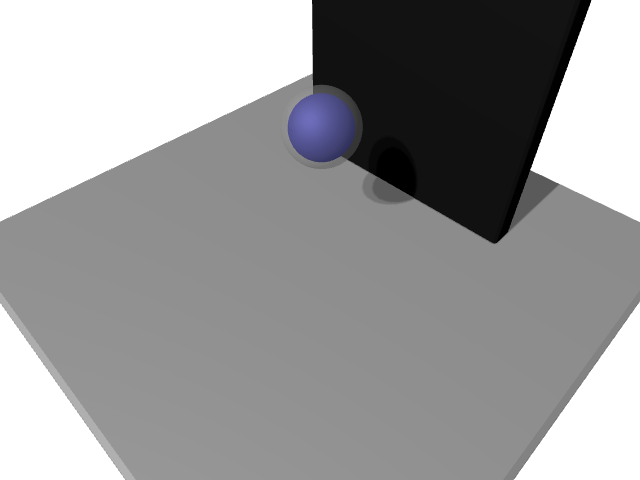}
		\includegraphics[width=.3\linewidth]{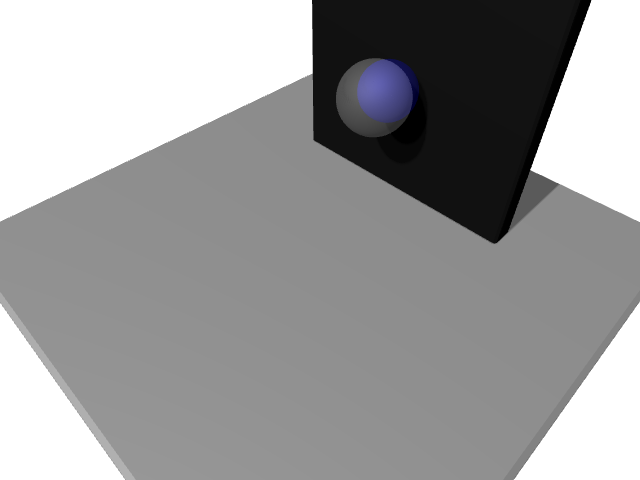}
		\includegraphics[width=.3\linewidth]{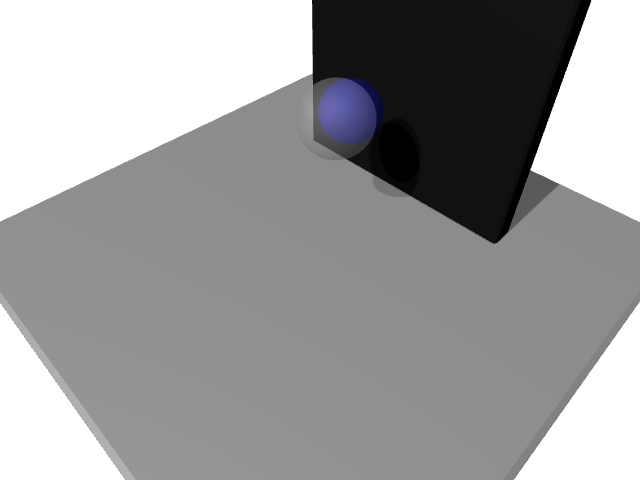}\\
		\includegraphics[width=.3\linewidth]{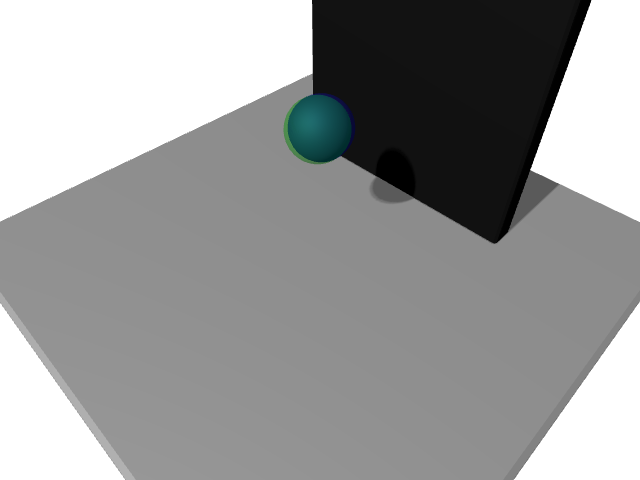}
		\includegraphics[width=.3\linewidth]{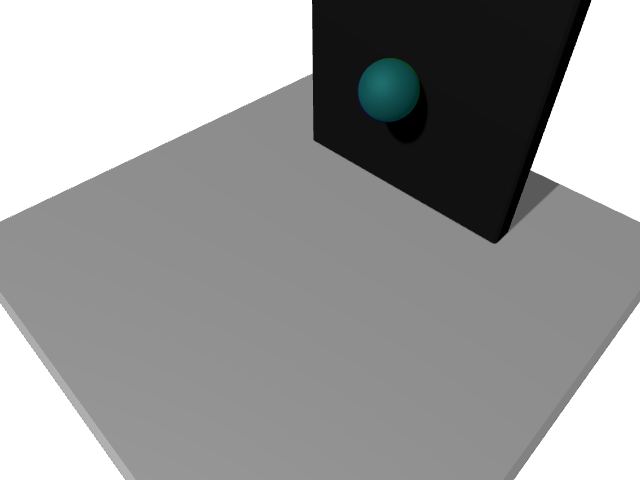}
		\includegraphics[width=.3\linewidth]{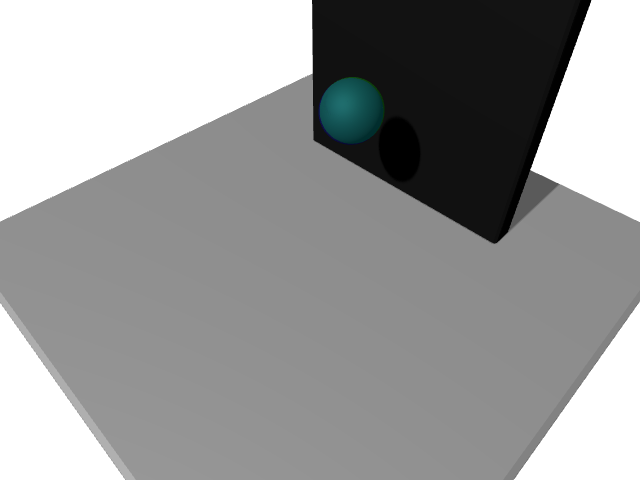}
	\end{subfigure}
	\begin{subfigure}{.49\linewidth}
		\centering
		\includegraphics[width=.3\linewidth]{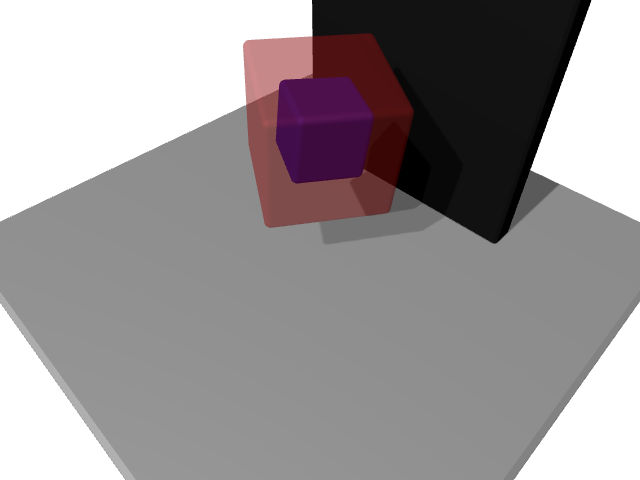}
		\includegraphics[width=.3\linewidth]{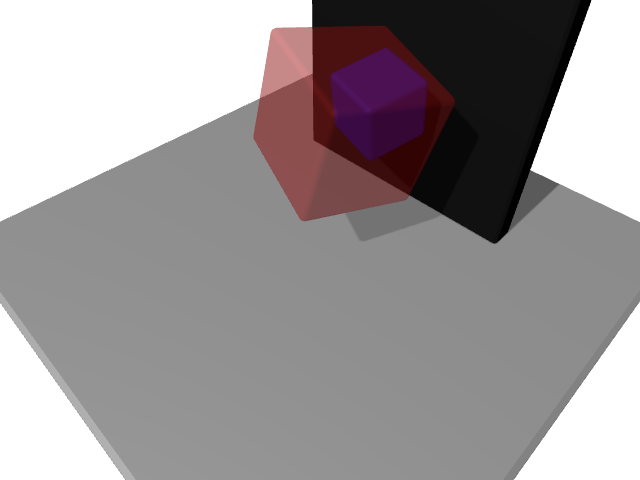}
		\includegraphics[width=.3\linewidth]{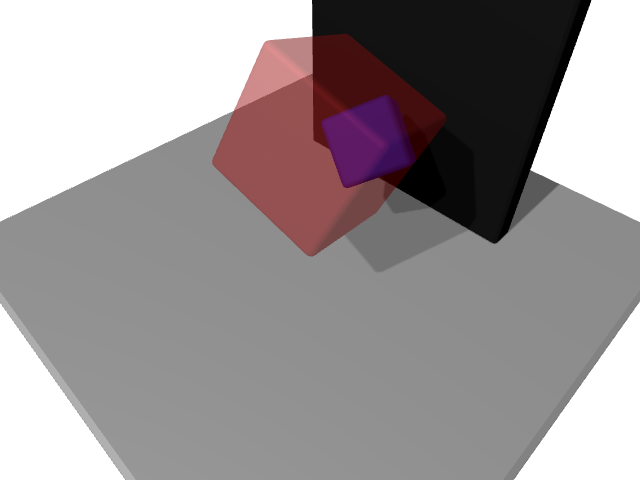}\\
		\includegraphics[width=.3\linewidth]{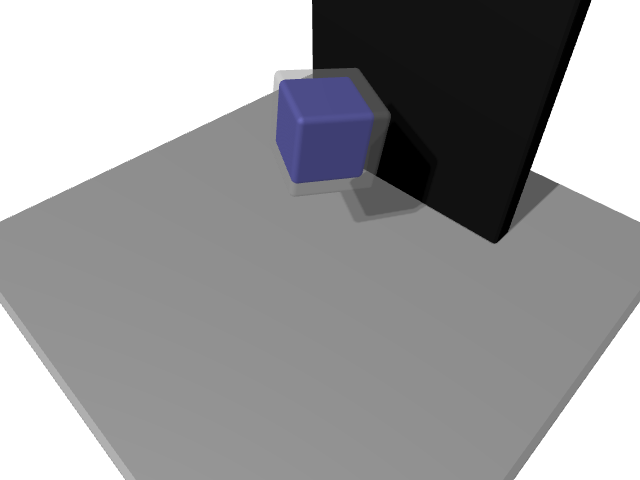}
		\includegraphics[width=.3\linewidth]{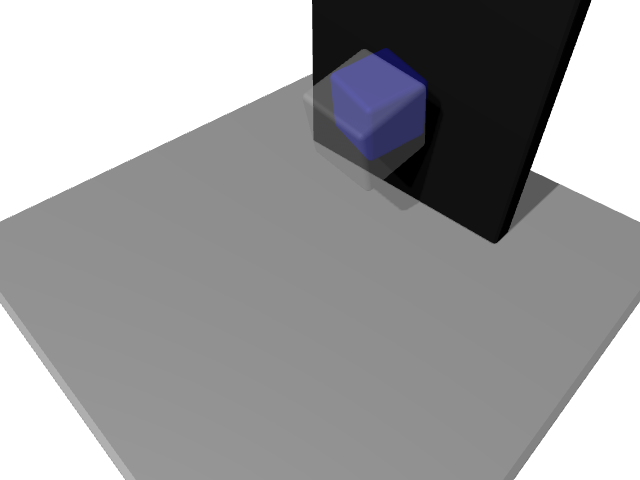}
		\includegraphics[width=.3\linewidth]{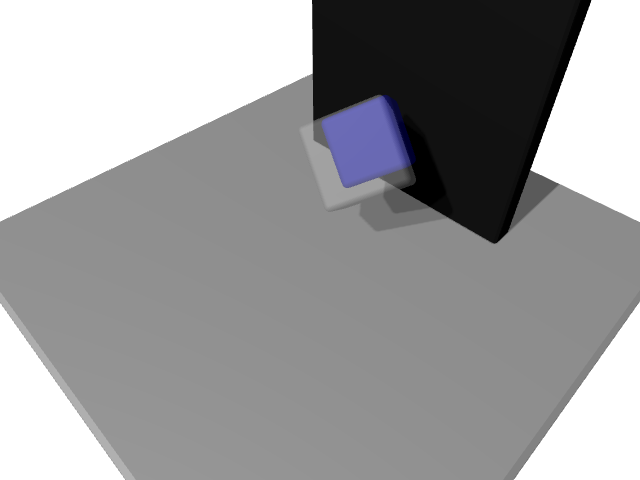}\\
		\includegraphics[width=.3\linewidth]{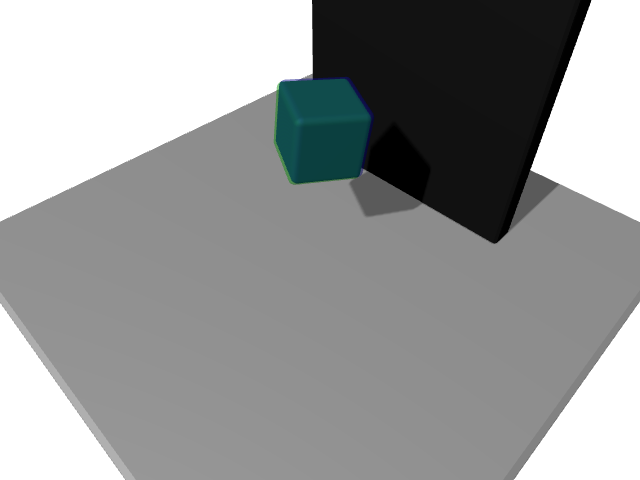}
		\includegraphics[width=.3\linewidth]{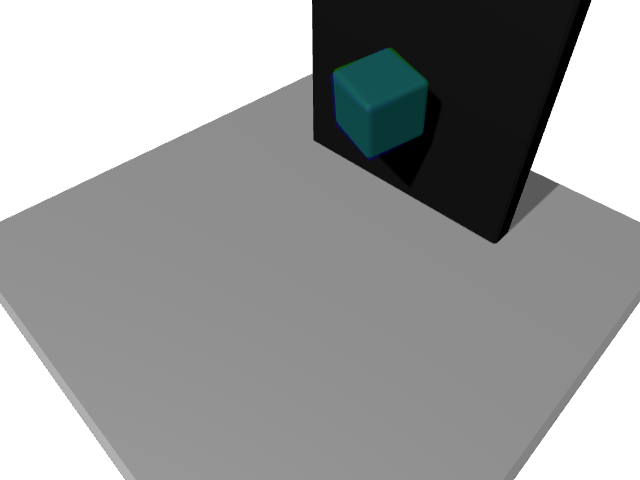}
		\includegraphics[width=.3\linewidth]{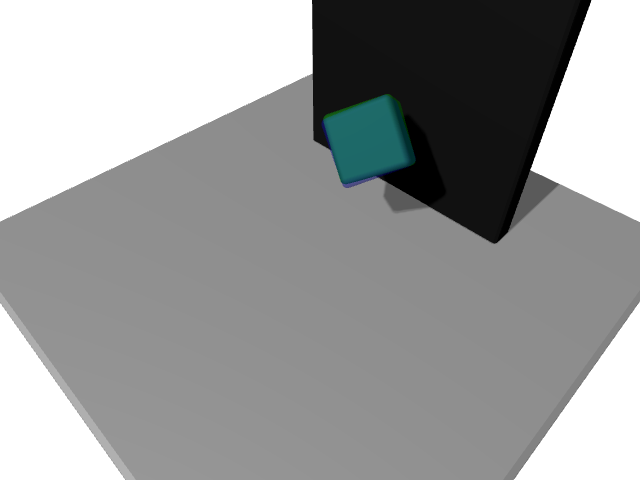}
	\end{subfigure}
	\caption{Fitting to depth observations. The input for this experiments are depth and segmentation masks (illustrated in the top left below a rendering of the scene) at 3 time steps from the beginning, middle and end of a 1.5s trajectory. Optimization of initialization pose and the size of the object is then carried out in 2 stages. From an initialization (red overlay over the blue target), pose and shape of the SDF are first fit to the first depth frame (gray overlay over blue target). This optimization can fall into local optima, \eg by overestimating the size and putting the object further back. This can be seen for the cube in the first frame in the second row. This error is recovered by our optimization using our differentiable simulation (green overlay over blue target).}
	\label{fig:depth_fitting}
\end{figure*}

The first stage of the optimization adapts the object size and pose by optimizing the squared mean SDF error for a point cloud generated from the first observation frame:
\begin{equation}
	E_{sdf}(\theta, \mathbf{T}) = \frac{1}{\|\mathcal{P}\|}\sum_{\mathbf{p}\in\mathcal{P}} \phi(\theta, \mathbf{T}\mathbf{p})^2,
	\label{eq:sdf_err}
\end{equation}
where $\theta$ is the object size parameter, and $\mathbf{T}$ is the SE(3) transformation matrix transforming the points to the object frame and $\phi(\theta, \mathbf{p})$ is the object's SDF.
This stage is prone to local minima and often overestimates the object size, as can be seen in the middle rows of Fig.~\ref{fig:depth_fitting}.
Fig.~\ref{fig:pointcloud_fit_local_optimum} illustrates this problem. Fitting the shape and position of the objects (gray) to the point clouds (yellow) yields wrong local optima.
For the cube, this local optimum is a perfect fit, while the actual object would be smaller, ending where the point cloud ends and having a position closer to the camera.
For the sphere, the local optimum ends up with some points outside and some inside the object.

The second stage optimizes the error function in eq.~\eqref{eq:sdf_err} cumulated for every frame of a simulated trajectory using our engine with the point cloud generated from the corresponding depth frame.
We use the result of the first stage for initialization of the second stage.
The qualitative results in the last rows of the scenes in Fig.~\ref{fig:depth_fitting} demonstrate that this strategy manages to recover well from the initial error.

For the real data experiment, we reproduce a setting similar to the synthetic depth fitting experiment by throwing a tennis ball against a wall and recording it with an Intel RealSense D455 camera at a resolution of $640\times 480$ with 30 FPS.
The camera comes with an IMU, which gives the gravity direction.
We segment the planes and the ball from the point cloud by combined geometric and color segmentation.
We initialize the ball radius and position in the first two frames based on the radius (2.96cm) and position of the point cloud segments.
We then fit the first two positions and the radius to the point clouds of the first two frames.
The result (3.68cm) is a slight over-estimate of the ball's radius (gt: 3.24cm).

Then, we compute the initial velocity from the positions in the first two frames.
We then optimize restitution, friction, initial velocity and position, and radius using the depth fitting objective like in the experiment in section 5.3 in the main paper.
This optimization problem has many degrees of freedom, hence, fitting these parameters on a single trajectory is prone to local minima.
Still our approach is able to recover the radius and trajectory well in this experiment.
The initial radius is improved to 3.13cm by our approach.

\subsection{Shape Spaces}
\begin{figure*}
	\centering
	\begin{subfigure}{0.24\linewidth}
		\centering
		\includegraphics[trim=0 40 0 60,clip,width=\linewidth]{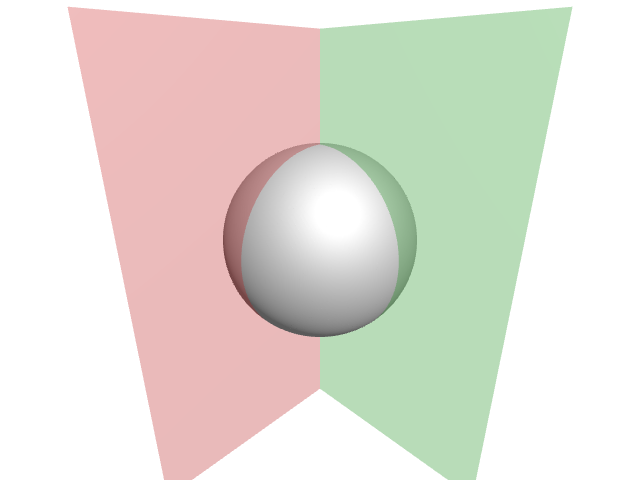}\\
		\includegraphics[trim=25 25 25 25, clip,width=.49\linewidth]{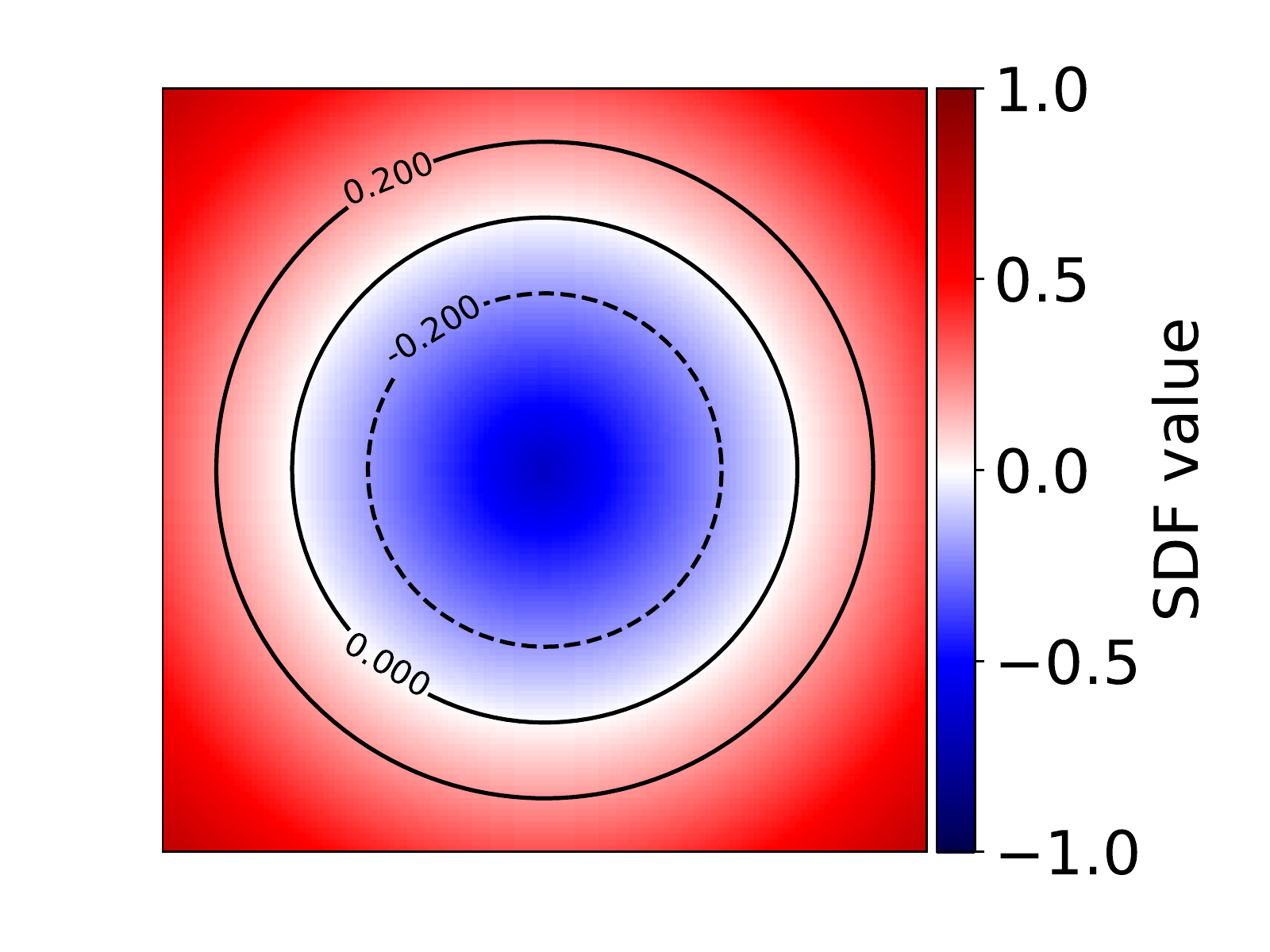}\includegraphics[trim=25 25 25 25, clip,width=.49\linewidth]{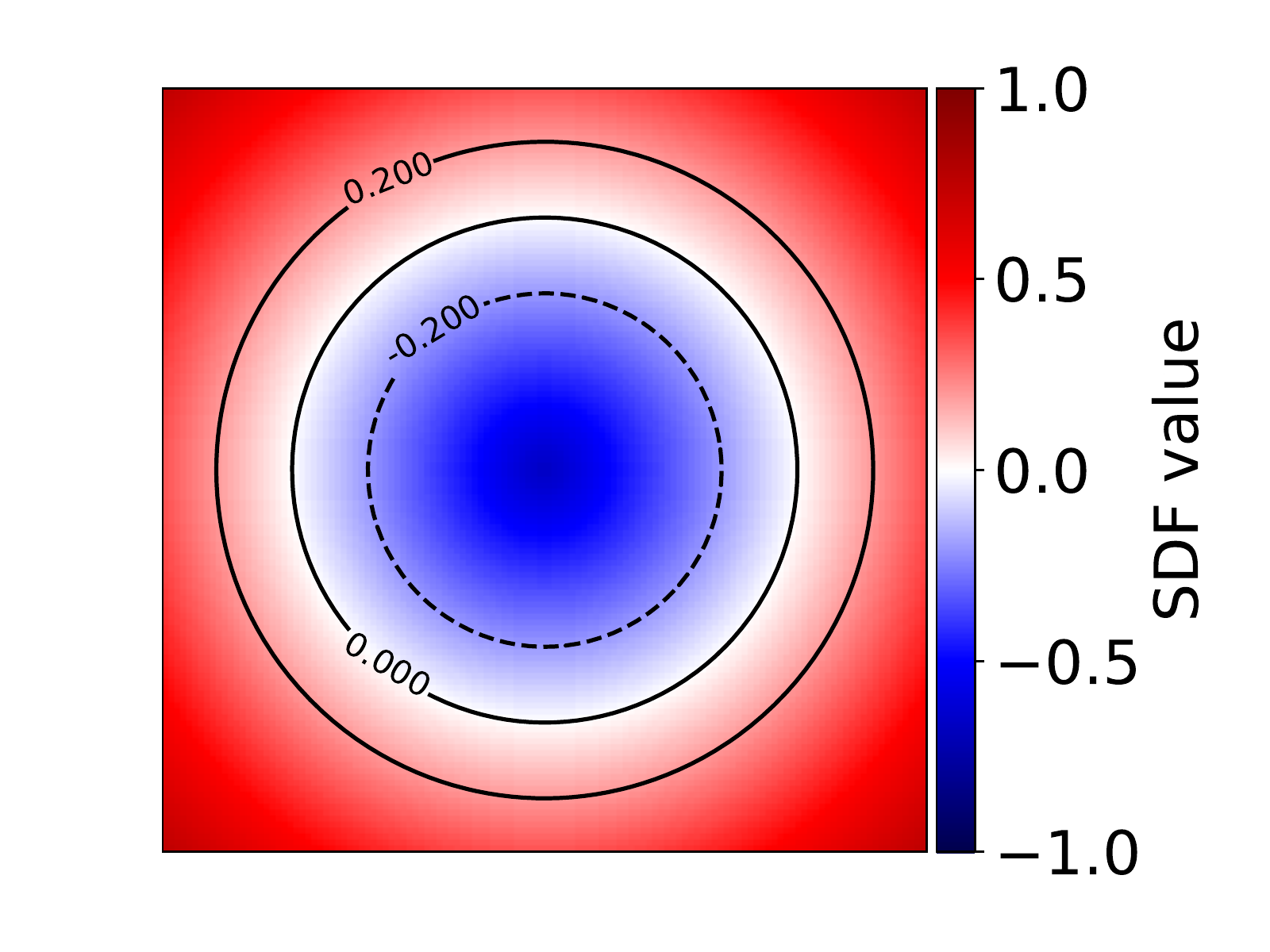}
	\end{subfigure}
	\begin{subfigure}{0.24\linewidth}
		\centering
		\includegraphics[trim=0 40 0 60,clip,width=\linewidth]{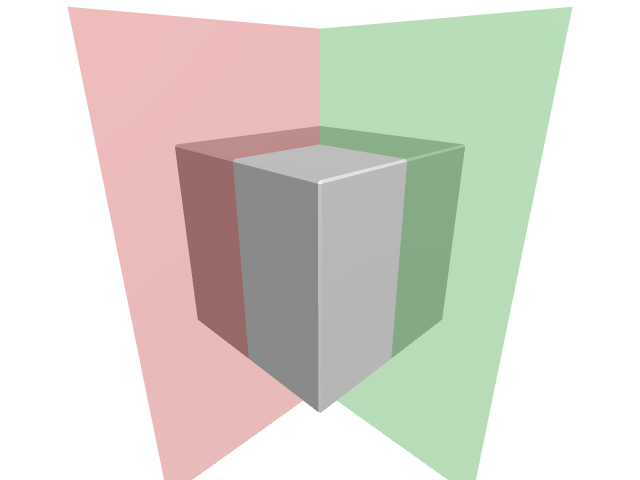}\\
		\includegraphics[trim=25 25 25 25, clip,width=.49\linewidth]{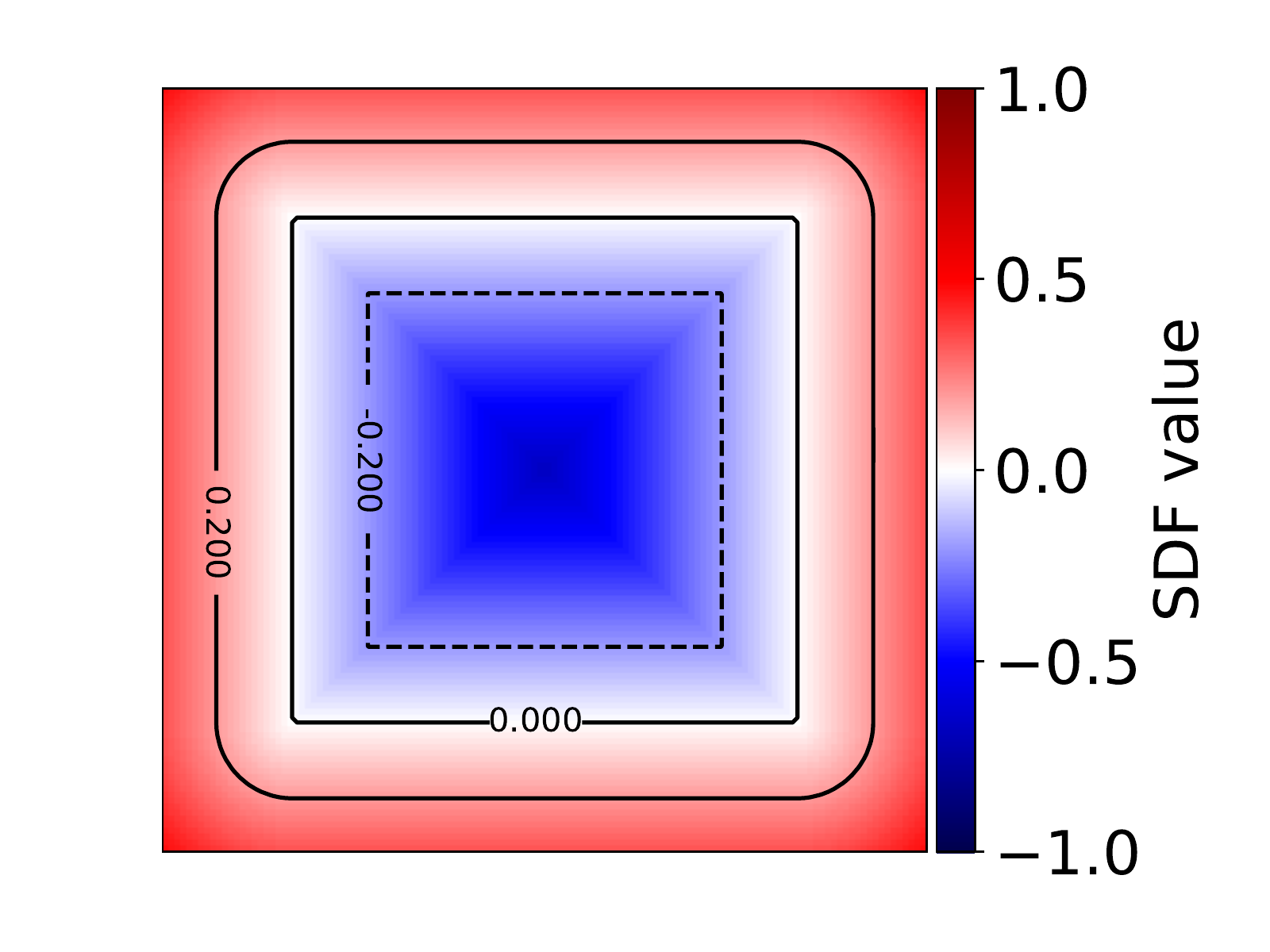}\includegraphics[trim=25 25 25 25, clip,width=.49\linewidth]{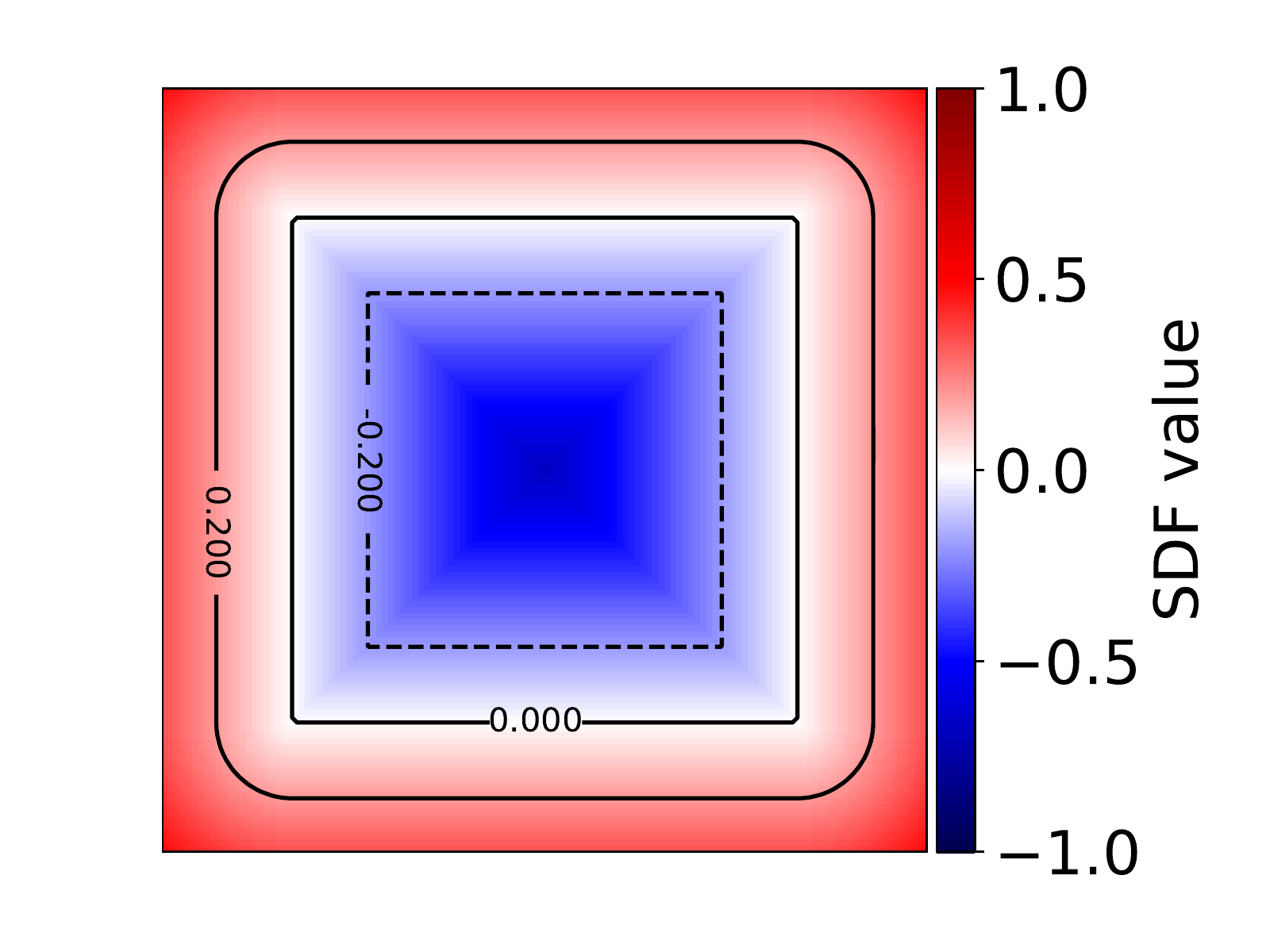}
	\end{subfigure}
	\begin{subfigure}{0.24\linewidth}
		\centering
		\includegraphics[trim=0 40 0 60,clip,width=\linewidth]{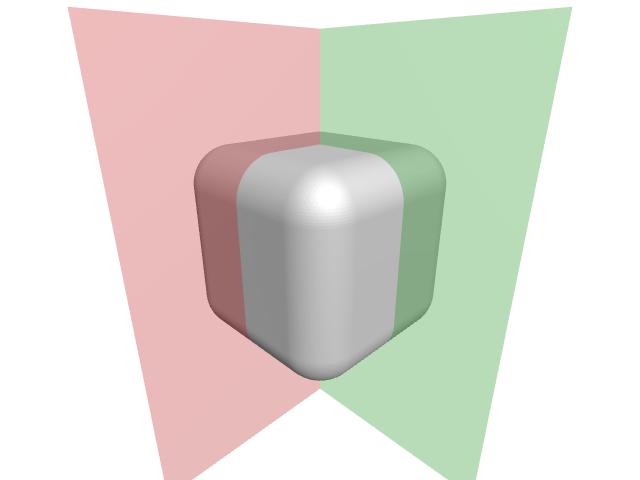}\\
		\includegraphics[trim=25 25 25 25, clip,width=.49\linewidth]{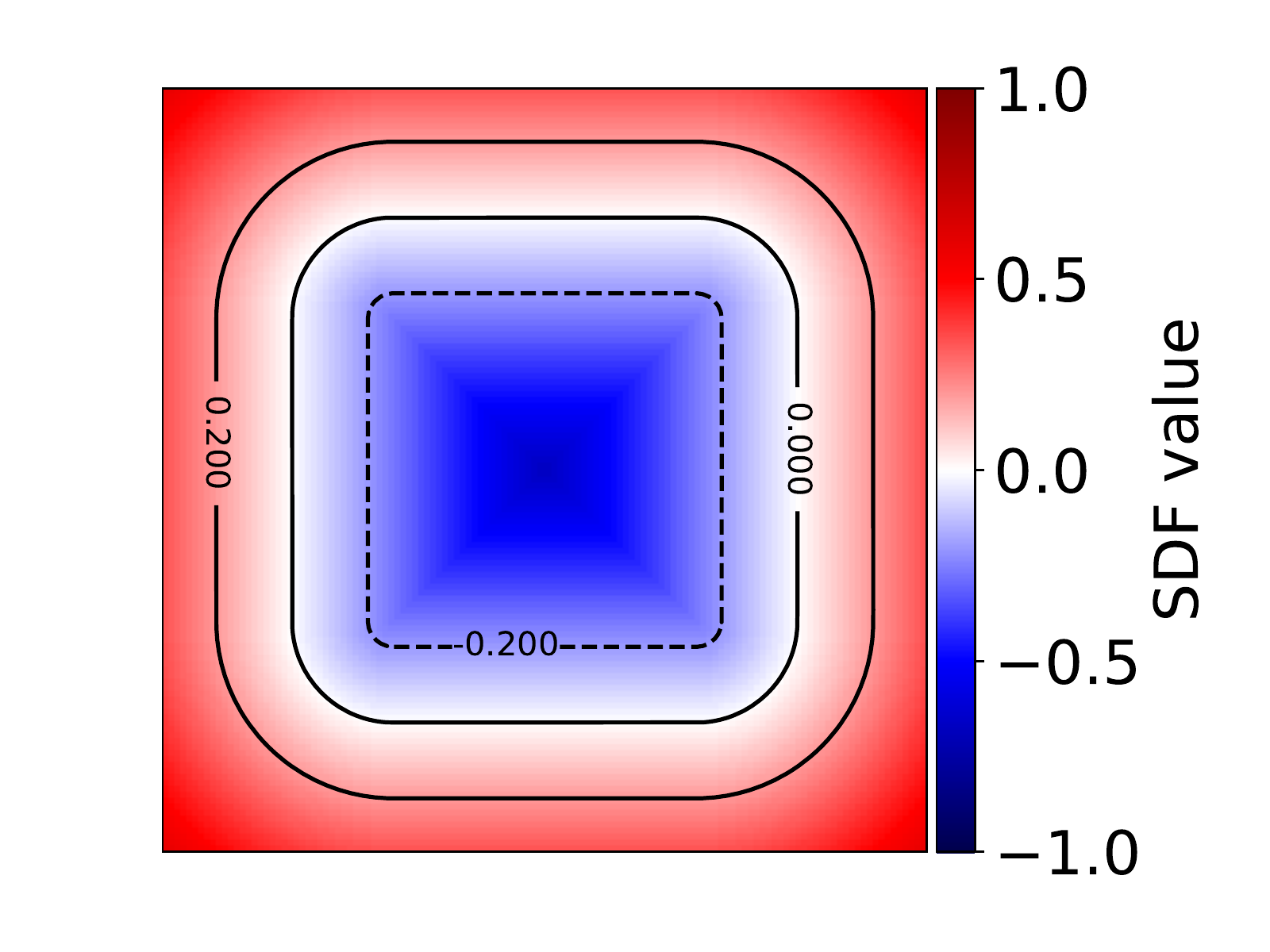}\includegraphics[trim=25 25 25 25, clip,width=.49\linewidth]{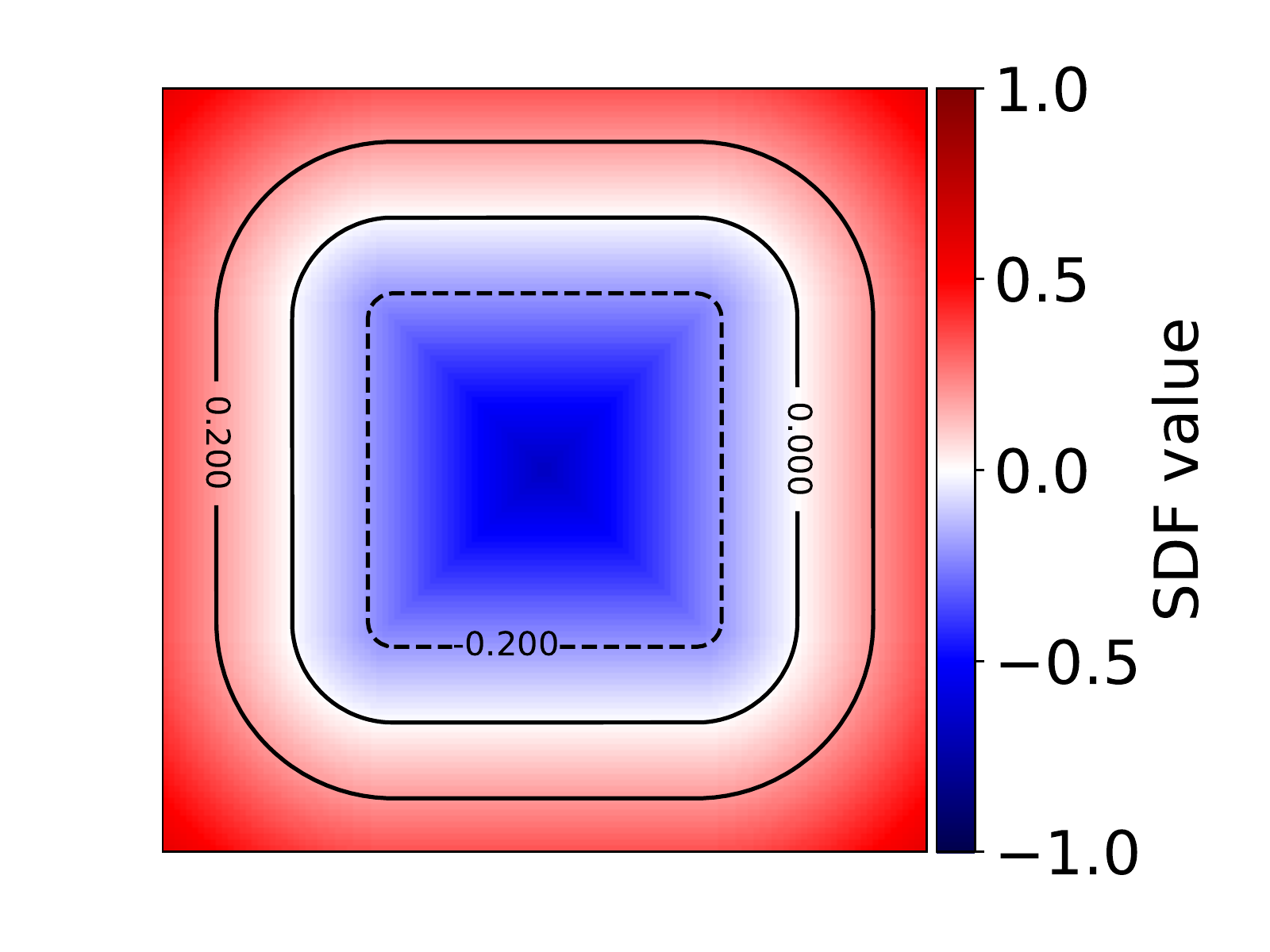}
	\end{subfigure}
	\begin{subfigure}{0.24\linewidth}
		\centering
		\includegraphics[trim=0 40 0 60,clip,width=\linewidth]{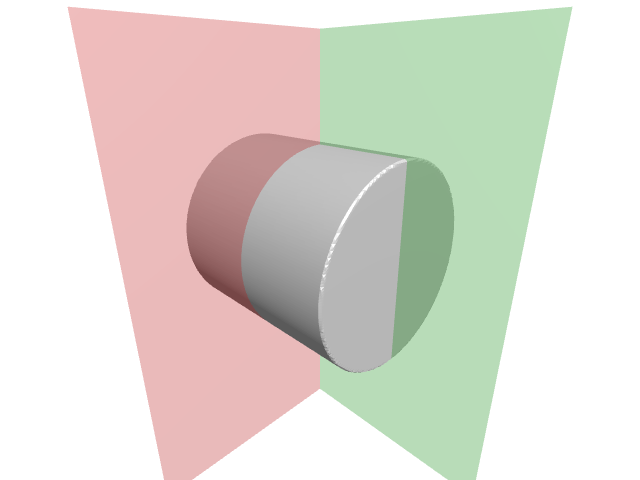}\\
		\includegraphics[trim=25 25 25 25, clip,width=.49\linewidth]{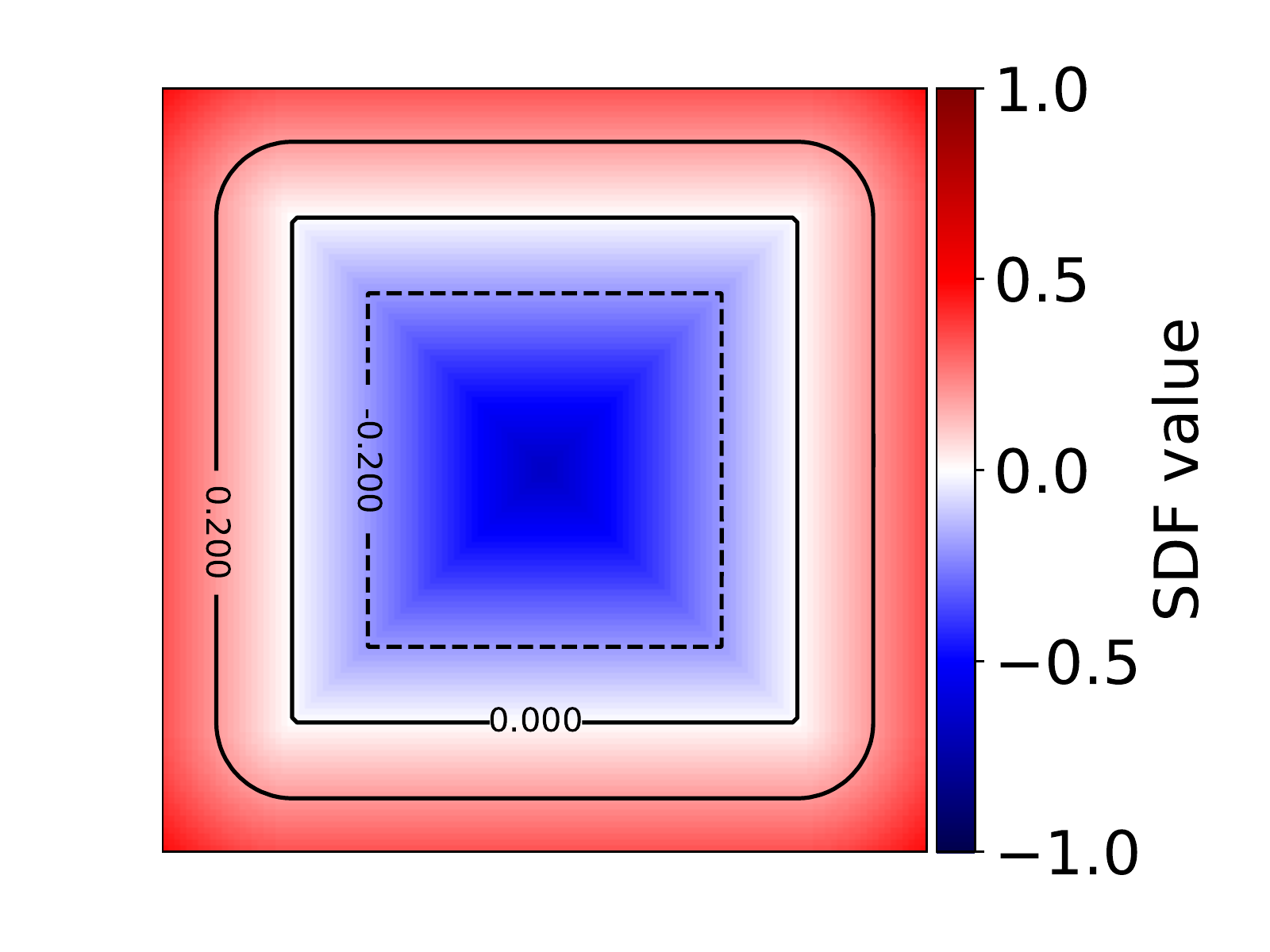}\includegraphics[trim=25 25 25 25, clip,width=.49\linewidth]{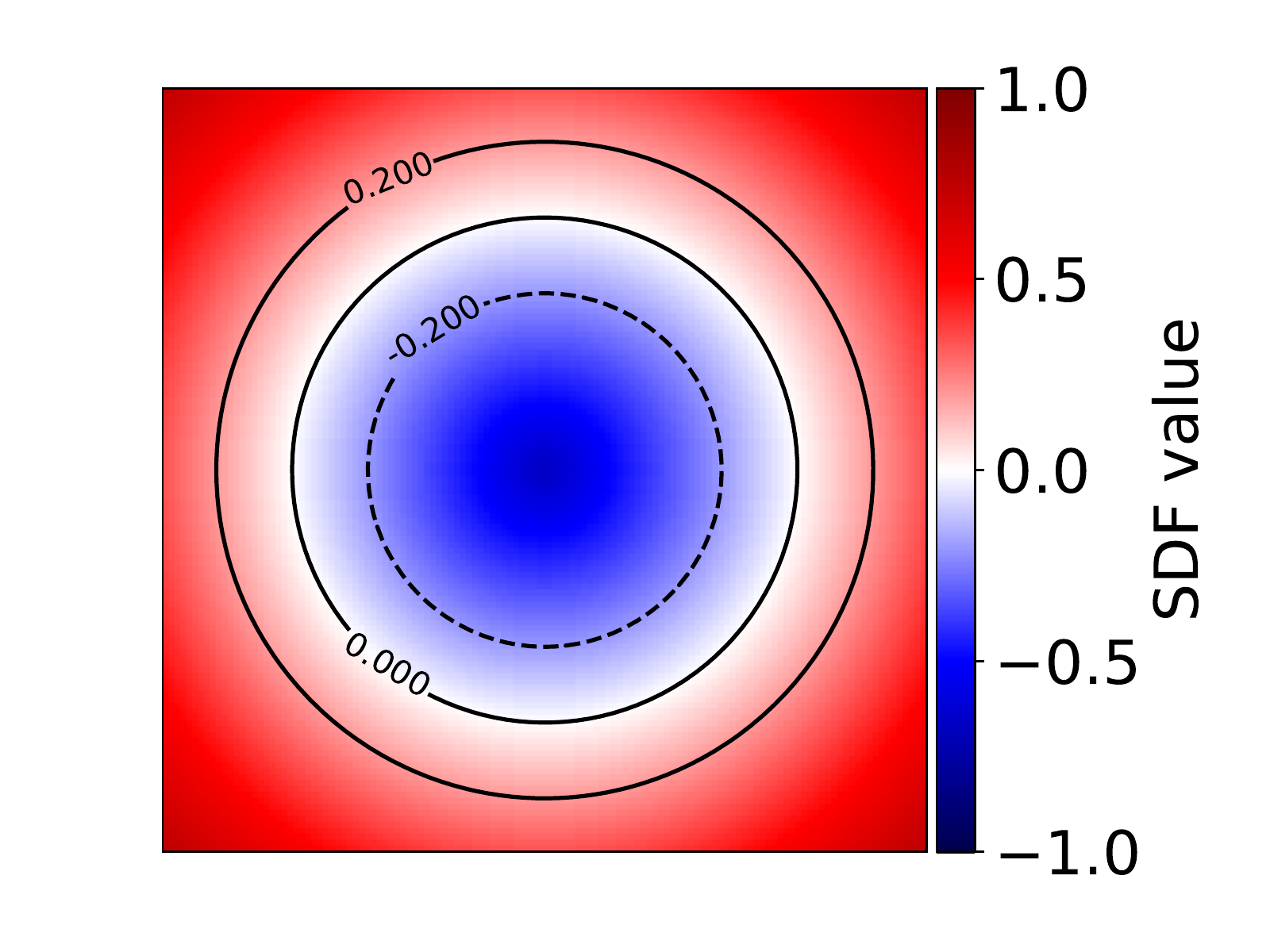}
	\end{subfigure}
	\caption{Example shapes for the sphere, box, rounded box and cylinder shape spaces from left to right. Mesh renderings are shown on the top and the bottom rows illustrate cuts through the respective SDFs along the $yz$- (left; green in the rendering) and $xy$-planes (right; red in the rendering). Note that for all rendered but the cylinder these cuts are identical. Note further, that the rounded box SDF is basically a box SDF for a smaller box with the 0-level set shifted further outside.}
	\label{fig:primitive_sdfs}
\end{figure*}

\paragraph{Primitive shapes}
In Fig. \ref{fig:primitive_sdfs}, we show example objects as rendered meshes as well as cuts through the SDFs for the primitive shapes used in the experiments.

\begin{figure*}
	\begin{subfigure}{.35\linewidth}
		\centering
		\includegraphics[width=\linewidth]{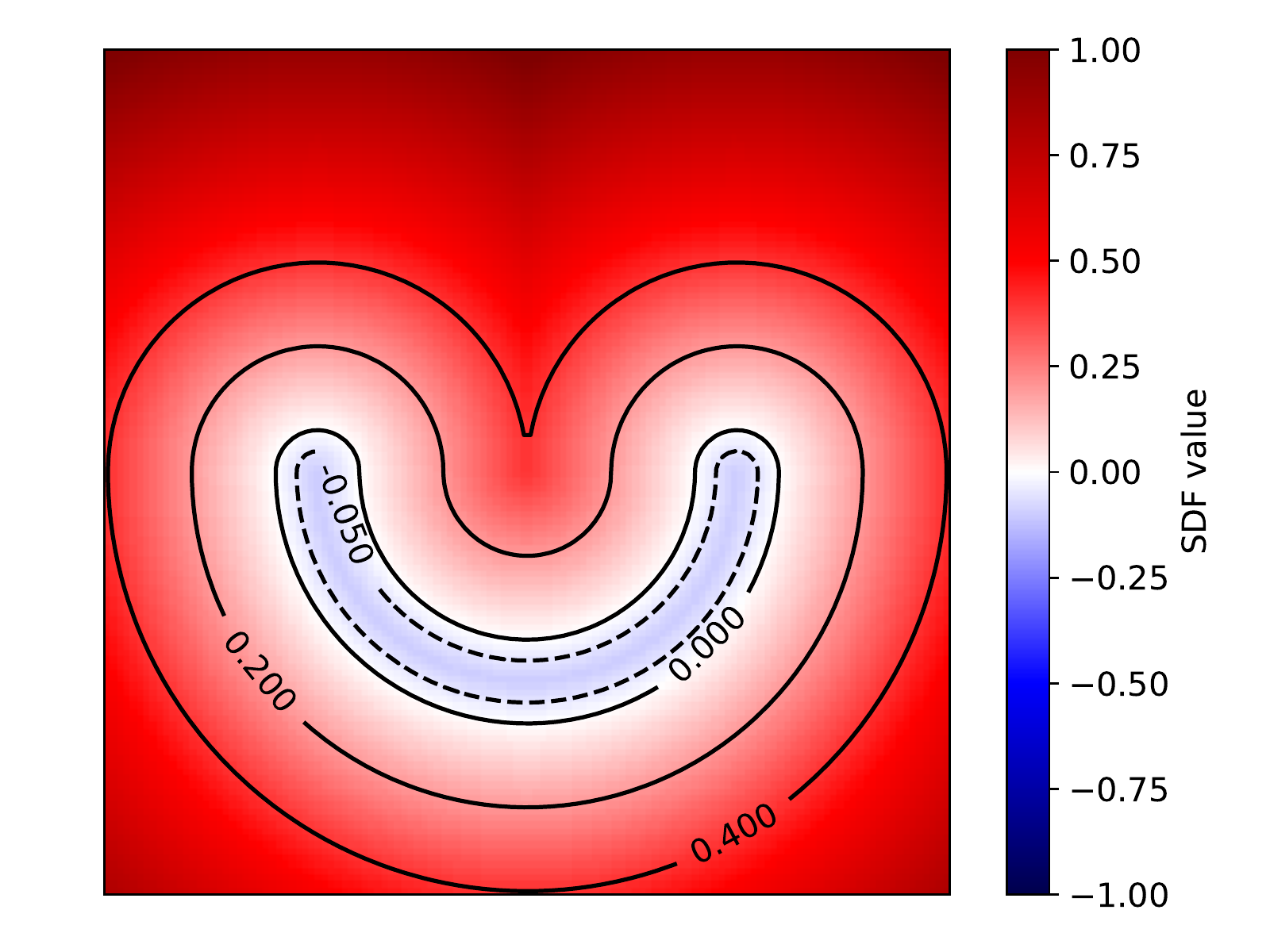}
	\end{subfigure}	
	\begin{subfigure}{.65\linewidth}
		\centering
		\includegraphics[trim=80 40 100 60,clip,width=0.25\linewidth]{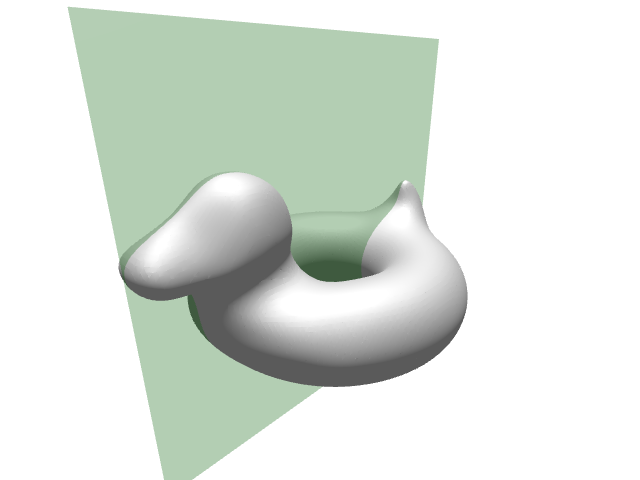}\hspace{0.05\linewidth}
		\includegraphics[trim=80 40 100 60,clip,width=0.25\linewidth]{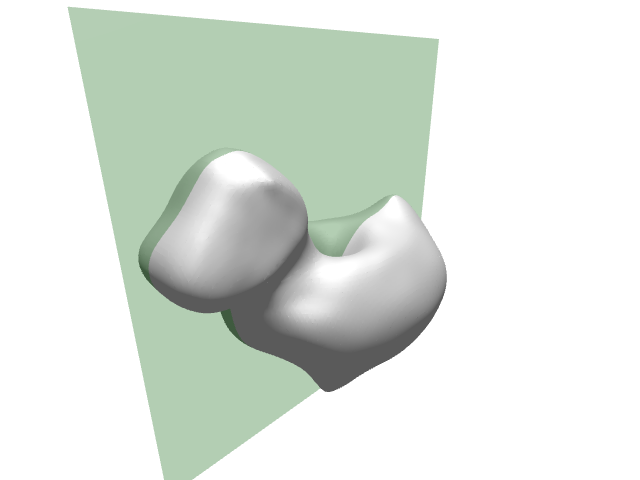}\hspace{0.05\linewidth}
		\includegraphics[trim=80 40 100 60,clip,width=0.25\linewidth]{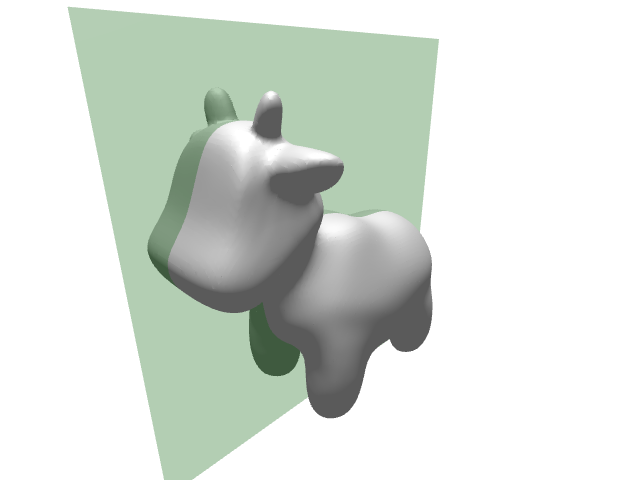}\\
		\includegraphics[width=0.3\linewidth]{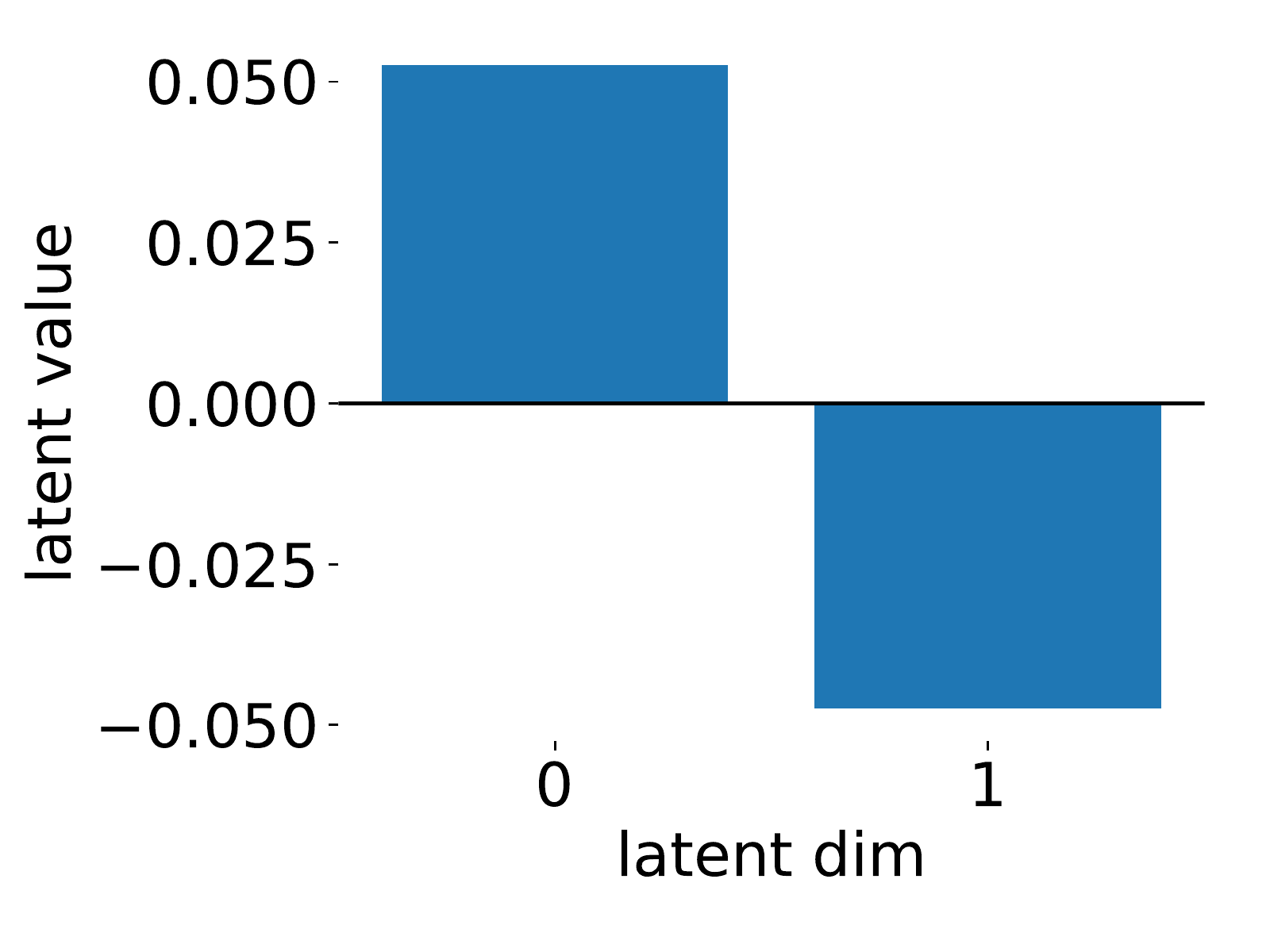}
		\includegraphics[width=0.3\linewidth]{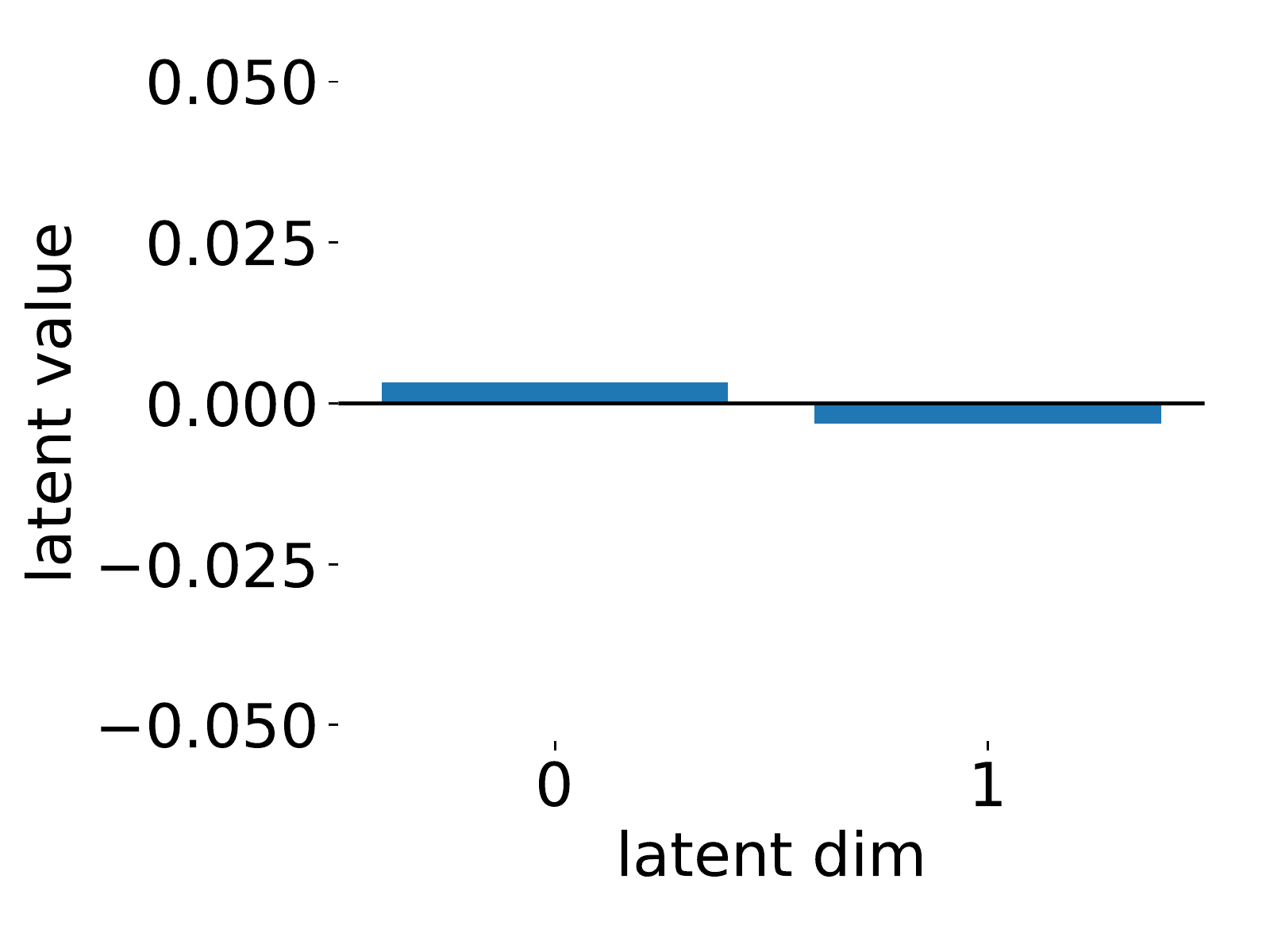}
		\includegraphics[width=0.3\linewidth]{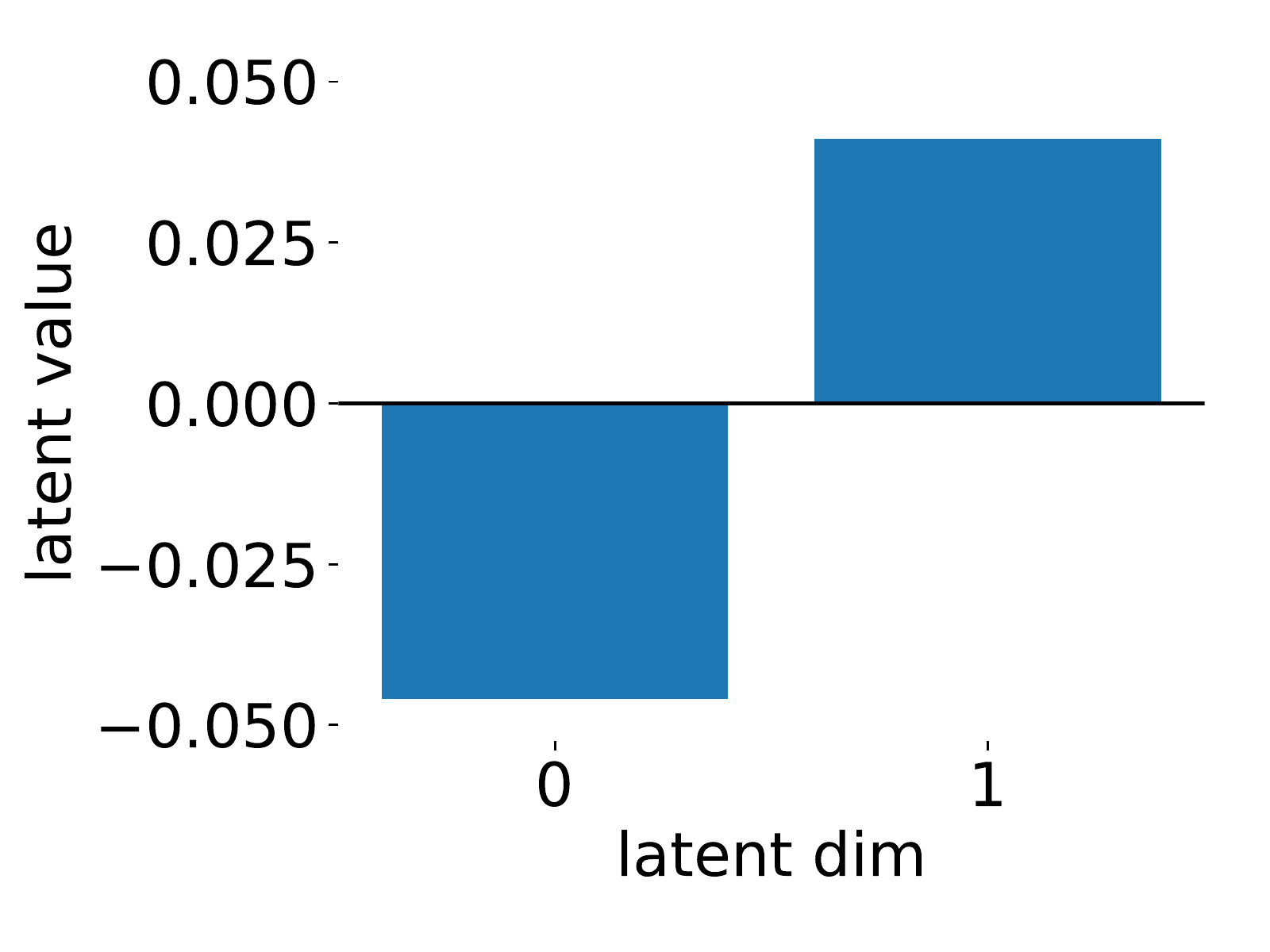}\\
		\includegraphics[width=0.3\linewidth]{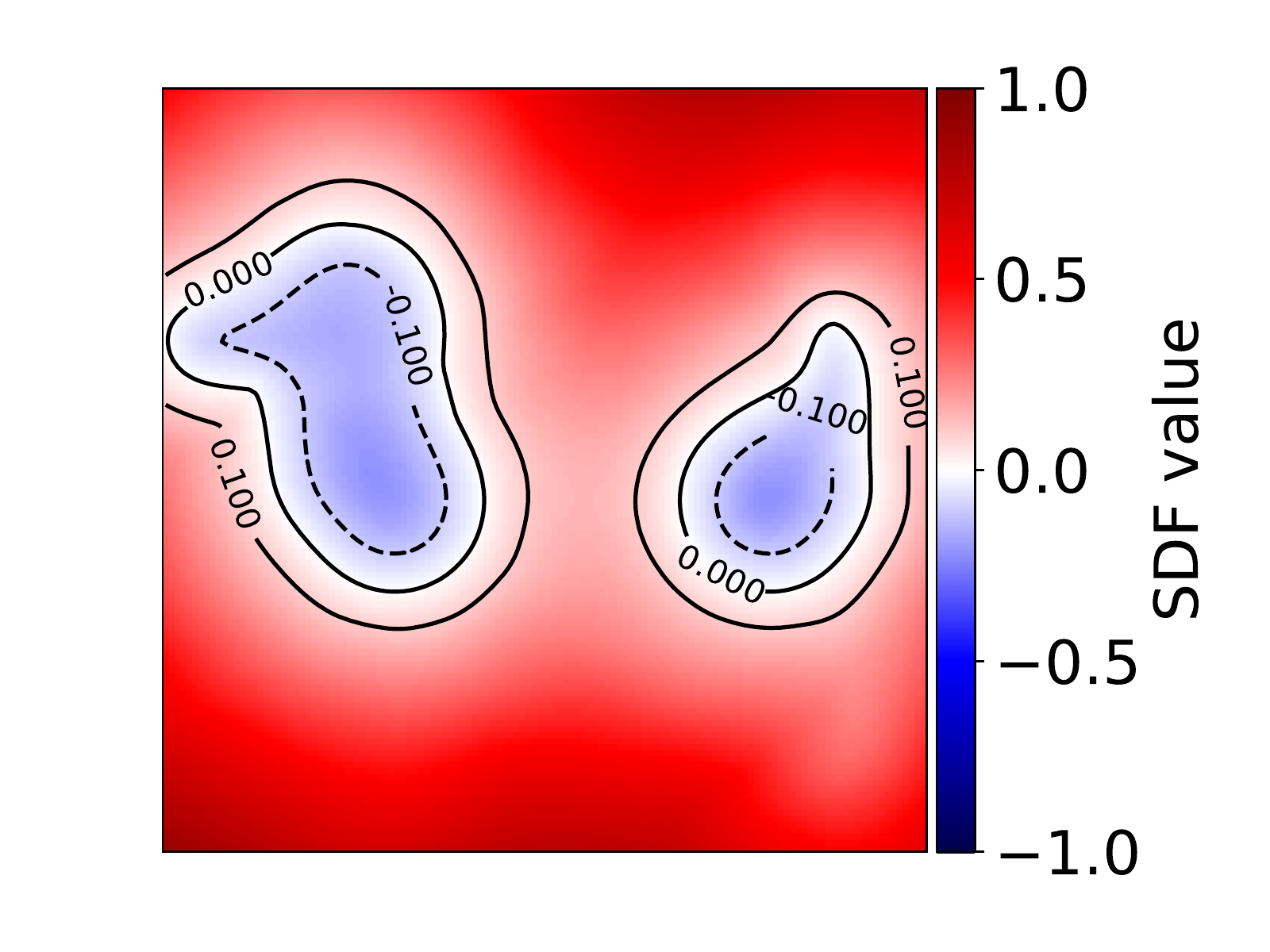}
		\includegraphics[width=0.3\linewidth]{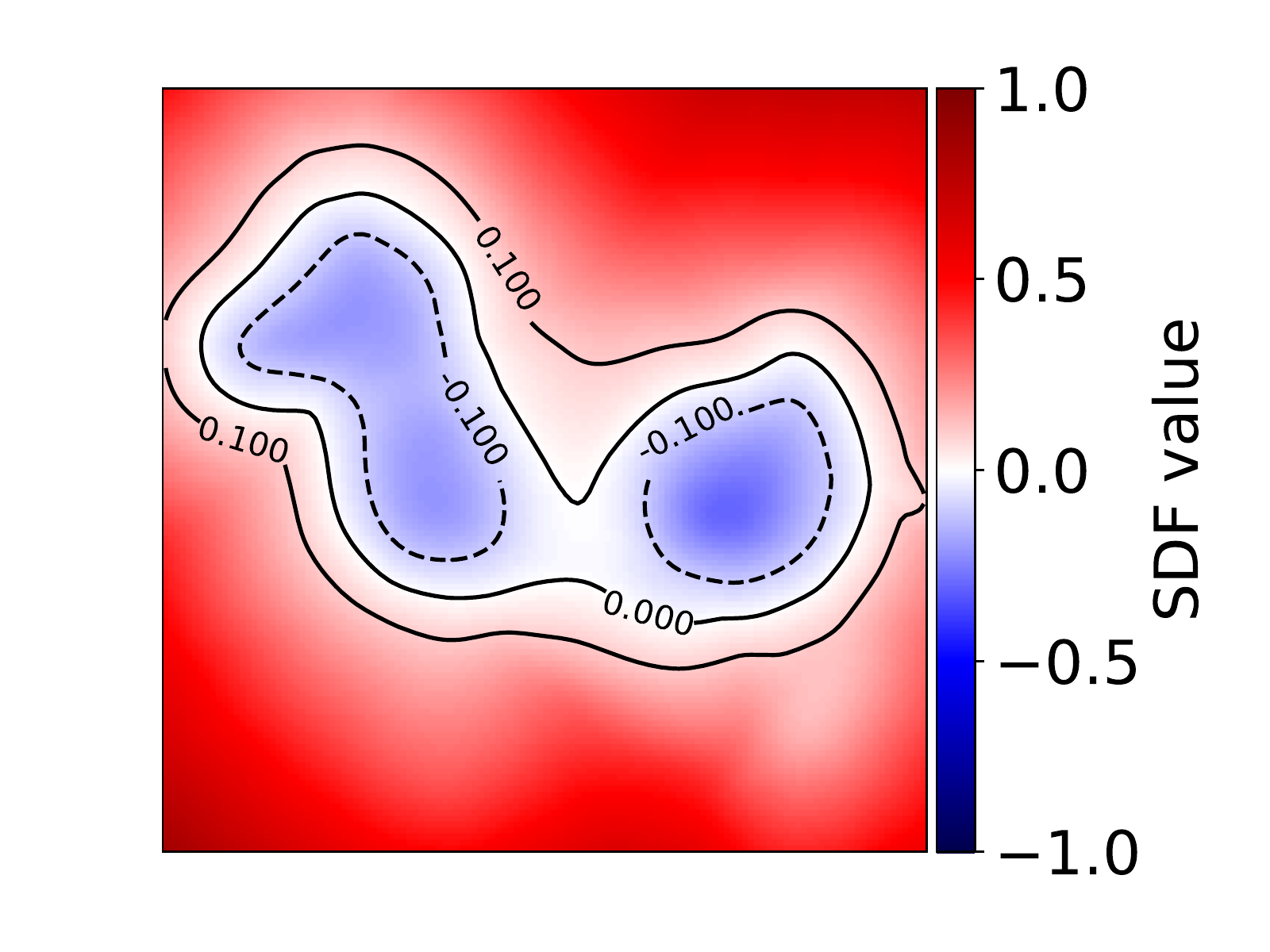}
		\includegraphics[width=0.3\linewidth]{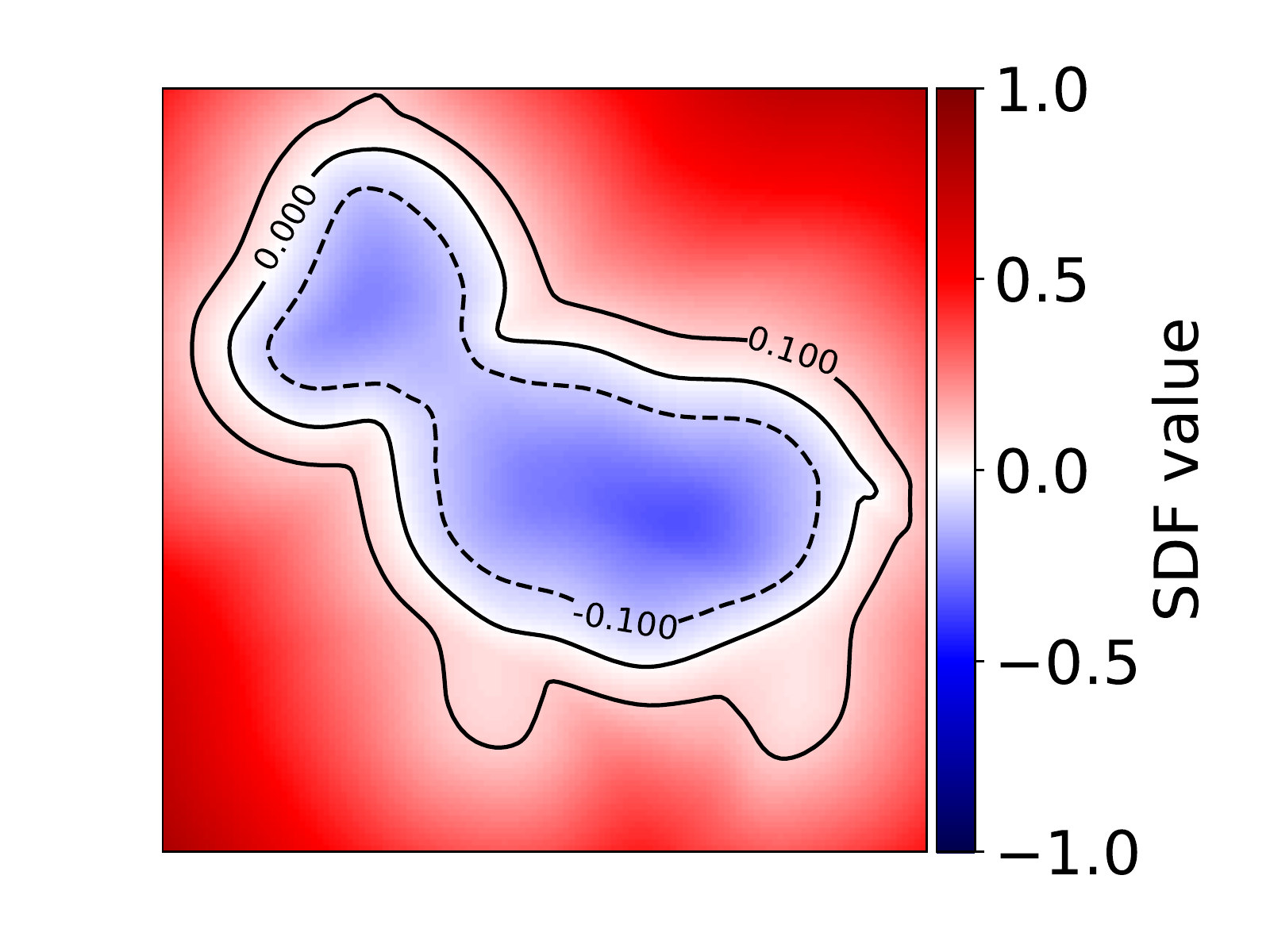}
	\end{subfigure}
	\caption{SDF shape representation. Left: cross section of a signed distance function of a bowl shape (surface is zero-level set). Right: Shapes (top) represented in an examplary two-dimenional SDF shape embedding with corresponding latent codes (middle) and cuts through the SDF along the $xy$-plane (bottom; the plane is indicated in green in the rendering on the top).}
	\label{fig:sdfs}
\end{figure*}

\paragraph{Learned shape spaces}
Fig. \ref{fig:sdfs} (left) illustrates how SDFs can encode nonconvex shapes like a bowl with information about the distance to the closest surface.
In Fig. \ref{fig:sdfs} (right) we show the generated meshes for the two training shapes ``bob'' and ``spot'' together with their two-dimensional encodings in their learned shape space.
We also show the ``mean shape'' for this shape space, \ie the result of interpolating half-way between bob and spot.
This shape space has latent size 2 and the DeepSDF \cite{park2019_deepsdf} auto-decoder has 8 layers with a hidden dimension of 128.

Figs. \ref{fig:renderings_can}, \ref{fig:renderings_camera} and \ref{fig:renderings_mug} show the generated meshes for the training shapes of the can, camera and mug classes, respectively.
These shape spaces are trained with a latent size of 4 and the DeepSDF \cite{park2019_deepsdf} auto-decoder has 8 layers with a hidden dimension of 256.

All learned shape spaces were trained using the implicit geometric regularization losses from \cite{gropp2020_implicit}, to encourage the network to learn actual distances to the surface.

\begin{figure*}
	\centering
	\input{figures/shapespace_renderings/can}
	\caption{Renderings (left) and cuts through the $xy$-plane (right; the plane is indicated in green in the rendering) for the 24 can objects in their learned shapespace. Each object is represented by a four-dimensional latent code.}
	\label{fig:renderings_can}
\end{figure*}

\begin{figure*}
	\centering
	\input{figures/shapespace_renderings/camera}
	\caption{Renderings (left) and cuts through the $xy$-plane (right; the plane is indicated in green in the rendering) for the 22 can objects in their learned shapespace. Each object is represented by a four-dimensional latent code.}
	\label{fig:renderings_camera}
\end{figure*}

\begin{figure*}
	\centering
	\input{figures/shapespace_renderings/mug}
	\caption{Renderings (left) and cuts through the $xy$-plane (right; the plane is indicated in green in the rendering) for the 22 can objects in their learned shapespace. Each object is represented by a four-dimensional latent code.}
	\label{fig:renderings_mug}
\end{figure*}

%% file: figures/shapespace_renderings/can.tex
\begin{subfigure}{.22500\linewidth}
\includegraphics[trim=80 30 100 30,clip,width=.39\linewidth]{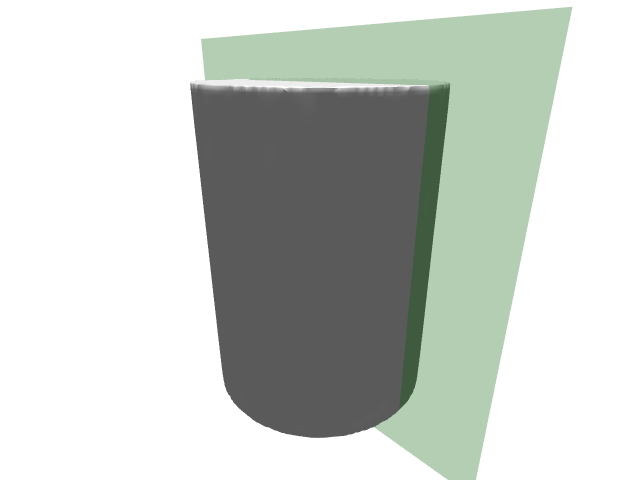}
\includegraphics[width=.59\linewidth]{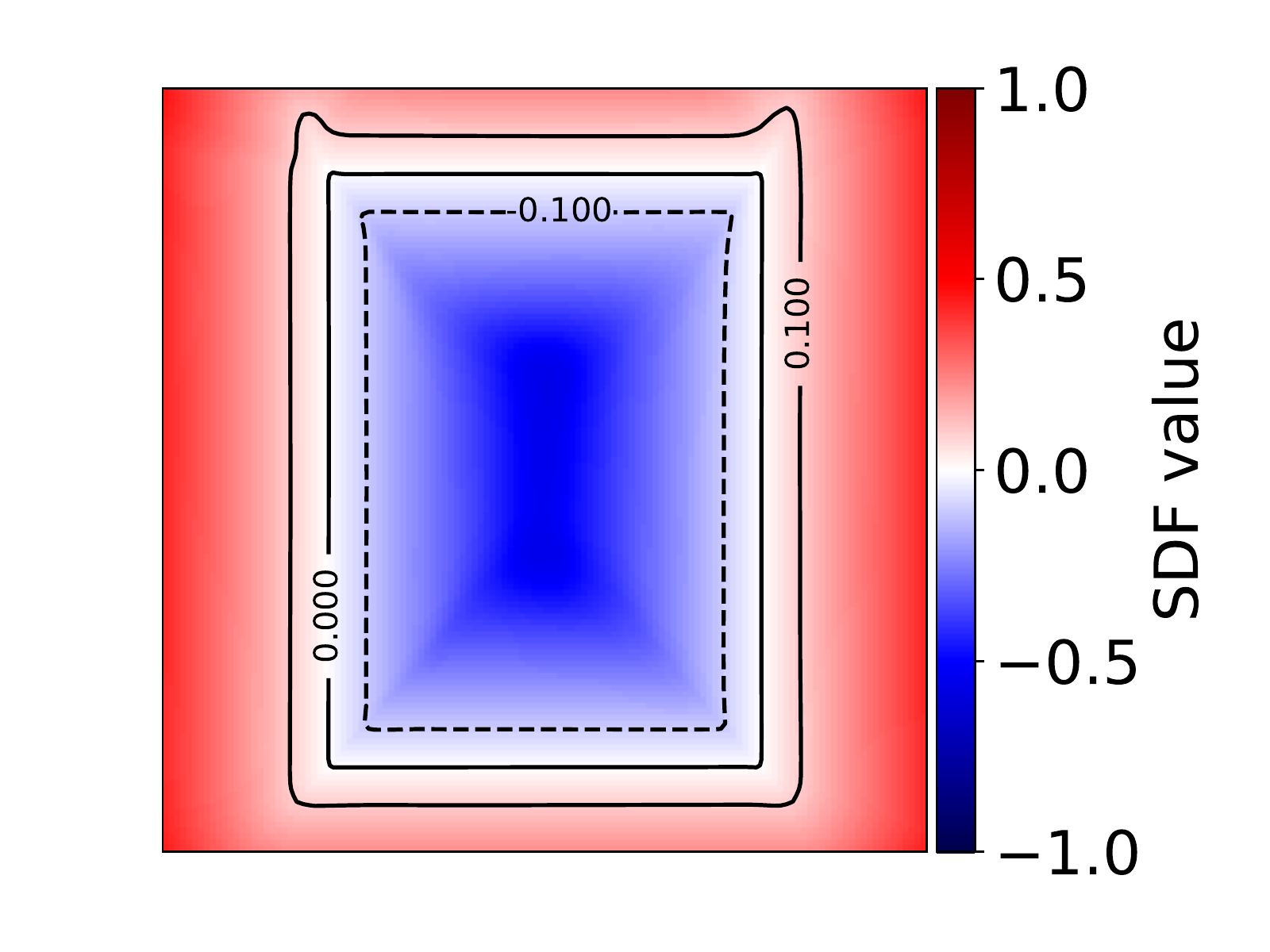}
\end{subfigure}
\begin{subfigure}{.22500\linewidth}
\includegraphics[trim=80 30 100 30,clip,width=.39\linewidth]{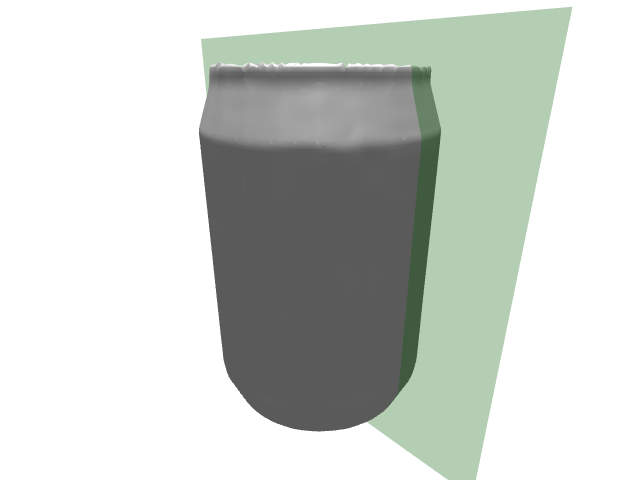}
\includegraphics[width=.59\linewidth]{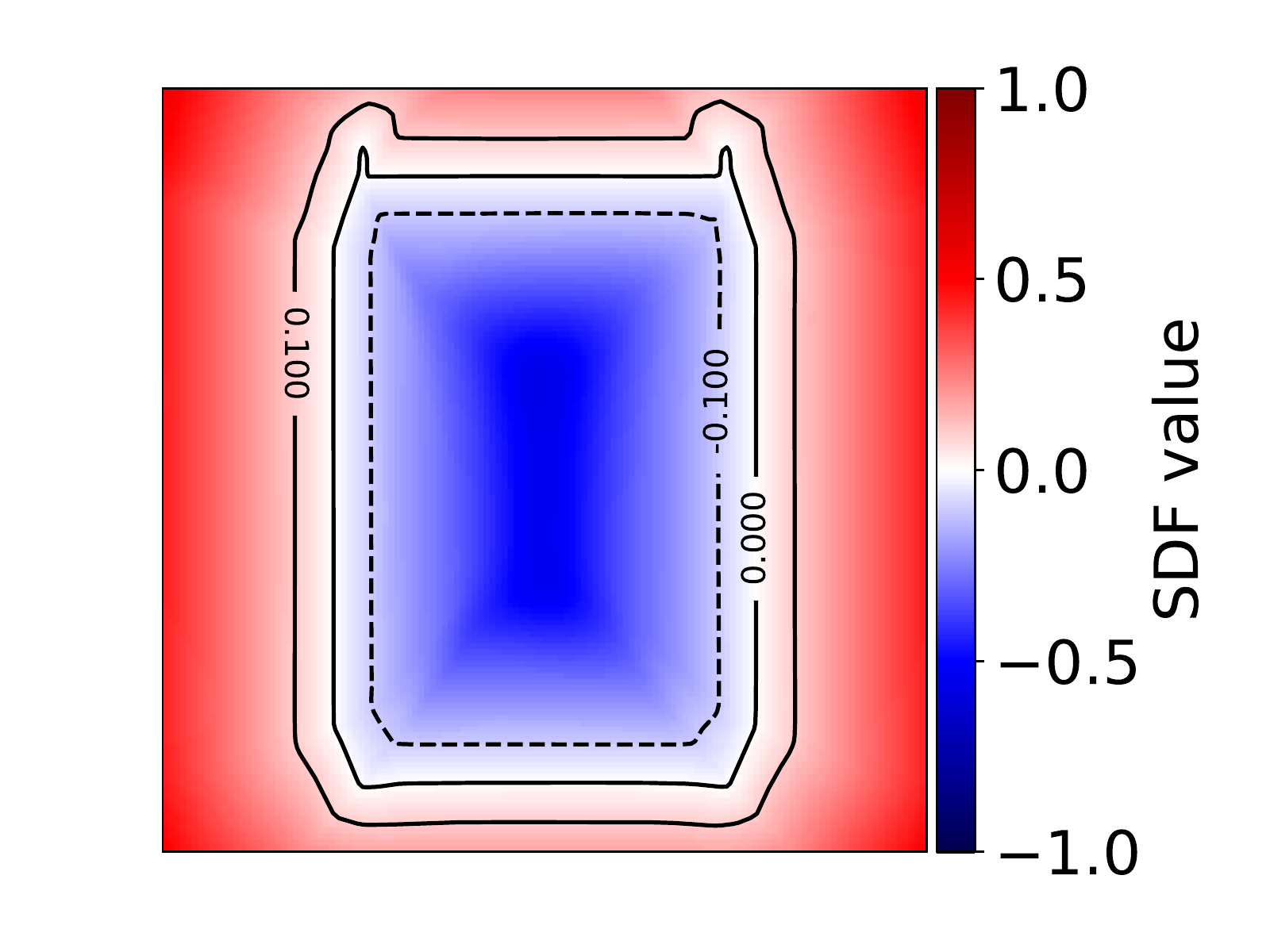}
\end{subfigure}
\begin{subfigure}{.22500\linewidth}
\includegraphics[trim=80 30 100 30,clip,width=.39\linewidth]{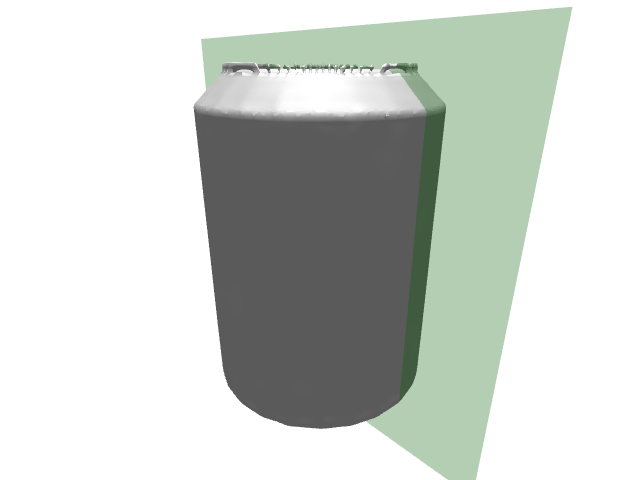}
\includegraphics[width=.59\linewidth]{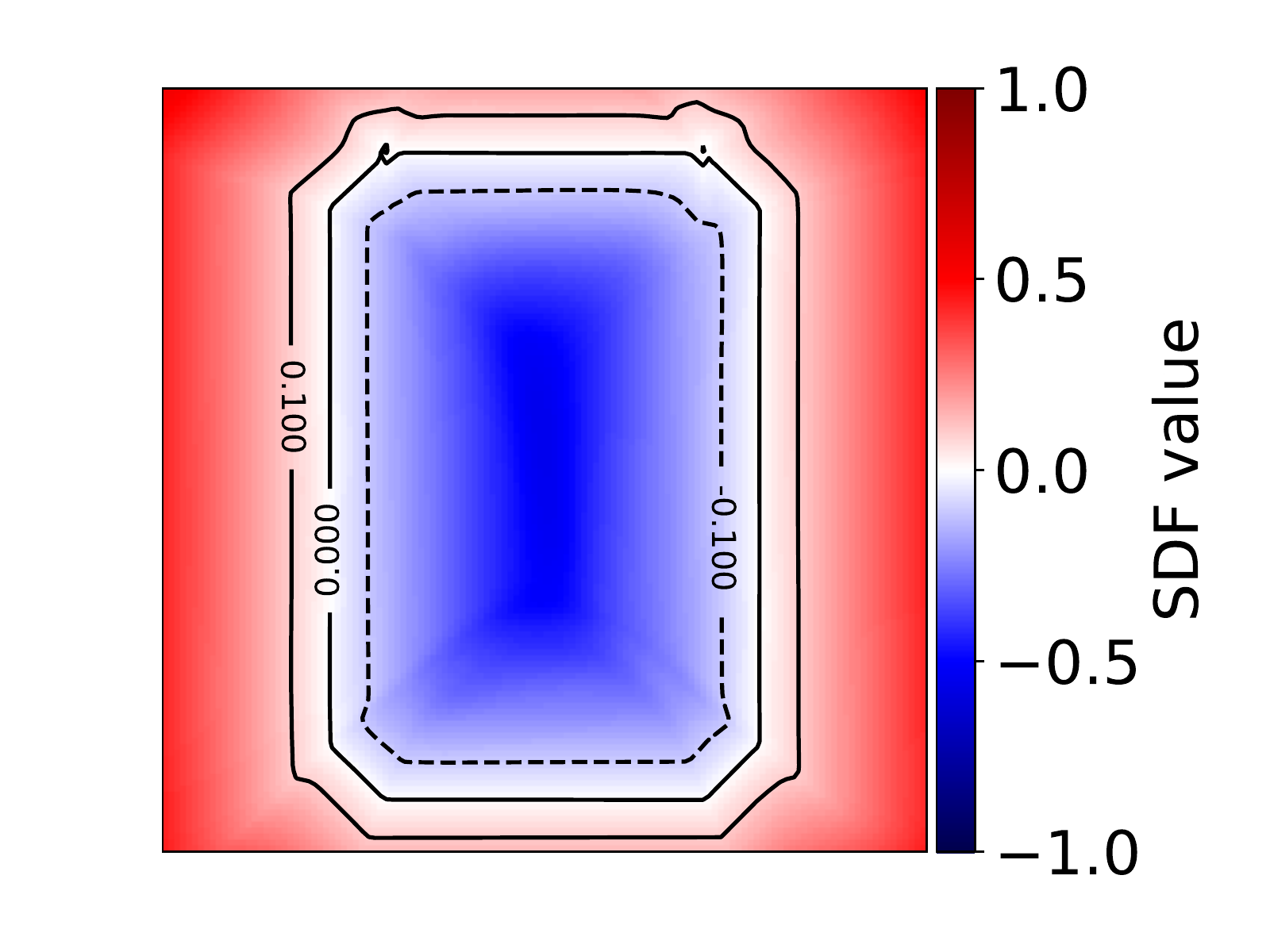}
\end{subfigure}
\begin{subfigure}{.22500\linewidth}
\includegraphics[trim=80 30 100 30,clip,width=.39\linewidth]{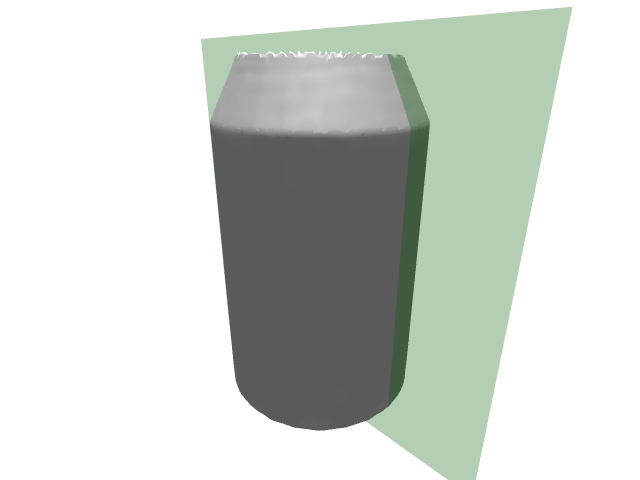}
\includegraphics[width=.59\linewidth]{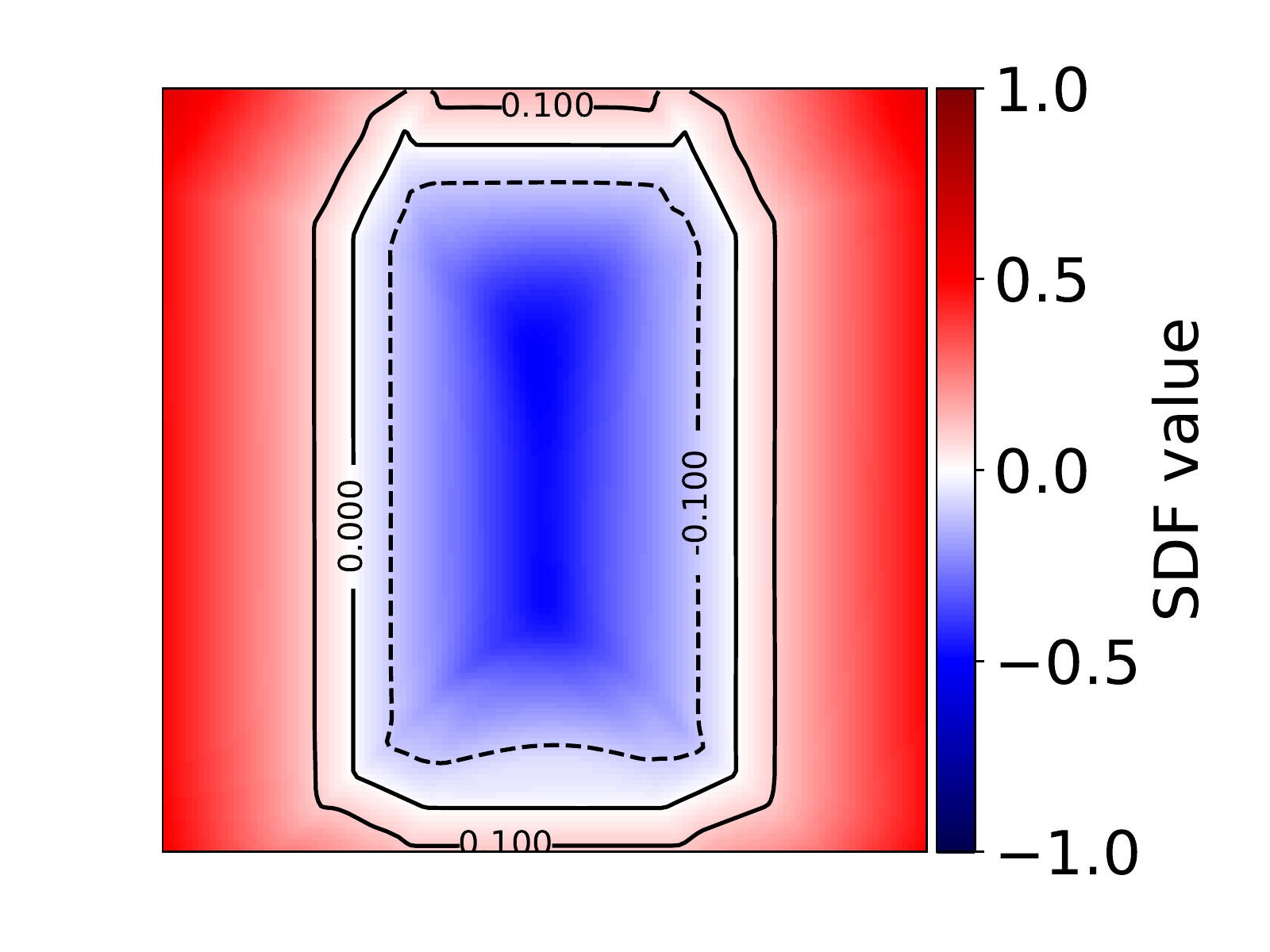}
\end{subfigure}
\begin{subfigure}{.22500\linewidth}
\includegraphics[trim=80 30 100 30,clip,width=.39\linewidth]{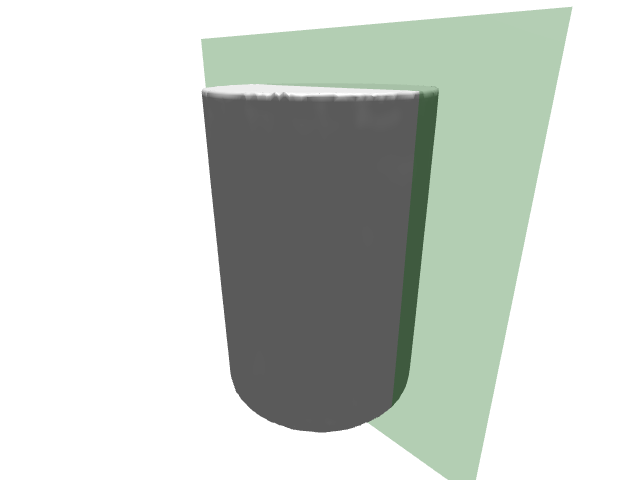}
\includegraphics[width=.59\linewidth]{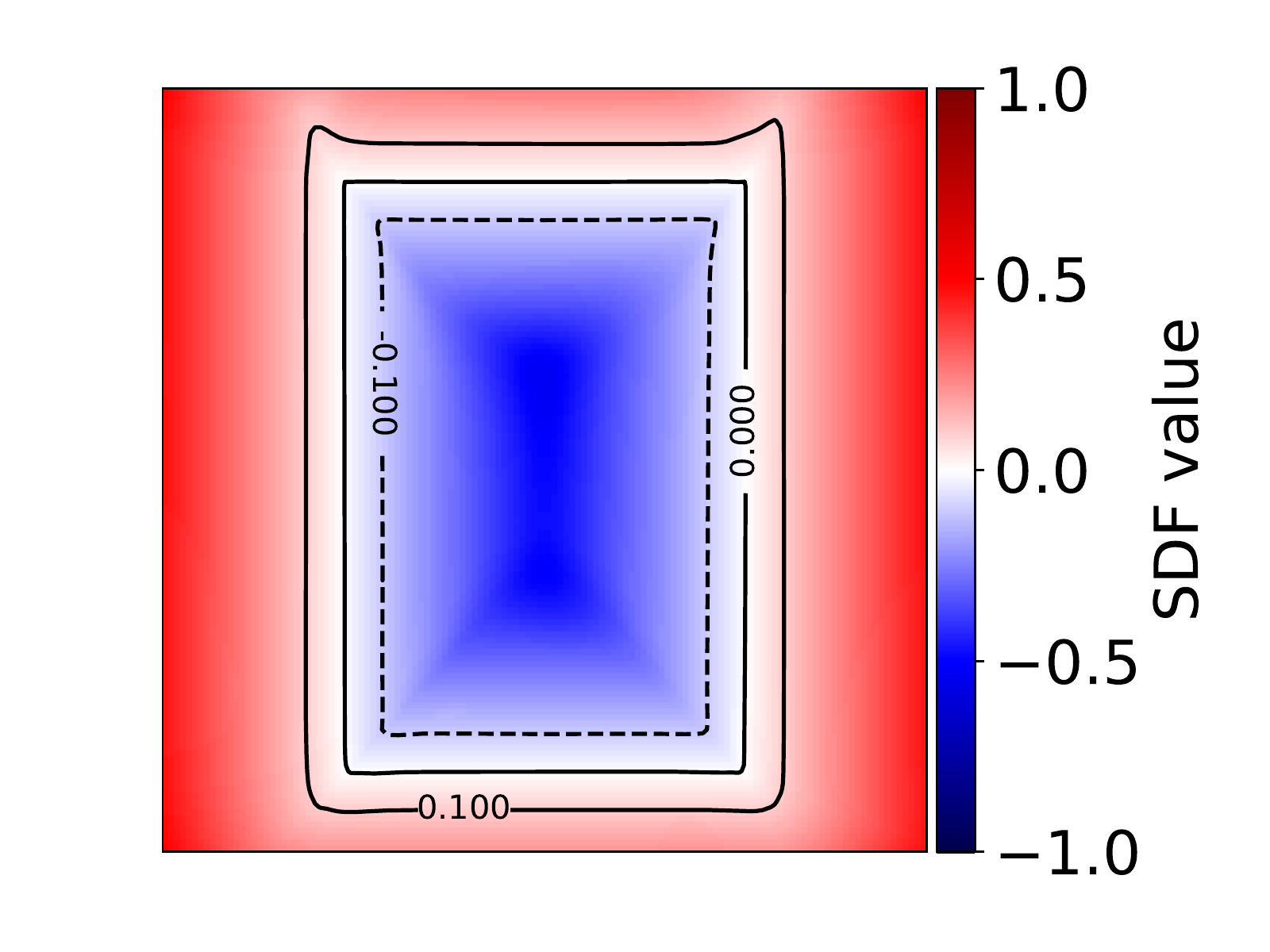}
\end{subfigure}
\begin{subfigure}{.22500\linewidth}
\includegraphics[trim=80 30 100 30,clip,width=.39\linewidth]{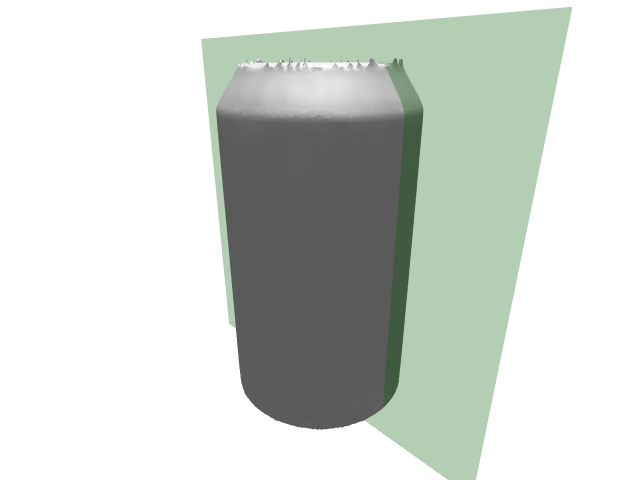}
\includegraphics[width=.59\linewidth]{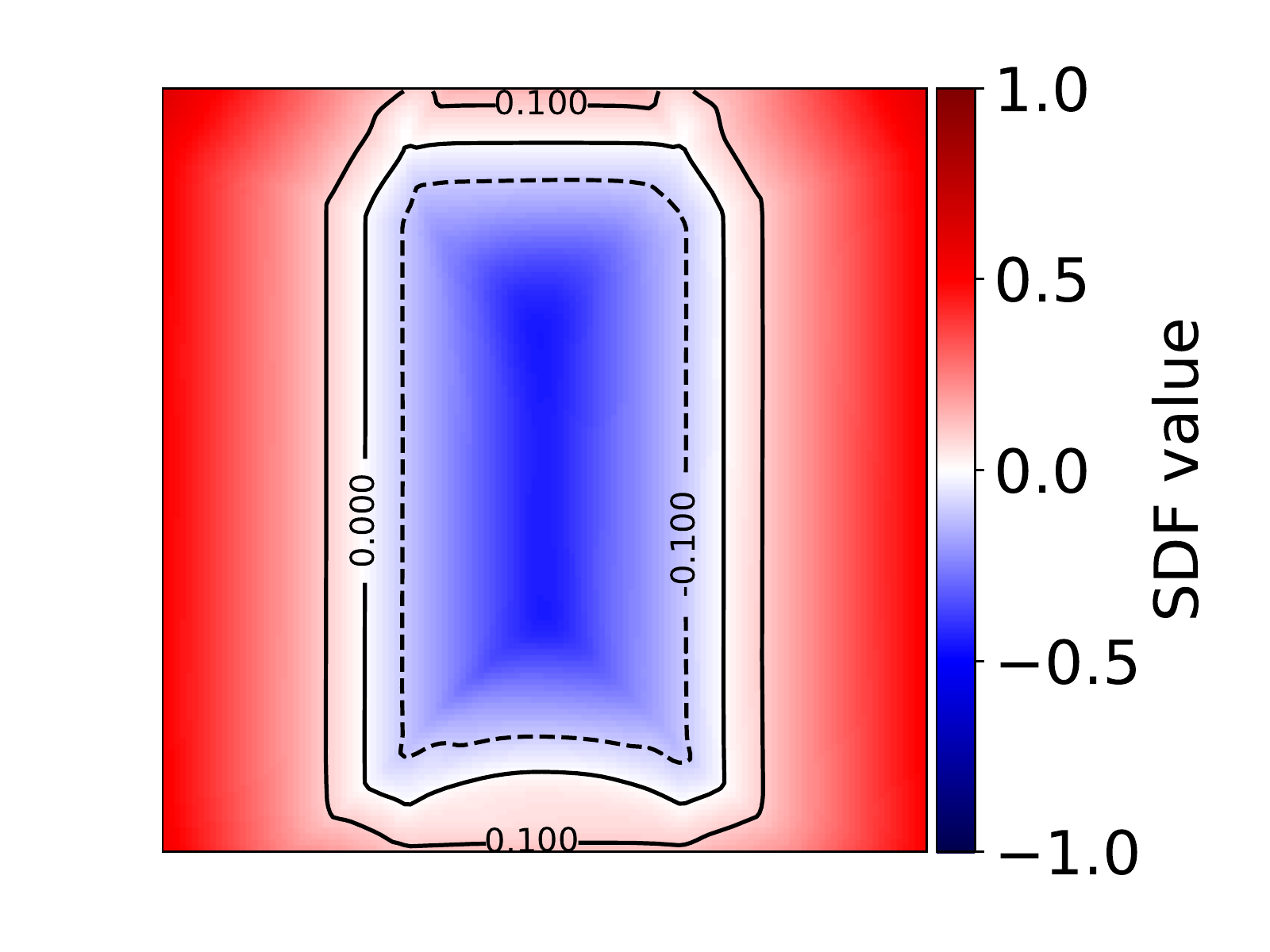}
\end{subfigure}
\begin{subfigure}{.22500\linewidth}
\includegraphics[trim=80 30 100 30,clip,width=.39\linewidth]{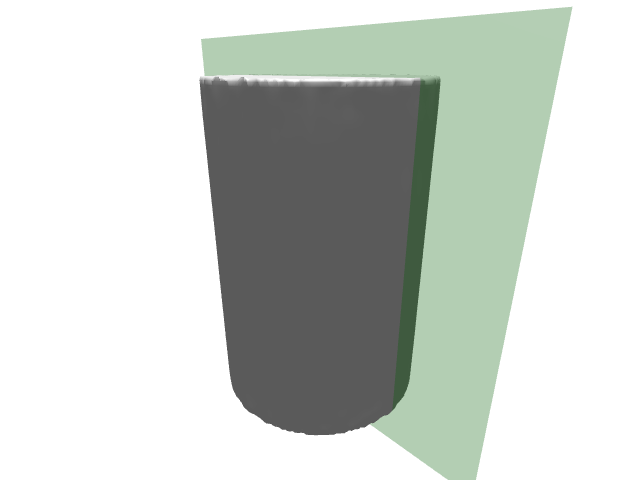}
\includegraphics[width=.59\linewidth]{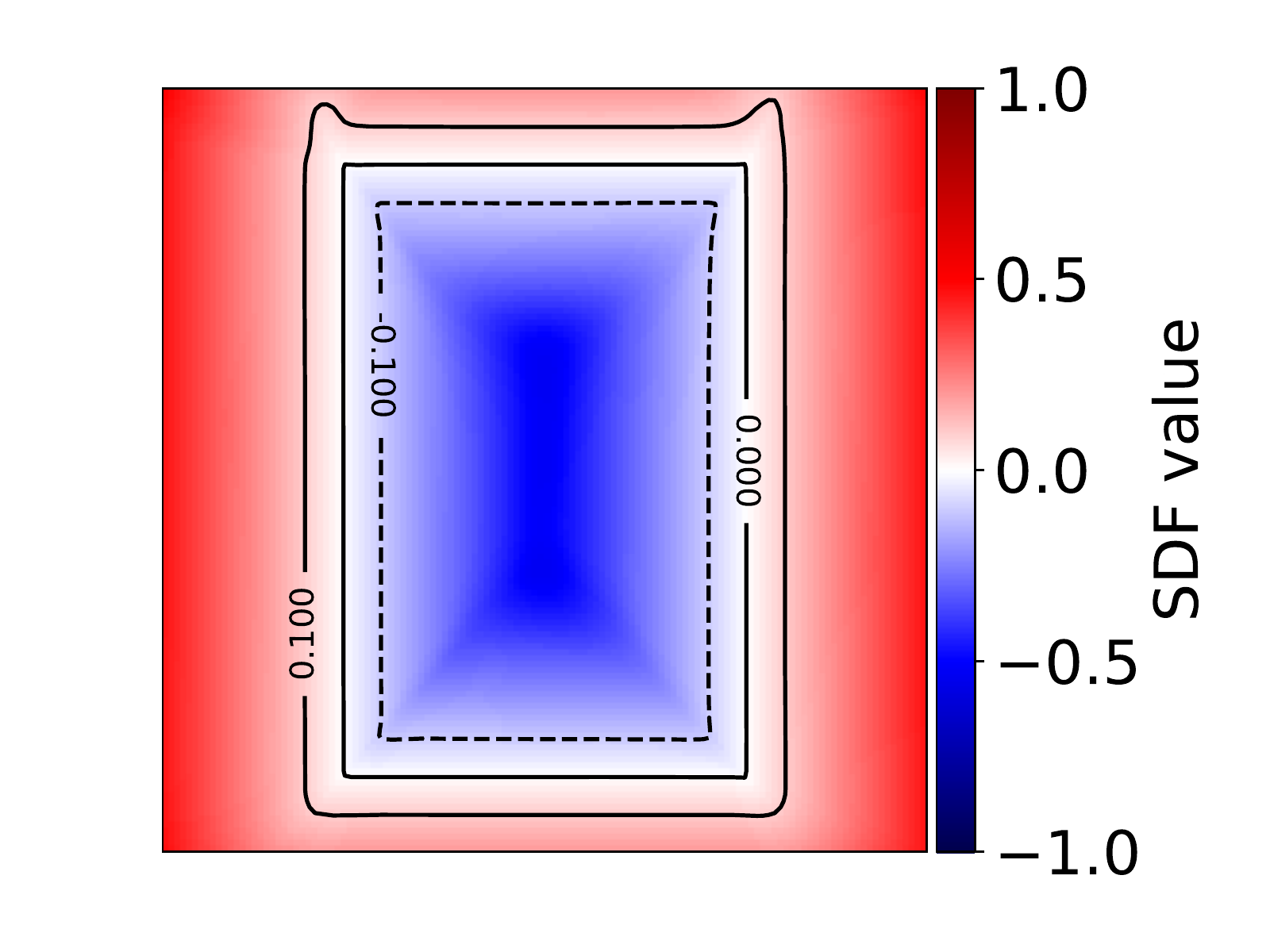}
\end{subfigure}
\begin{subfigure}{.22500\linewidth}
\includegraphics[trim=80 30 100 30,clip,width=.39\linewidth]{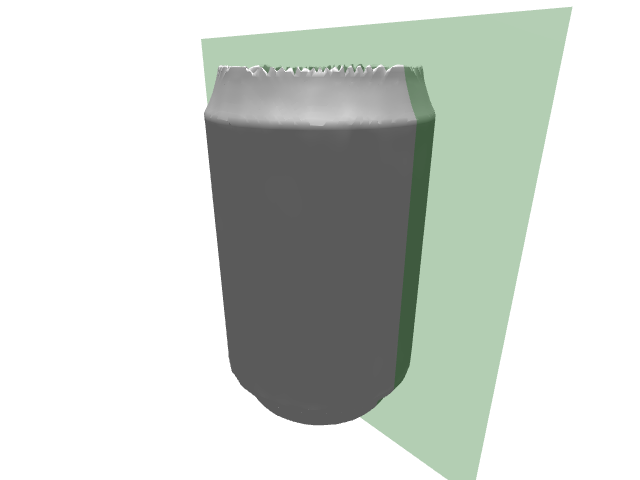}
\includegraphics[width=.59\linewidth]{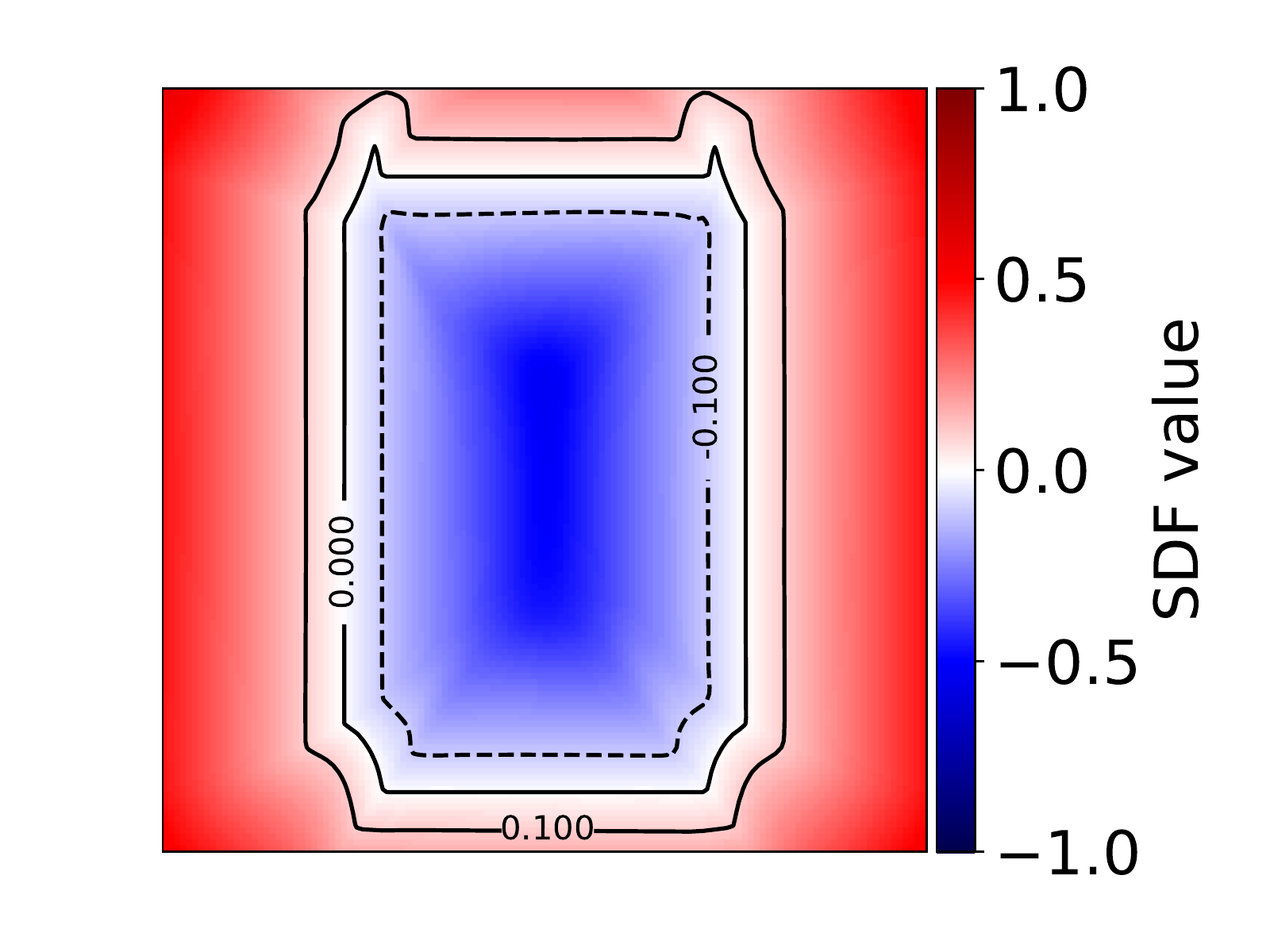}
\end{subfigure}
\begin{subfigure}{.22500\linewidth}
\includegraphics[trim=80 30 100 30,clip,width=.39\linewidth]{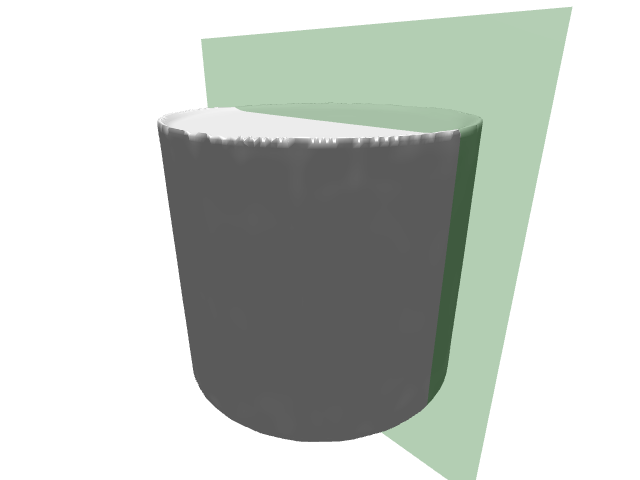}
\includegraphics[width=.59\linewidth]{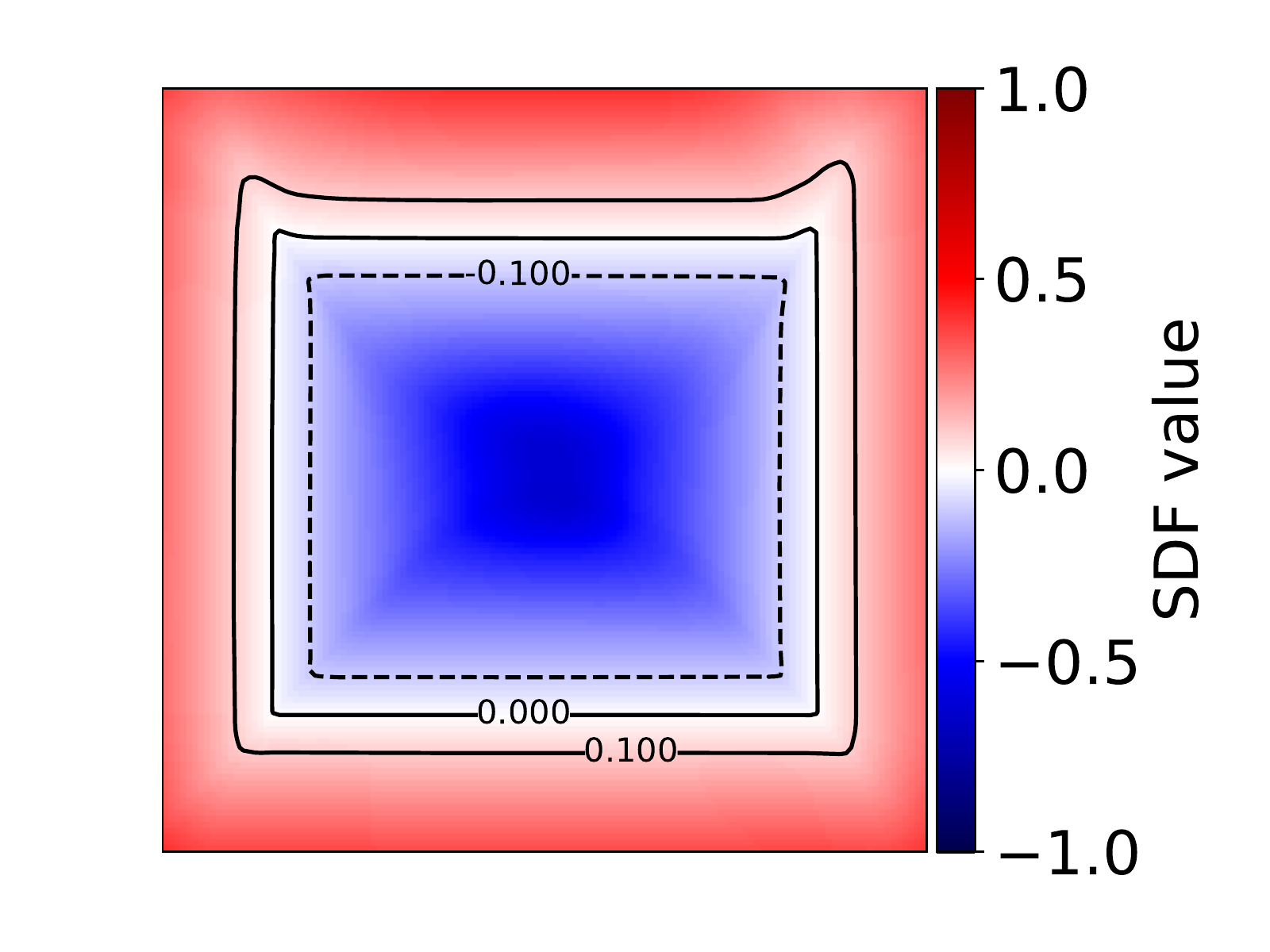}
\end{subfigure}
\begin{subfigure}{.22500\linewidth}
\includegraphics[trim=80 30 100 30,clip,width=.39\linewidth]{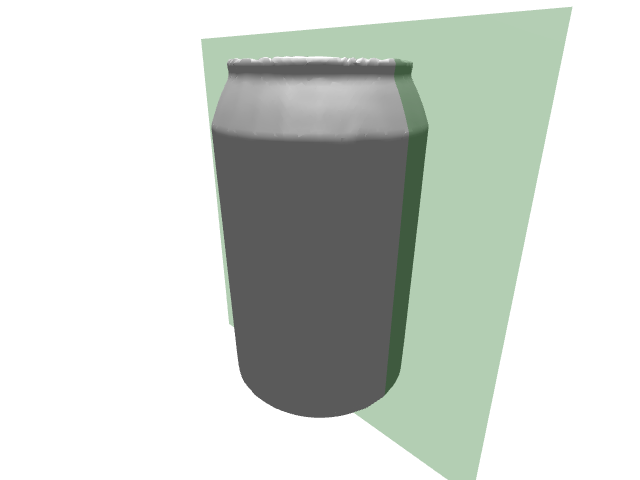}
\includegraphics[width=.59\linewidth]{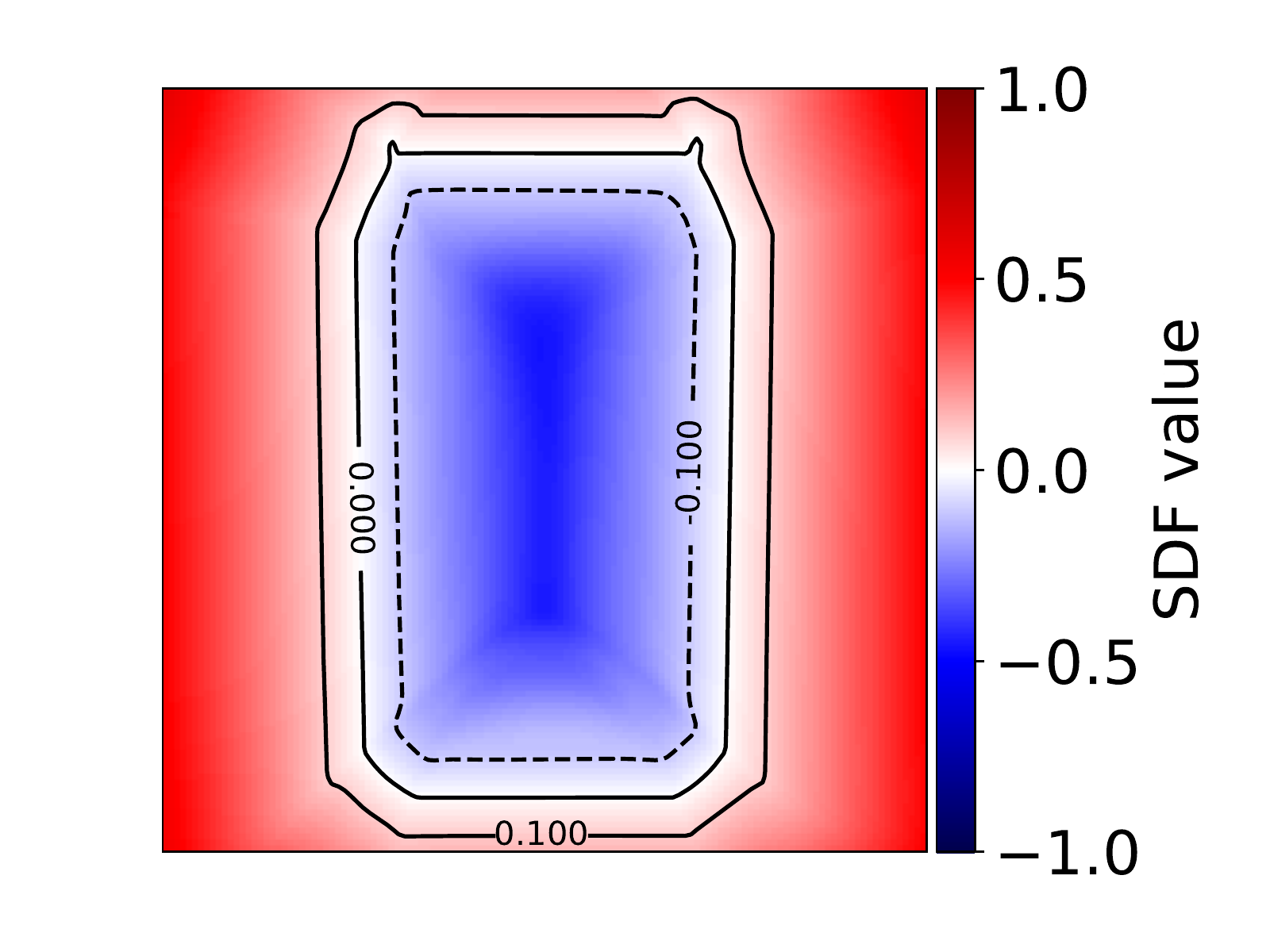}
\end{subfigure}
\begin{subfigure}{.22500\linewidth}
\includegraphics[trim=80 30 100 30,clip,width=.39\linewidth]{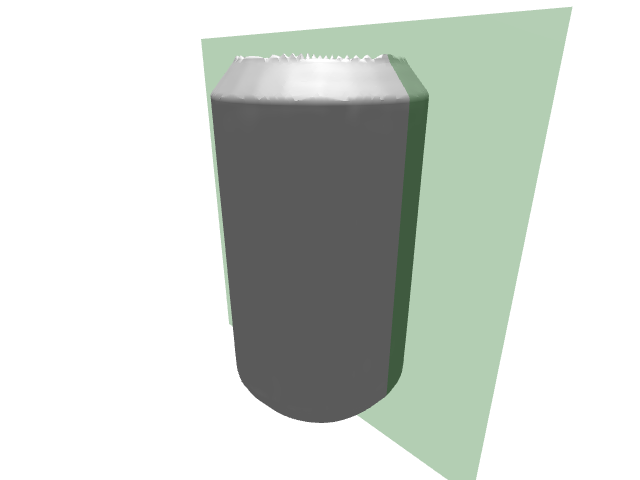}
\includegraphics[width=.59\linewidth]{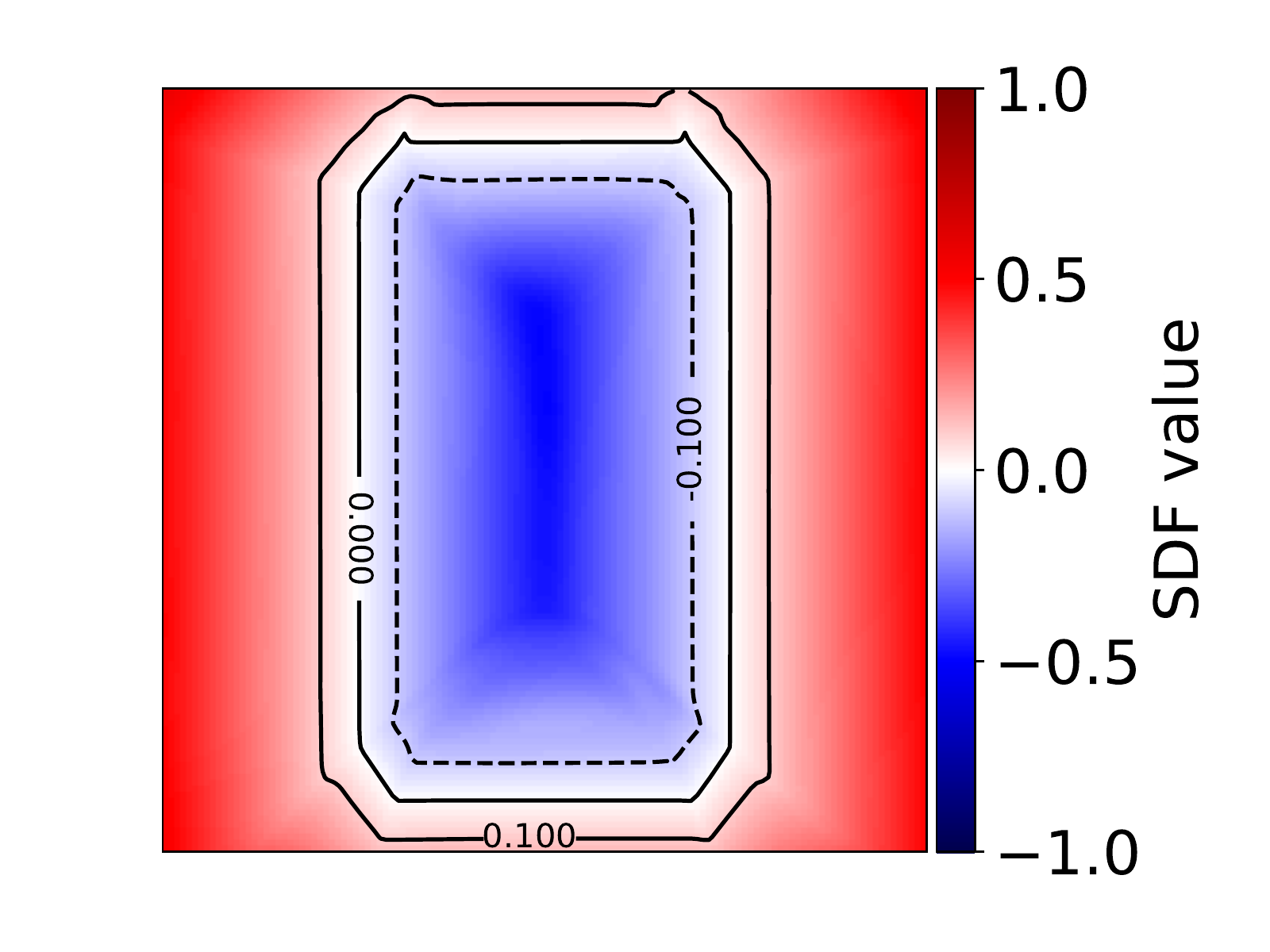}
\end{subfigure}
\begin{subfigure}{.22500\linewidth}
\includegraphics[trim=80 30 100 30,clip,width=.39\linewidth]{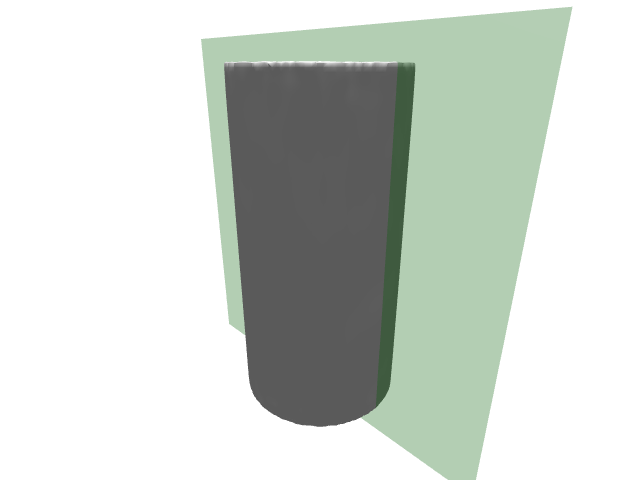}
\includegraphics[width=.59\linewidth]{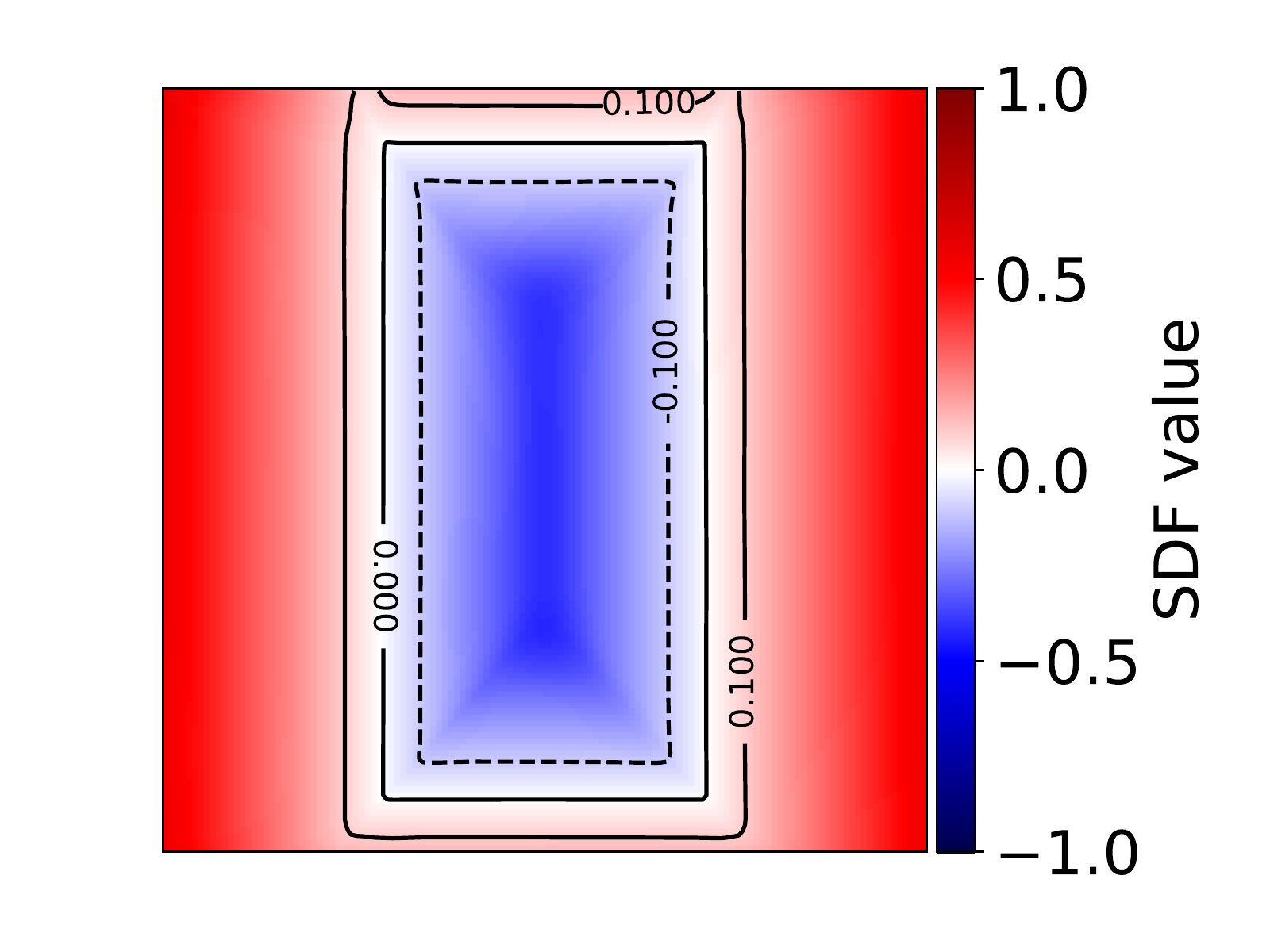}
\end{subfigure}
\begin{subfigure}{.22500\linewidth}
\includegraphics[trim=80 30 100 30,clip,width=.39\linewidth]{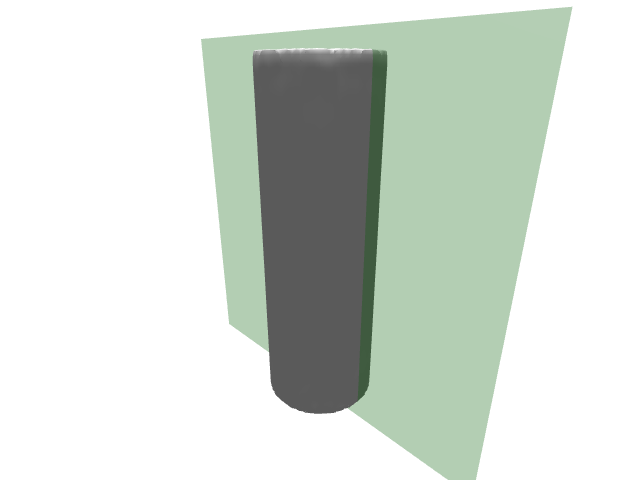}
\includegraphics[width=.59\linewidth]{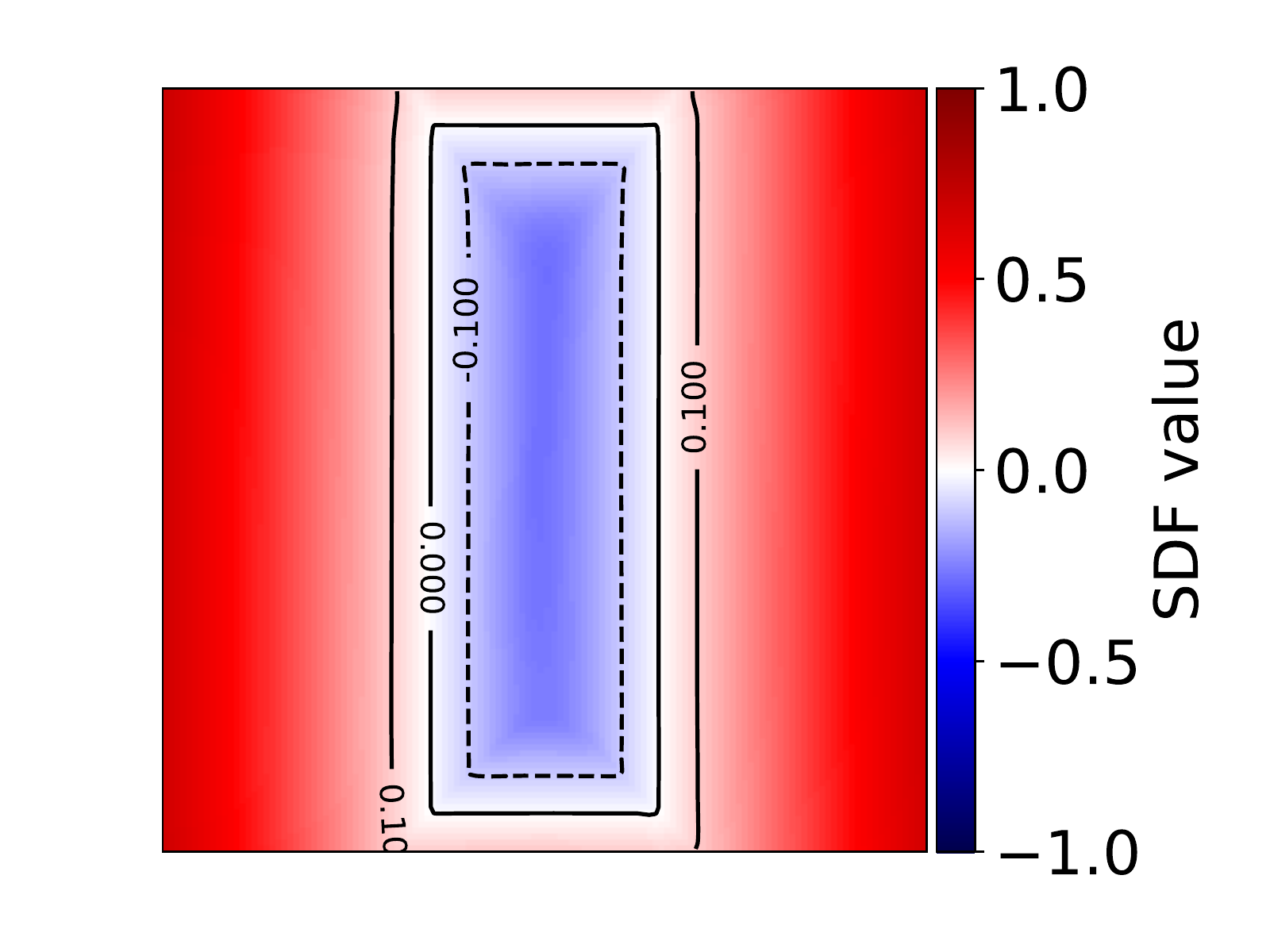}
\end{subfigure}
\begin{subfigure}{.22500\linewidth}
\includegraphics[trim=80 30 100 30,clip,width=.39\linewidth]{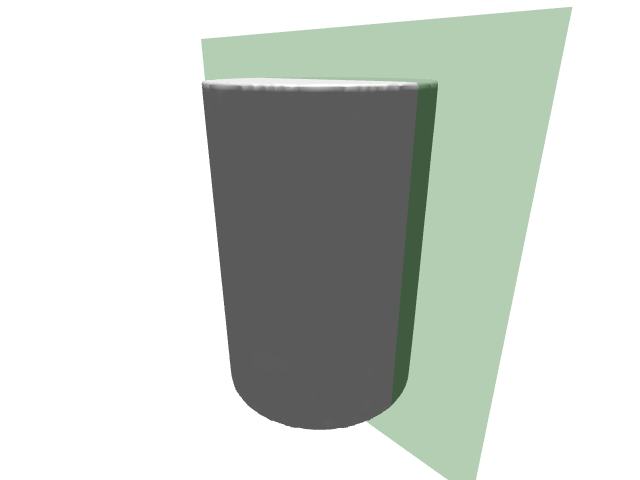}
\includegraphics[width=.59\linewidth]{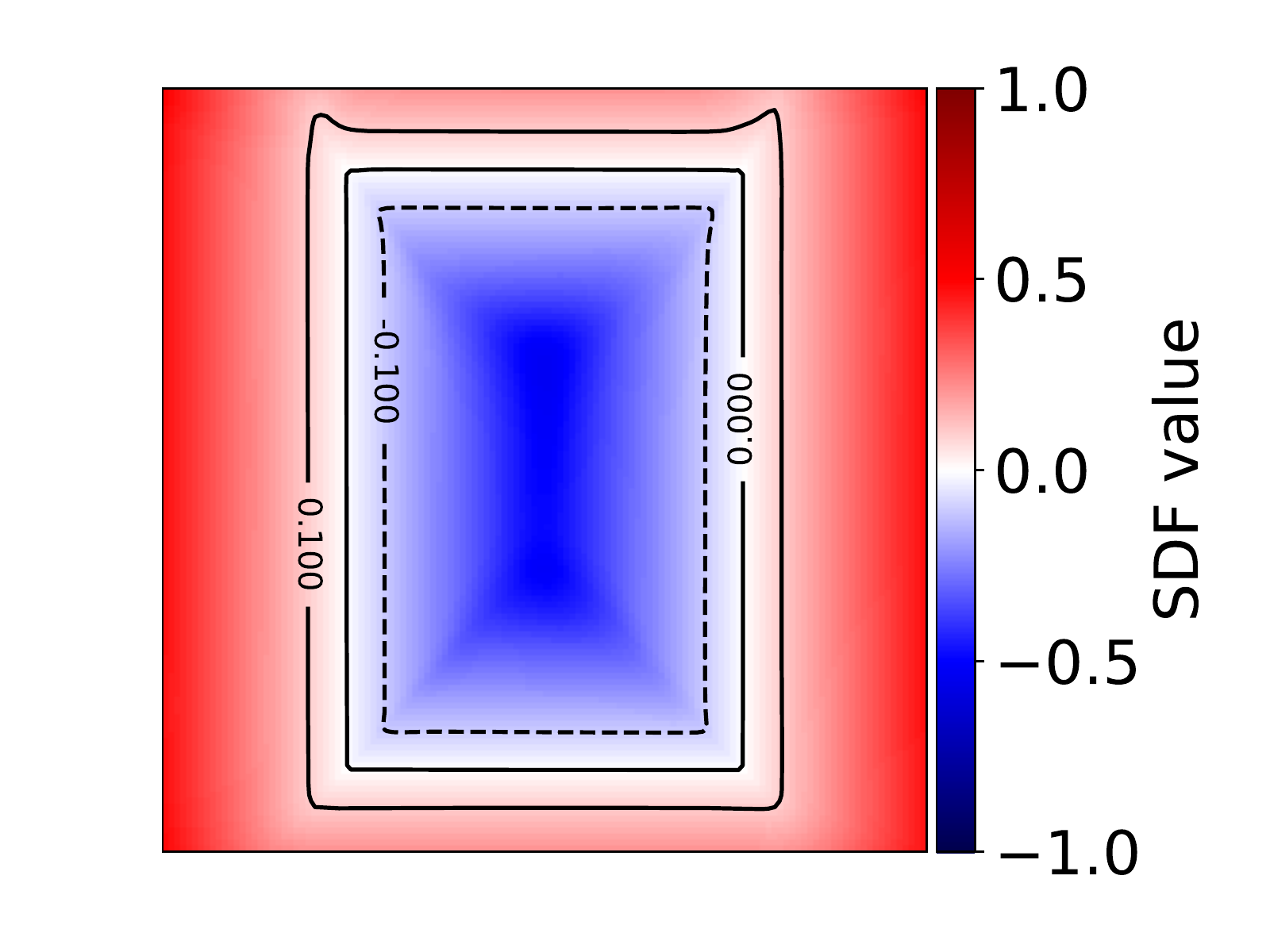}
\end{subfigure}
\begin{subfigure}{.22500\linewidth}
\includegraphics[trim=80 30 100 30,clip,width=.39\linewidth]{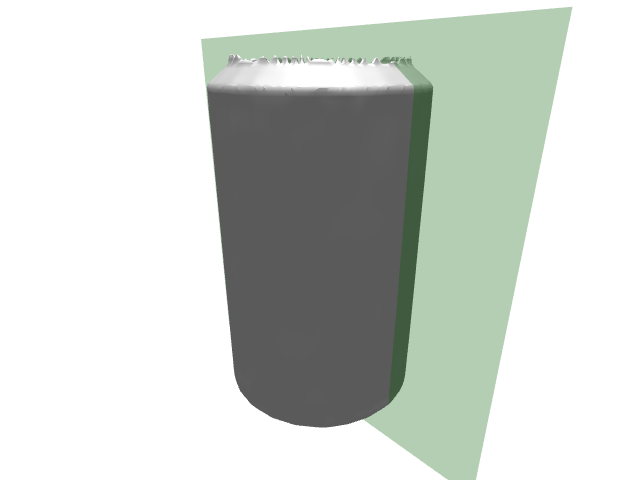}
\includegraphics[width=.59\linewidth]{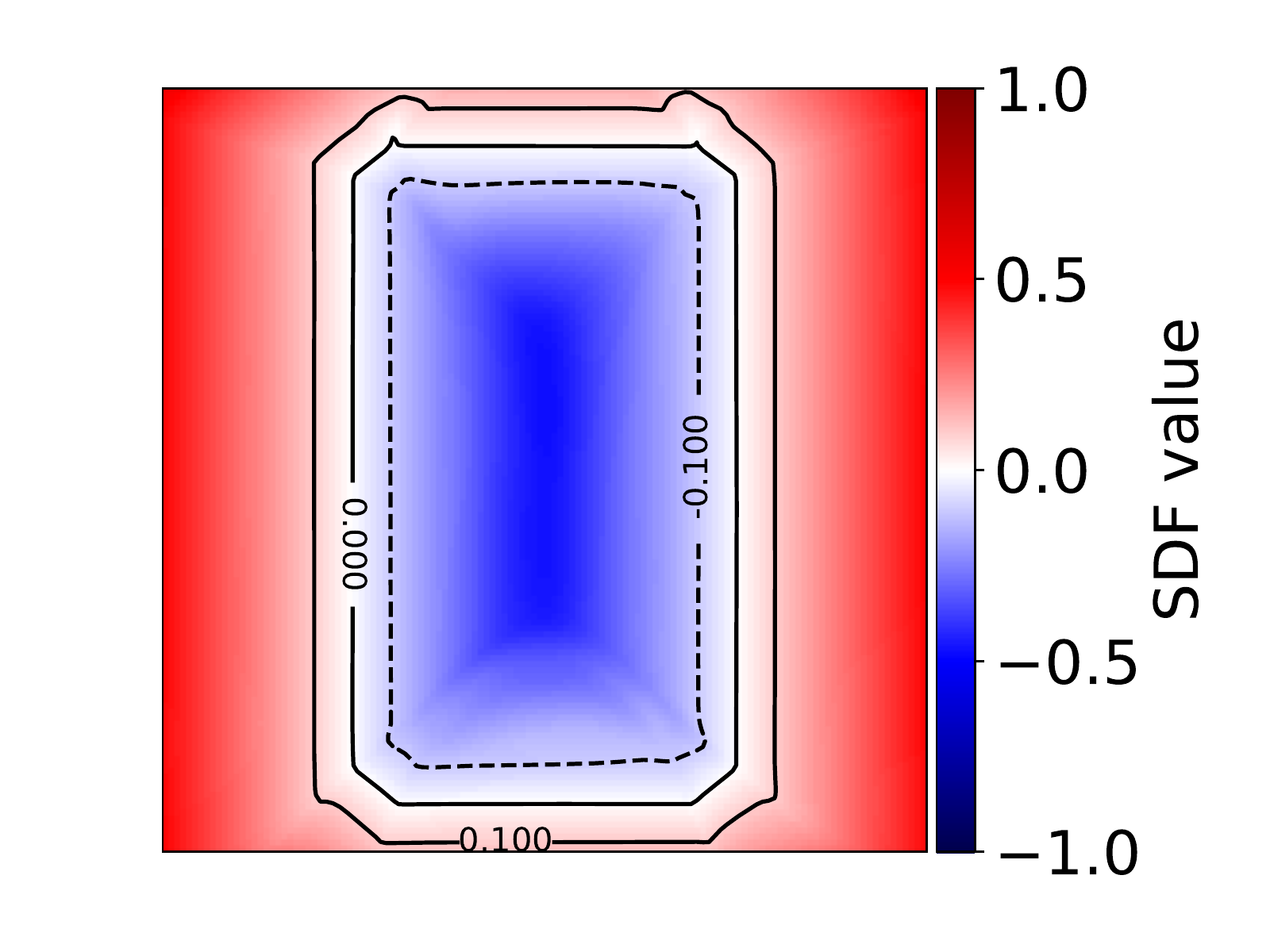}
\end{subfigure}
\begin{subfigure}{.22500\linewidth}
\includegraphics[trim=80 30 100 30,clip,width=.39\linewidth]{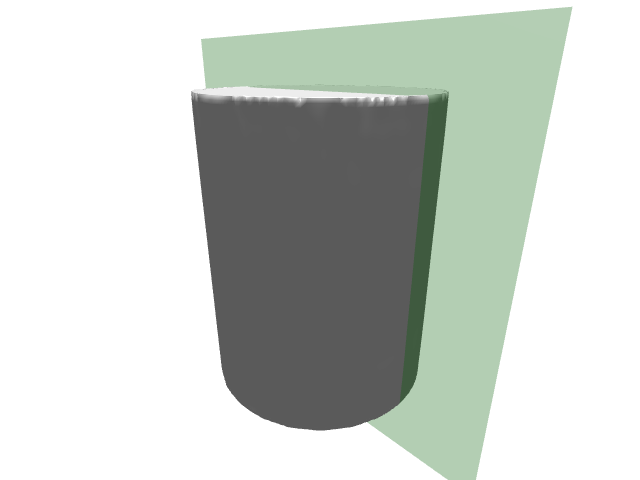}
\includegraphics[width=.59\linewidth]{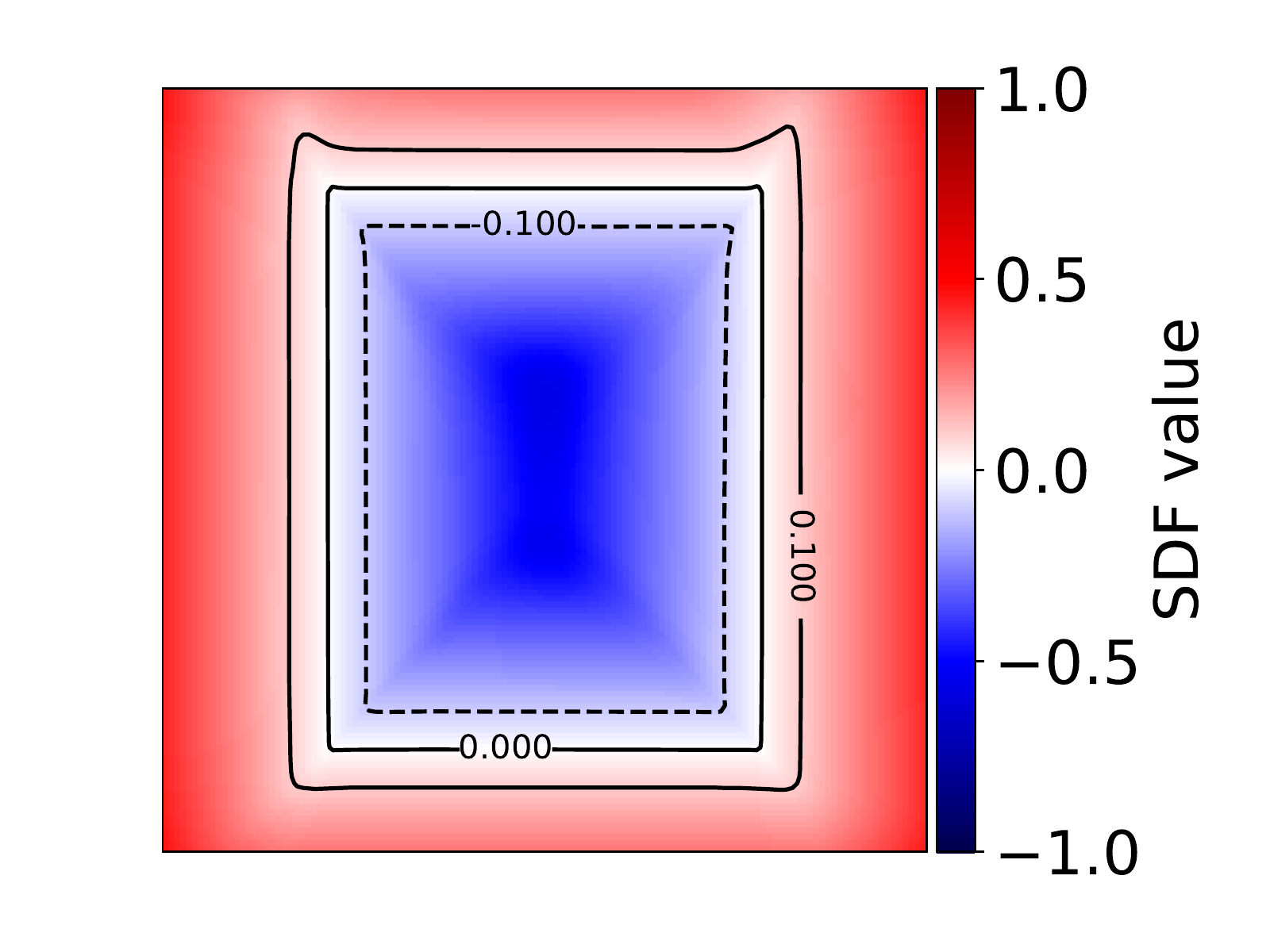}
\end{subfigure}
\begin{subfigure}{.22500\linewidth}
\includegraphics[trim=80 30 100 30,clip,width=.39\linewidth]{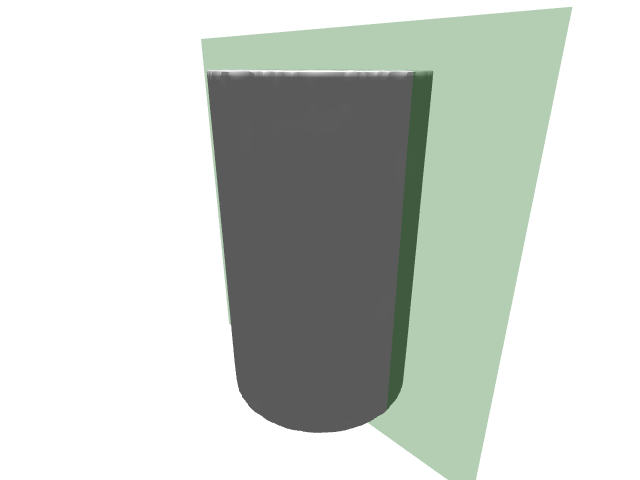}
\includegraphics[width=.59\linewidth]{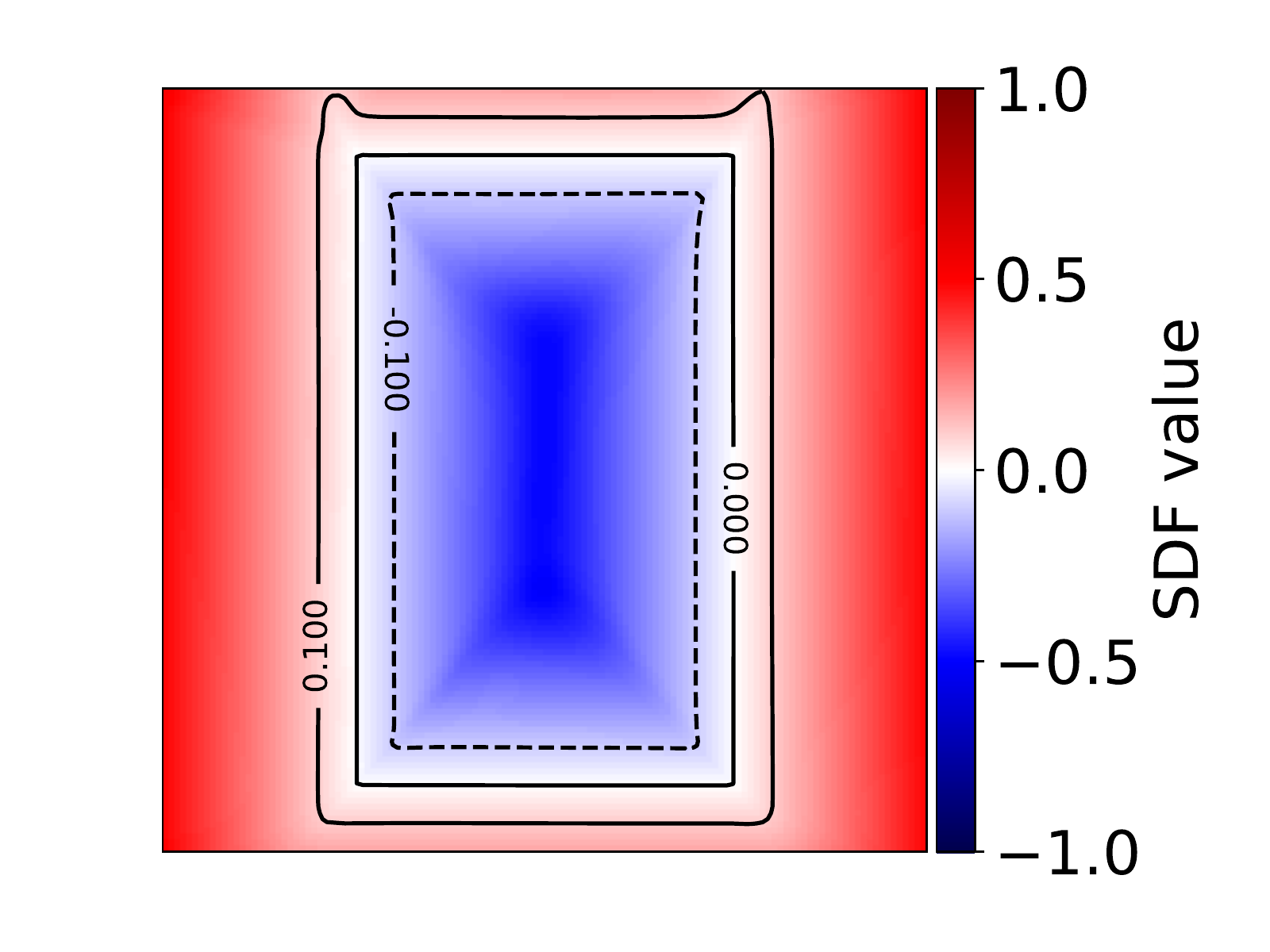}
\end{subfigure}
\begin{subfigure}{.22500\linewidth}
\includegraphics[trim=80 30 100 30,clip,width=.39\linewidth]{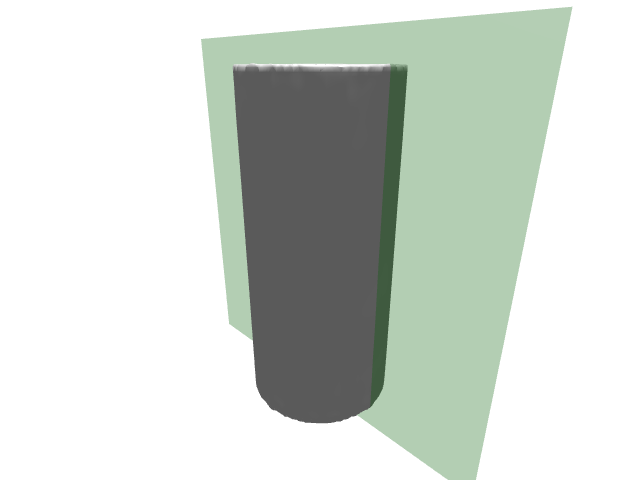}
\includegraphics[width=.59\linewidth]{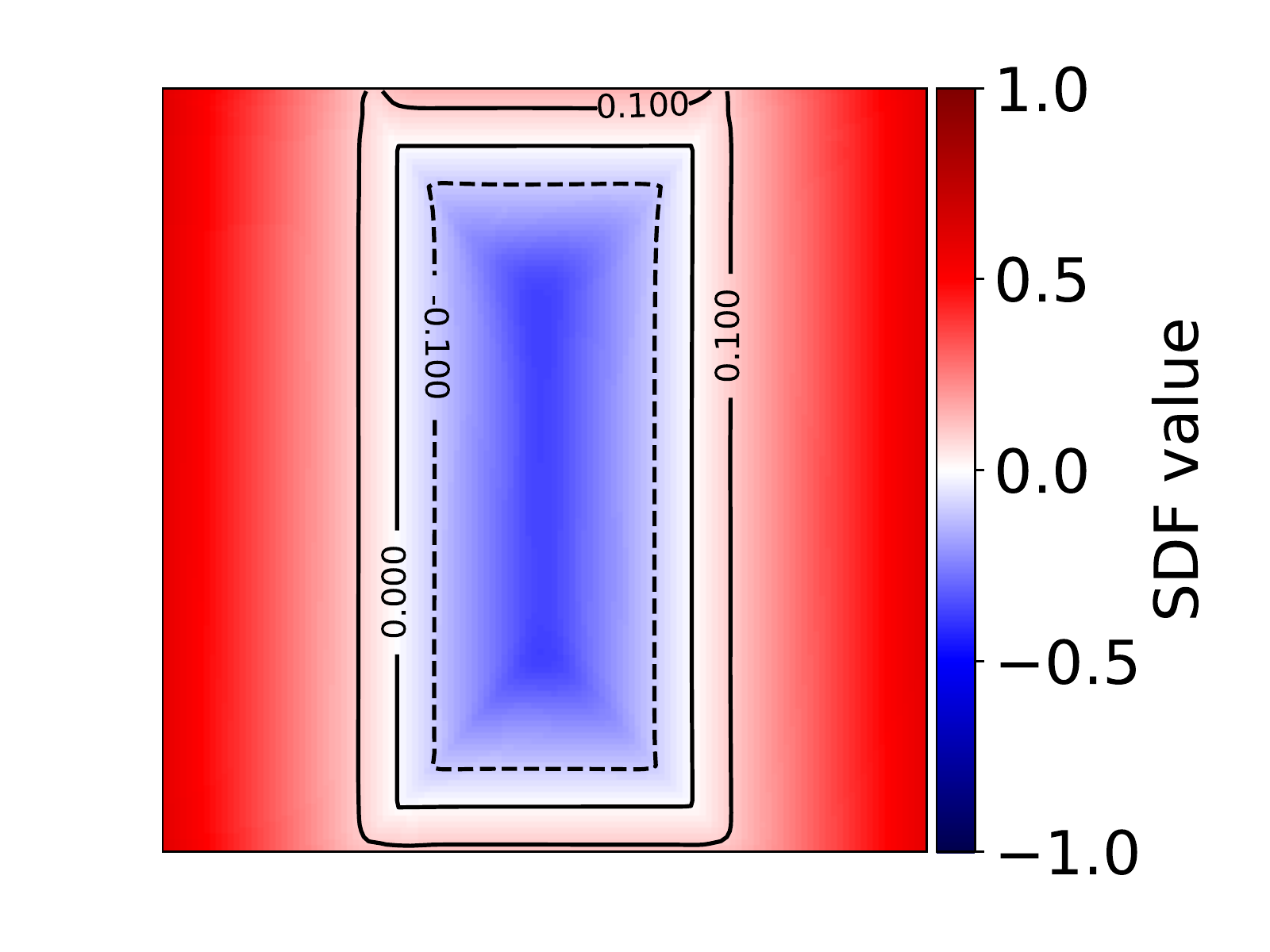}
\end{subfigure}
\begin{subfigure}{.22500\linewidth}
\includegraphics[trim=80 30 100 30,clip,width=.39\linewidth]{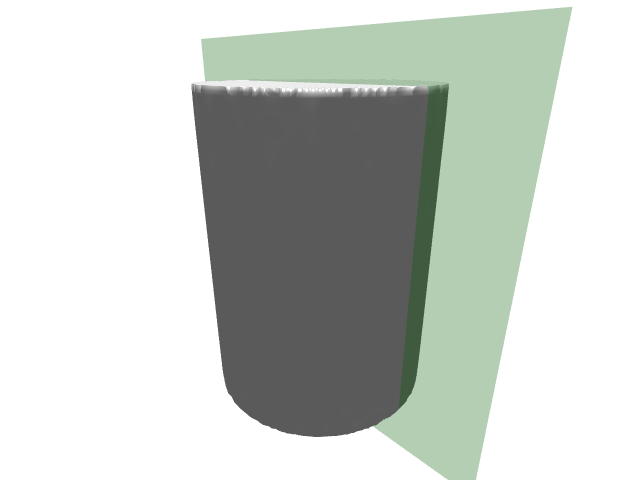}
\includegraphics[width=.59\linewidth]{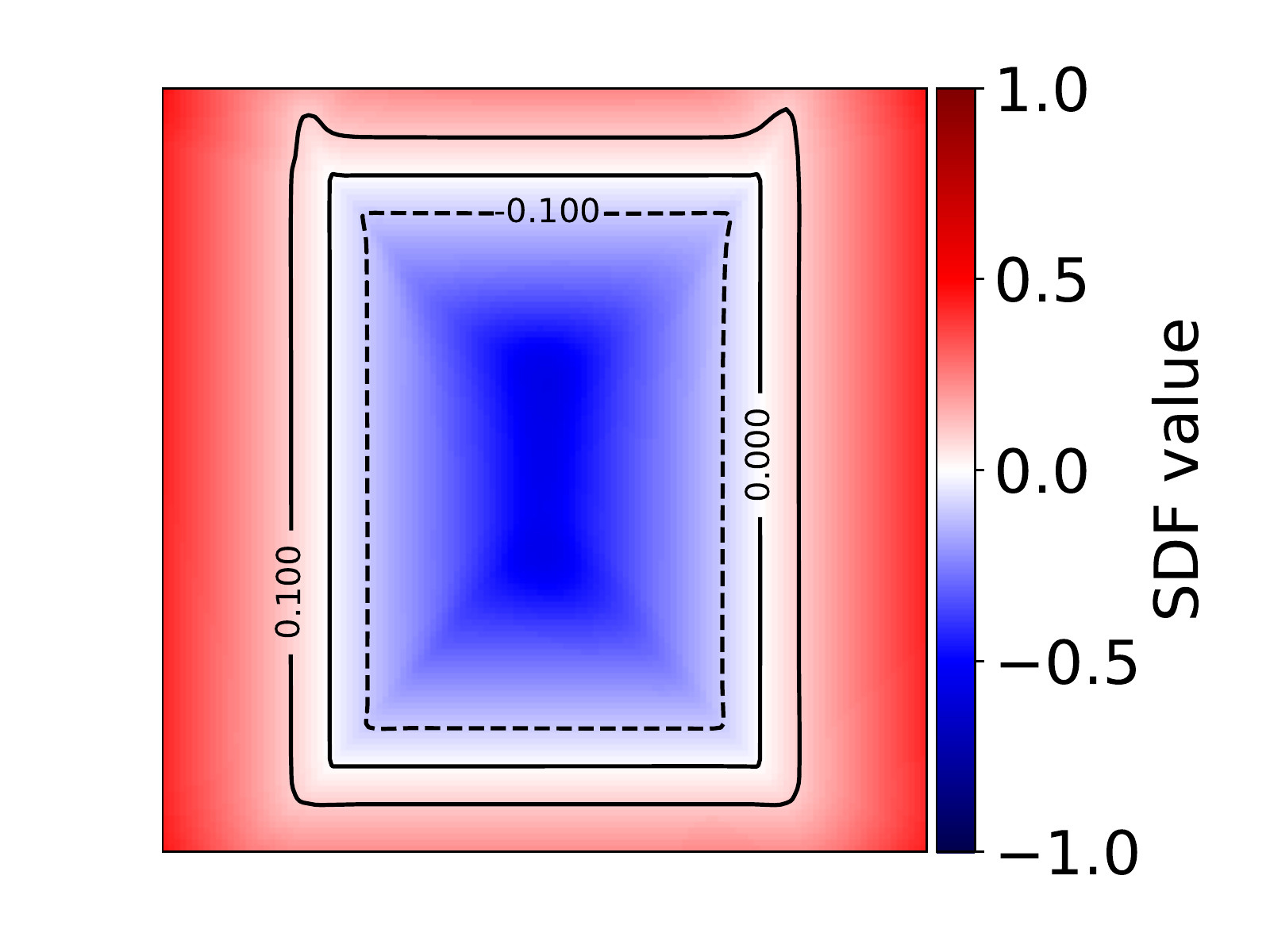}
\end{subfigure}
\begin{subfigure}{.22500\linewidth}
\includegraphics[trim=80 30 100 30,clip,width=.39\linewidth]{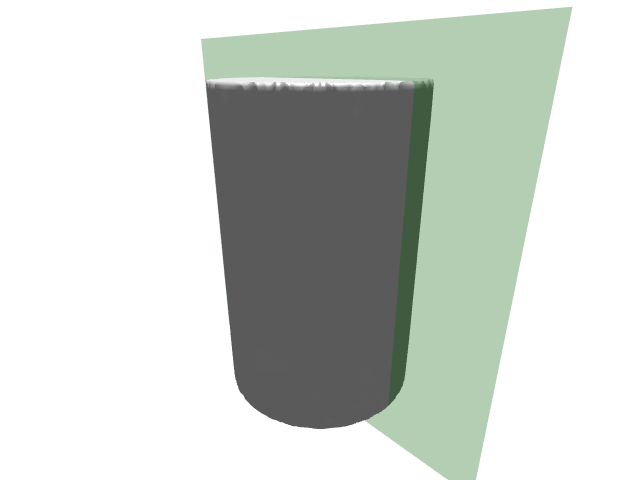}
\includegraphics[width=.59\linewidth]{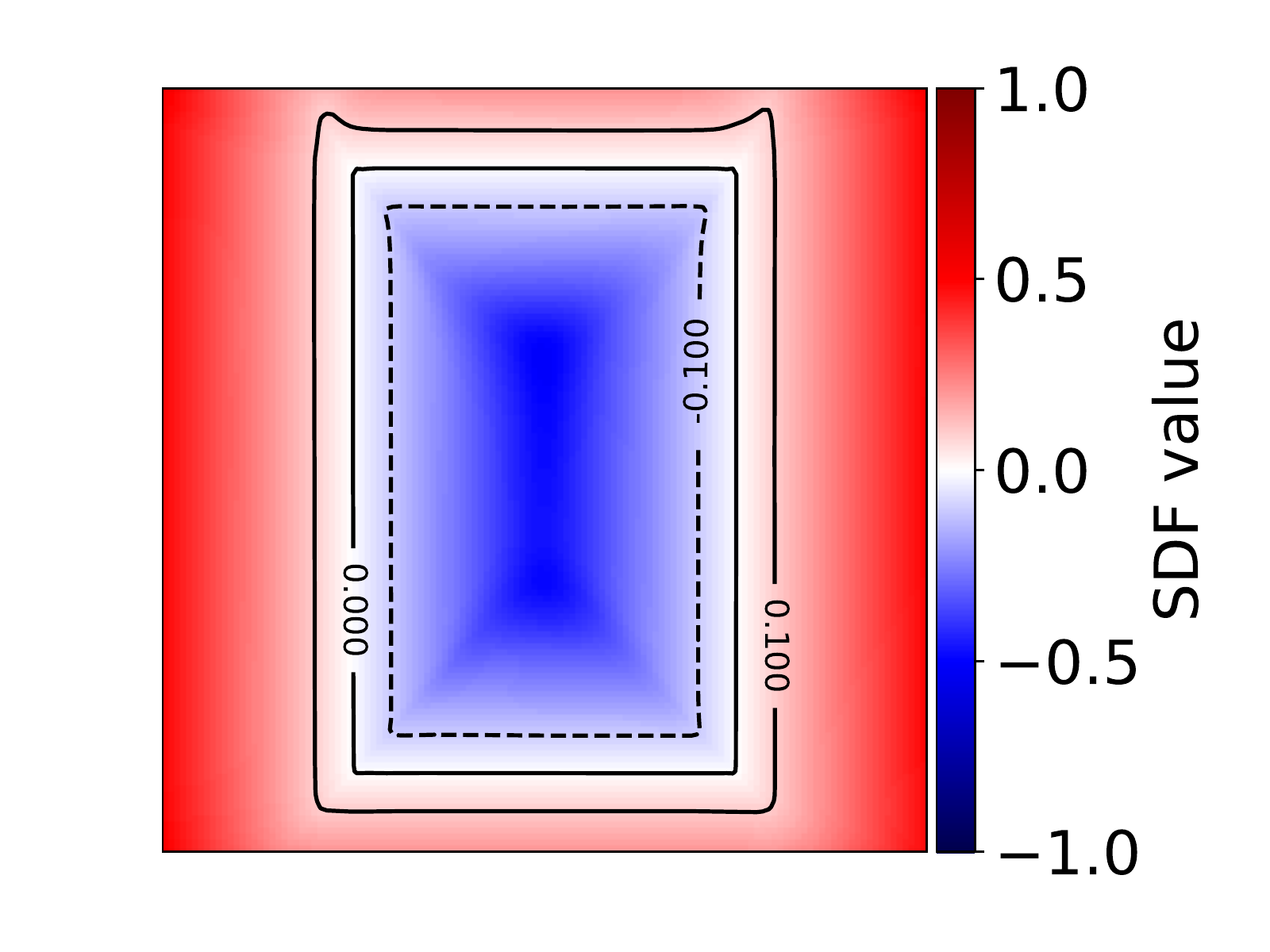}
\end{subfigure}
\begin{subfigure}{.22500\linewidth}
\includegraphics[trim=80 30 100 30,clip,width=.39\linewidth]{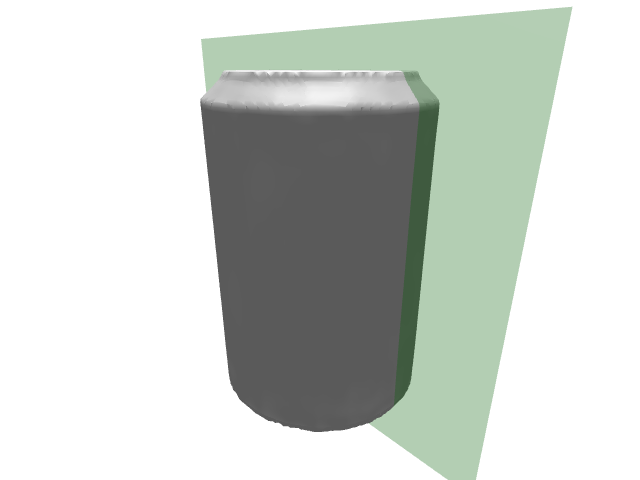}
\includegraphics[width=.59\linewidth]{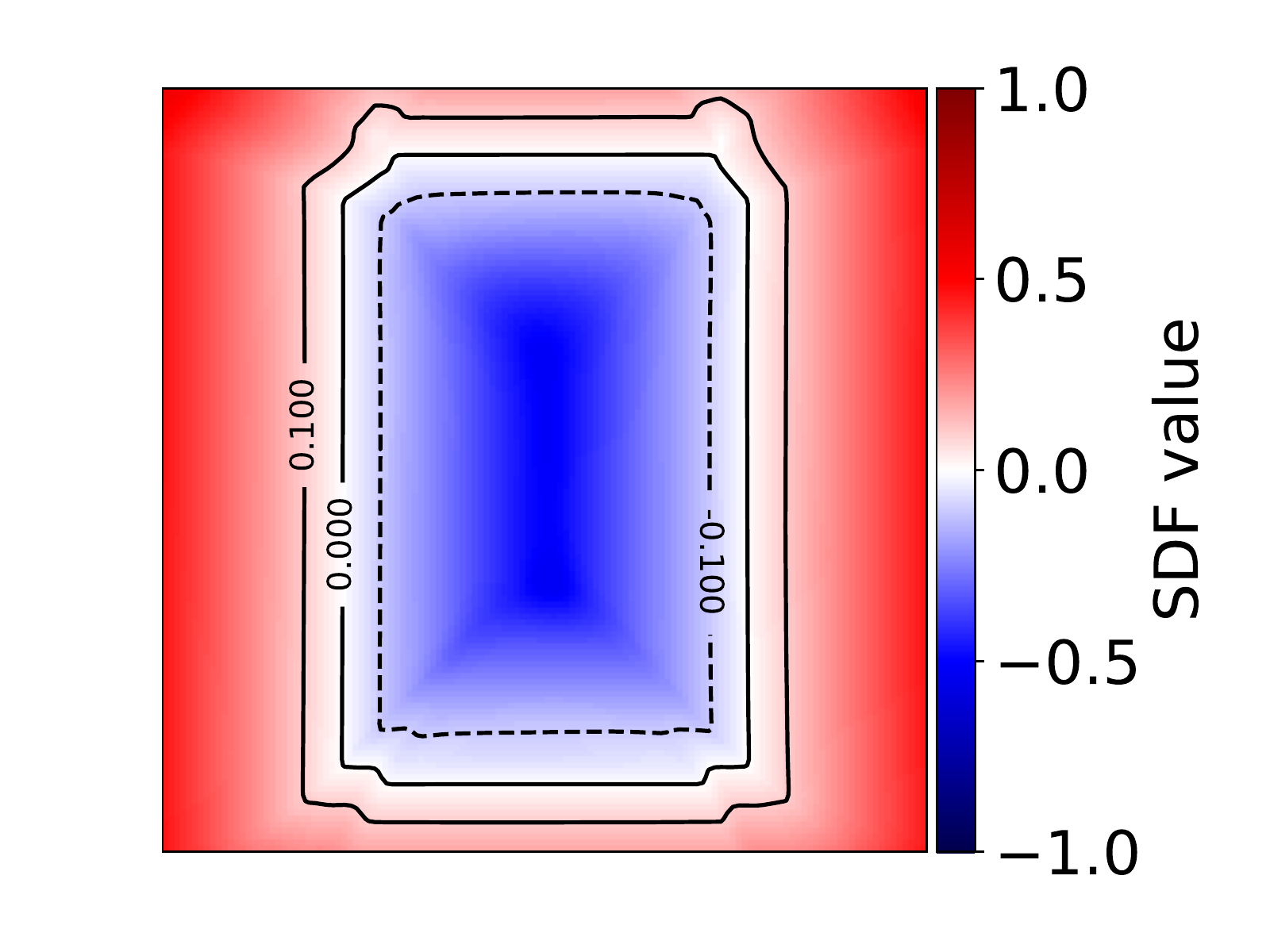}
\end{subfigure}
\begin{subfigure}{.22500\linewidth}
\includegraphics[trim=80 30 100 30,clip,width=.39\linewidth]{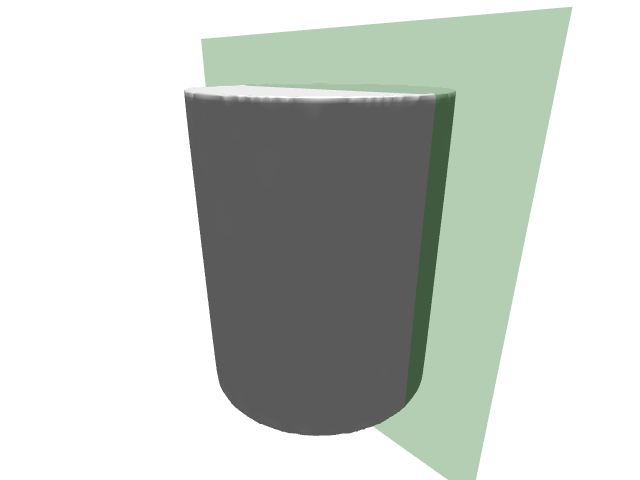}
\includegraphics[width=.59\linewidth]{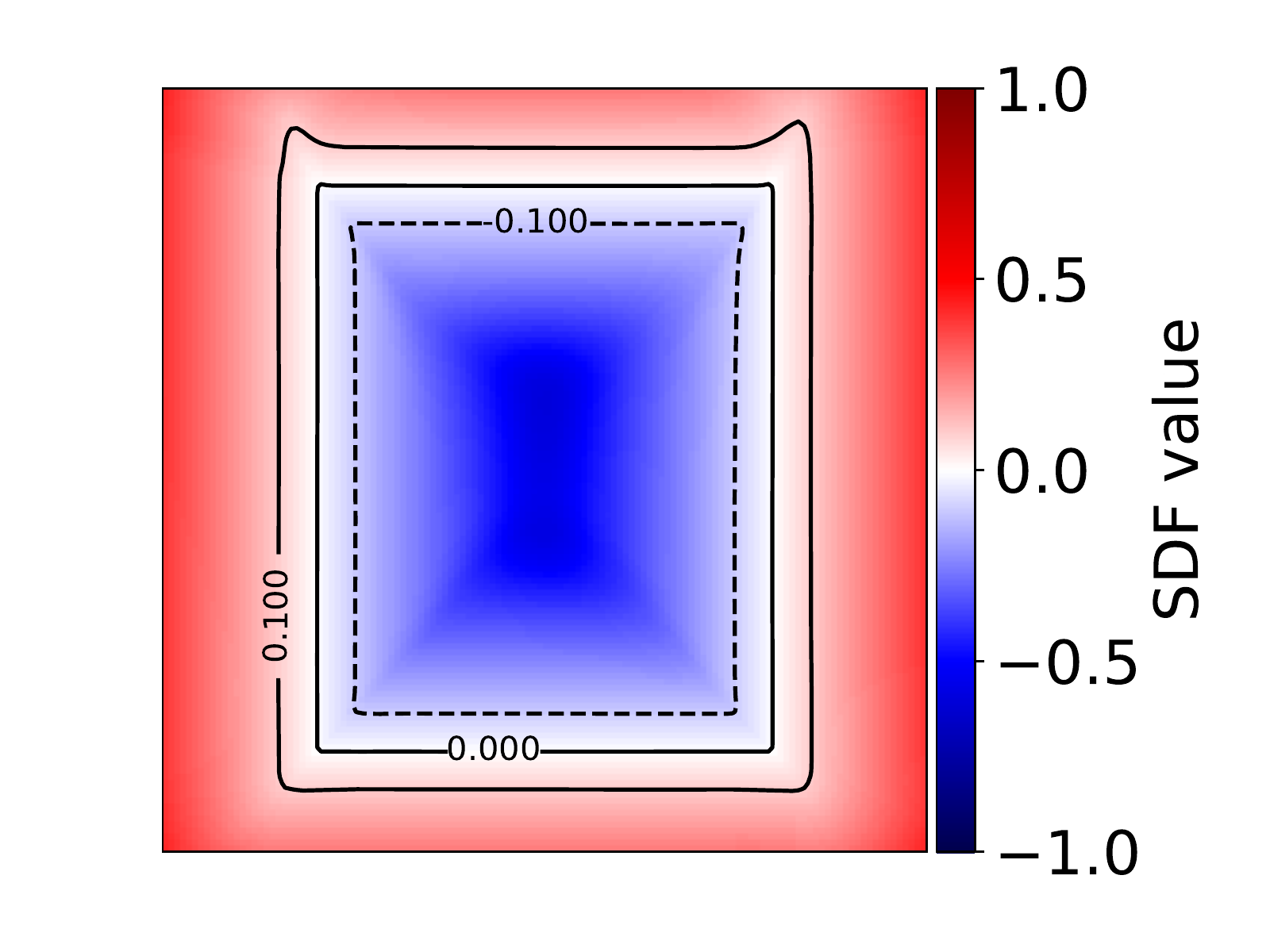}
\end{subfigure}
\begin{subfigure}{.22500\linewidth}
\includegraphics[trim=80 30 100 30,clip,width=.39\linewidth]{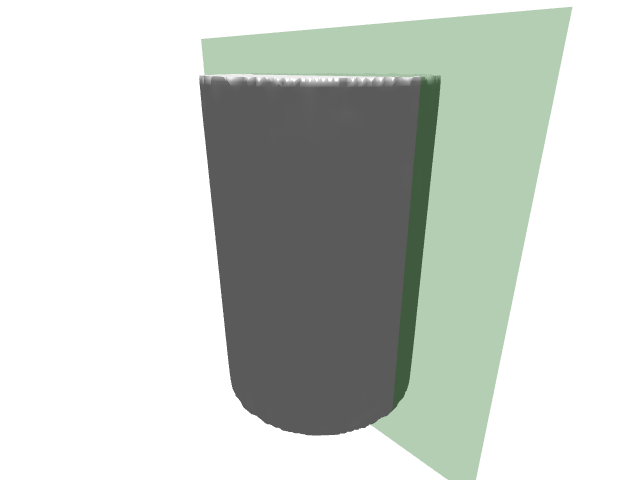}
\includegraphics[width=.59\linewidth]{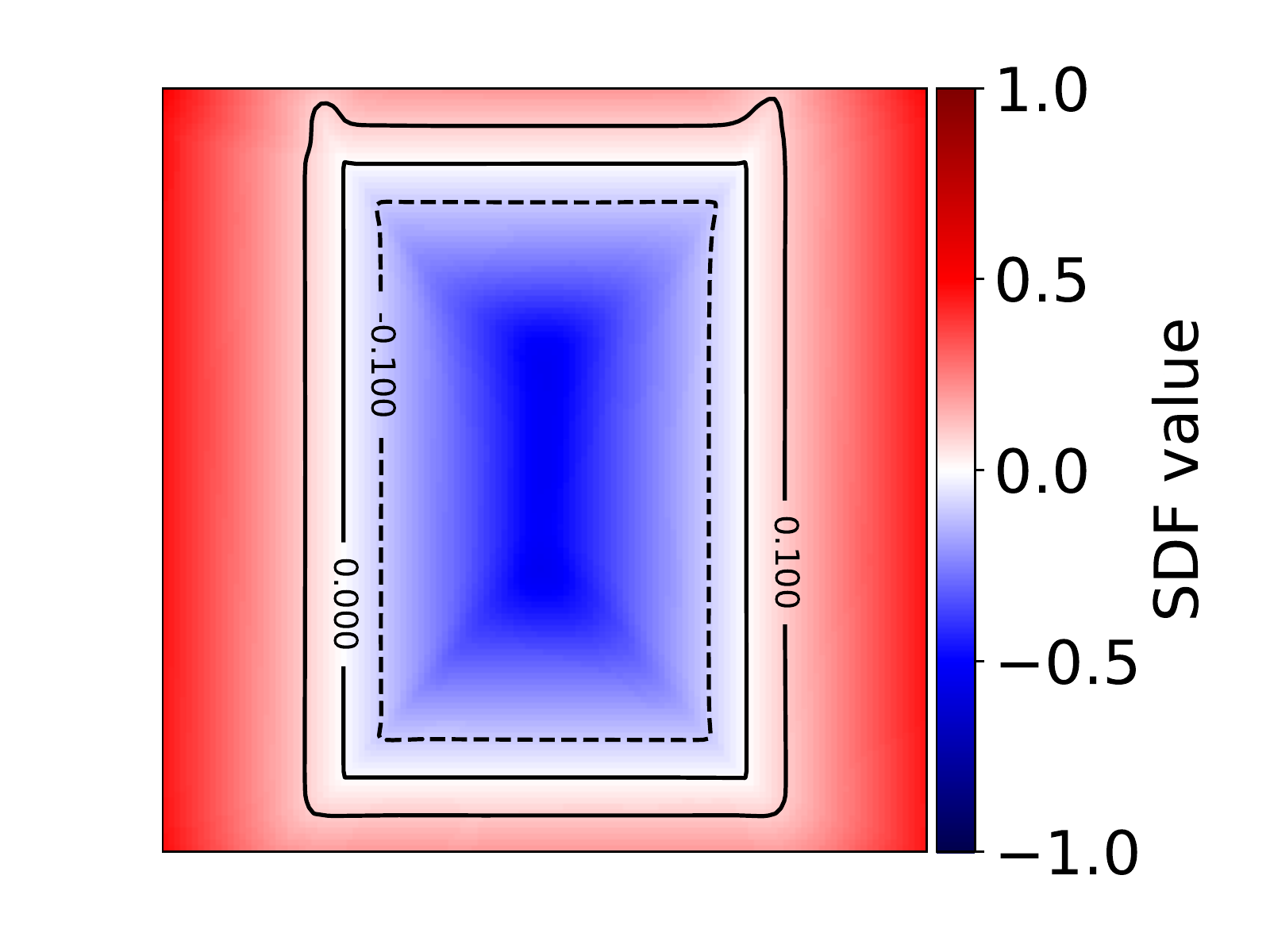}
\end{subfigure}
\begin{subfigure}{.22500\linewidth}
\includegraphics[trim=80 30 100 30,clip,width=.39\linewidth]{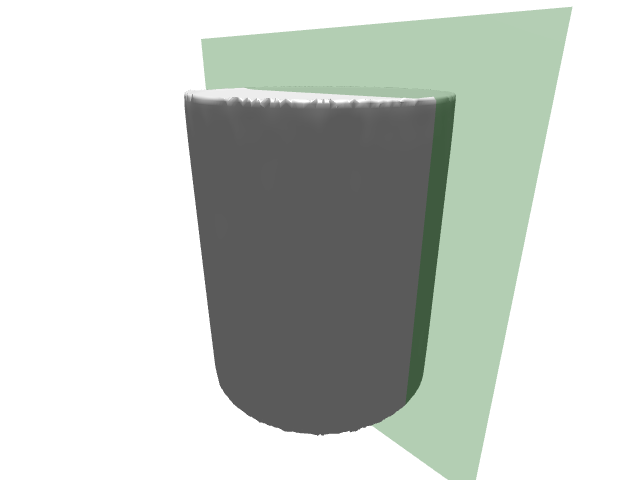}
\includegraphics[width=.59\linewidth]{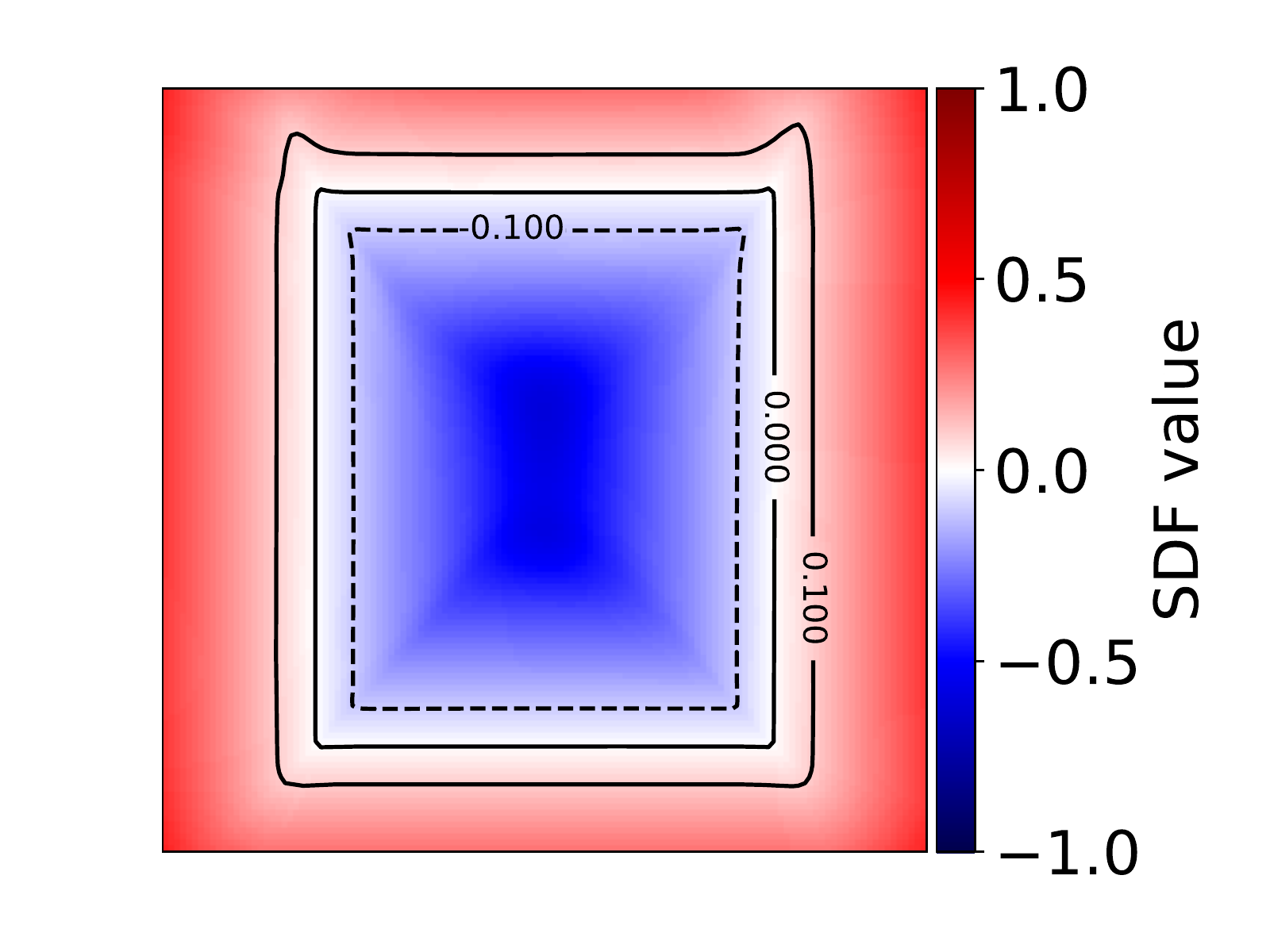}
\end{subfigure}

%% file: figures/shapespace_renderings/camera.tex
\begin{subfigure}{.24545\linewidth}
\includegraphics[trim=80 30 100 30,clip,width=.39\linewidth]{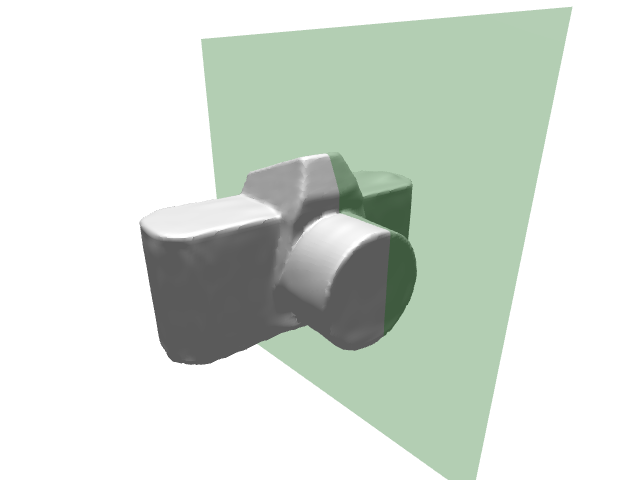}
\includegraphics[width=.59\linewidth]{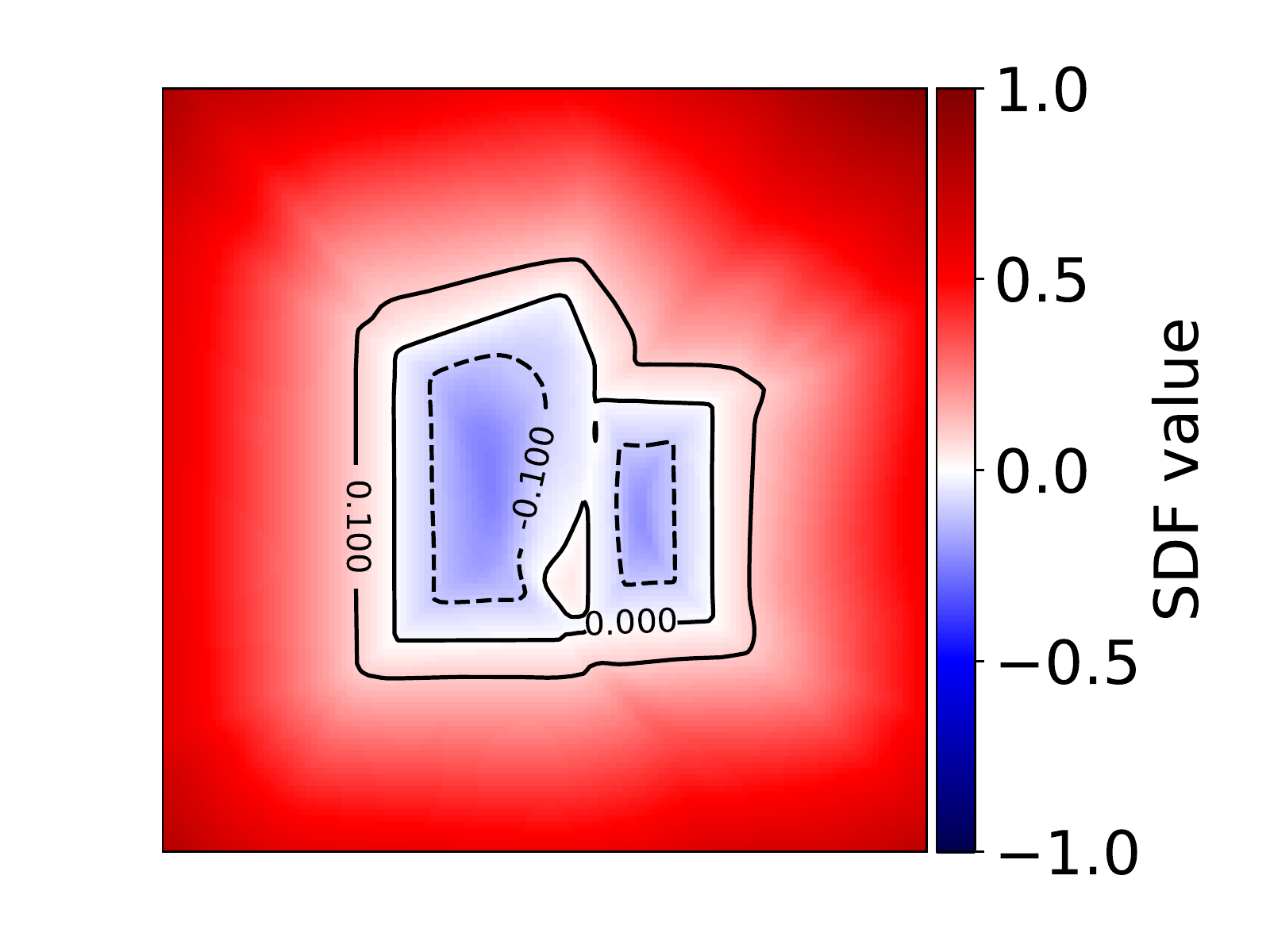}
\end{subfigure}
\begin{subfigure}{.24545\linewidth}
\includegraphics[trim=80 30 100 30,clip,width=.39\linewidth]{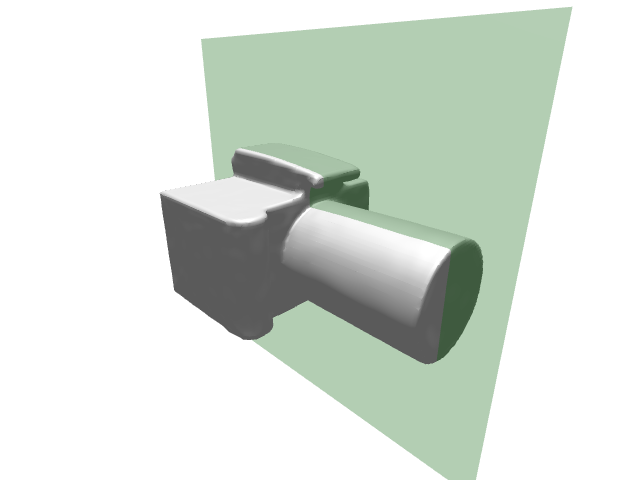}
\includegraphics[width=.59\linewidth]{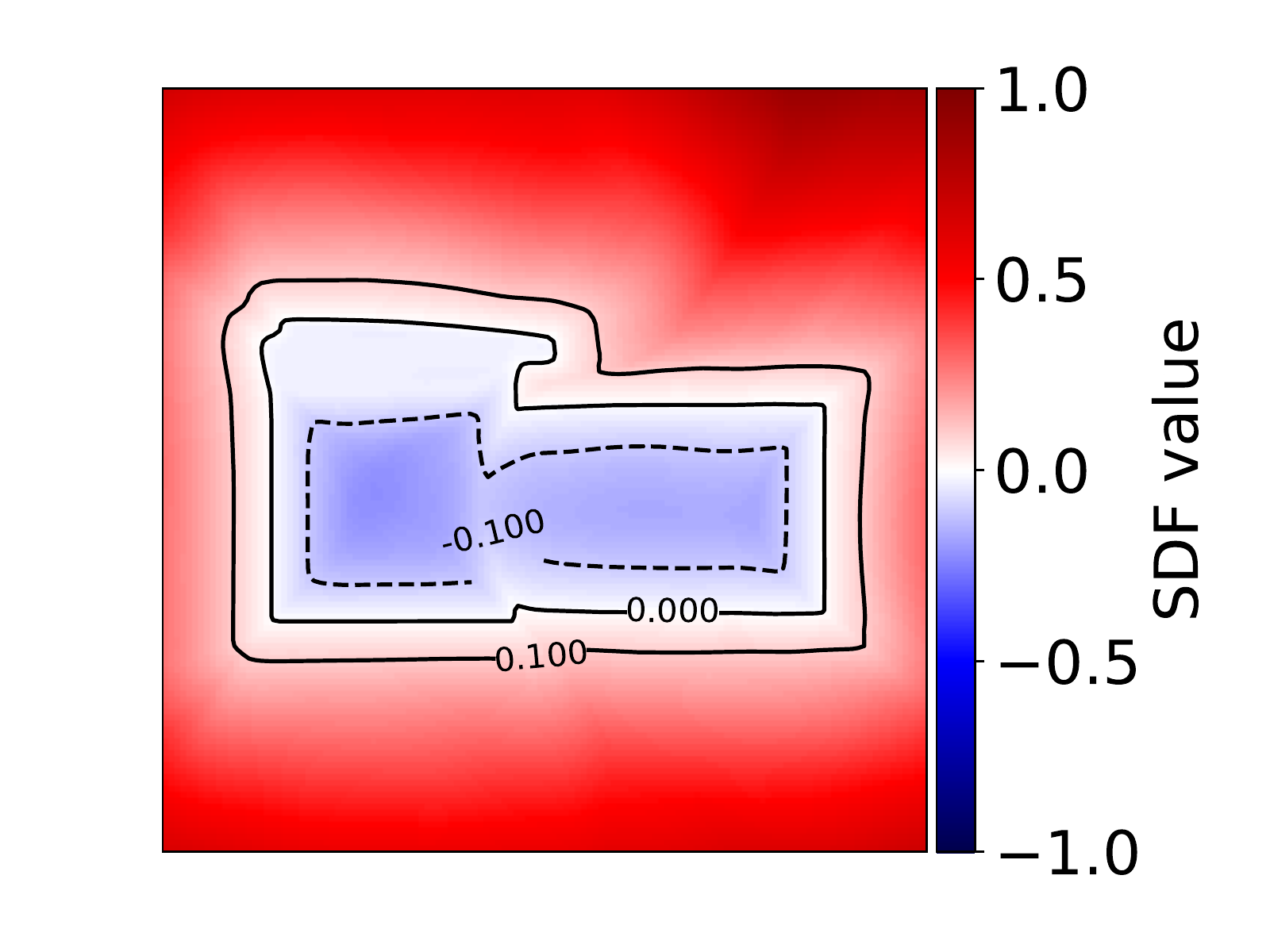}
\end{subfigure}
\begin{subfigure}{.24545\linewidth}
\includegraphics[trim=80 30 100 30,clip,width=.39\linewidth]{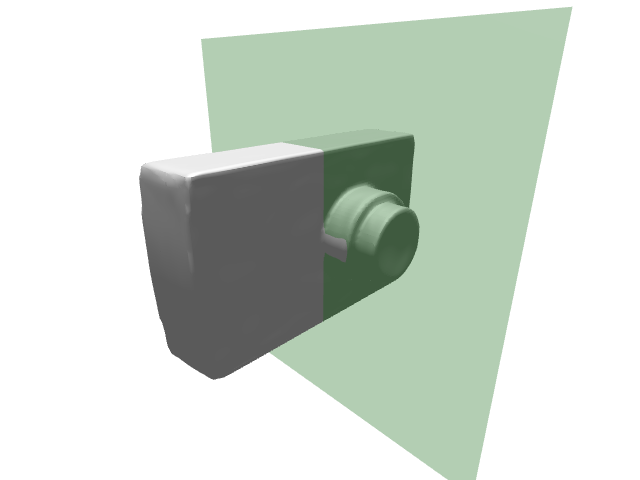}
\includegraphics[width=.59\linewidth]{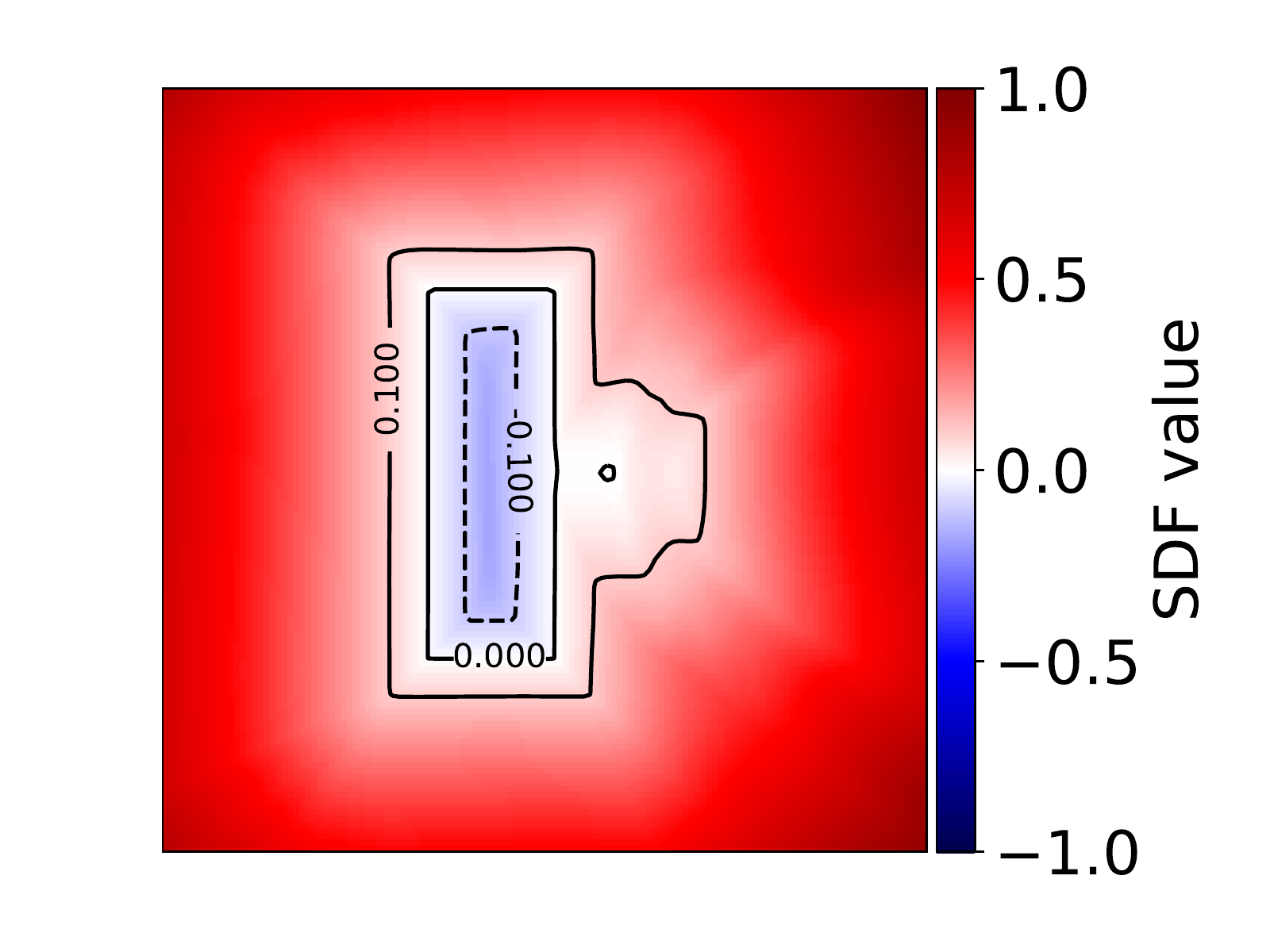}
\end{subfigure}
\begin{subfigure}{.24545\linewidth}
\includegraphics[trim=80 30 100 30,clip,width=.39\linewidth]{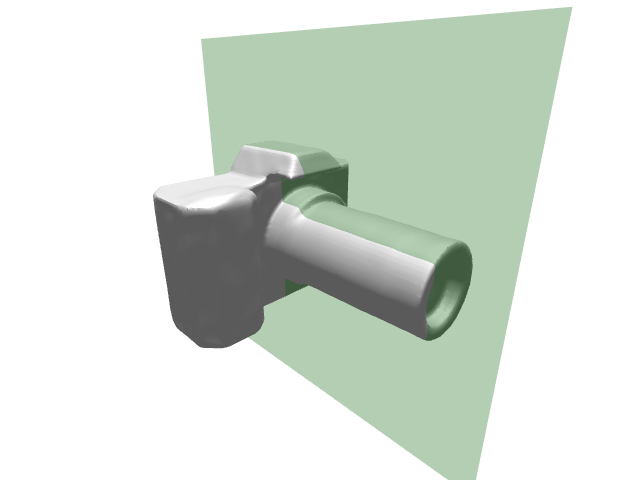}
\includegraphics[width=.59\linewidth]{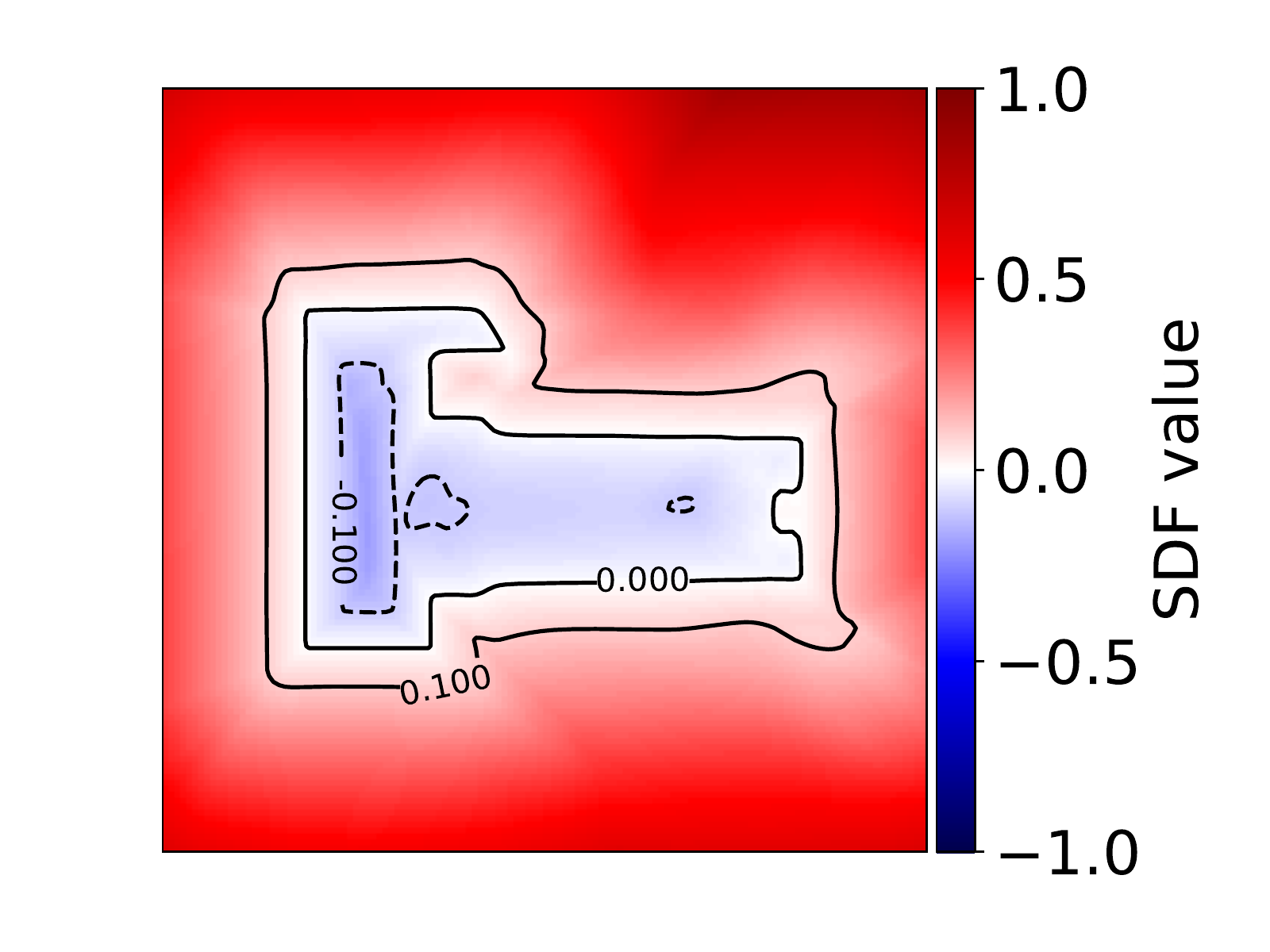}
\end{subfigure}
\begin{subfigure}{.24545\linewidth}
\includegraphics[trim=80 30 100 30,clip,width=.39\linewidth]{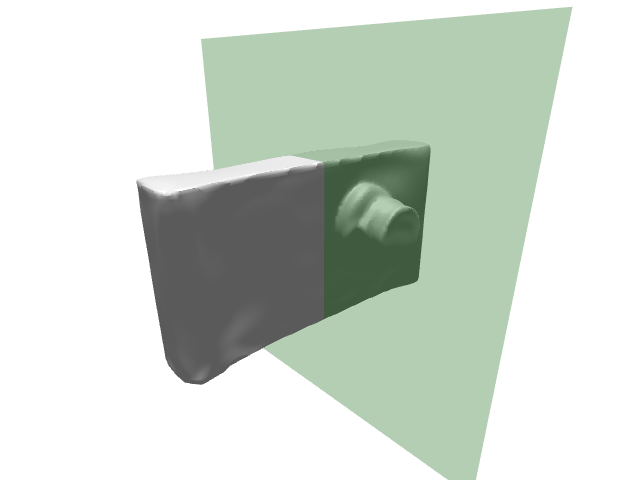}
\includegraphics[width=.59\linewidth]{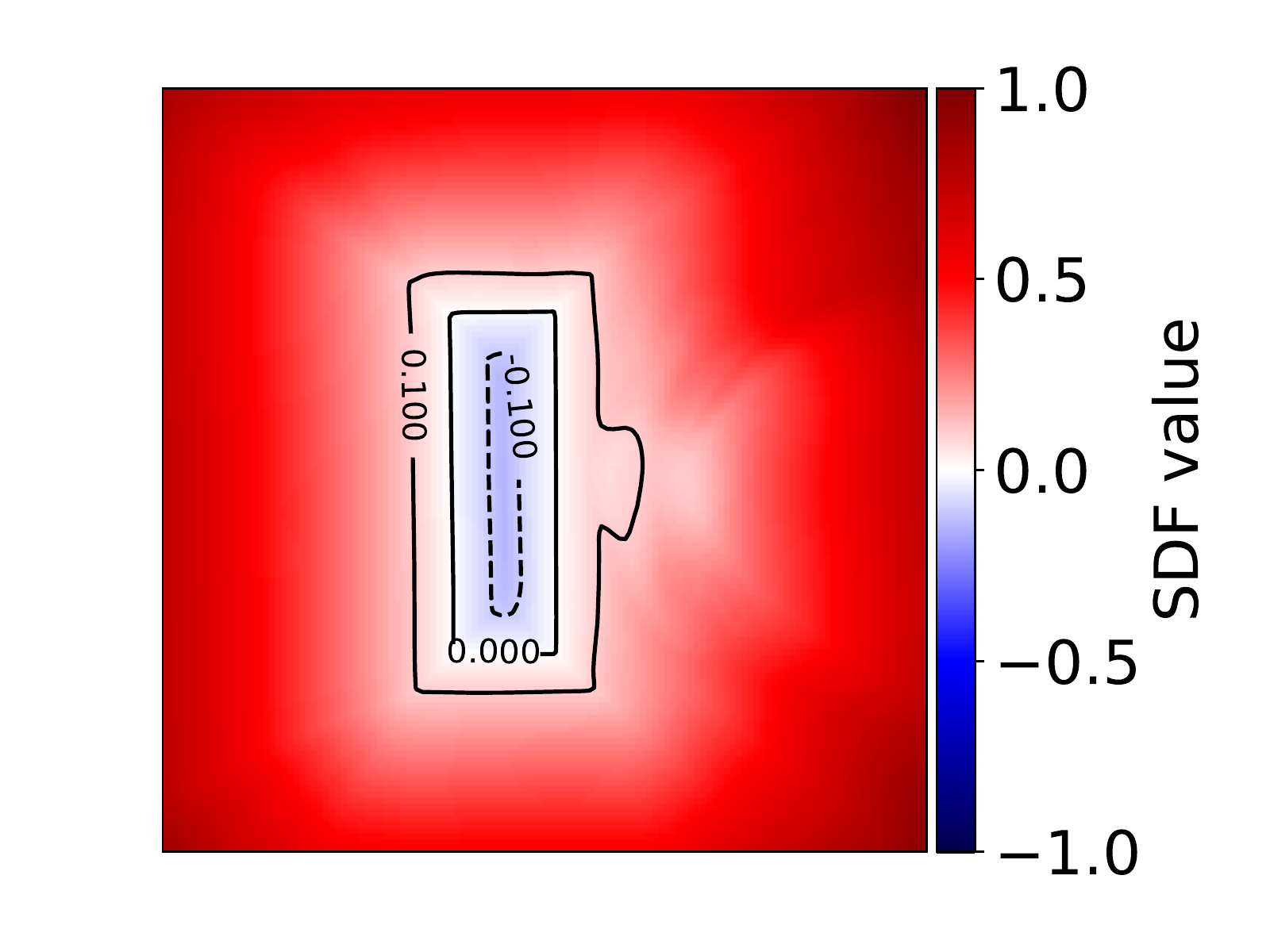}
\end{subfigure}
\begin{subfigure}{.24545\linewidth}
\includegraphics[trim=80 30 100 30,clip,width=.39\linewidth]{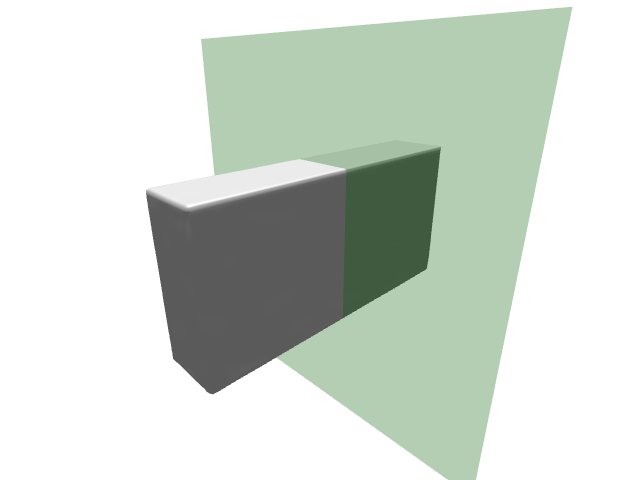}
\includegraphics[width=.59\linewidth]{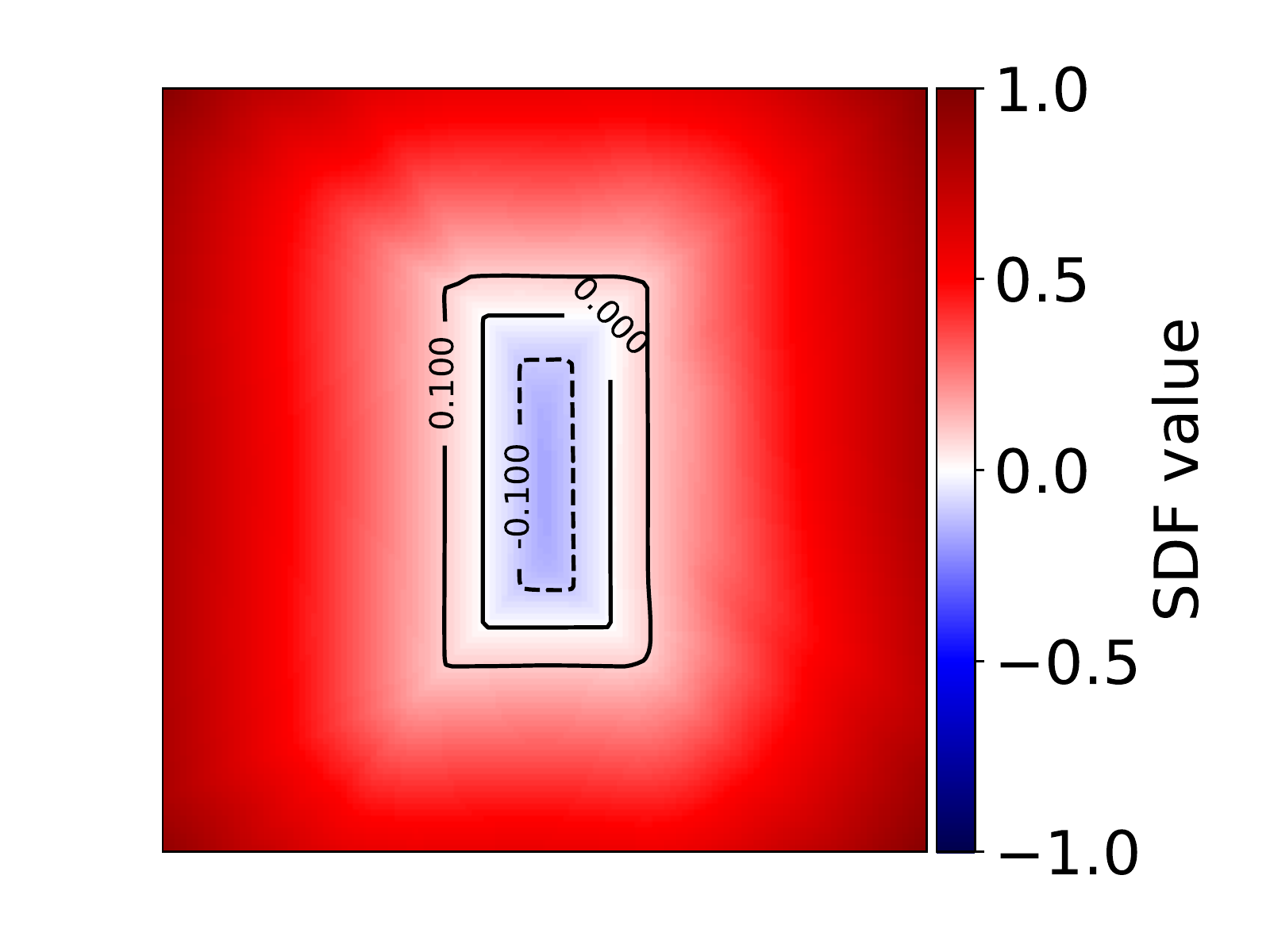}
\end{subfigure}
\begin{subfigure}{.24545\linewidth}
\includegraphics[trim=80 30 100 30,clip,width=.39\linewidth]{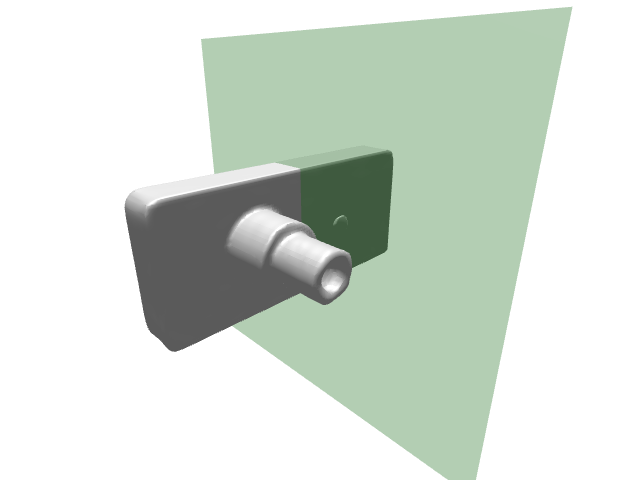}
\includegraphics[width=.59\linewidth]{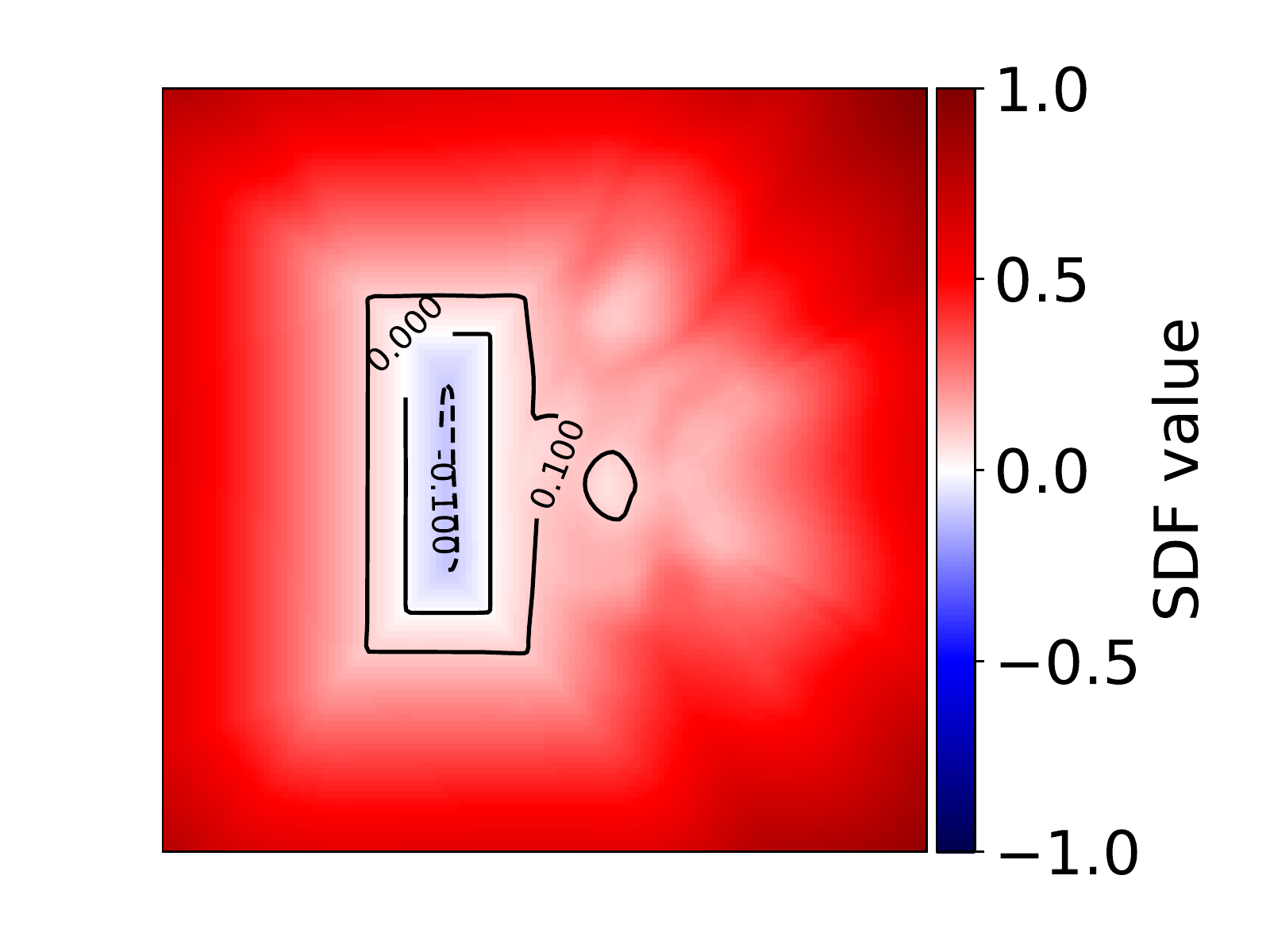}
\end{subfigure}
\begin{subfigure}{.24545\linewidth}
\includegraphics[trim=80 30 100 30,clip,width=.39\linewidth]{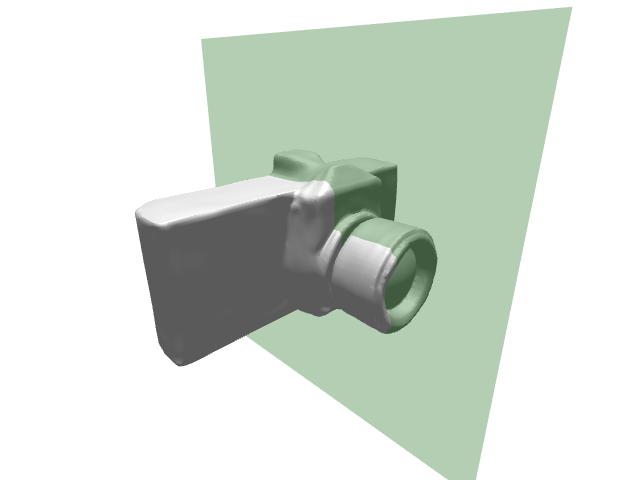}
\includegraphics[width=.59\linewidth]{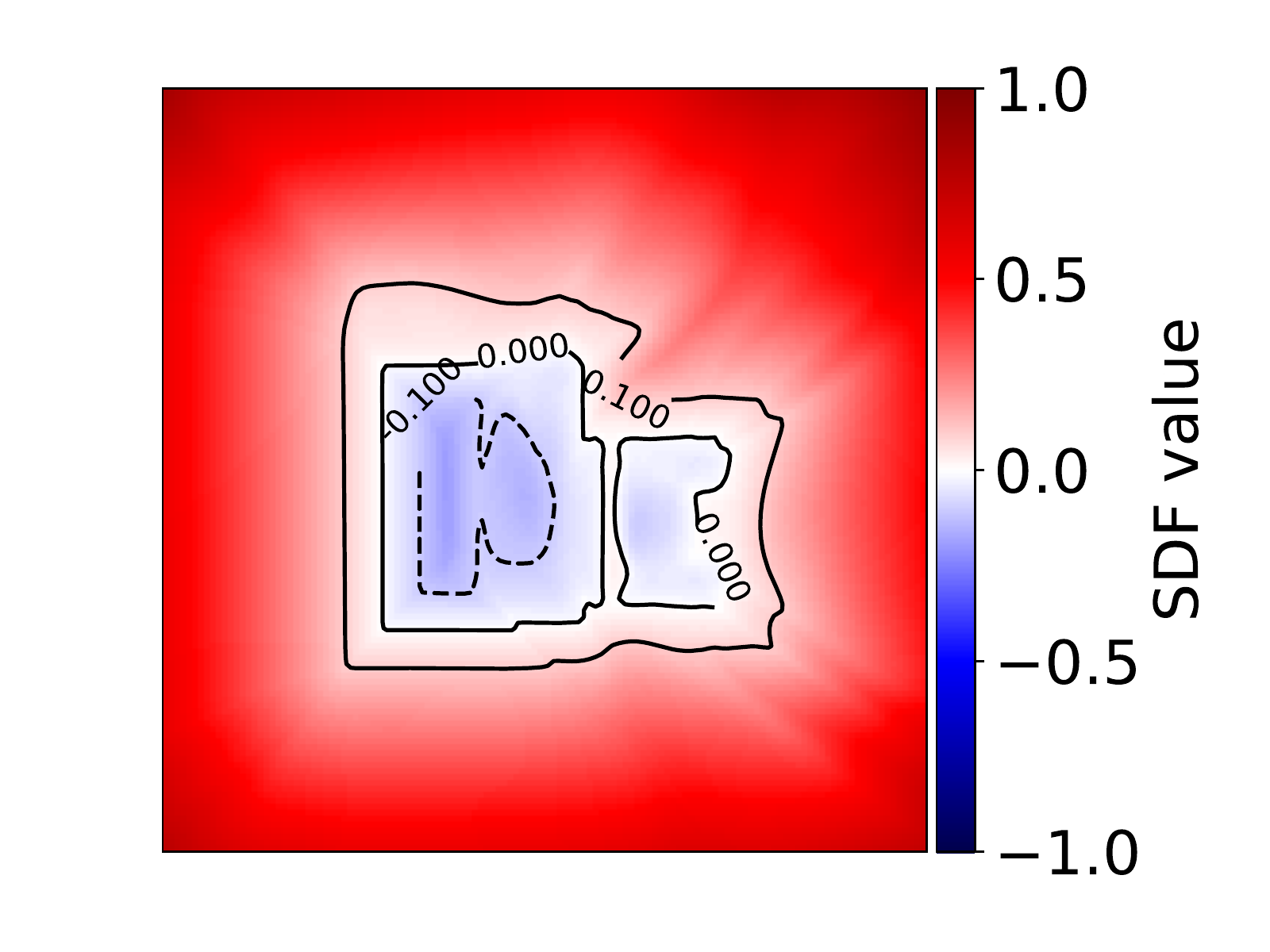}
\end{subfigure}
\begin{subfigure}{.24545\linewidth}
\includegraphics[trim=80 30 100 30,clip,width=.39\linewidth]{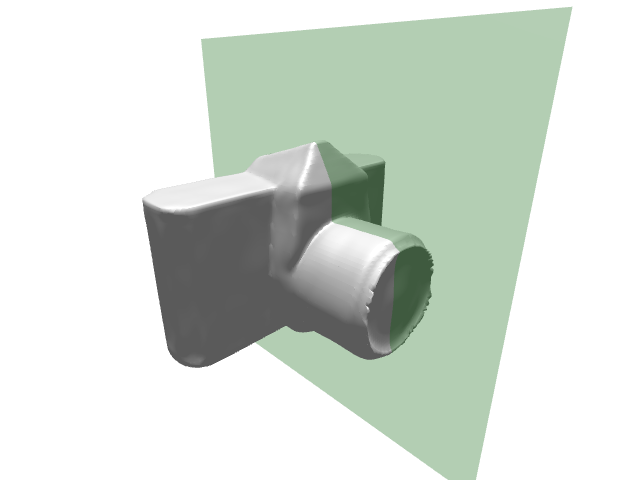}
\includegraphics[width=.59\linewidth]{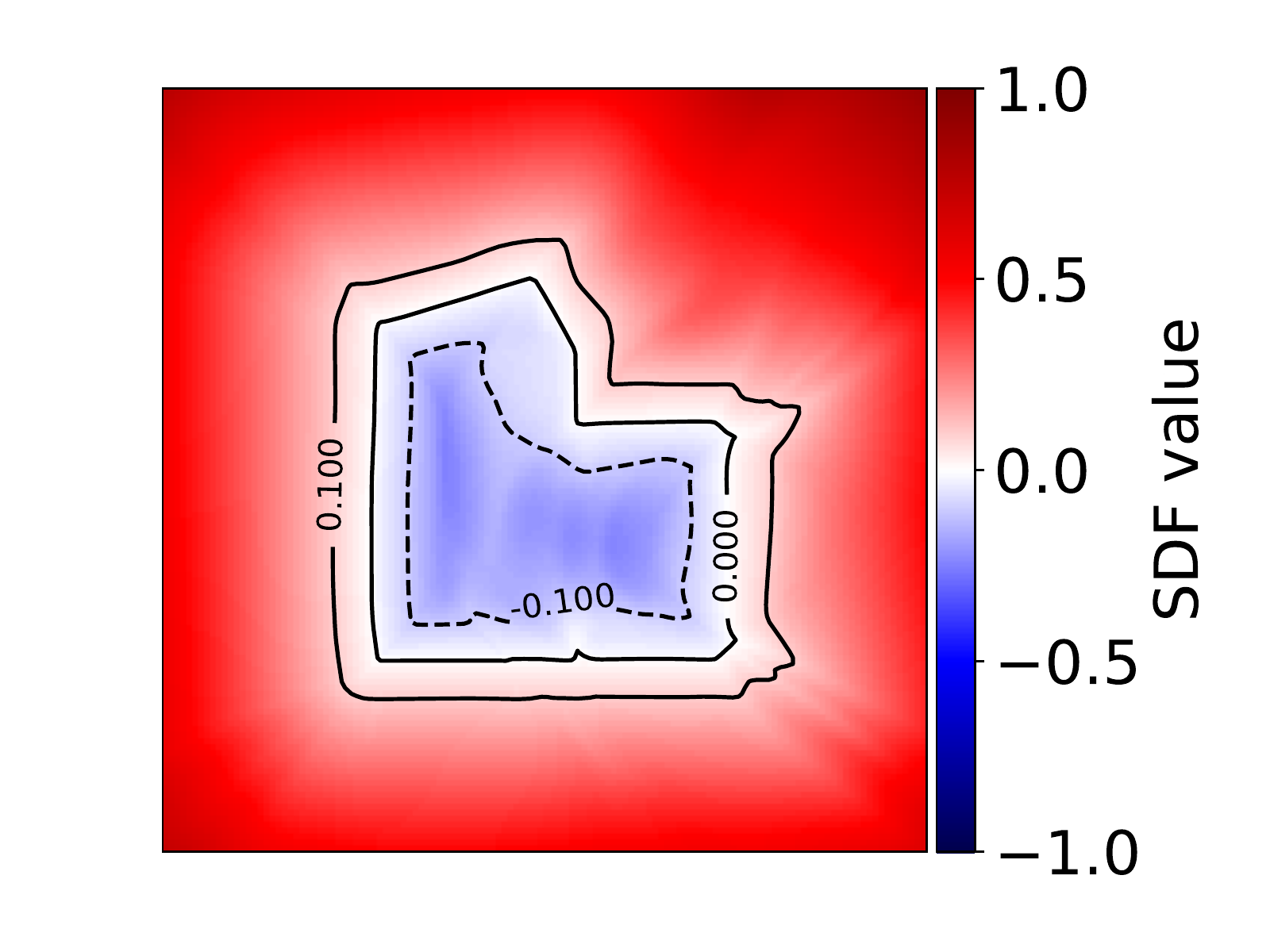}
\end{subfigure}
\begin{subfigure}{.24545\linewidth}
\includegraphics[trim=80 30 100 30,clip,width=.39\linewidth]{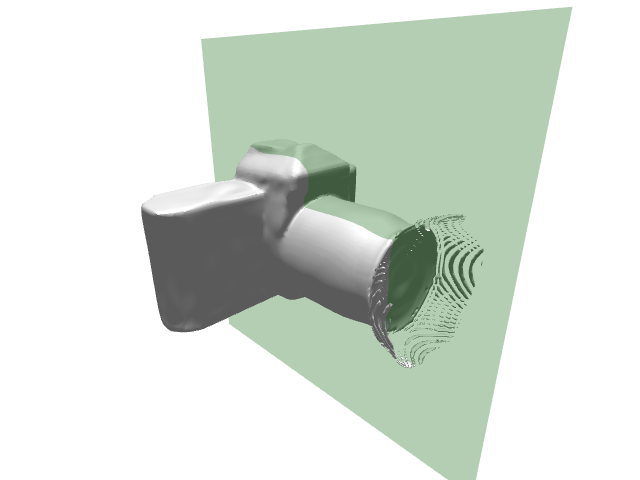}
\includegraphics[width=.59\linewidth]{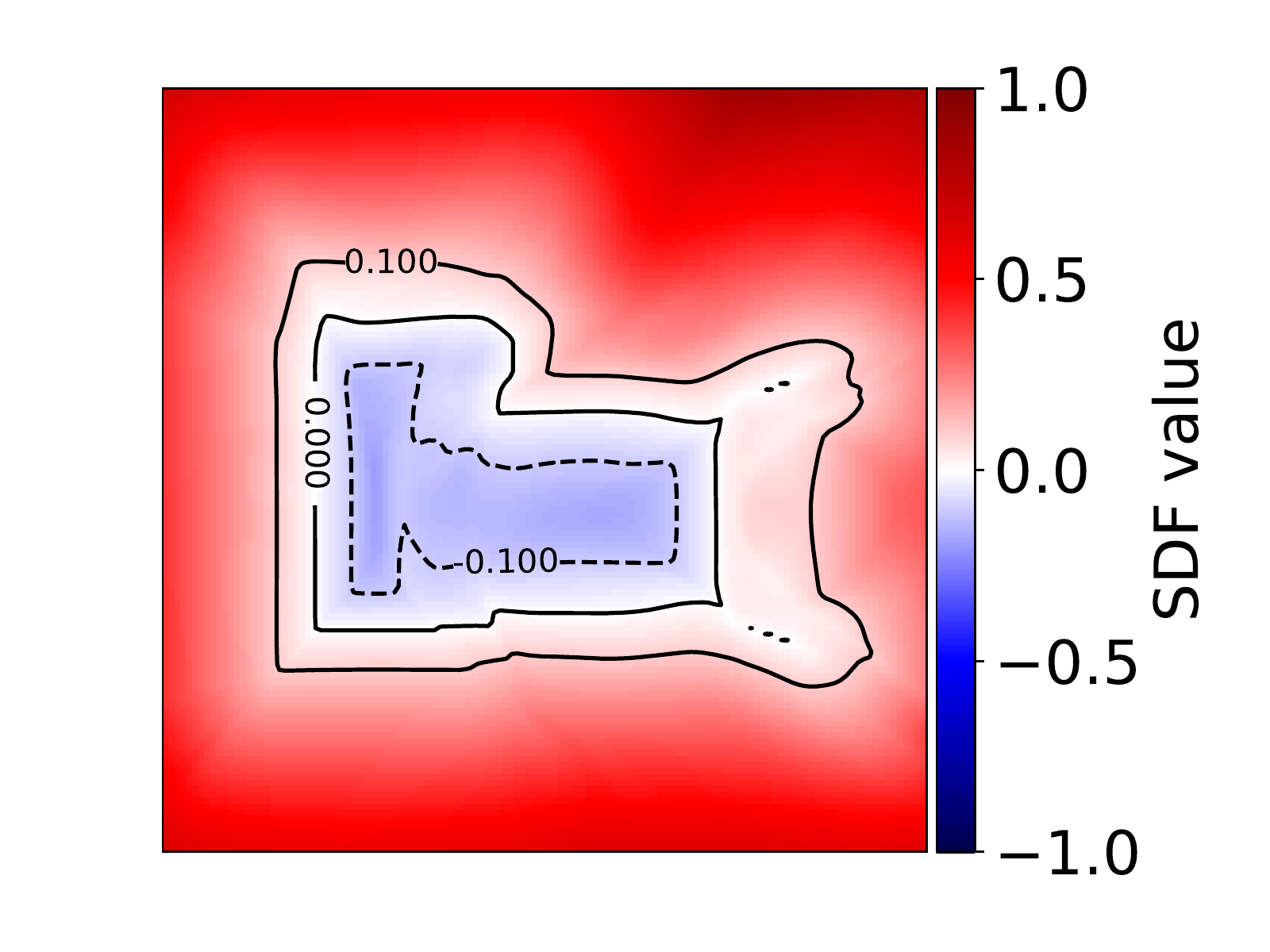}
\end{subfigure}
\begin{subfigure}{.24545\linewidth}
\includegraphics[trim=80 30 100 30,clip,width=.39\linewidth]{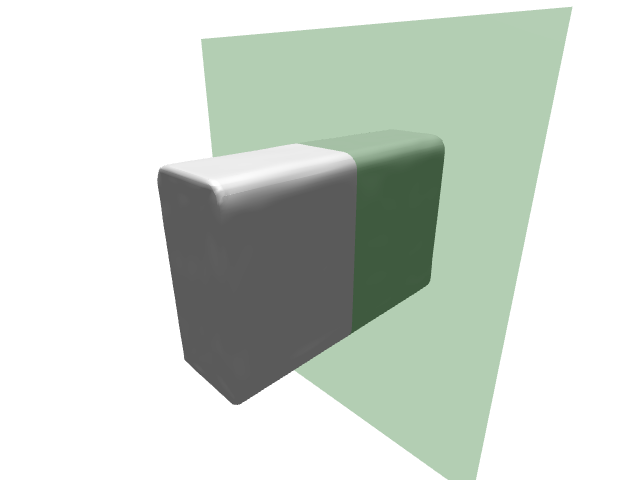}
\includegraphics[width=.59\linewidth]{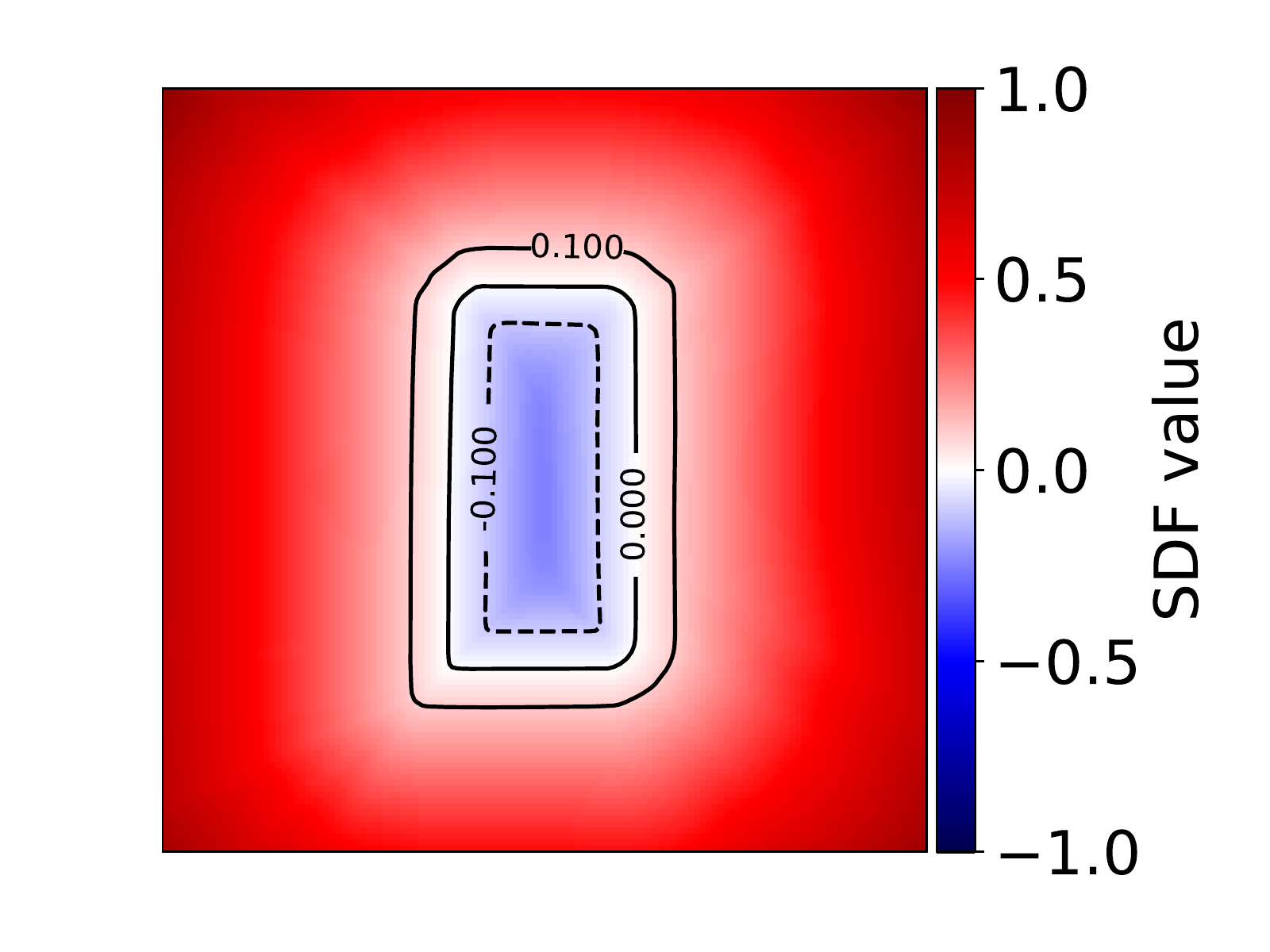}
\end{subfigure}
\begin{subfigure}{.24545\linewidth}
\includegraphics[trim=80 30 100 30,clip,width=.39\linewidth]{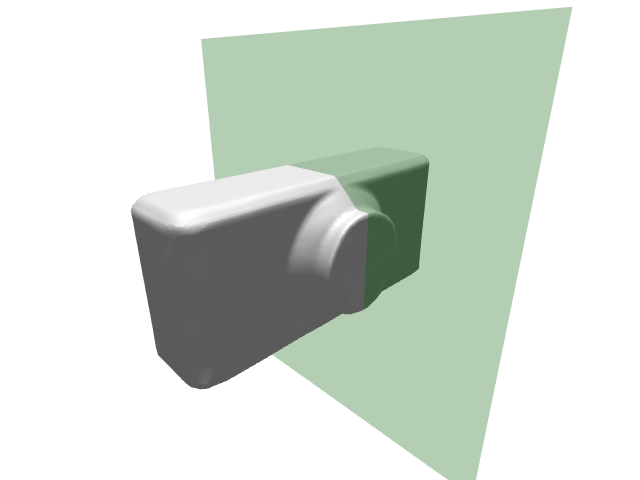}
\includegraphics[width=.59\linewidth]{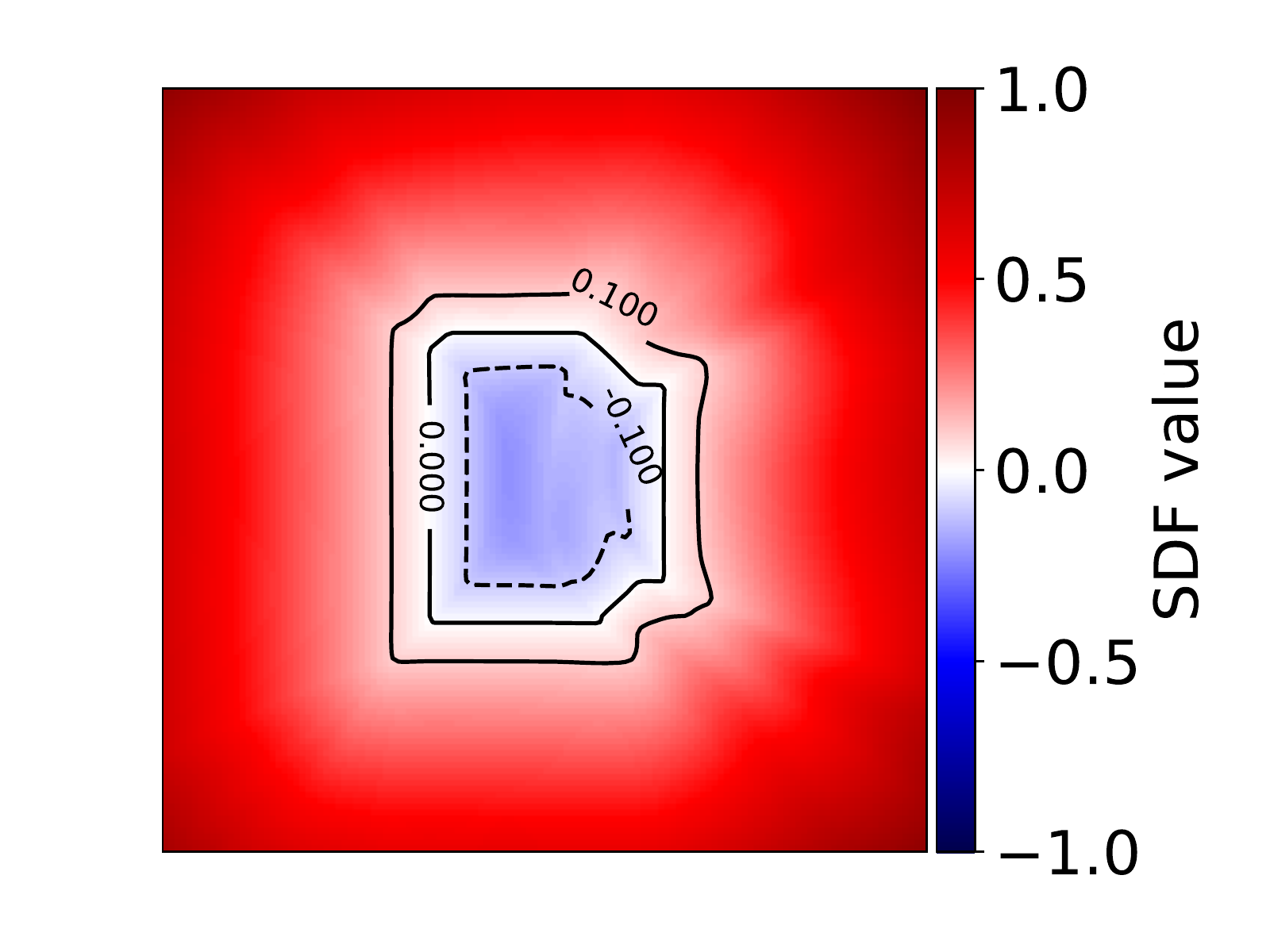}
\end{subfigure}
\begin{subfigure}{.24545\linewidth}
\includegraphics[trim=80 30 100 30,clip,width=.39\linewidth]{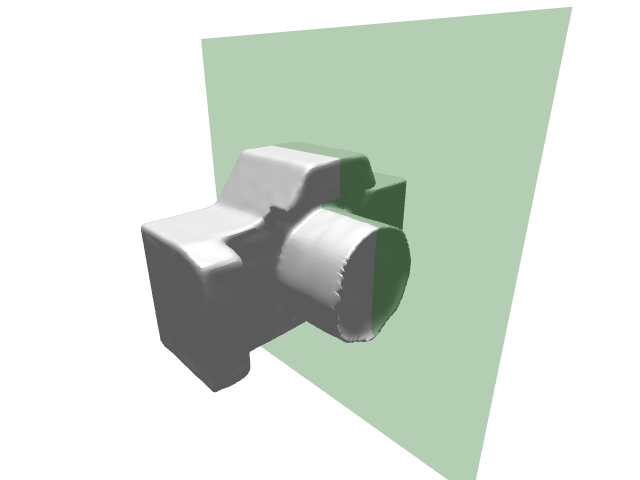}
\includegraphics[width=.59\linewidth]{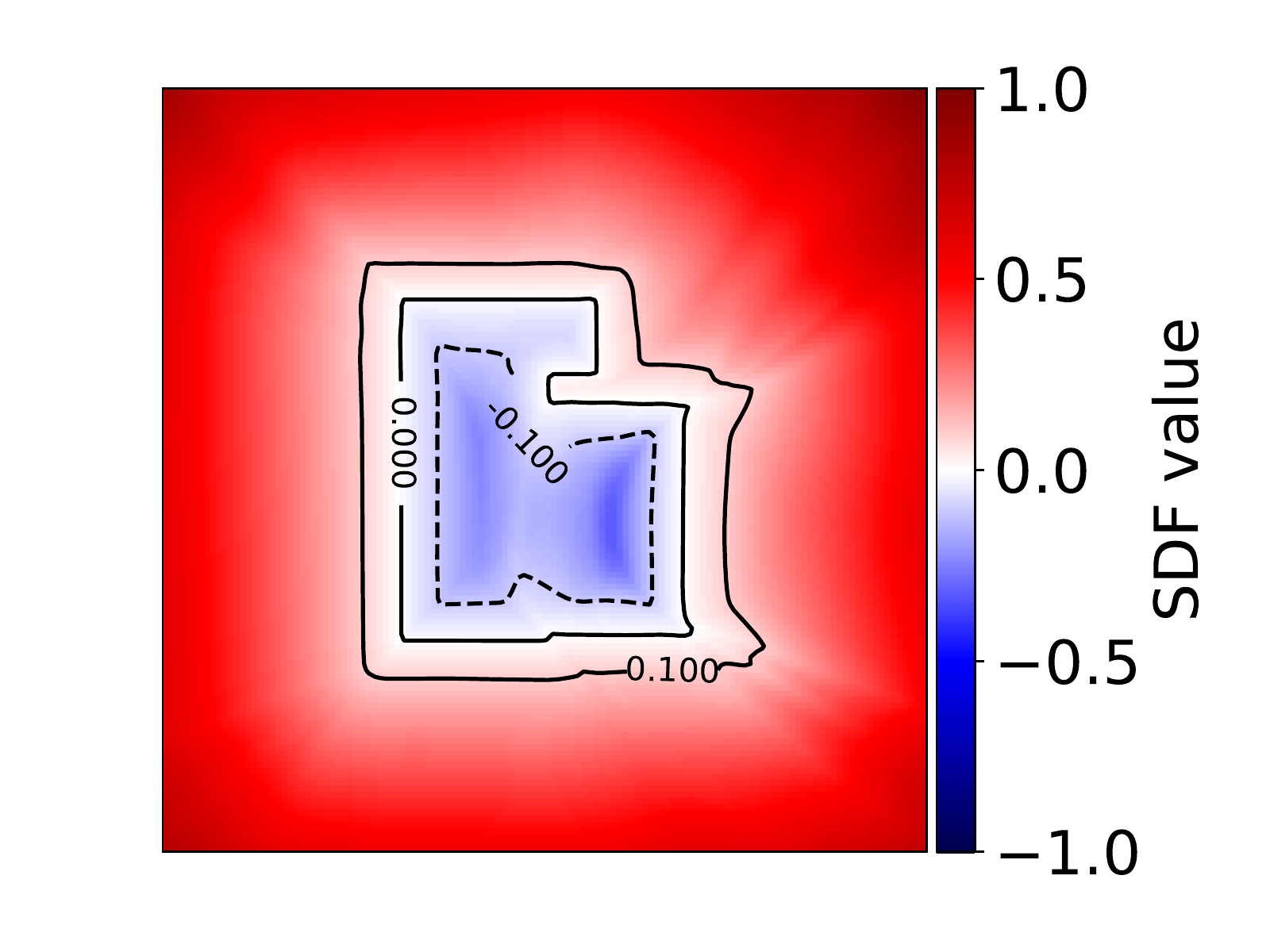}
\end{subfigure}
\begin{subfigure}{.24545\linewidth}
\includegraphics[trim=80 30 100 30,clip,width=.39\linewidth]{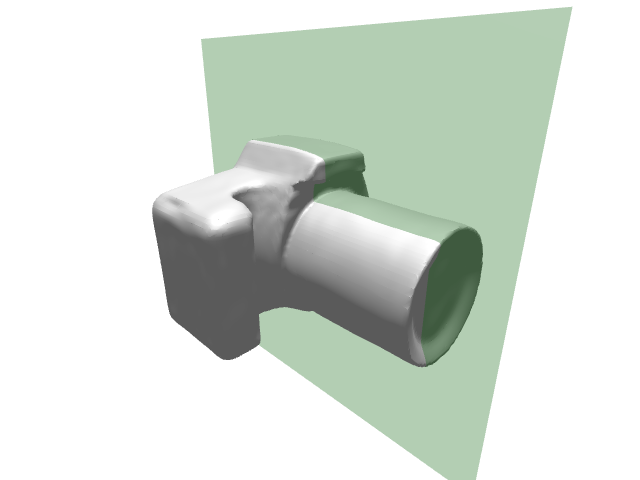}
\includegraphics[width=.59\linewidth]{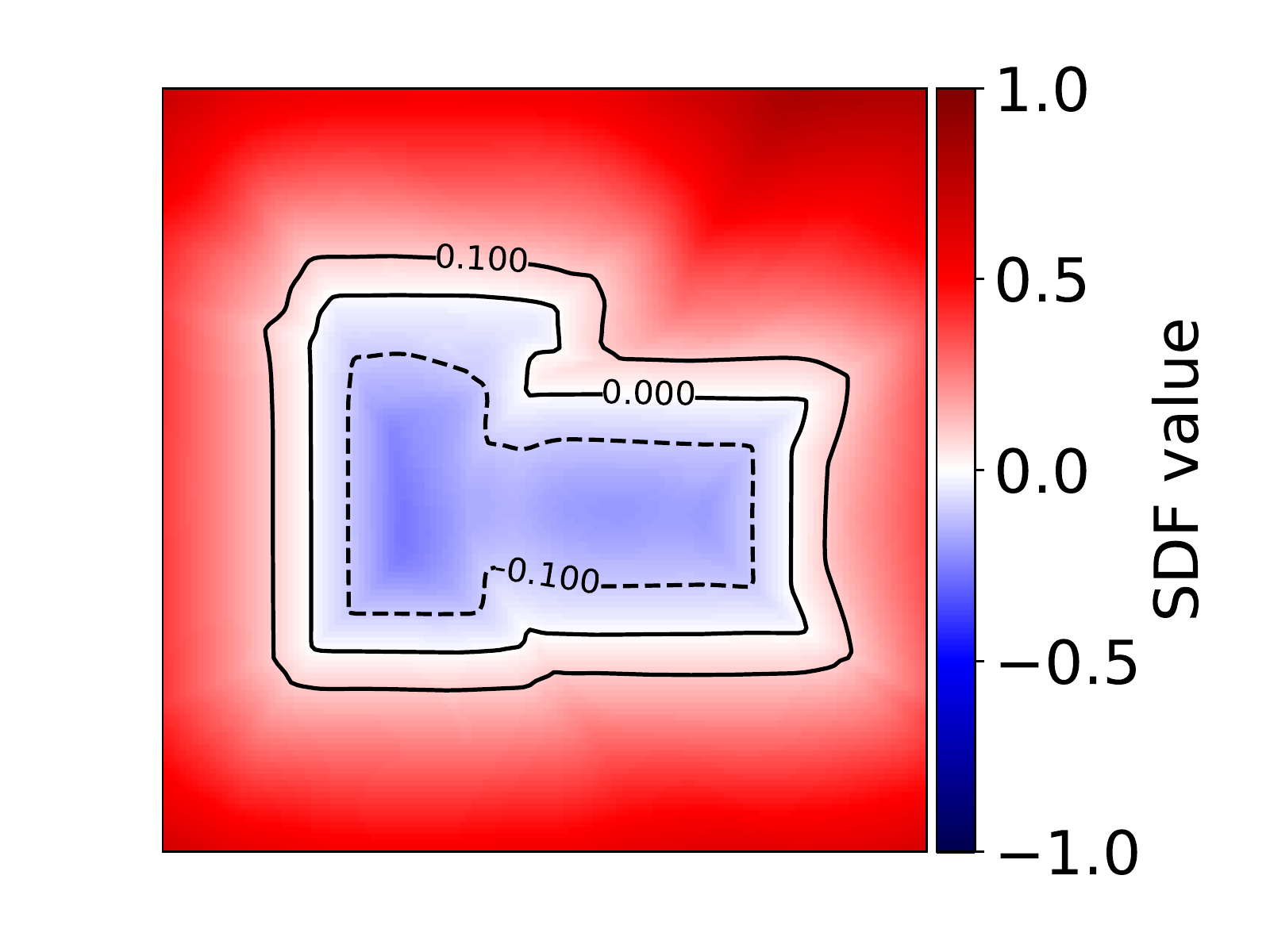}
\end{subfigure}
\begin{subfigure}{.24545\linewidth}
\includegraphics[trim=80 30 100 30,clip,width=.39\linewidth]{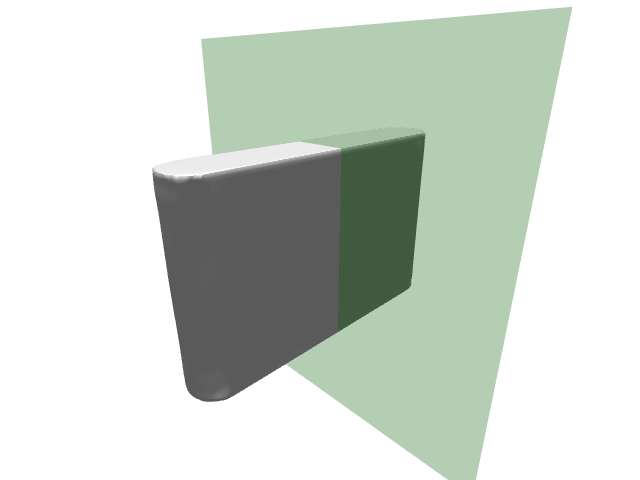}
\includegraphics[width=.59\linewidth]{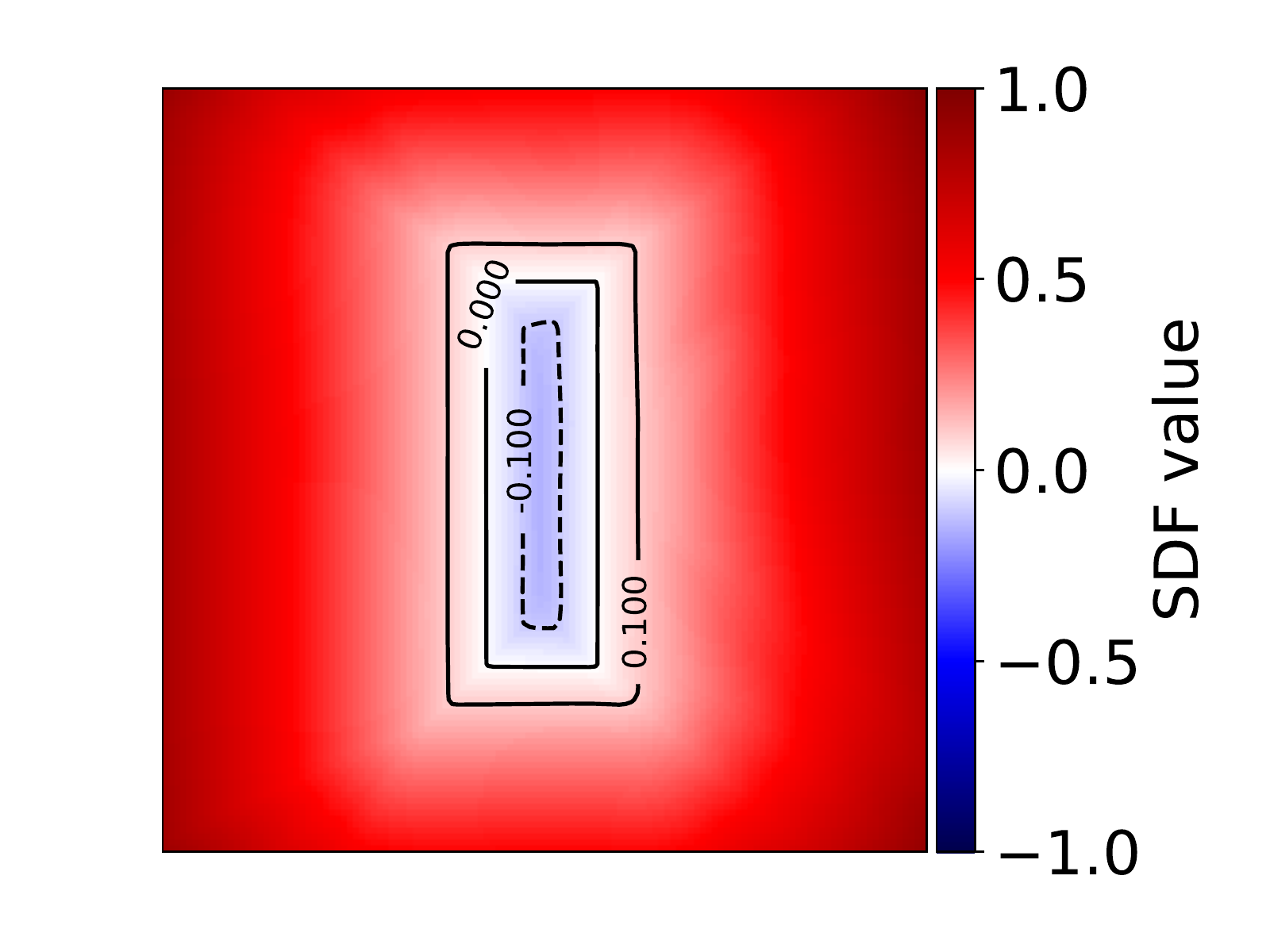}
\end{subfigure}
\begin{subfigure}{.24545\linewidth}
\includegraphics[trim=80 30 100 30,clip,width=.39\linewidth]{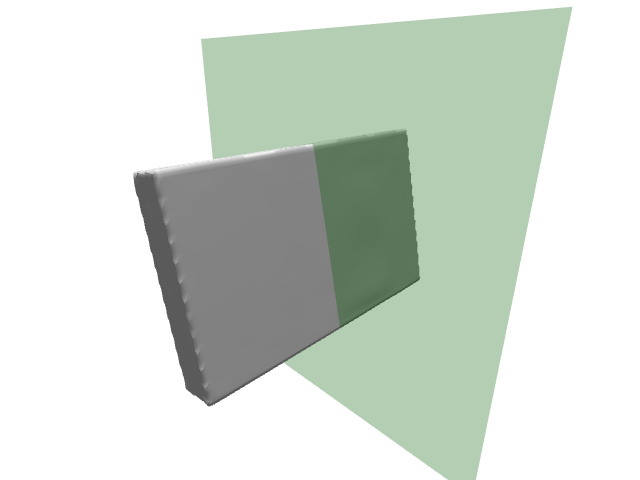}
\includegraphics[width=.59\linewidth]{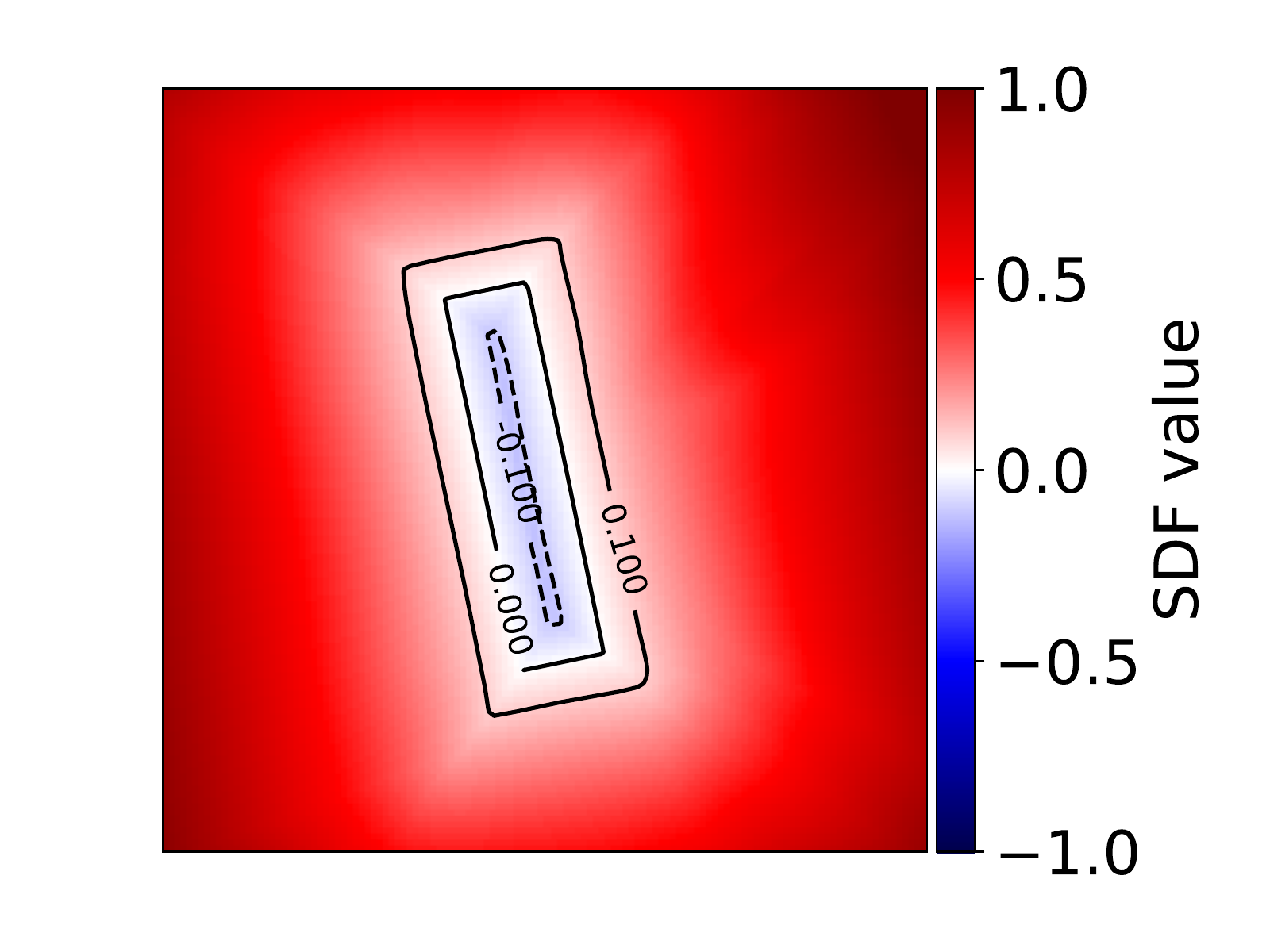}
\end{subfigure}
\begin{subfigure}{.24545\linewidth}
\includegraphics[trim=80 30 100 30,clip,width=.39\linewidth]{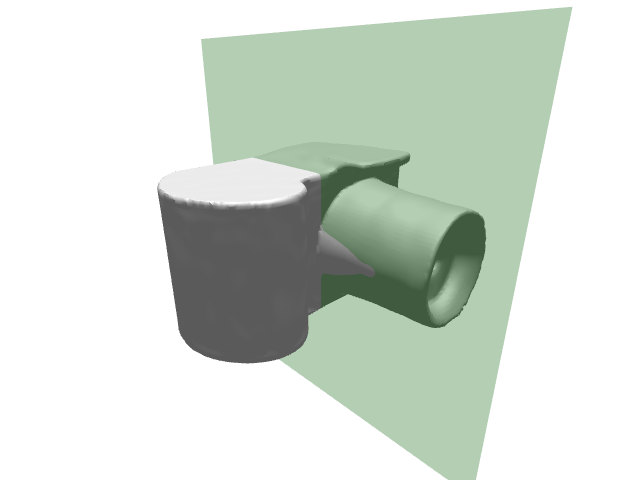}
\includegraphics[width=.59\linewidth]{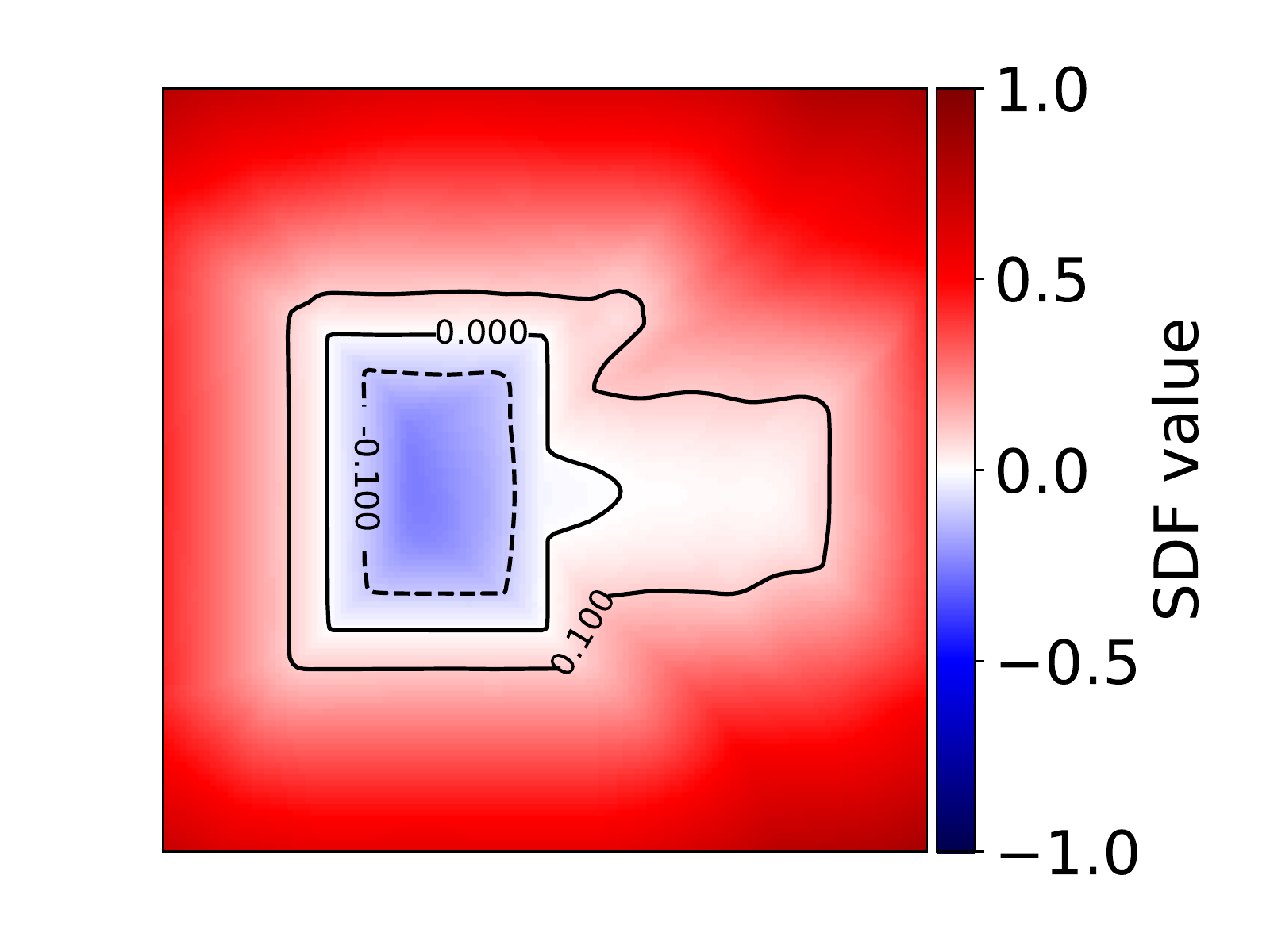}
\end{subfigure}
\begin{subfigure}{.24545\linewidth}
\includegraphics[trim=80 30 100 30,clip,width=.39\linewidth]{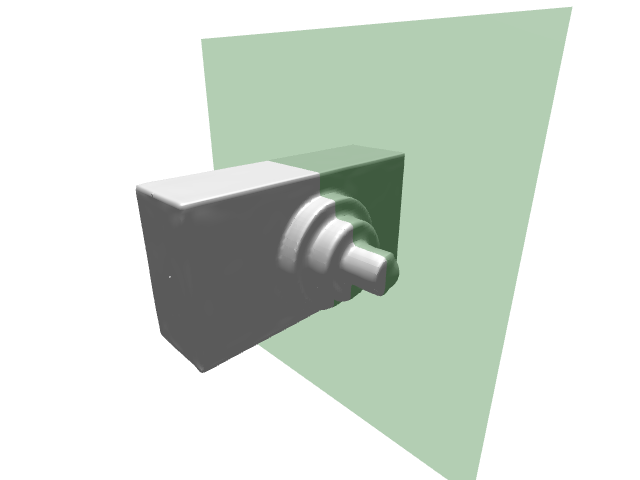}
\includegraphics[width=.59\linewidth]{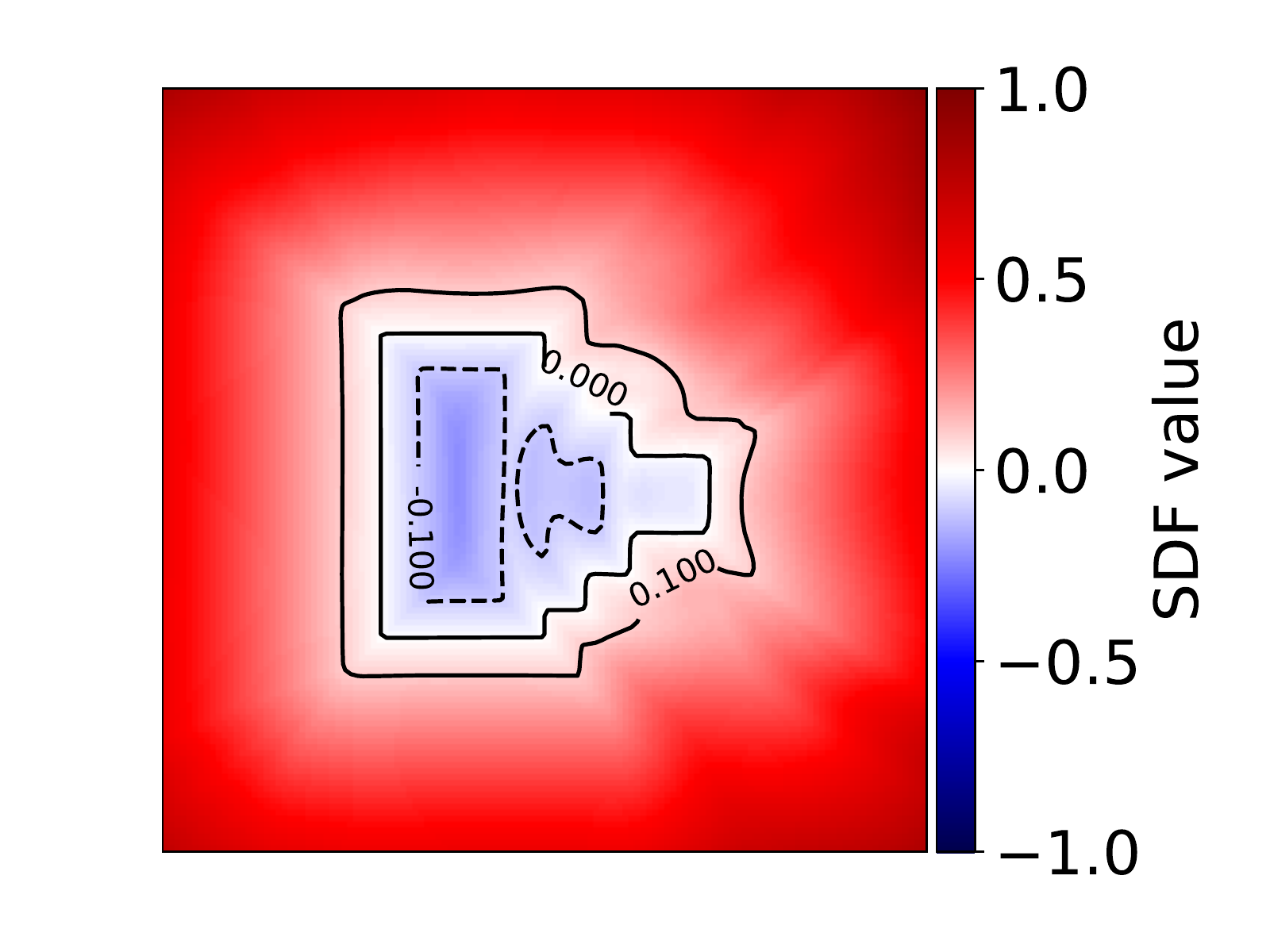}
\end{subfigure}
\begin{subfigure}{.24545\linewidth}
\includegraphics[trim=80 30 100 30,clip,width=.39\linewidth]{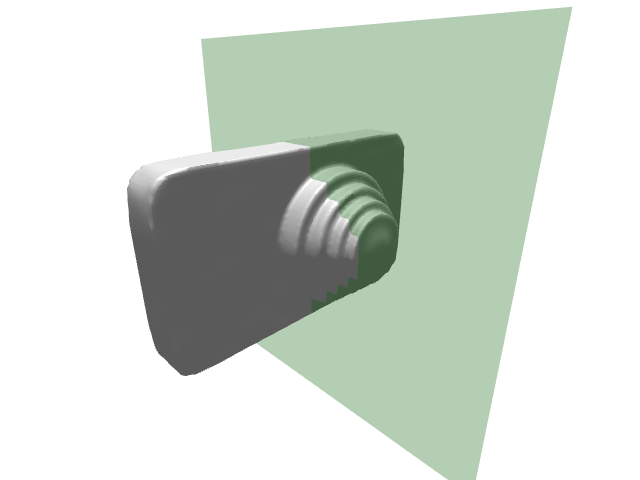}
\includegraphics[width=.59\linewidth]{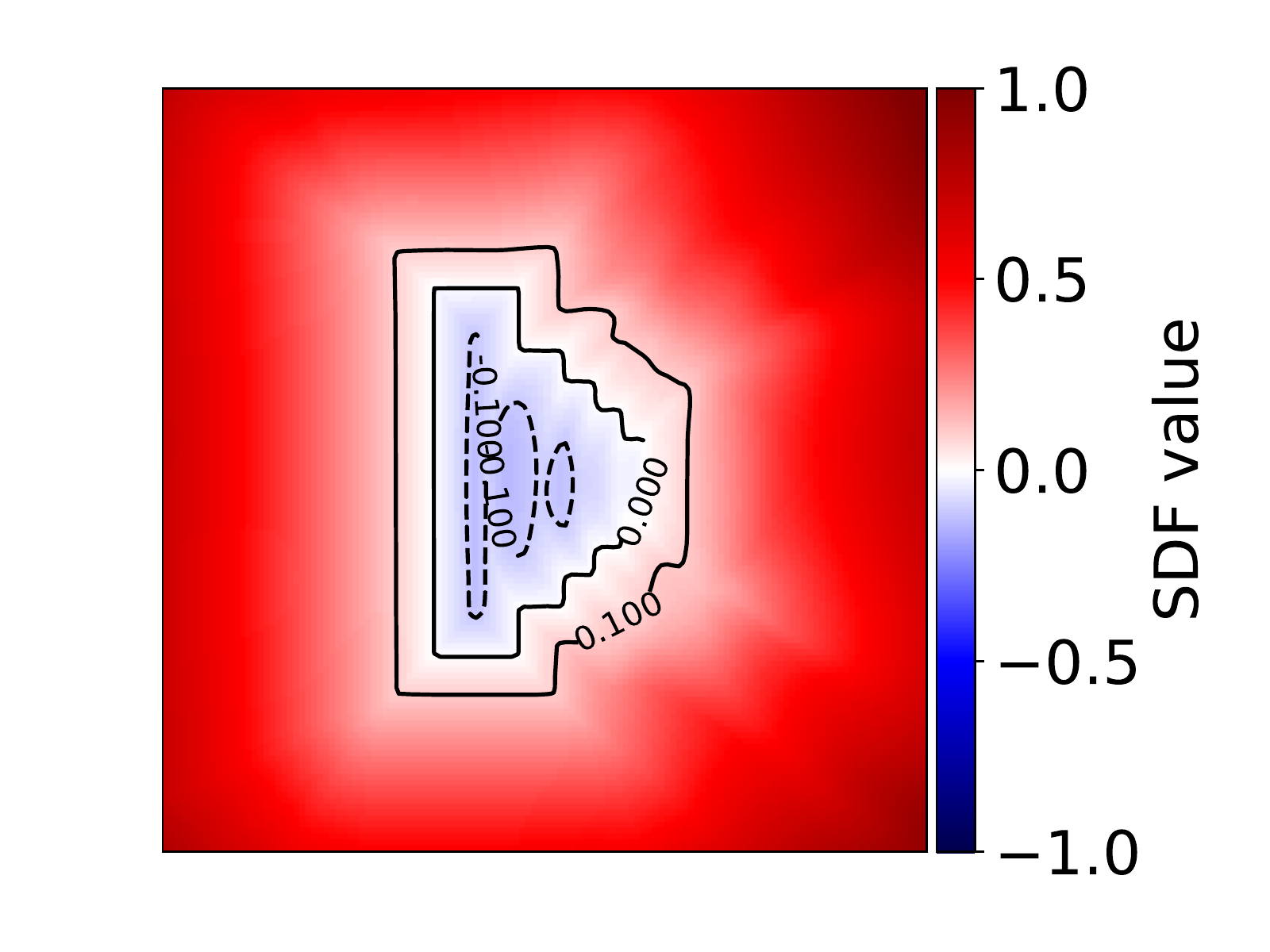}
\end{subfigure}
\begin{subfigure}{.24545\linewidth}
\includegraphics[trim=80 30 100 30,clip,width=.39\linewidth]{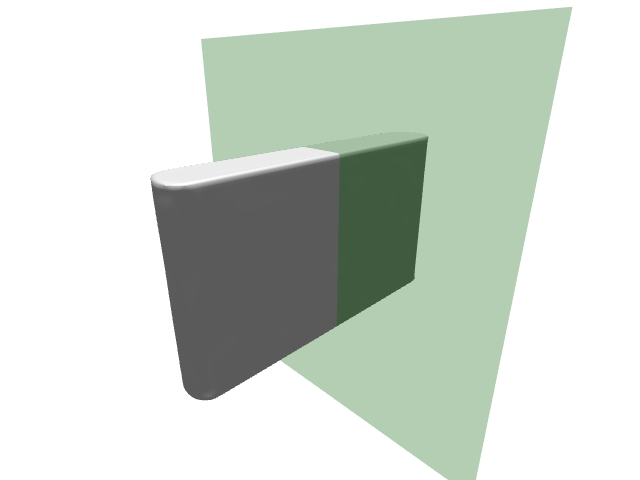}
\includegraphics[width=.59\linewidth]{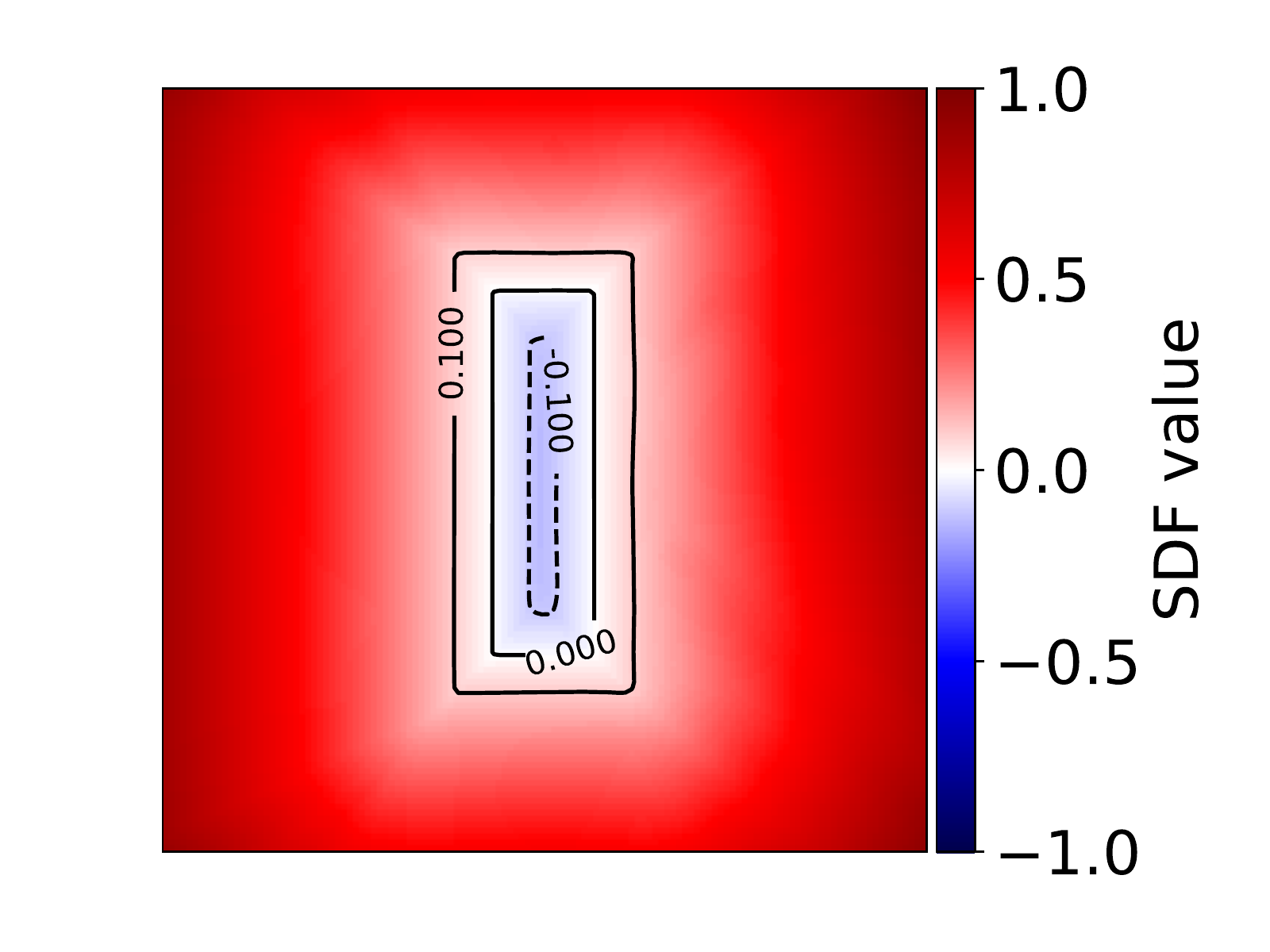}
\end{subfigure}
\begin{subfigure}{.24545\linewidth}
\includegraphics[trim=80 30 100 30,clip,width=.39\linewidth]{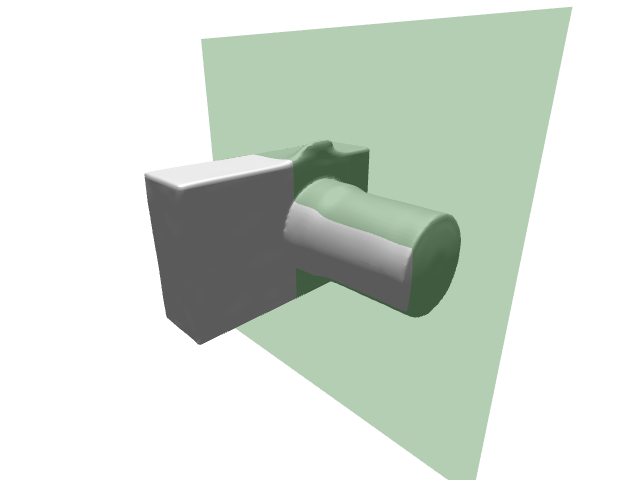}
\includegraphics[width=.59\linewidth]{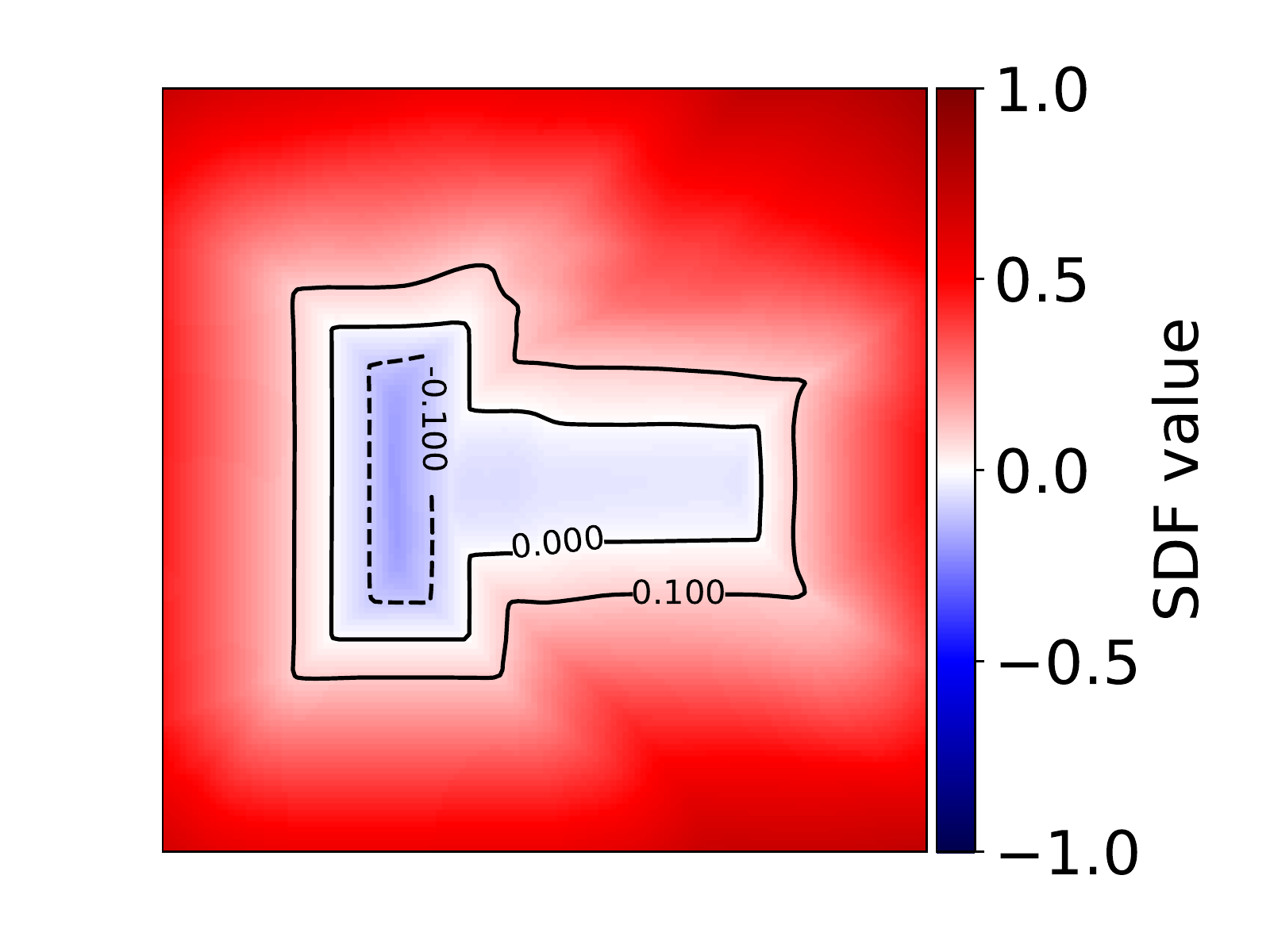}
\end{subfigure}
\begin{subfigure}{.24545\linewidth}
\includegraphics[trim=80 30 100 30,clip,width=.39\linewidth]{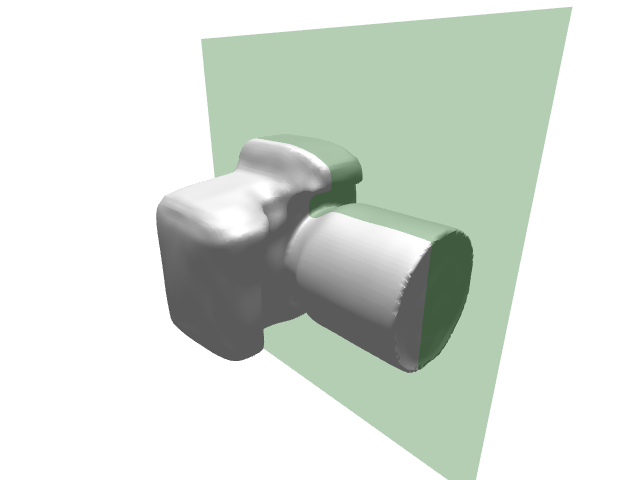}
\includegraphics[width=.59\linewidth]{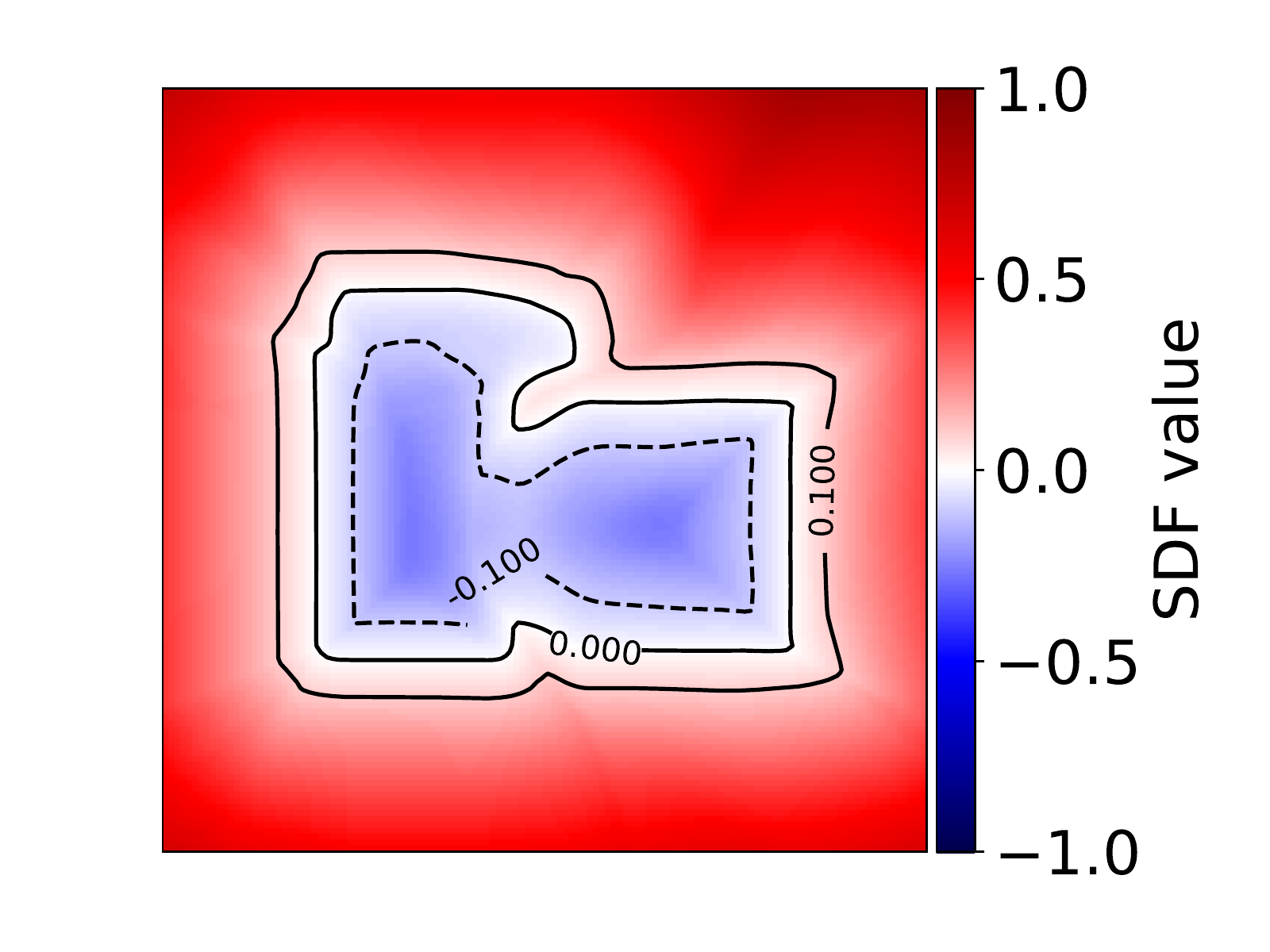}
\end{subfigure}

%% file: figures/shapespace_renderings/mug.tex
\begin{subfigure}{.24545\linewidth}
\includegraphics[trim=80 30 100 30,clip,width=.39\linewidth]{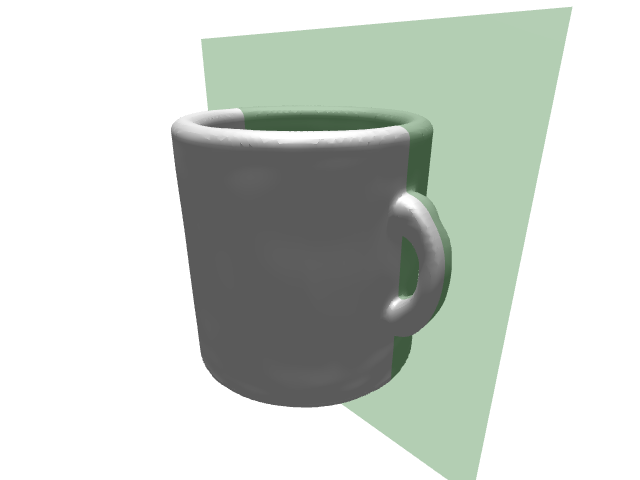}
\includegraphics[width=.59\linewidth]{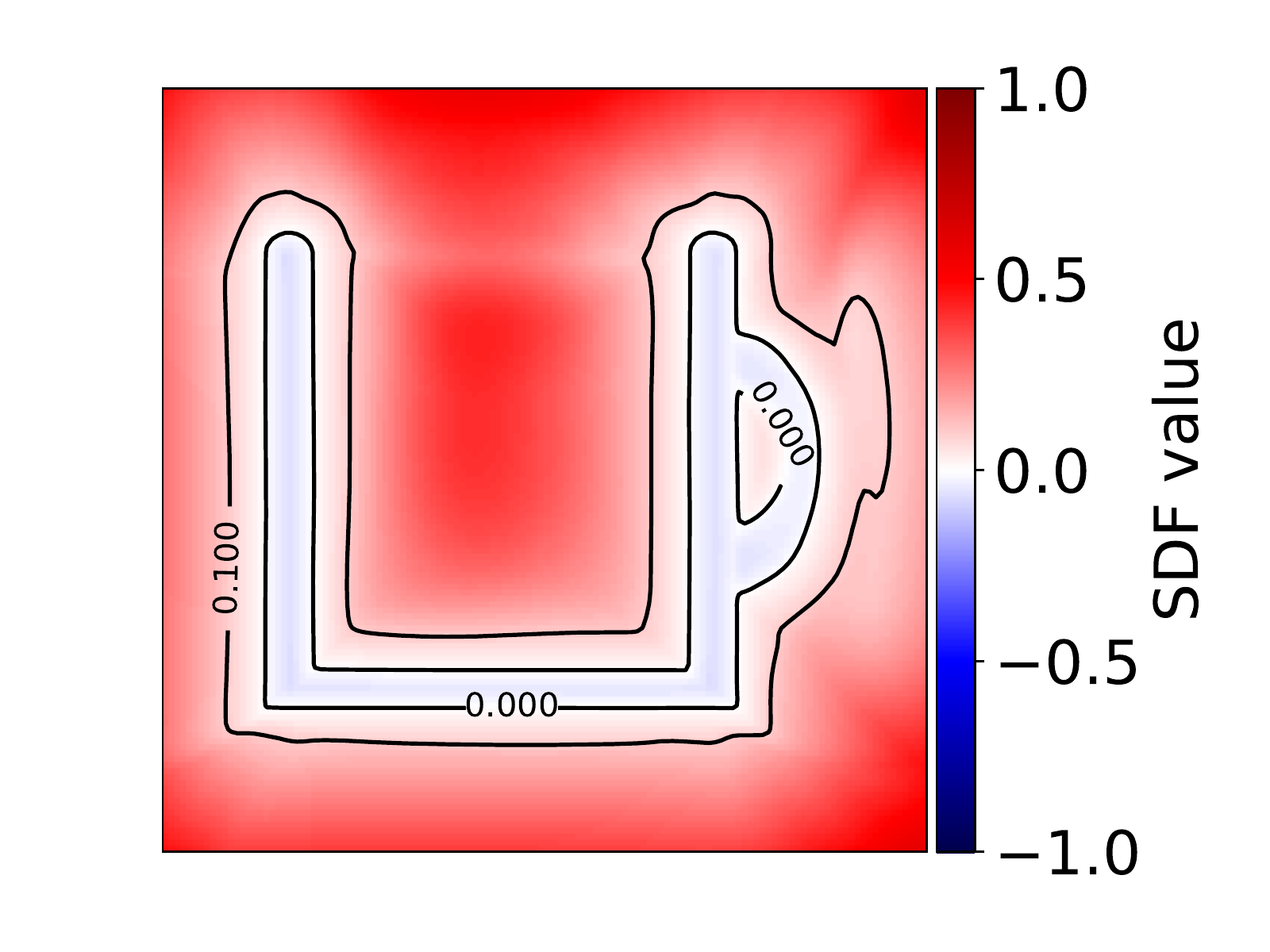}
\end{subfigure}
\begin{subfigure}{.24545\linewidth}
\includegraphics[trim=80 30 100 30,clip,width=.39\linewidth]{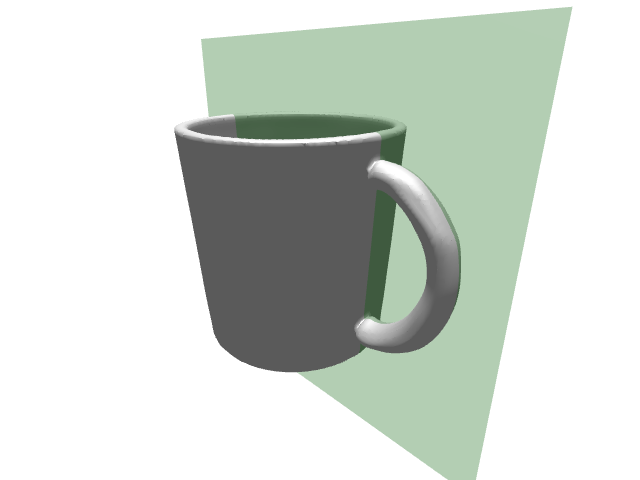}
\includegraphics[width=.59\linewidth]{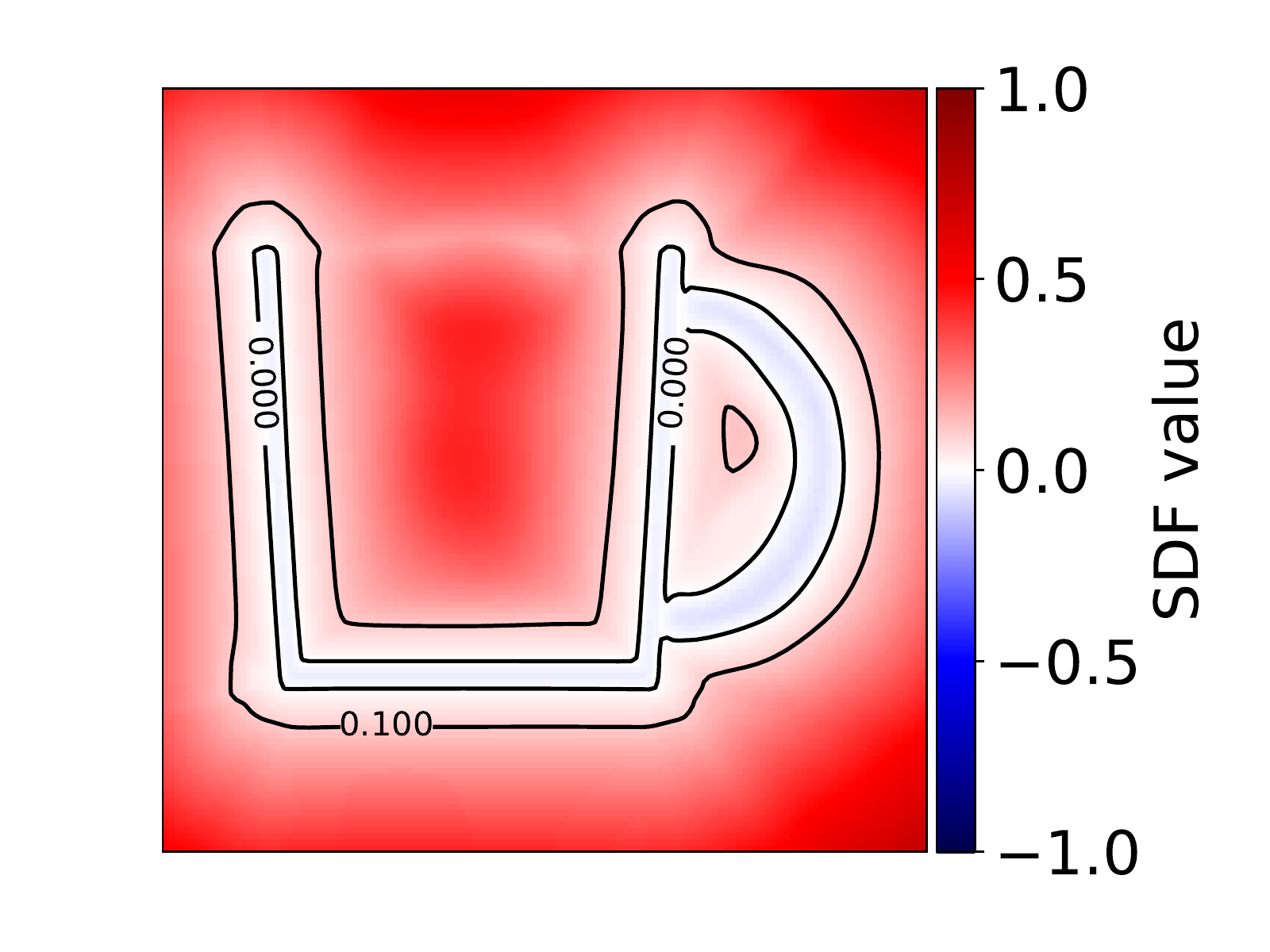}
\end{subfigure}
\begin{subfigure}{.24545\linewidth}
\includegraphics[trim=80 30 100 30,clip,width=.39\linewidth]{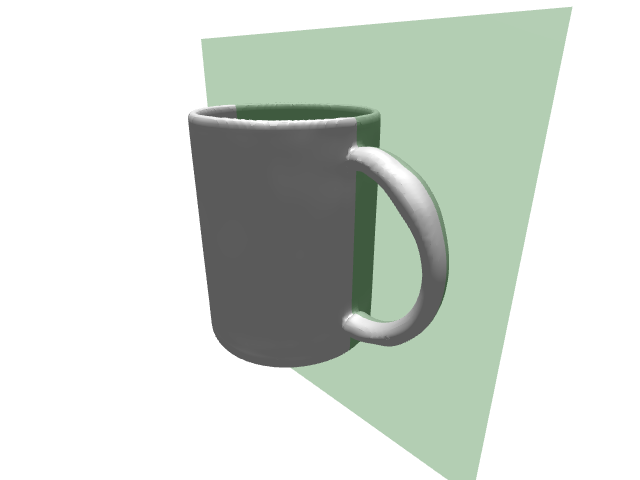}
\includegraphics[width=.59\linewidth]{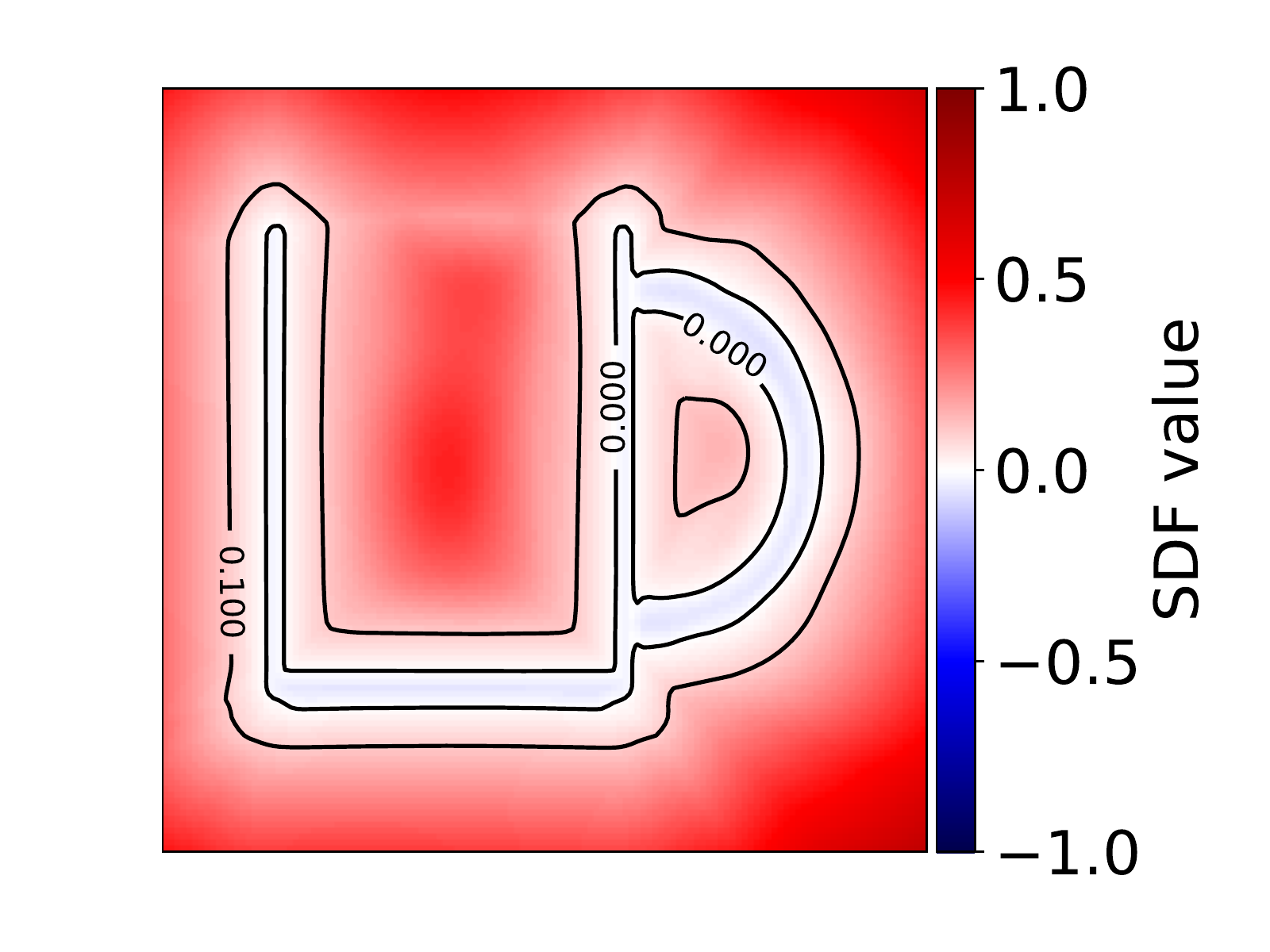}
\end{subfigure}
\begin{subfigure}{.24545\linewidth}
\includegraphics[trim=80 30 100 30,clip,width=.39\linewidth]{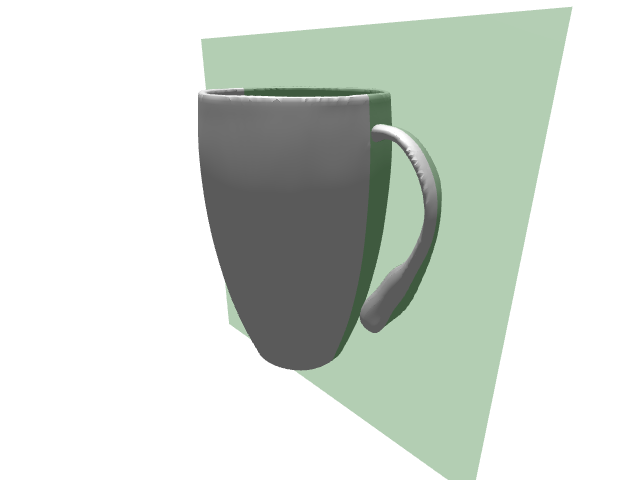}
\includegraphics[width=.59\linewidth]{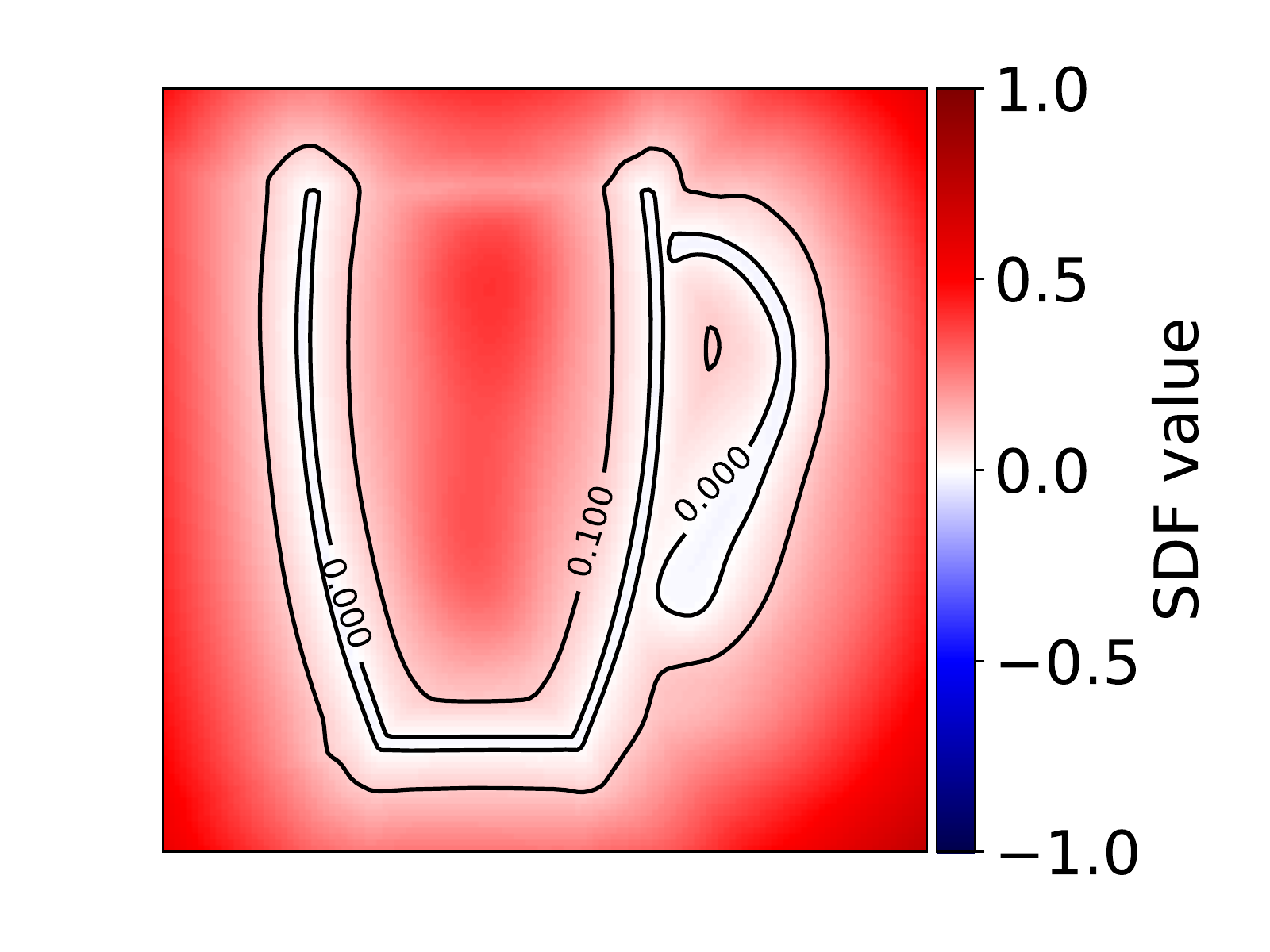}
\end{subfigure}
\begin{subfigure}{.24545\linewidth}
\includegraphics[trim=80 30 100 30,clip,width=.39\linewidth]{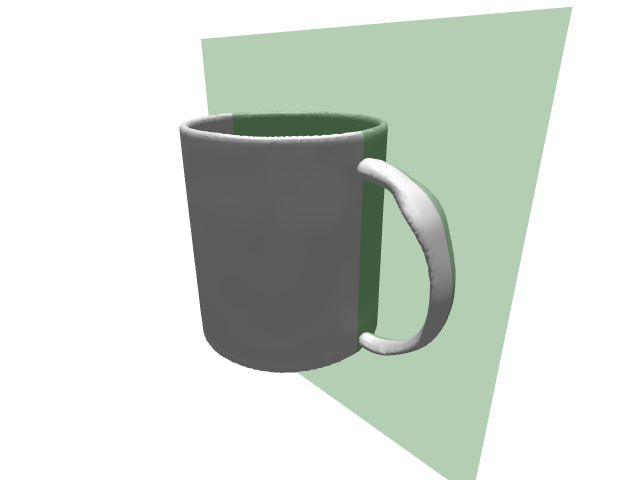}
\includegraphics[width=.59\linewidth]{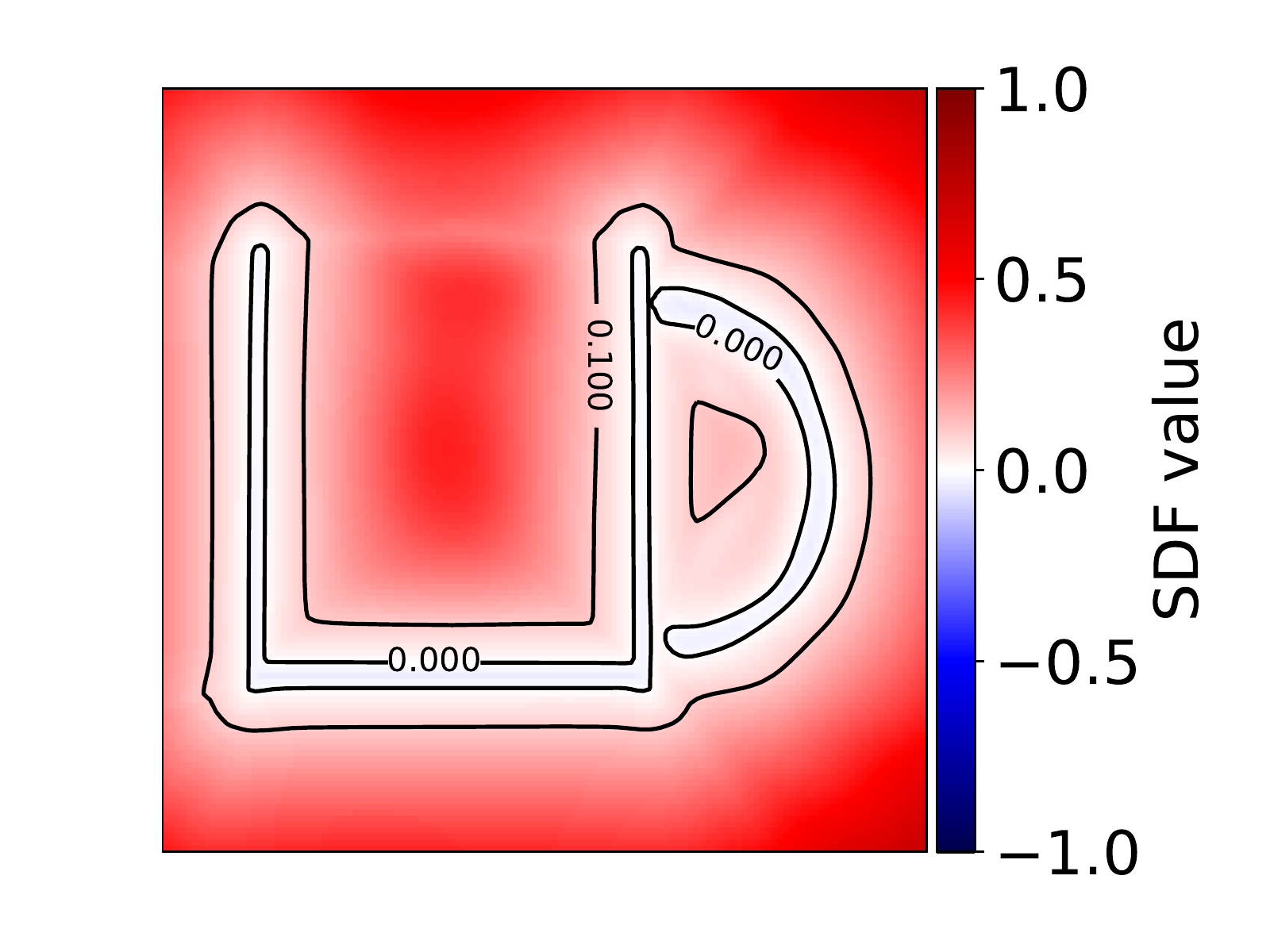}
\end{subfigure}
\begin{subfigure}{.24545\linewidth}
\includegraphics[trim=80 30 100 30,clip,width=.39\linewidth]{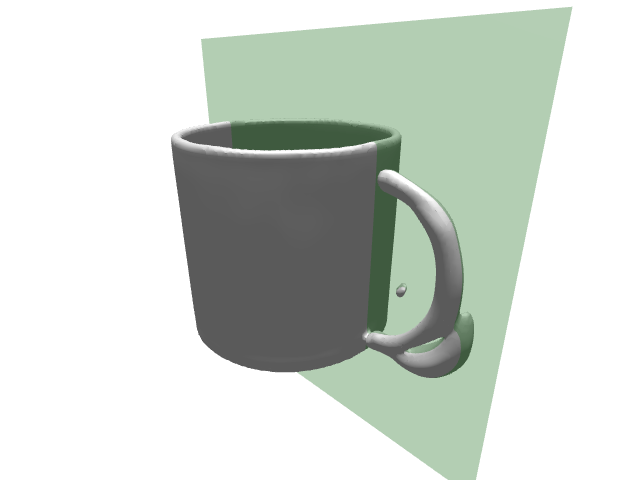}
\includegraphics[width=.59\linewidth]{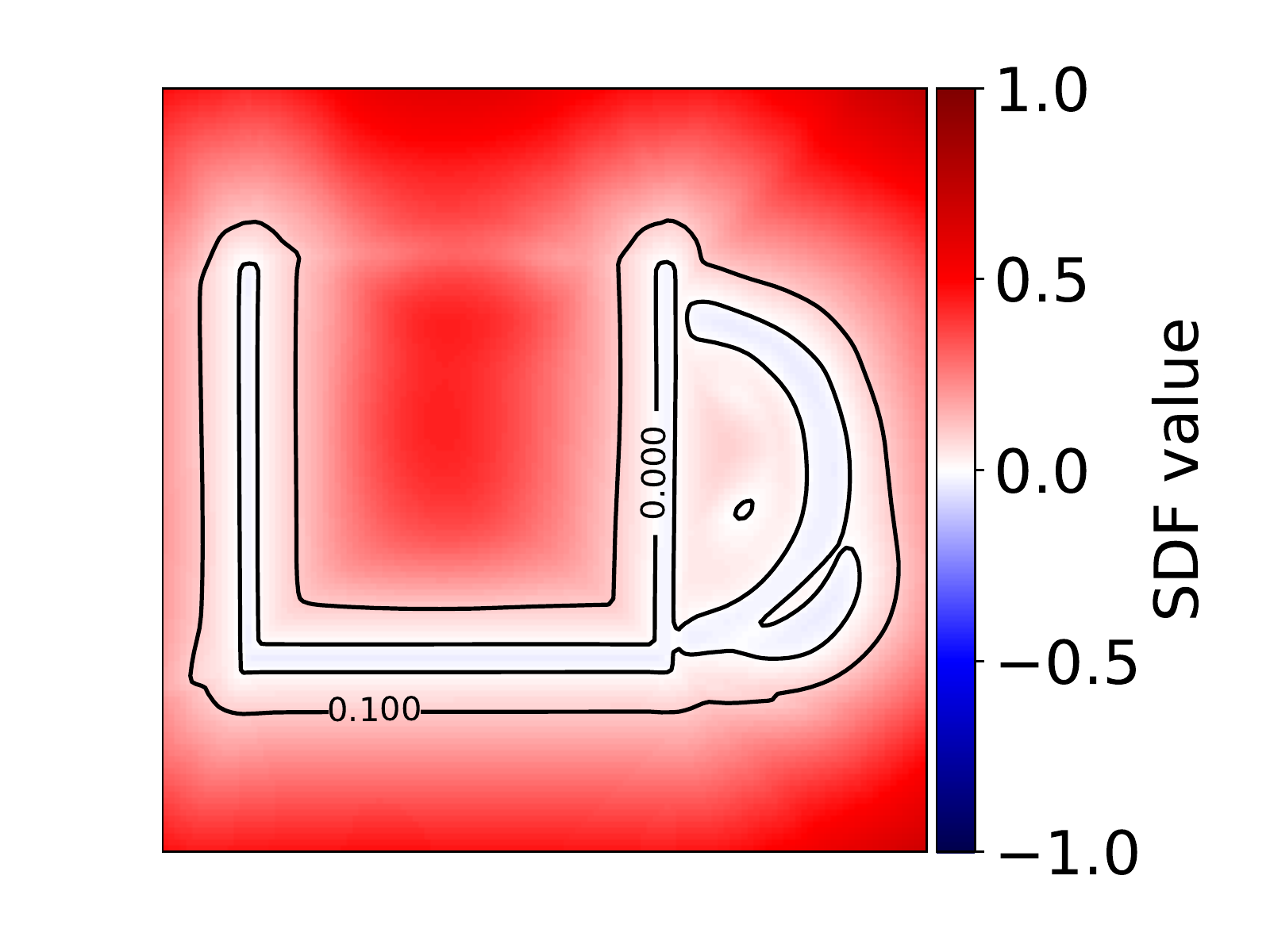}
\end{subfigure}
\begin{subfigure}{.24545\linewidth}
\includegraphics[trim=80 30 100 30,clip,width=.39\linewidth]{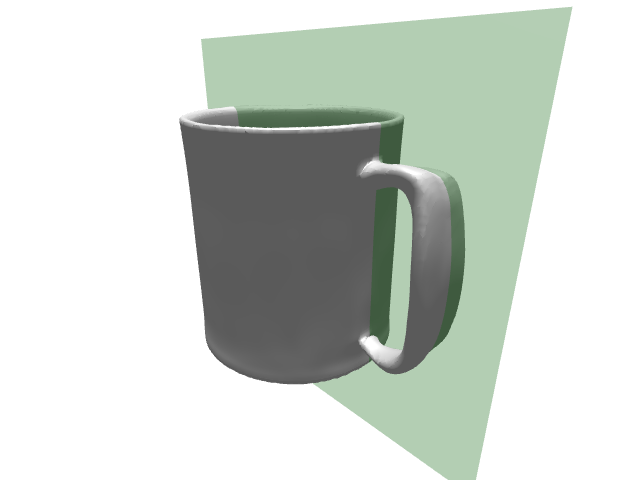}
\includegraphics[width=.59\linewidth]{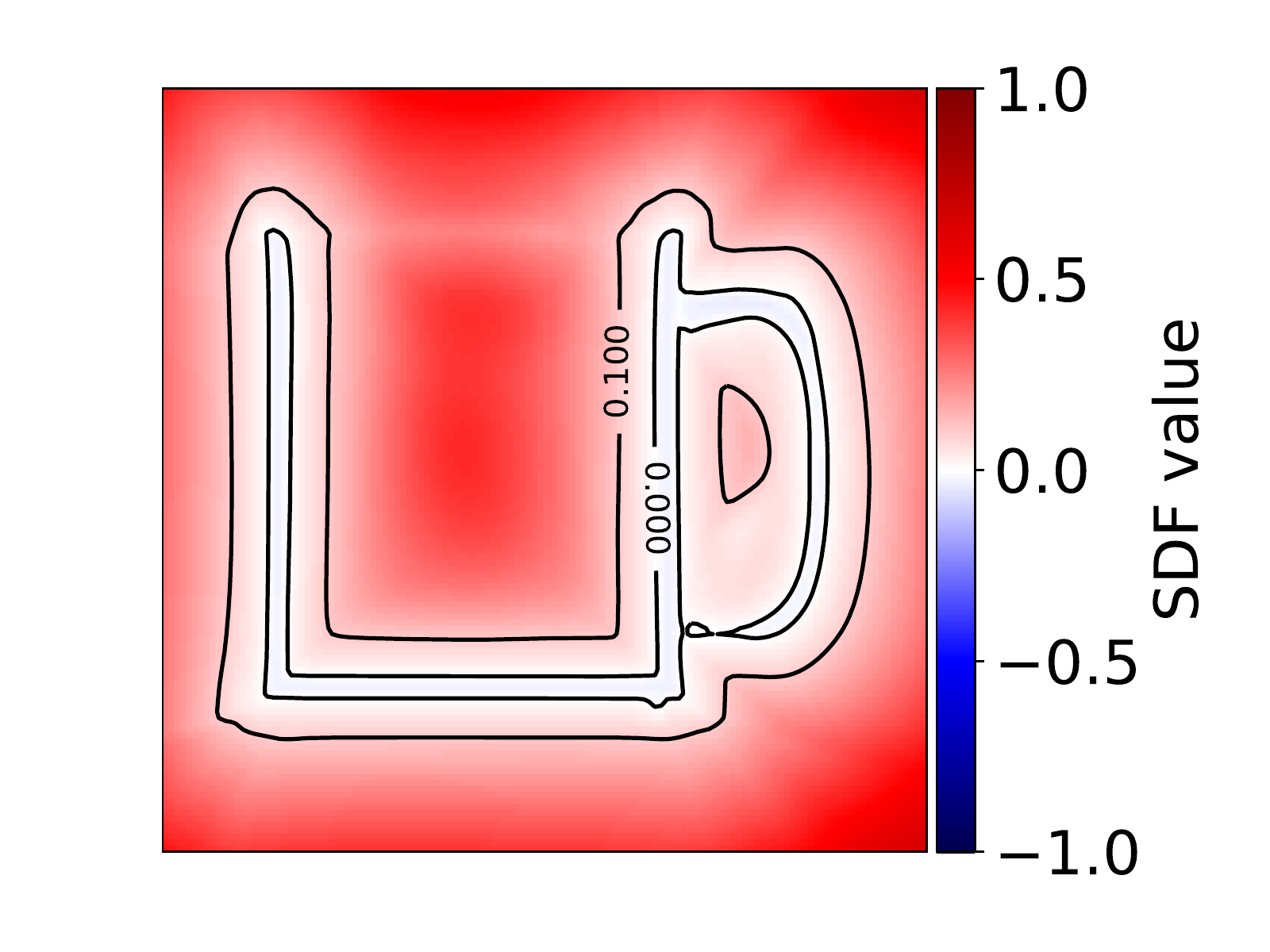}
\end{subfigure}
\begin{subfigure}{.24545\linewidth}
\includegraphics[trim=80 30 100 30,clip,width=.39\linewidth]{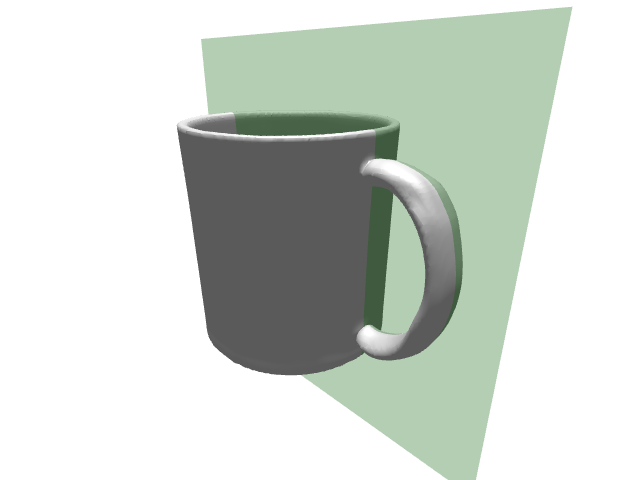}
\includegraphics[width=.59\linewidth]{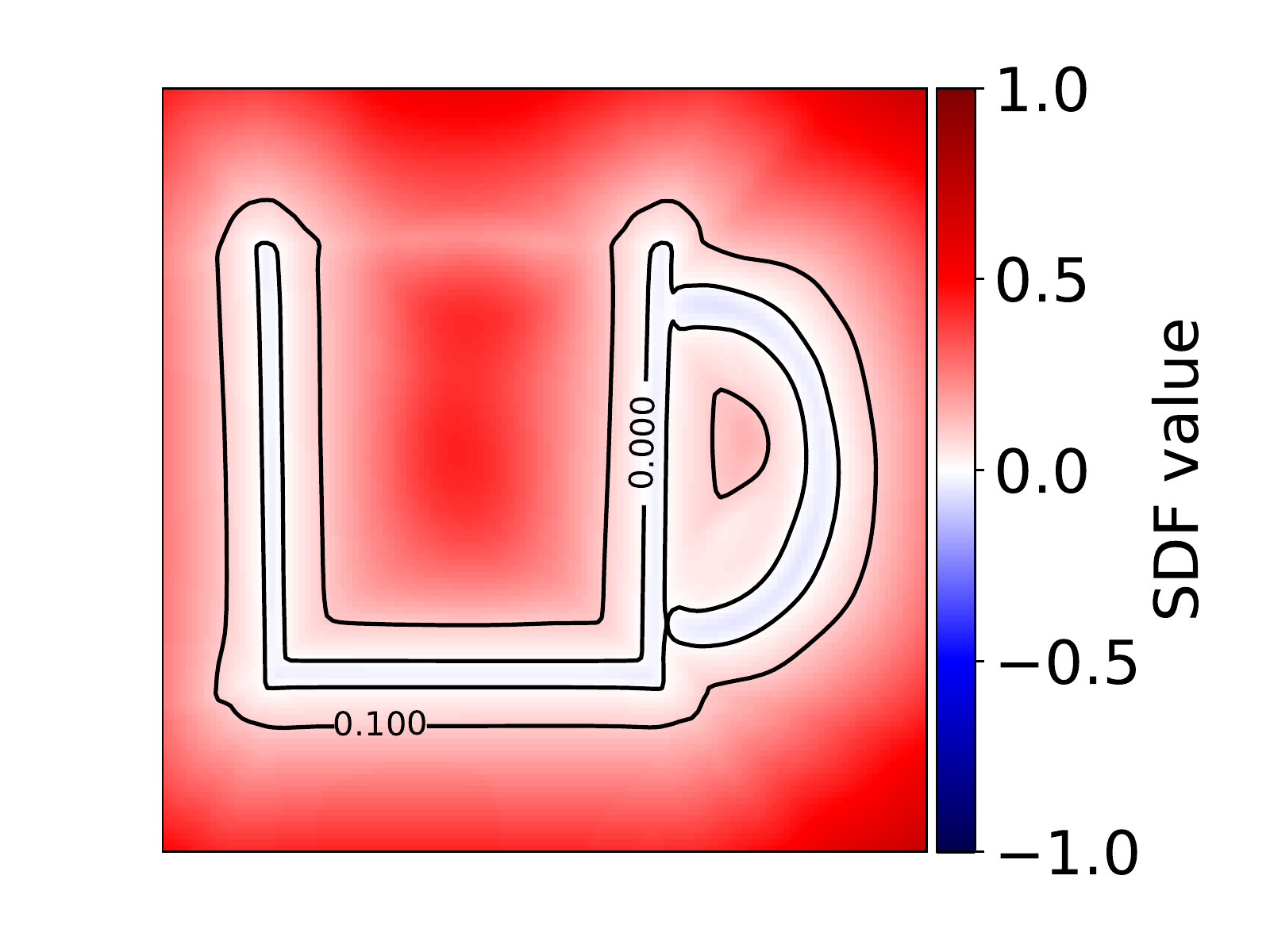}
\end{subfigure}
\begin{subfigure}{.24545\linewidth}
\includegraphics[trim=80 30 100 30,clip,width=.39\linewidth]{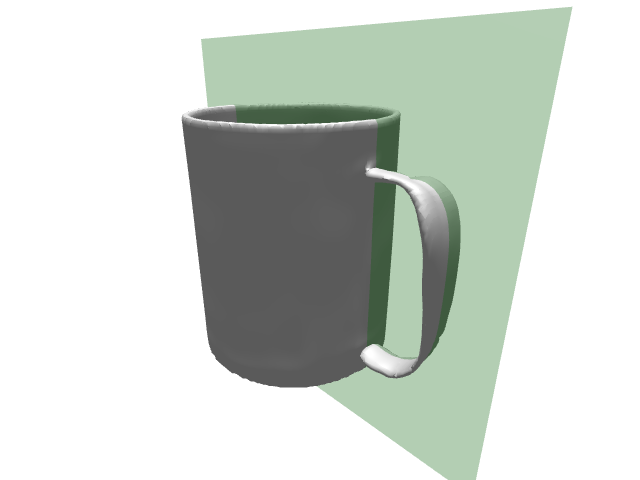}
\includegraphics[width=.59\linewidth]{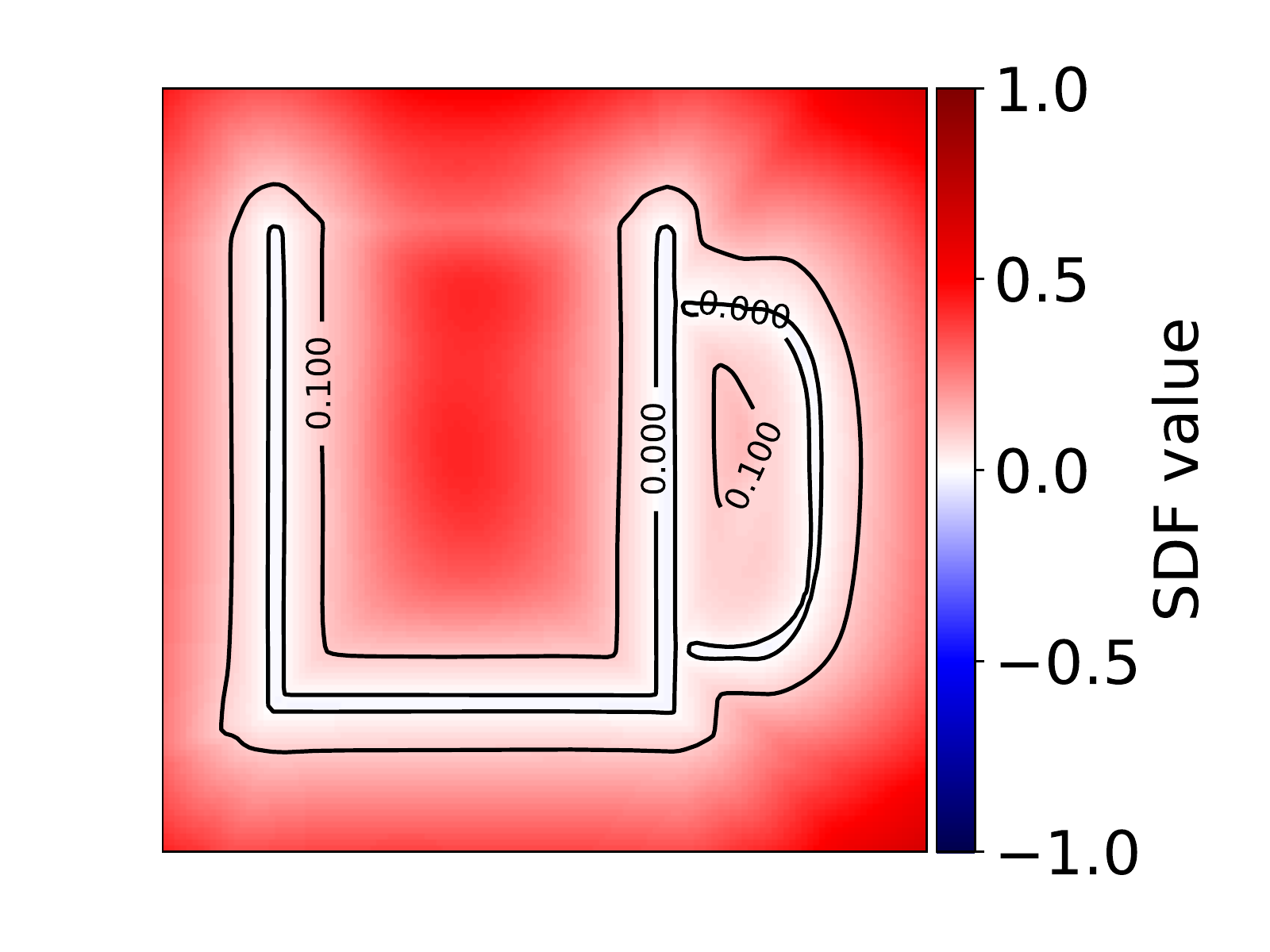}
\end{subfigure}
\begin{subfigure}{.24545\linewidth}
\includegraphics[trim=80 30 100 30,clip,width=.39\linewidth]{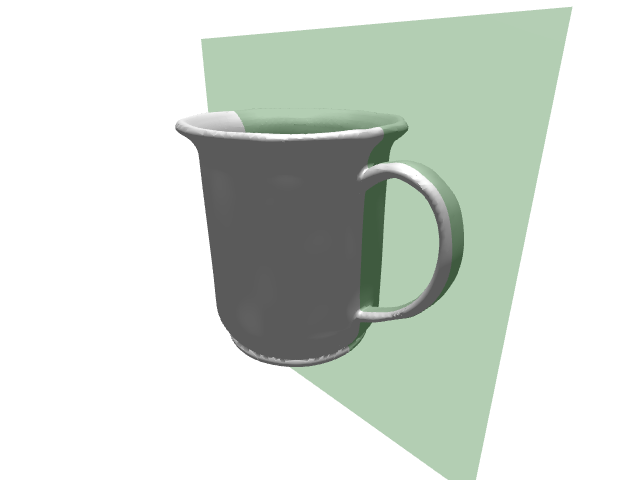}
\includegraphics[width=.59\linewidth]{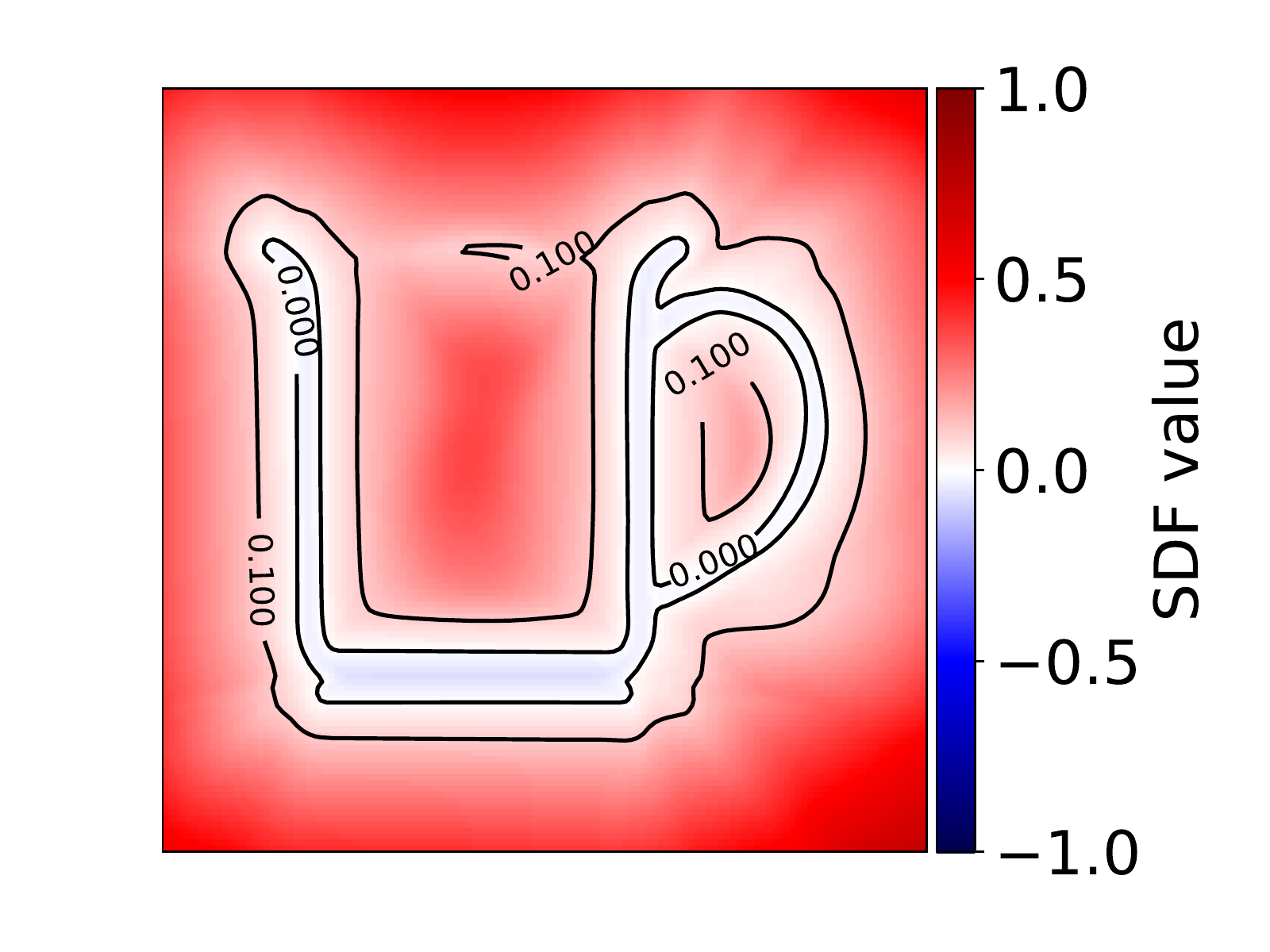}
\end{subfigure}
\begin{subfigure}{.24545\linewidth}
\includegraphics[trim=80 30 100 30,clip,width=.39\linewidth]{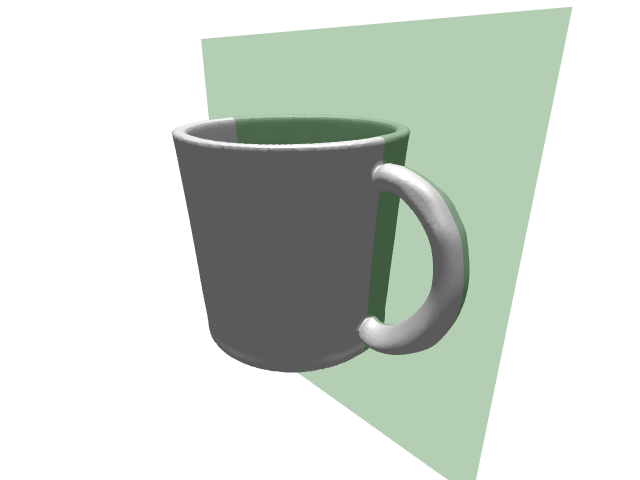}
\includegraphics[width=.59\linewidth]{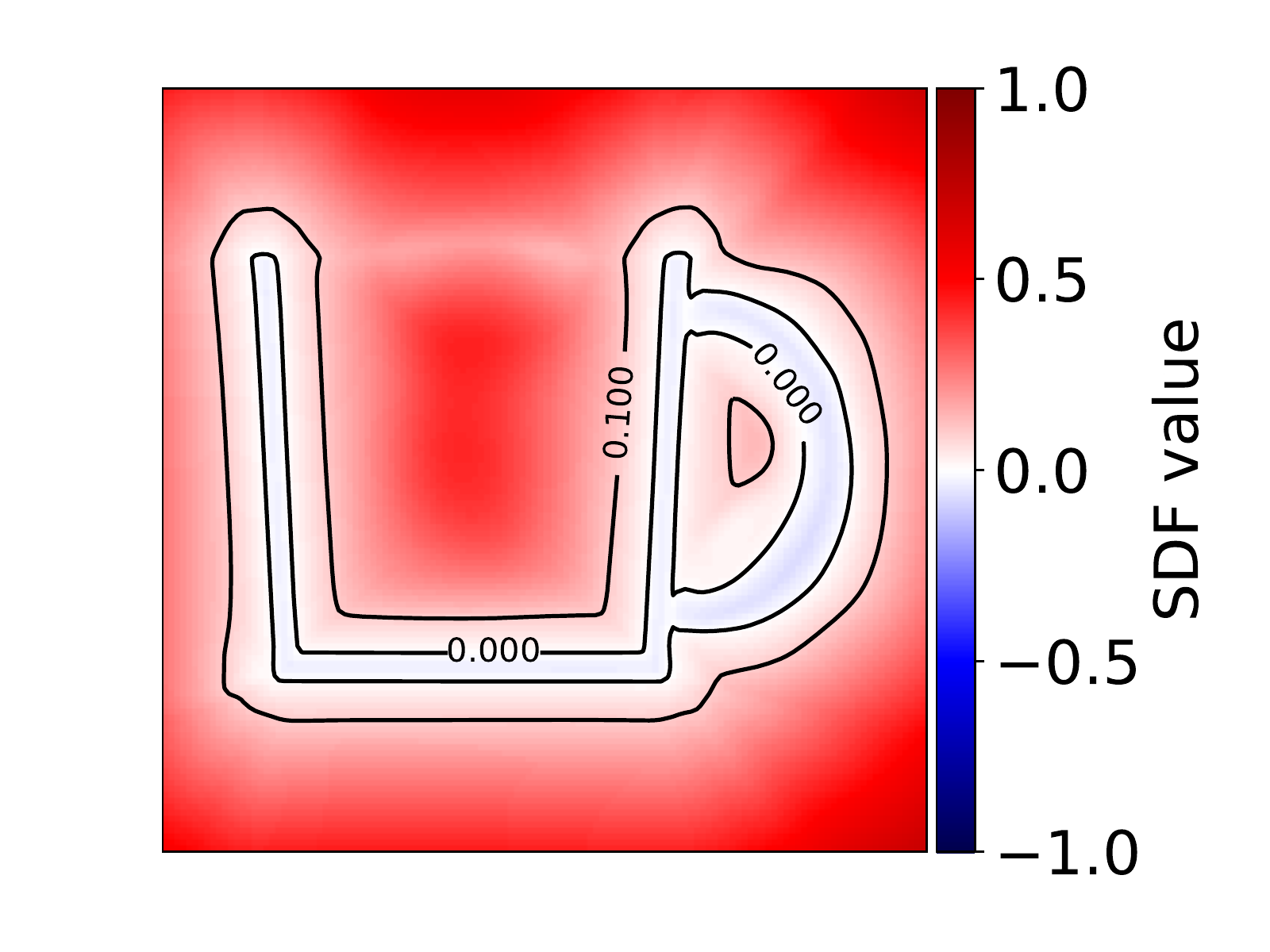}
\end{subfigure}
\begin{subfigure}{.24545\linewidth}
\includegraphics[trim=80 30 100 30,clip,width=.39\linewidth]{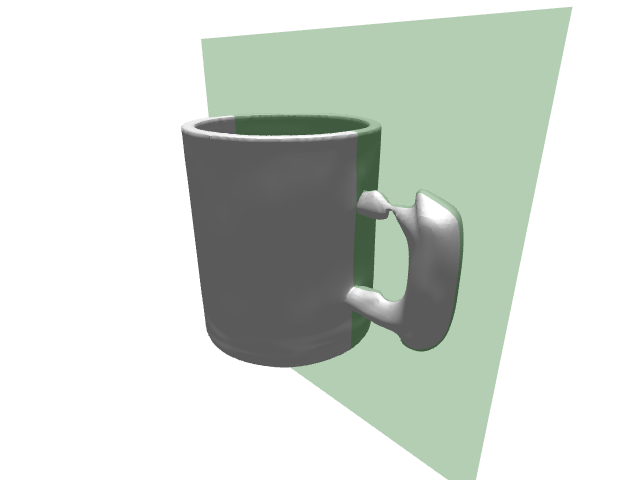}
\includegraphics[width=.59\linewidth]{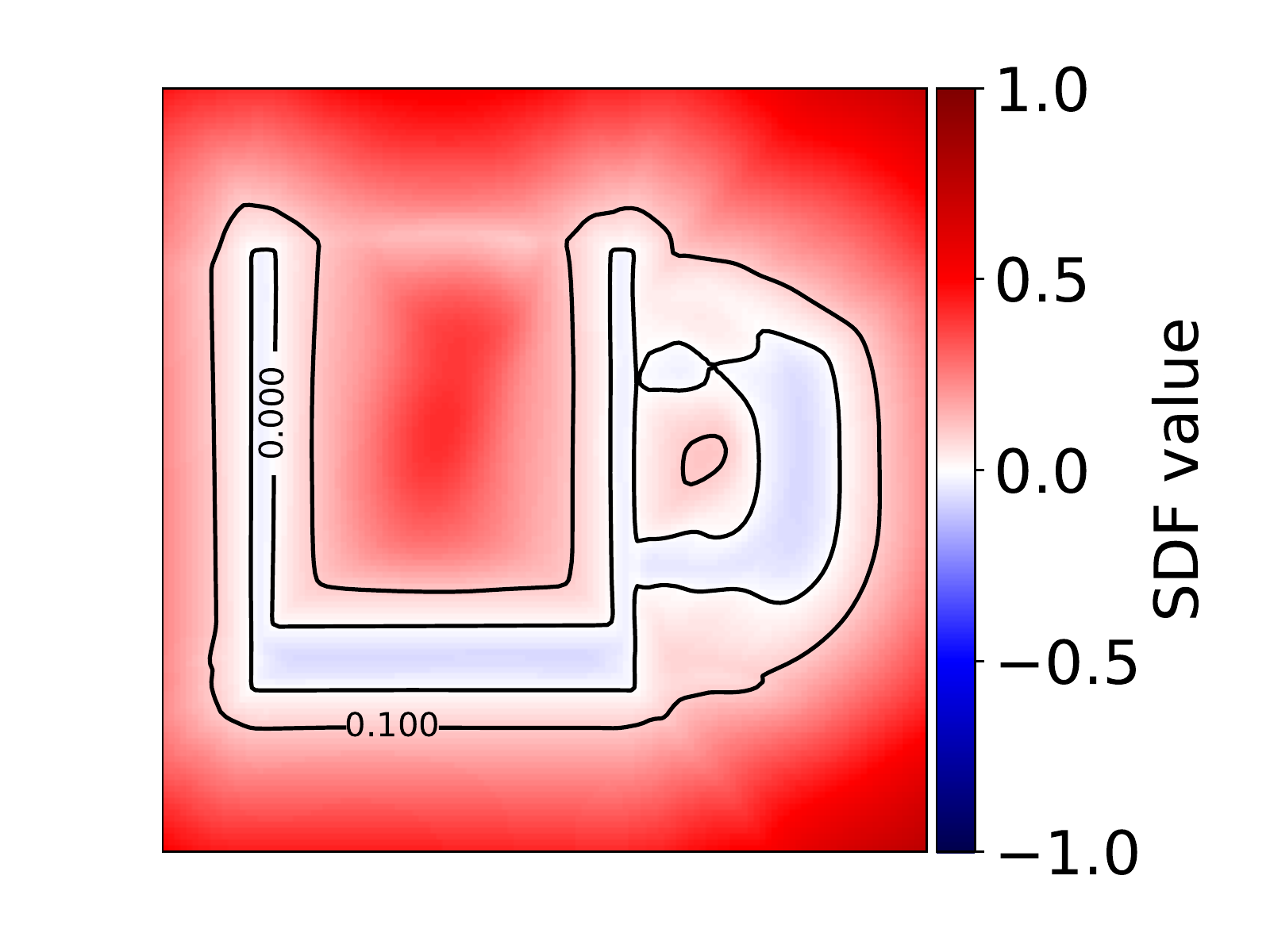}
\end{subfigure}
\begin{subfigure}{.24545\linewidth}
\includegraphics[trim=80 30 100 30,clip,width=.39\linewidth]{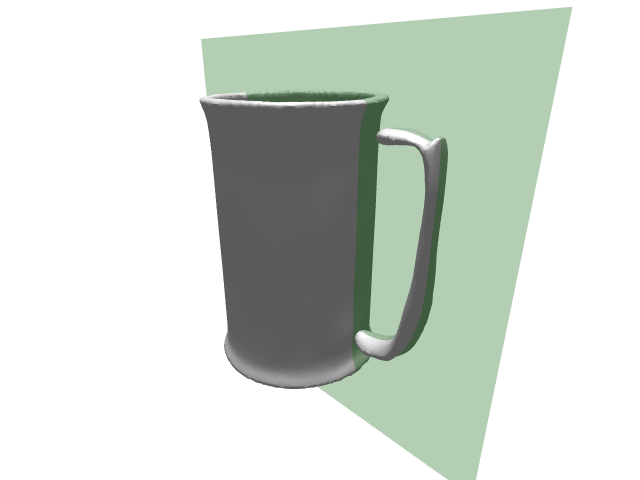}
\includegraphics[width=.59\linewidth]{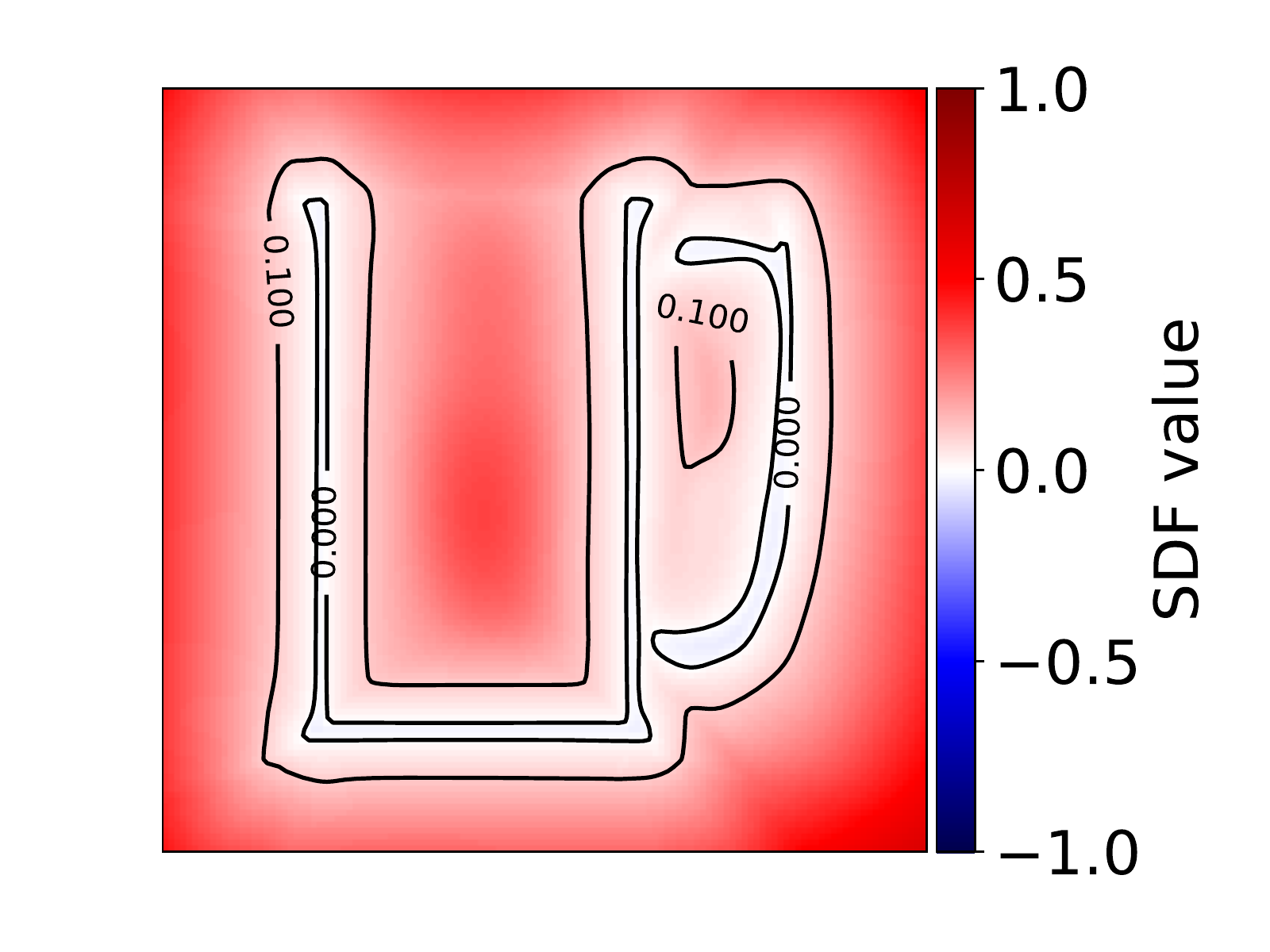}
\end{subfigure}
\begin{subfigure}{.24545\linewidth}
\includegraphics[trim=80 30 100 30,clip,width=.39\linewidth]{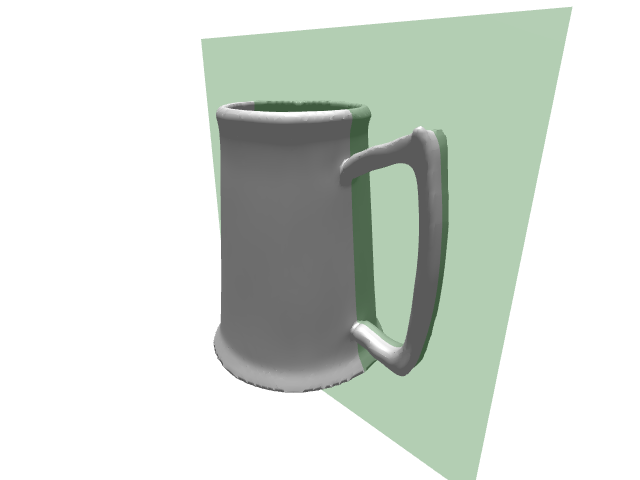}
\includegraphics[width=.59\linewidth]{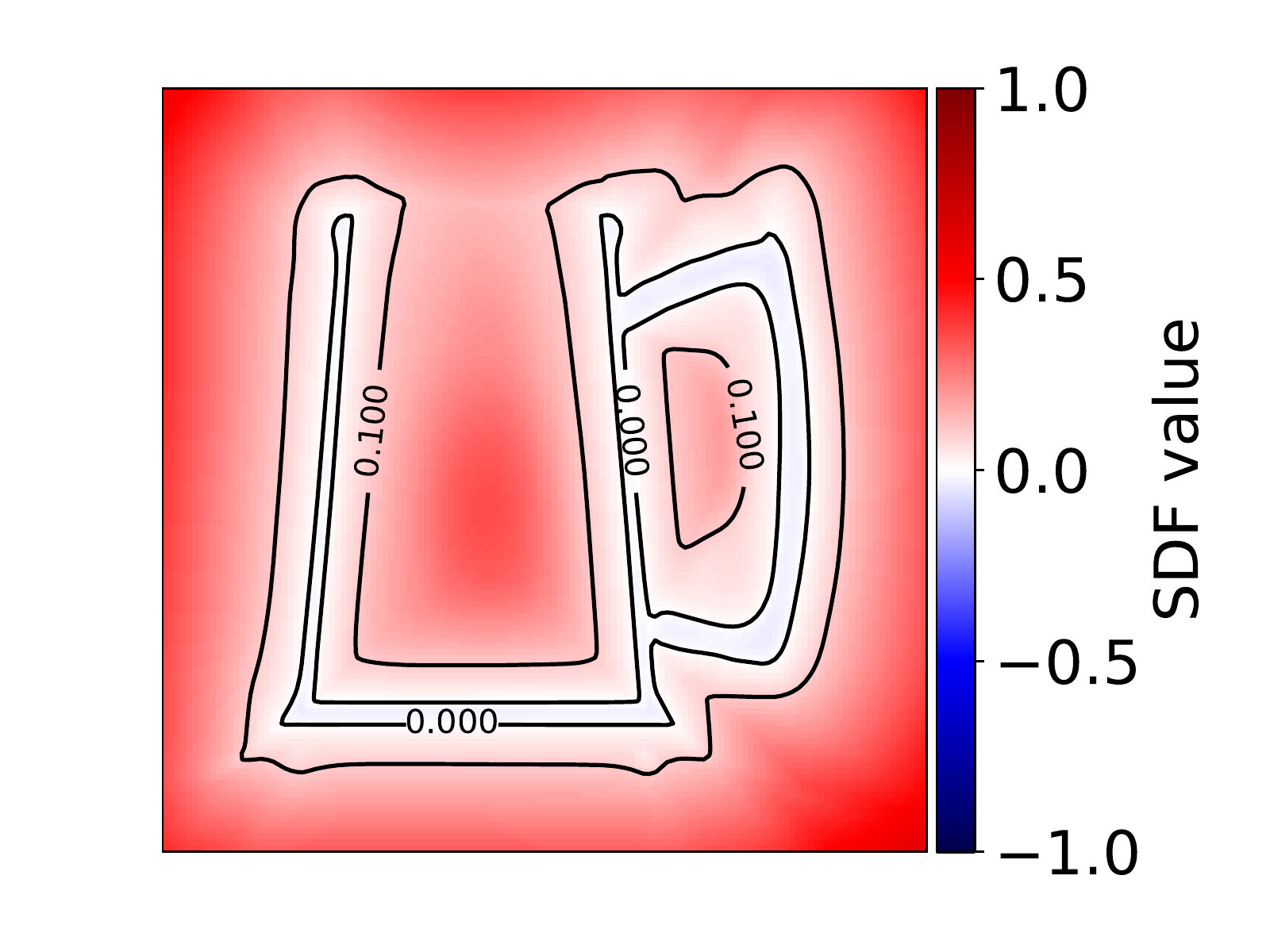}
\end{subfigure}
\begin{subfigure}{.24545\linewidth}
\includegraphics[trim=80 30 100 30,clip,width=.39\linewidth]{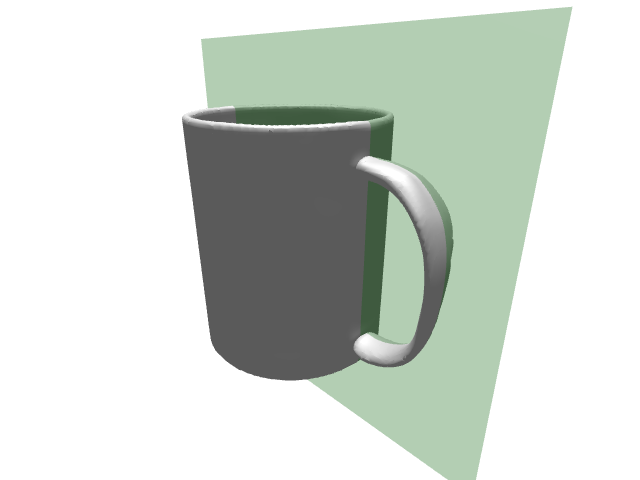}
\includegraphics[width=.59\linewidth]{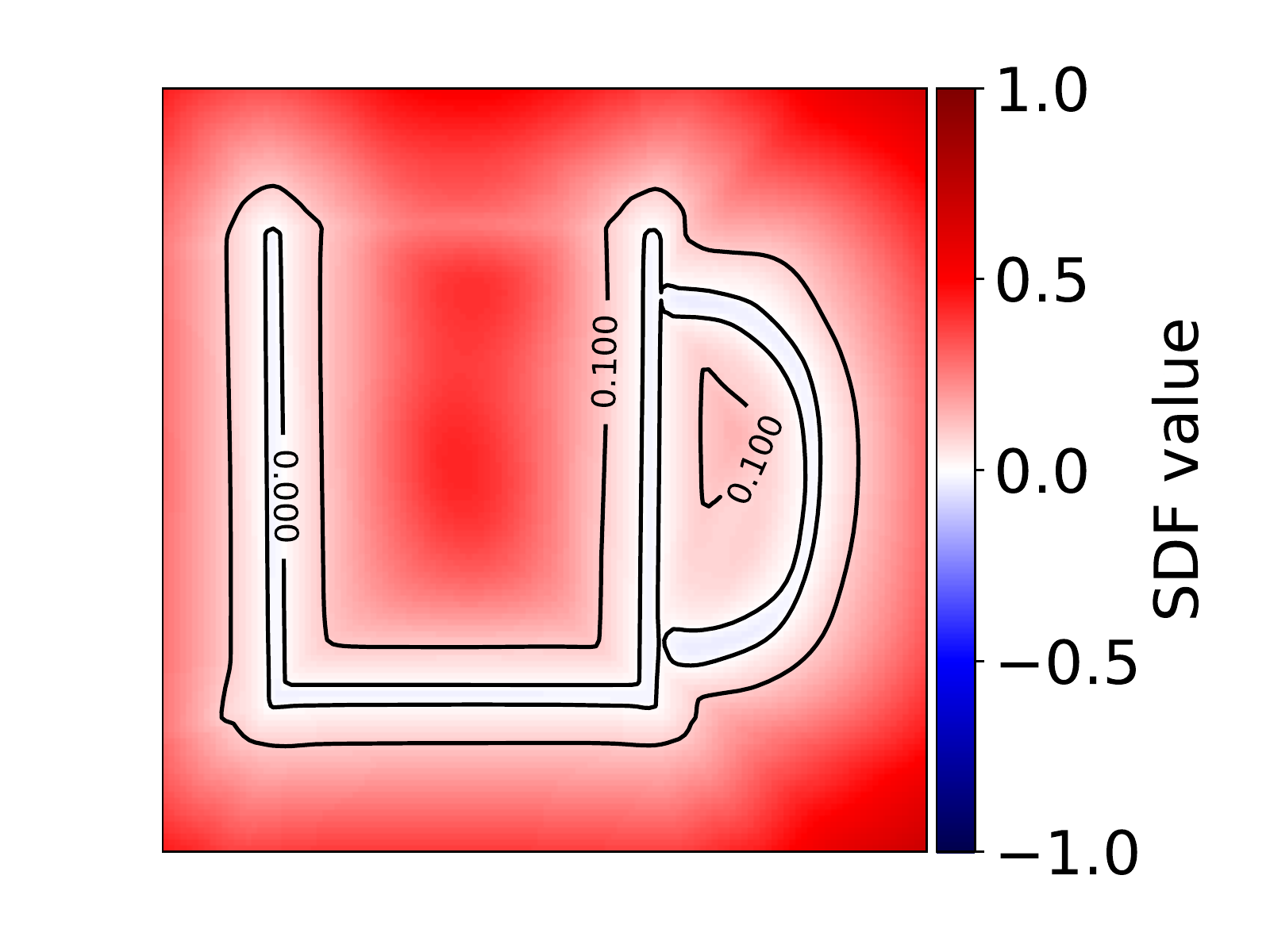}
\end{subfigure}
\begin{subfigure}{.24545\linewidth}
\includegraphics[trim=80 30 100 30,clip,width=.39\linewidth]{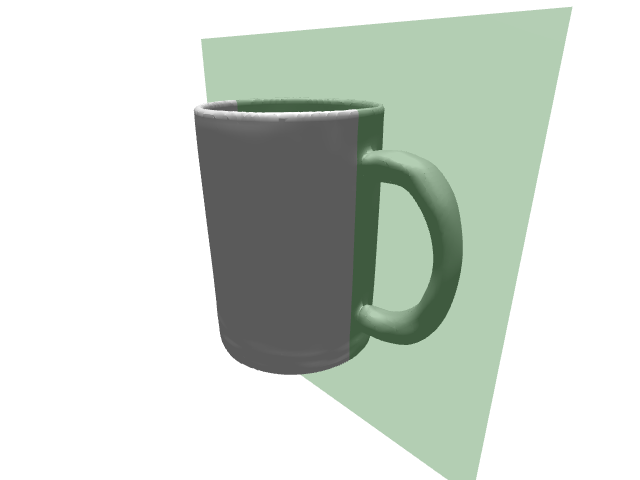}
\includegraphics[width=.59\linewidth]{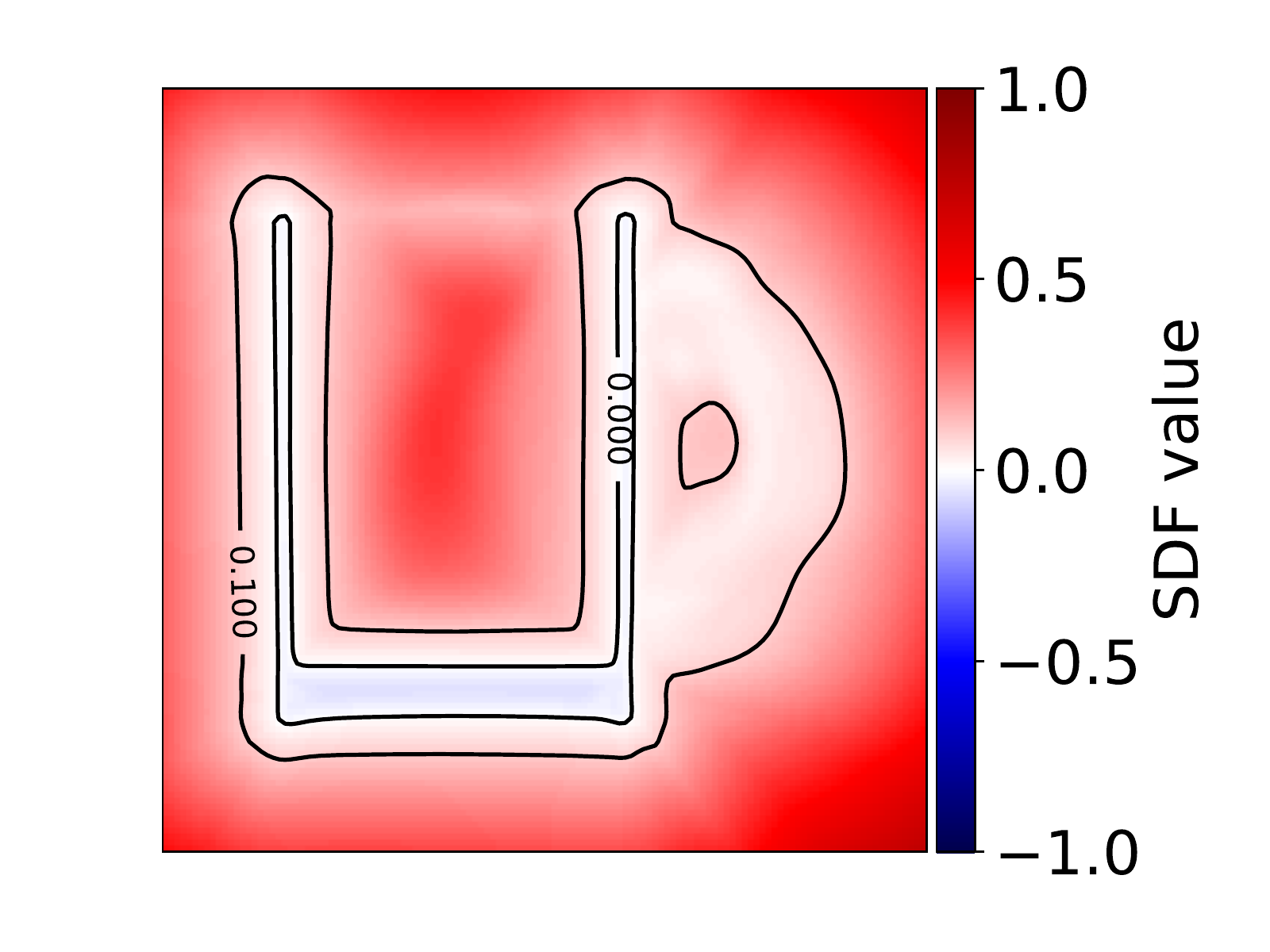}
\end{subfigure}
\begin{subfigure}{.24545\linewidth}
\includegraphics[trim=80 30 100 30,clip,width=.39\linewidth]{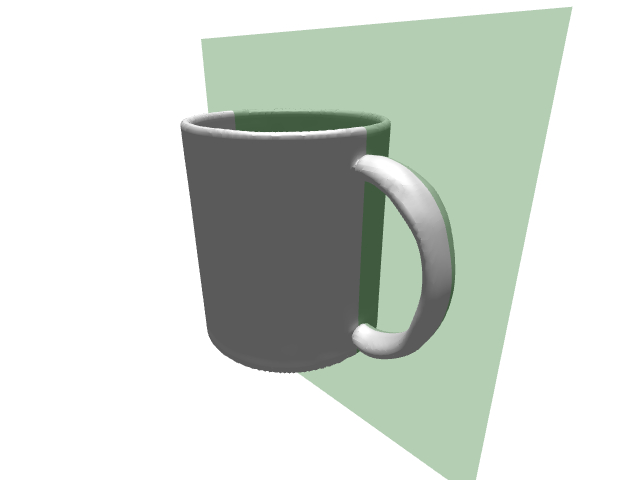}
\includegraphics[width=.59\linewidth]{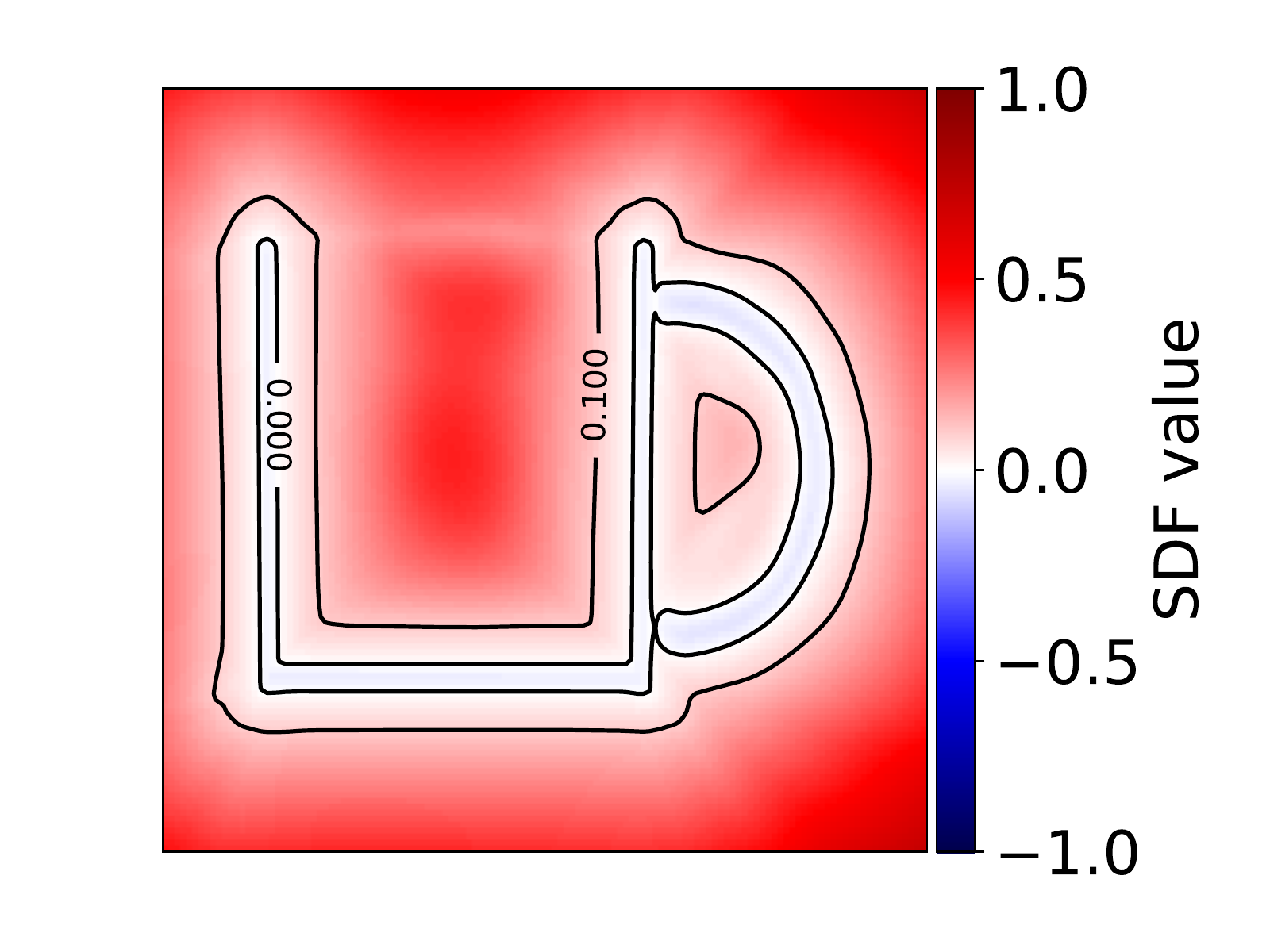}
\end{subfigure}
\begin{subfigure}{.24545\linewidth}
\includegraphics[trim=80 30 100 30,clip,width=.39\linewidth]{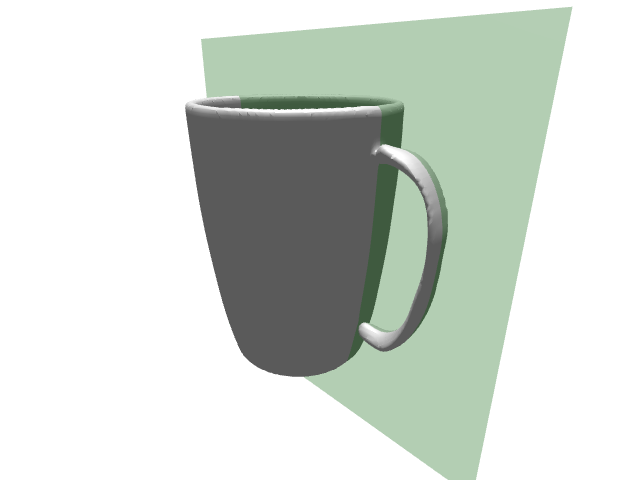}
\includegraphics[width=.59\linewidth]{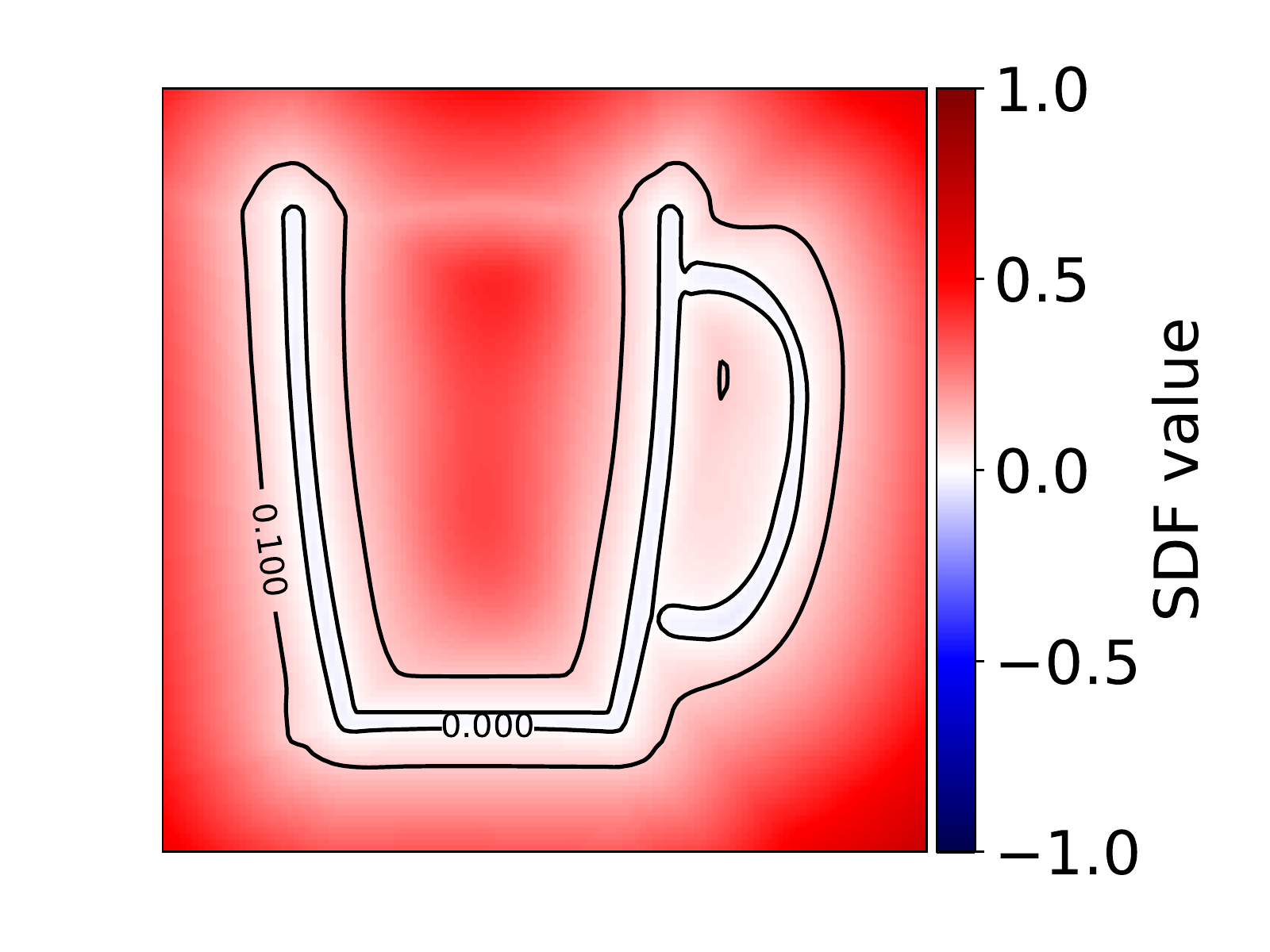}
\end{subfigure}
\begin{subfigure}{.24545\linewidth}
\includegraphics[trim=80 30 100 30,clip,width=.39\linewidth]{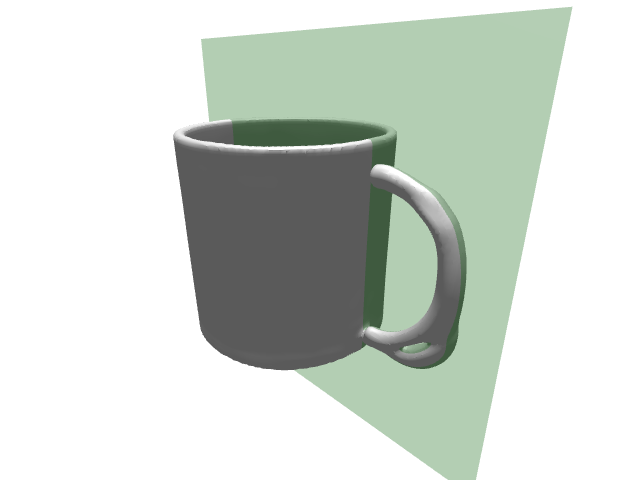}
\includegraphics[width=.59\linewidth]{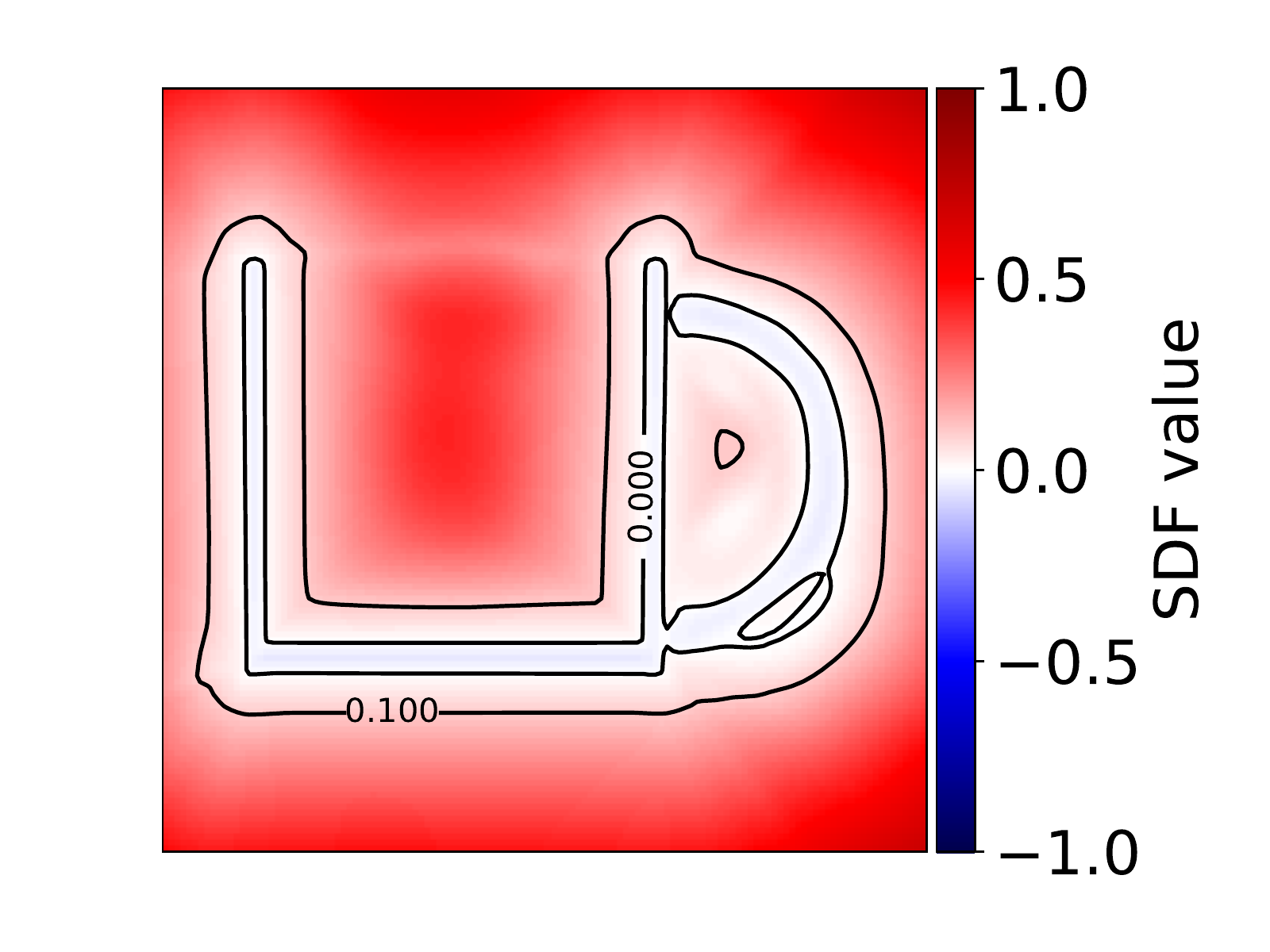}
\end{subfigure}
\begin{subfigure}{.24545\linewidth}
\includegraphics[trim=80 30 100 30,clip,width=.39\linewidth]{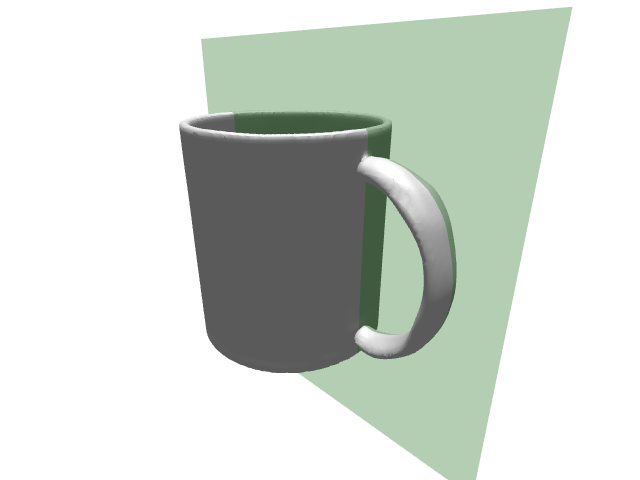}
\includegraphics[width=.59\linewidth]{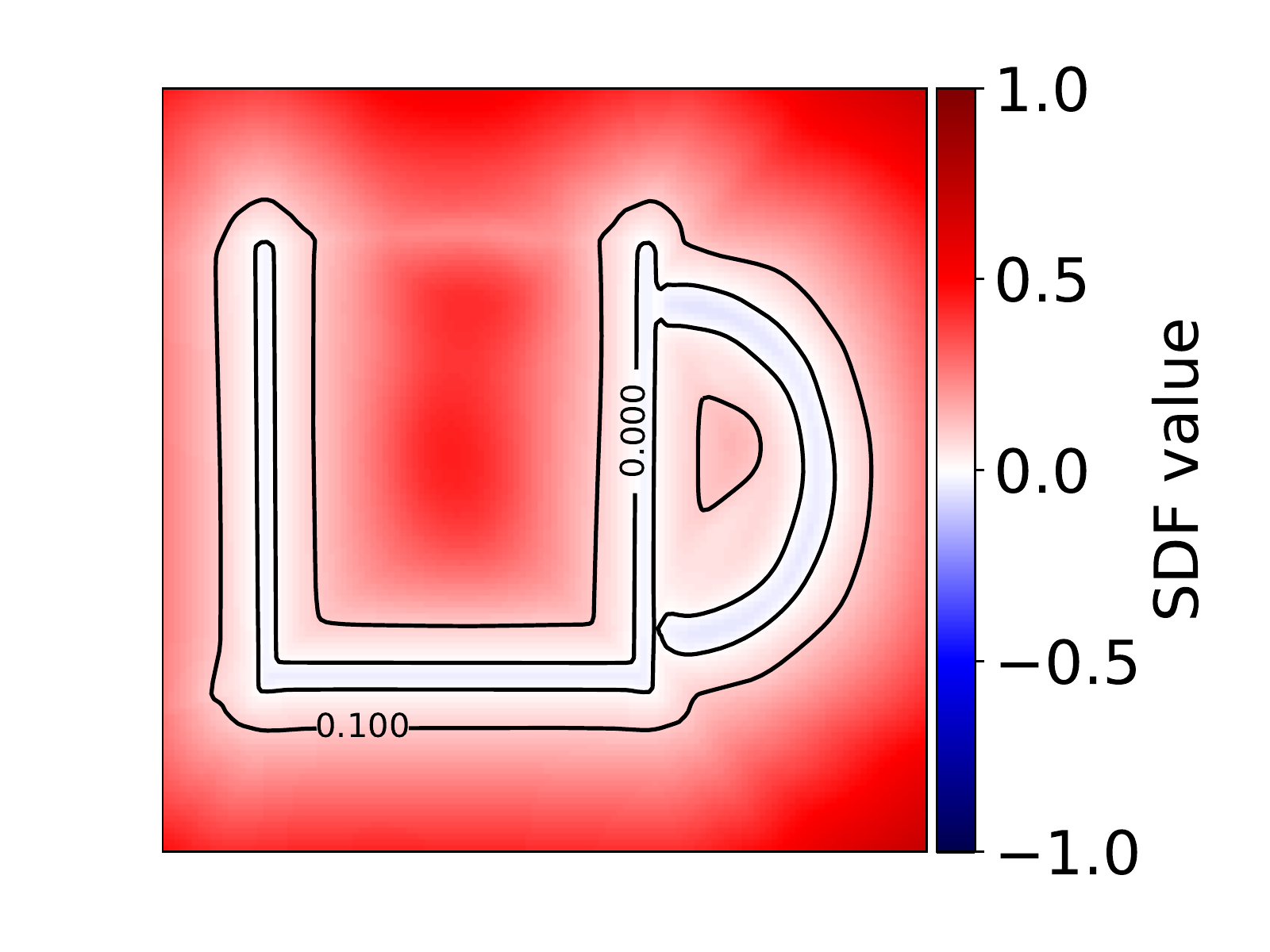}
\end{subfigure}
\begin{subfigure}{.24545\linewidth}
\includegraphics[trim=80 30 100 30,clip,width=.39\linewidth]{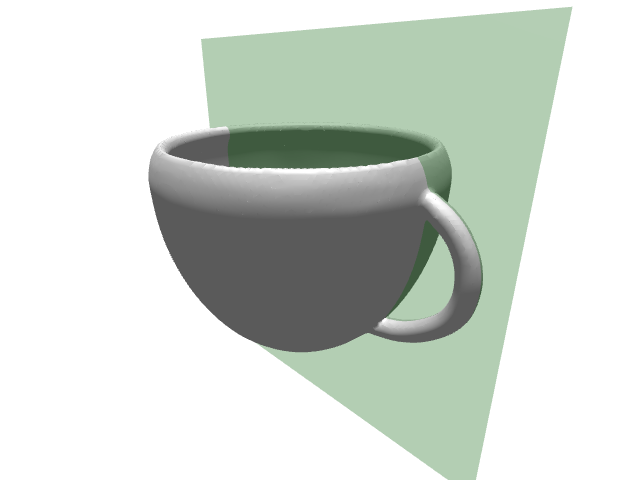}
\includegraphics[width=.59\linewidth]{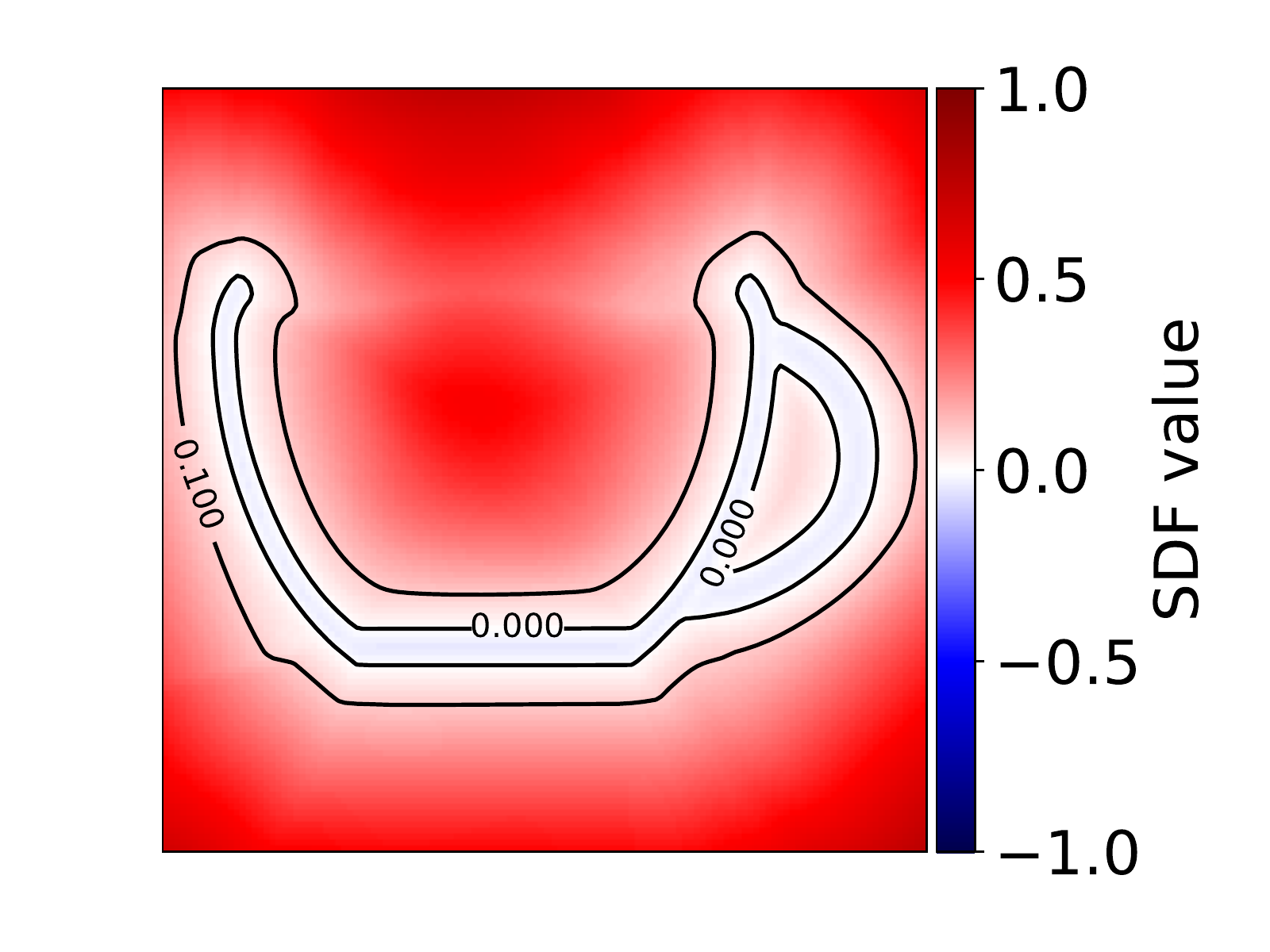}
\end{subfigure}
\begin{subfigure}{.24545\linewidth}
\includegraphics[trim=80 30 100 30,clip,width=.39\linewidth]{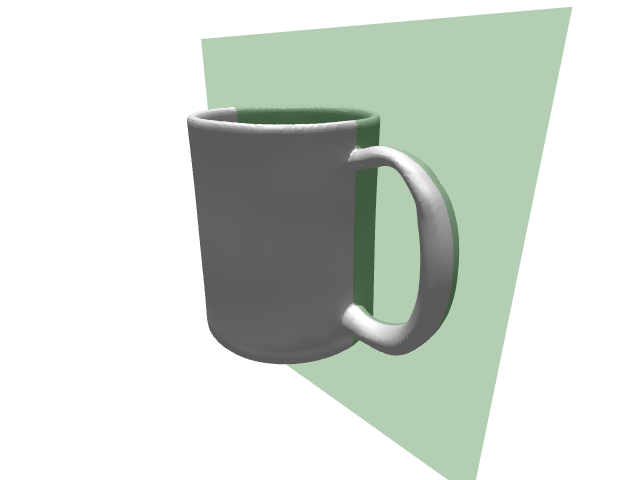}
\includegraphics[width=.59\linewidth]{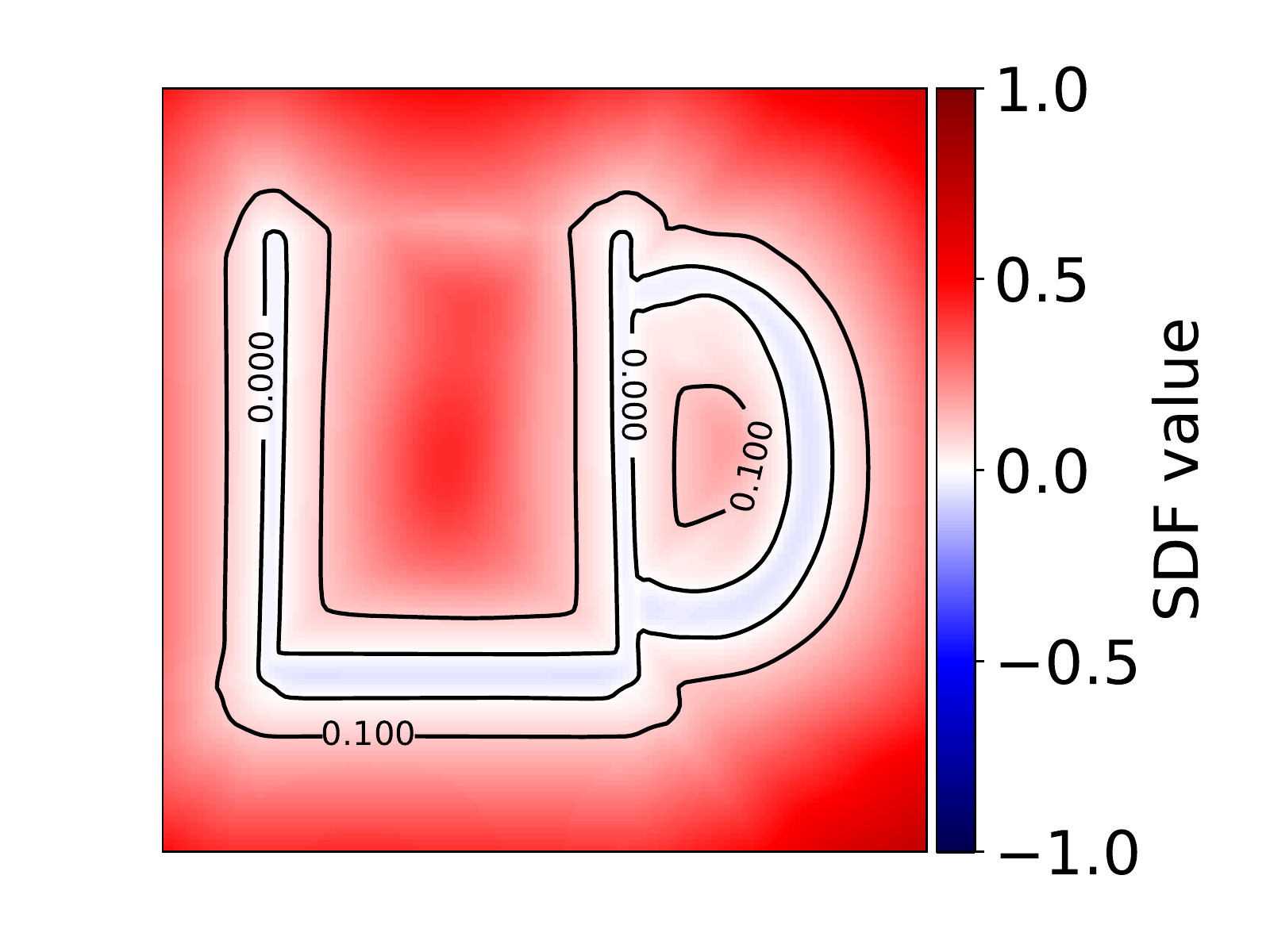}
\end{subfigure}